\documentclass[journal]{vgtc}     
\onlineid{0}
\vgtccategory{Research}

\title{Visualizing High-Dimensional Graph Embeddings via Informed Multi-View Projections}

\abstract{Graphs are commonly visualized in 2D, where humans readily interpret spatial relationships, yet such layouts often distort higher-dimensional structure. We propose to embed graphs in high-dimensional space and search for informative 2D viewpoints that optimize aesthetic and readability metrics (e.g., edge crossings and angular resolution), 
enabled by a novel differentiable surrogate for edge crossings. Numerical experiments show that these viewpoints consistently outperform standard 2D layouts, and can even surpass methods explicitly designed to optimize these metrics. We further introduce \DataFly, an interactive system for exploring multiple candidate viewpoints through seamless navigation. A usability study demonstrates that our approach reveals structural patterns that remain hidden in conventional 2D visualizations.}


\teaser{
  \centering
    \includegraphics[width=\linewidth]{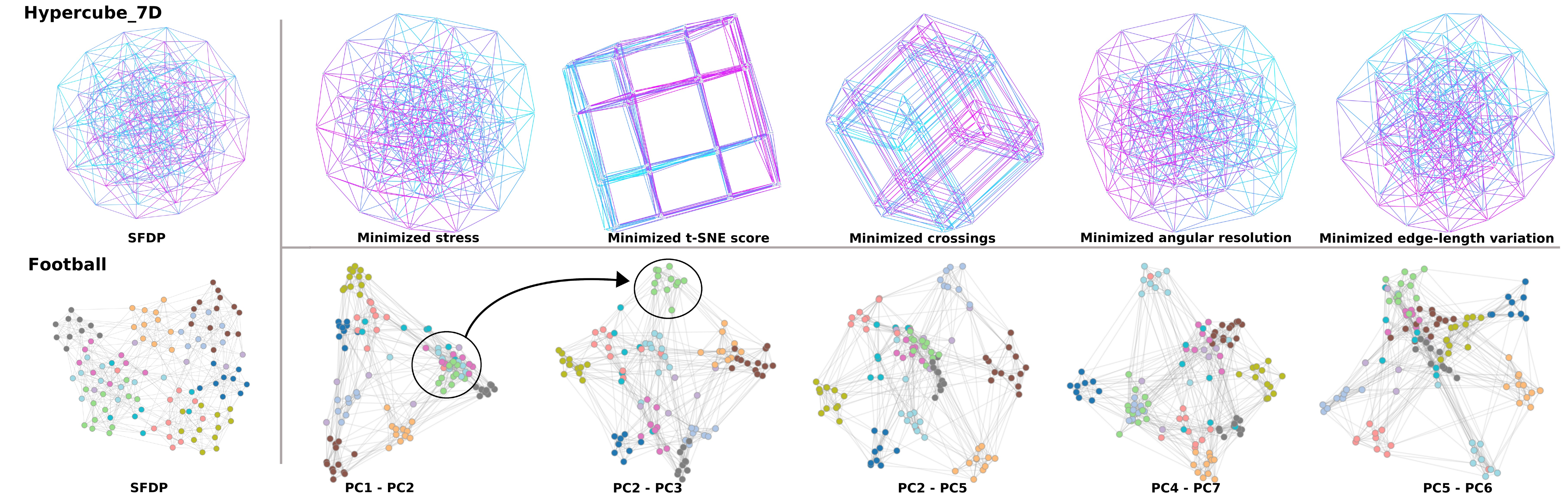}
    \caption{Example visualizations produced from high-dimensional projections. (Top) shows a 7D hypercube embedded via \sfdp in 10D and projected to optimize stress, t-SNE score, edge crossings, angular resolution, and edge length variance, respectively. 
    (Bottom) shows the classic football network embedded via a spectral embedding in 10 dimensions, from five different axis-aligned viewpoints. 
    The set of projections reveals that class memberships are spatially salient in some projections and disappear in others, highlighted by the circled cluster in the first two projections.}
  \label{fig:teaser}
}

\graphicspath{{figs/}{figures/}{pictures/}{./images/}{./}} 

\usepackage{tabu}
\usepackage{booktabs}
\usepackage{ccicons}
\usepackage{mathptmx}
\usepackage[T1]{fontenc}
\usepackage{dfadobe}
\usepackage{cite}
\usepackage{tabularx}
\usepackage{graphicx}
\usepackage{siunitx}
\usepackage[utf8]{inputenc}
\usepackage{array}
\usepackage[table]{xcolor}
\usepackage{xspace}
\usepackage{makecell}
\usepackage{lineno}
\usepackage{todonotes}
\usepackage{amsmath}
\usepackage{amssymb}
\usepackage{multirow}
\usepackage{soul}
\usepackage{colortbl}
\usepackage{comment}

\newcommand{\graymidrule}{%
  \noalign{\vskip -0.6ex}%
  \arrayrulecolor{gray!60}\midrule
  \arrayrulecolor{black}%
  \noalign{\vskip -0.6ex}%
}

\definecolor{mygray}{gray}{0.9}

\newcommand{\HighD}{$K$}
\newcommand{\highD}{K}
\newcommand{\smartgd}{SmartGD}
\newcommand{\neato}{\texttt{neato}\xspace}
\newcommand{\neatoNoSpace}{\texttt{neato}}
\newcommand{\sgd}{$\text{SGD}^2$\xspace}
\newcommand{\sfdp}{\texttt{sfdp}\xspace}
\newcommand{\sfdpNoSpace}{\texttt{sfdp}}
\newcommand{\pivotmds}{\texttt{PivotMDS}\xspace}
\newcommand{\spectral}{\texttt{spectral}\xspace}
\newcommand{\DataFly}{\texttt{DataFly}\xspace}
\newcommand{\dataFly}{\texttt{DataFly}}

\newcommand{\OptProj}{DF${}^{\text{neato}}_{\text{opt}}$}
\newcommand{\OptProjSfdp}{DF${}^{\text{sfdp}}_{\text{opt}}$}

\newcommand{\OptPCASfdp}{DF${}_{\text{PCA}}^{\text{sfdp}}$}
\newcommand{\SPC}{SPC\xspace}
\newcommand{\sPC}{SPC}
\newcommand{\mysubsubsection}[1]{\smallskip\noindent\textbf{#1}}

\AtBeginDocument{
 
}

\newcommand{\circmark}[1]{\tikz[baseline=(char.base)]{\node[shape=circle,draw,inner sep=0.5pt](char){{\scriptsize #1}};}}

\begin{document}

\author{
\authororcid{Ya Ji}{0009-0009-2542-6095},
\authororcid{Xuefeng Li}{0009-0007-6059-9747},
\authororcid{Timo Brand}{0009-0004-3111-2045},
\authororcid{Jacob Miller}{0000-0002-0567-785X},
\authororcid{Peng Zhang}{0009-0002-7858-3691},
\authororcid{Stephen Kobourov}{0000-0002-0477-2724},
 and
\authororcid{Yifan Hu}{0000-0003-2017-924X}
}

\shortauthortitle{Ji \MakeLowercase{\textit{et al.}}: Visualizing High-Dimensional Graph Embeddings}

\authorfooter{
  \item Ya Ji, Xuefeng Li, Peng Zhang and Yifan Hu are with the Khoury College of Computer Sciences, Northeastern University, Seattle, WA, USA. E-mail: \{ji.ya1 | li.xuefen | zhang.peng2, yif.hu\}@northeastern.edu.
    
  \item Timo Brand, Jacob Miller and Stephen Kobourov are with the School of Computation, Information and Technology, Technical University of Munich, Heilbronn, Germany. E-mail: \{timo.brand | jacob.miller | stephen.kobourov\}@tum.de.
}

\firstsection{Introduction}

\maketitle

\thispagestyle{plain}
\pagestyle{plain}
\setcounter{page}{1}

Graph visualization converts abstract relationships into readable graphics, enabling users to understand complex relational data, uncover patterns, and make informed decisions. Traditionally, graphs are visualized in 2D, reflecting human-interpretable spatial relationships. However, large real-world graphs often exhibit characteristics reminiscent of high-dimensional data: the small-world phenomenon, in which most node pairs are connected by short paths, mirrors the distance concentration observed in high dimensions, and real-world graphs, like high-dimensional point clouds, tend to be locally dense and globally sparse. Two-dimensional layouts struggle to preserve these higher-dimensional relationships. Just as a cube cannot be projected onto a 2D plane with equal edge lengths, complex networks contain local connectivity patterns, such as dense cliques or overlapping communities, that no single 2D projection can faithfully represent. It follows that a single 2D graph layout, although intuitive and accessible, risks distorting or obscuring the higher-dimensional structure of the data.

However, high-dimensional layouts pose a practical challenge: presenting them to human readers requires selecting a 2D projection, or viewpoint, at any time.
Thus, effective viewpoints are essential for high-dimensional layouts as visualization tools, yet this has seen limited study. Many methods compute a single projection optimal under some criterion~\cite{Gajer_Goodrich_Kobourov_2000,PivotGraph,spectral}.
Van Wageningen et al.~\cite{Wageningen2025} study the viewpoint problem for 3D drawings, but whether effective projections of higher-dimensional layouts can be achieved remains unclear.

These observations motivate the following research question: \textit{Do high-dimensional graph embeddings contain 2D projections that reveal structural patterns otherwise obscured by traditional static layouts?} To investigate this, we develop a computational pipeline to systematically explore high-dimensional graph layouts, which produces a family of candidate projections.
We further introduce a gradient-based optimization scheme that finds near-optimal
projections for a given quality metric, such as stress, edge crossings, or angular resolution. 
Our findings show that such projections can outperform state-of-the-art algorithms that are explicitly designed to optimize these same metrics directly in 2D. Throughout the paper, we use the term \emph{optimal projection} as shorthand for the result of this gradient-based optimization, noting that it converges to a local rather than global optimum.

To make these high-dimensional embeddings accessible and explorable, we provide \DataFly, an interactive online tool for real-time visualization of high-dimensional graph layouts. \DataFly enables principled exploration of viewpoints: users can inspect optimal projections for specific metrics, or transition between them by ``flying over'' the high-dimensional space, observing how one projection morphs into another. Because all viewpoints derive from a single shared embedding, diverse visual perspectives on the same network coexist within a coherent spatial structure, allowing users to discover structural patterns that remain hidden in any single 2D layout.

Our main contributions are as follows:

\begin{itemize}
\item We show empirically that an optimal 2D projection of a high-dimensional graph embedding can yield better metric values than state-of-the-art benchmarks that directly optimize them in 2D.
        
\item  We propose a differentiable loss function, SigmoidX, which enables the effective minimization of edge crossings, and demonstrate that the optimal projection using this loss function achieves significantly fewer crossings than \sgd and \smartgd. 
        
\item We provide a working prototype of \DataFly, an interactive system for exploring high-dimensional graph layouts and identifying viewpoints of interest, validated by a small usability study with both expert and non-expert participants.     
\end{itemize}

\DataFly, together with a demonstration video, is available at \textbf{\href{https://w3id.org/datafly}{https://w3id.org/datafly}}.
In the remainder of the paper, \dataFly\ refers to both the interactive system and the associated optimal-projection framework.  Throughout the paper, we use two edge-coloring schemes. By default, the \DataFly system colors edges according to a breadth-first search (BFS) traversal, from blue to red, providing a stable visual anchor that helps preserve the user’s mental map during interaction. For quantitative results illustrating the effect of optimal projections (e.g., minimizing edge-length variance), edges are instead colored by length using a reversed Turbo colormap (red–green–blue), where warmer colors indicate shorter edges and cooler colors indicate longer edges.

\section{Related Work}
\label{sec:related}

Since 1963~\cite{tutte_1963}, numerous graph drawing algorithms have been proposed. A widely used class of methods for generating straight-line drawings relies on physical models that minimize the energy of the system. 
Examples include stress-based models, which minimize the stress energy of springs connecting nodes~\cite{kamada_kawai_1989, neato, zheng-gd2}, and force-directed systems of springs and electrical charges, in which springs along edges contract while electrical charges on nodes repel each other~\cite{Eades_1984,Fruchterman_Reingold_1991,kobourov_2013}. 

\subsection{Layout algorithms for optimizing specific metrics} 
There are a number of layout methods that specifically target aesthetic metrics, such as minimizing edge crossings~\cite{xing-heuristic, spx},  maximizing crossing angles~\cite{Argyriou, Bekos-xangle}, or a combination~\cite{didimo}. Two notable algorithms are \sgd~\cite{sgd2} and \smartgd~\cite{wang_2023_smartgd}.

\sgd\ is a method that can jointly optimize multiple graph readability criteria. 
In particular, it can optimize any criterion that can be described by a differentiable function. 
When a criterion is not a differentiable function (e.g., edge crossings), a proxy objective is used instead.
\sgd supports optimization of 
angular resolution, edge crossings, and desired edge lengths, among other criteria. 
\smartgd~\cite{wang_2023_smartgd} is a deep-learning framework for graph drawing based on generative adversarial networks. 
It can learn from examples of a ``good'' layout, where ``good'' may be defined by even non-differentiable metrics, e.g., edge crossings. While $SGD^2$ and \smartgd\ are among the first graph embedding methods that can simultaneously optimize multiple criteria, neither works well beyond small graphs ($|V| \approx 100$). The optimal projection algorithm proposed in this paper performs well on both small and large graphs (over 1000 nodes) while supporting multiple criteria.

While joint optimization seeks to be effective at minimizing a given metric
while preserving the aesthetic appeal of the layout,
it will not necessarily generate a drawing with any one criterion optimal. 
Of particular interest to the algorithmic community are algorithms that optimize a specific metric, possibly at the expense of readability. 
One such algorithm is \textit{Vertex Movement}\cite{xing-heuristic} (VM), which optimizes edge crossings while ignoring other aesthetic criteria.
The algorithm starts with an initial layout, iterates over the vertices, and tries to find the crossing-minimal position for the current vertex. 
While not guaranteed to find an optimal drawing, it achieves strong performance in finding drawings with few edge crossings. 
We use VM as a state-of-the-art baseline for crossing minimization, 
but note that this algorithm has high computational complexity, and cannot be applied to large graphs.

\subsection{Understanding high-dimensional data through projection.} Paulovich et al.~\cite{paulovich2025dimensionality} recently linked dimensionality reduction and graph drawing, showing their close relationship. Algorithms and interactive systems for exploring high-dimensional data via projections have long been studied, with early work in the statistics and physics communities~\cite{buja:1996:XGobi, cook1995grand, cutura2018viscoder, Dang2014ScagExplorer, laa2020high, huh2002visualization, Morariu_2023}. Notable examples include the R-based visualization tool XGobi~\cite{buja:1996:XGobi} and the Embedding Projector~\cite{smilkov2016embedding}. 
GraphDice~\cite{GraphDice2010} is a relevant tool for exploring and analyzing graph data. It supports exploration of multivariate social network graphs through animated transitions between nearby views. 
However, it operates on observed actor and edge attributes rather than on geometric projections of a high-dimensional graph embedding.
EvoGraphDice~\cite{EvoGraphDice2012} extends this line of work to general multidimensional data, using interactive evolutionary search to propose specific views based on user feedback. 
It is therefore relevant as a visual analytics approach to guided viewpoint discovery, but it is not designed for graph layouts or for optimizing graph-drawing metrics.
As far as we are aware, there is little prior work focused on understanding high-dimensional graph embeddings (beyond 2/3D) via optimal projection. Gajer et al.~\cite{Gajer_Goodrich_Kobourov_2000} demonstrated that laying out a graph in a high-dimensional space and randomly projecting it back to 2D can produce interesting drawings, illustrated using a Möbius strip example; however, their work does not aim to identify optimal views.
Also related is work by Gaertler and Krug, who used animations between different viewpoints of a spectral graph embedding~\cite{gaertler05}.

Li et al.~\cite{li2020visualizing} investigated methods for visualizing multidimensional data arising from neural networks. Although they examined nonlinear dimensionality reduction techniques such as t-SNE and UMAP, linear projections were favored due to their predictable response to changes in the original data. 

Etemadpour et al.~\cite{Etemadpour2015} conducted a perception-based user study on multidimensional projections, showing that projection performance depends on tasks and data characteristics. The study shows that no single projection method yields universally optimal layouts, which supports the argument for 
generating and evaluating multiple views of high-dimensional graph layouts. 
Recently, Lekschas and Abdennur~\cite{lekschas2026dtour} provide grand and guided tours of multidimensional data with dtour, but these are not designed for graph inputs.

Martins et al.~\cite{Martins2012MDProjection} introduce multidimensional projection methods for social network visualization, where node placement is guided by 
node similarity to produce meaningful 2D layouts without heavy force-based optimization. The work demonstrates that high-dimensional embeddings and projection-driven views can reveal structural patterns beyond conventional graph drawing, reinforcing the motivation for multi-view exploration of high-dimensional graph layouts.

It has been shown that 2D viewpoint drawings acquired from 3D drawings can be of higher quality than 2D drawings produced with the same algorithm~\cite{Wageningen2025}.
This result is achieved through  optimization including gradient descent and evolutionary-inspired metaheuristics, that automatically find high-quality viewpoints for 3D graph drawings, replacing slow brute-force sampling methods that require evaluating thousands of viewpoints. Although this viewpoint optimization is similar to the optimal projection found in \DataFly, we aim to find good viewpoints for a \highD-D graph embedding instead. Recent findings~\cite{joos2025show} show that, after creating such a 3D drawing, users prefer to view it from angles that yield 2D projections with high values for several 2D quality metrics. 
Similar findings were shown for Dimensionality Reduction (DR) plots~\cite{castelein2023based}. We primarily consider linear projections, though a rich set of non-linear DR techniques exist; see recent surveys~\cite{DBLP:journals/tvcg/NonatoA19,DBLP:journals/tvcg/EspadotoMKHT21}.

\section{Understanding Graphs through High-Dimensional Layouts}
\label{sec:overview}

Here, we give a high level overview of our methodology. We use standard graph definitions: $G = (V, E)$, where $V$ denotes the set of vertices and $E$ denotes the set of edges.
We posit that embedding graphs in a high-dimensional space enables a more faithful representation of their intrinsic structure, such as local neighborhoods. To this end, we first embed the graph in \HighD-dimensional space (e.g., \HighD{} = 10), using an off-the-shelf algorithm. 
What remains is to investigate potentially interesting 2D viewpoints of this embedding for visualization. 

\begin{figure}[t]
    \centering
    \includegraphics[width=\linewidth]{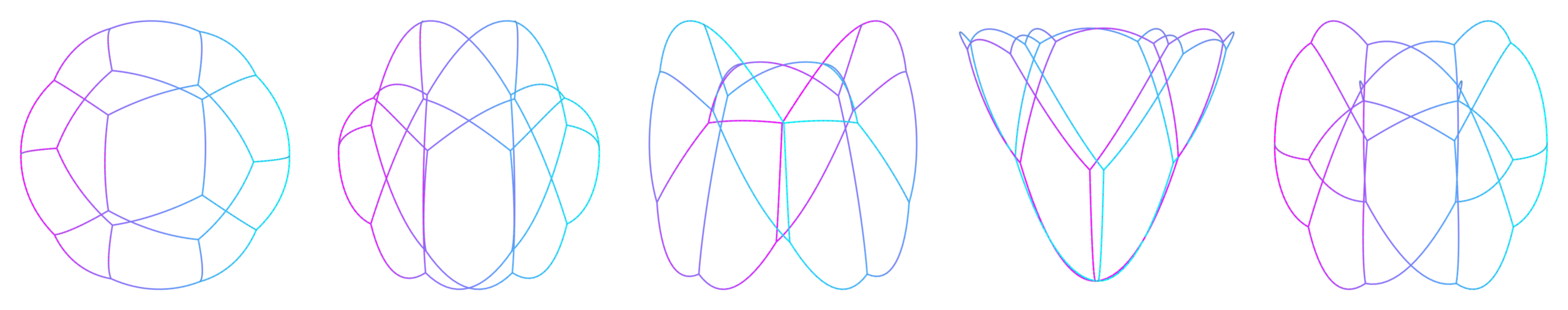}
    \caption{Several projections of a 10D spectral embedding of a subdivided dodecahedron graph. The leftmost projection looks like a typical embedding, but other viewpoints reveal the structure twists and wraps on itself in the ambient space.}
    \label{fig:dodecahedron}
\end{figure}

\Cref{fig:dodecahedron} illustrates how different projections of the same 10D embedding can reveal qualitatively different structures. However, it is not clear that such viewpoints offer any real advantages for real-world data. 
We posit the following research questions:

\noindent \textbf{RQ1}. Do high-dimensional graph embeddings, when combined with carefully designed projection methods, offer advantages over direct 2D layouts in optimizing specific aesthetic metrics? 

\noindent \textbf{RQ2}. Can exploring systematic projections of high-dimensional viewpoints help users understand graph structure? 

We address \textbf{RQ1} through the optimal viewpoint formulations developed in this section and the quantitative evaluation in~\cref{sec:results}; \textbf{RQ2} later through qualitative examples (\cref{sec:visual_comparison}) and by introducing an interactive tool, \DataFly, validated through a usability study (\cref{sec:case-study-results}).

In this section, we propose carefully designed 2D projections of high-dimensional layouts, referred to as \textit{viewpoints}, to address \textbf{RQ1}. 
Our goal is to identify viewpoints that are most likely to provide informative insights into the graph structure. 

\subsection{High-dimensional layouts} 

Many graph layout algorithms 
generalize naturally to high dimensions. In our work we consider four approaches: a stress model~\cite{neato}, a force-directed algorithm~\cite{Hu_2004}, the spectral embedding algorithm~\cite{Hall_1970_eigenplacement}, and PivotMDS~\cite{Brandes_Pich_2006}. However, the pipeline we describe works for any embedding algorithm in principle.

For the stress model, we use \neato~\cite{neato} from the Graphviz library~\cite{graphviz}. 
Stress-based optimization is one of the standard 2D graph embedding methods and is available in most software platforms, ranging from Graphviz and NetworkX~\cite{networkx}, to OGDF~\cite{Chimani2014OGDF} and yFiles~\cite{Wiese2004_yfiles}. Stress-based layouts are often included as a baseline when evaluating new methods, and the implementation in Graphviz is a trusted standard.
The notion of stress (\cref{eqn_stress}) is based on Euclidean distances and therefore well-defined in high dimensions. Human-subject studies~\cite{Chimani_2014_stress,Mooney25} have shown that lower stress is an important predictor of aesthetic preference in graph visualization. 

Perhaps the only graph embedding methods as popular as stress-based models are force-directed methods~\cite{kobourov_2013}. These methods typically minimize a spring-electrical energy (\cref{eqn_sfdp}) that captures attractive forces between pairs of nodes connected by edges, and repulsive forces between all pairs of nodes. As a representative of force-directed algorithms, we use \sfdp~\cite{Hu_2004} from the Graphviz library~\cite{graphviz}. This is an efficient implementation frequently used as a baseline for evaluation.

We evaluate two additional algorithms 
as they naturally lay out graphs in up to $|V|$ dimensions. 
The spectral algorithm~\cite{Hall_1970_eigenplacement} is based on taking the top \HighD-eigenvectors of the Laplacian corresponding to the lowest \HighD-eigenvalues. 
We use the implementation in the NetworkX library~\cite{networkx}. 
We also include the PivotMDS algorithm~\cite{Brandes_Pich_2006} (similar to HDE~\cite{harel_koren_2002}, and implemented based on \cite{Brandes_Pich_2006,hu_2012_large_networks}), 
a scalable variant of the classical MDS algorithm~\cite{Torgerson_MDS_1952}. 
It works by picking \HighD~pivot nodes, forming the $|V|\times\highD$ squared and double-centered distance matrix to the pivots, and performing an eigendecomposition on this matrix. 

Given the wide popularity of the stress metric, we present most of the results for projections of high-dimensional layouts produced by \neato, alongside corresponding results for \sfdp. Additional results are given in Appendix \cref{sec_appendix_additional_res_neato} and \cref{sec_appendix_additional_res_sfdp}. We also limit our evaluation to up to 10D embeddings, as Graphviz officially supports graph layouts in at most 10 dimensions (but see \cref{sec_appendix_30D} for preliminary results up to 30D).

Although higher dimensions provide more freedom for the layout to preserve connectivity, the final display of information to human readers is limited to the 2 dimensions of the computer screen. We must therefore rely on informative 2D projections to interpret these high-dimensional layouts. 

\subsection{Optimal Viewpoints} 

We define a \textit{viewpoint} as a specific 2D projection of a \HighD-D layout of a graph. While there are infinitely many possible viewpoints, some of them are more interesting than others.

An \textit{optimal viewpoint} is defined as the projection from \HighD-D to 2D that minimizes a specific metric $\mathcal{L}$. 
For brevity, we refer to locally optimized viewpoints as optimal viewpoints.
We collectively call these viewpoint-optimizing algorithms \DataFly.

Let $X \in \mathbb{R}^{|V| \times K}$ denote a high-dimensional embedding matrix, where each row represents the \HighD-D embedding of a node, obtained from an algorithm. 
The goal 
is then to learn a linear projection $P \in \mathbb{R}^{K \times 2}$ that maps $X$ to a two-dimensional space while minimizing a loss of the projected layout. 
Formally, we solve

\begin{equation}
    \min_{P \in \mathbb{R}^{K \times 2}} \ \mathcal{L}(X P) \label{eqn_optimal}
\end{equation}

\noindent where $\mathcal{L}$ is the graph metric, and $XP \in \mathbb{R}^{|V| \times 2}$
gives the 2D coordinates resulting from the projection. 
Note that this optimization involves only $2\highD$ variables (e.g., $\highD=10$), 
which is far fewer than the $2|V|$ variables that algorithms for traditional layouts 
must optimize over, making the problem easier and less computationally expensive to solve.

To find the optimal projection $P$ in \cref{eqn_optimal}, we use an iterative gradient descent algorithm, updated as:
\begin{equation}
    P_{t+1} = P_t - \eta \nabla_{P_t} \mathcal{L}(X P_t), \label{eqn_gd}
\end{equation}
\noindent where $\eta$ is the step size.

\subsection{Metrics}\label{sec:metrics}

Graph layout quality is inherently multi-objective, and no single metric consistently captures aesthetic quality across different graphs and tasks. Different metrics instead reflect different, and sometimes competing, objectives—for example, stress emphasizes structural faithfulness, whereas edge crossings emphasize readability. Therefore, we explore a diverse set of complementary metrics to evaluate viewpoint quality. 
We denote $x_i\in \mathbb{R}^d$ the coordinates of node $i$ in $d$ dimensions.

\mysubsubsection{Stress:} 
This metric measures the stress of the layout, 
\begin{equation}
\text{stress}=
\sum_{i, j \in V, i \neq j}
    w_{ij} \left(\alpha \|\mathbf{x}_i - \mathbf{x}_j\| - d_{ij}\right)^2 , \label{eqn_stress}
\end{equation}

\noindent where $d_{ij}$ is the desired distance between $i$ and $j$, and $w_{ij}$ is set to $1 / d_{ij}^2$. 
Here, for a given layout, we apply a scaling factor of $\alpha$ to normalize the stress (e.g., \cite{wang_2023_smartgd,smelser2026scale}):

\begin{equation}
\alpha = \frac{ \sum_{i,j \in V, i \neq j}
w_{ij} d_{ij} \| \mathbf{x}_i - \mathbf{x}_j \| }
{ \sum_{i,j \in V, i \neq j} w_{ij} \| \mathbf{x}_i - \mathbf{x}_j \|^2 }.\label{alpha}
\end{equation}

\mysubsubsection{Spring-Electrical Energy:} This metric treats edges as springs and nodes as electrically repelling particles and measures the energy of the system:
\begin{equation}
\text{se\_eng} = \sum_{(i,j) \in E} \frac{\| \mathbf{x}_i - \mathbf{x}_j \|^3}{3K_{s}}  
- \sum_{i \ne j} K_{r}^2\ln(\|\mathbf{x}_i - \mathbf{x}_j\|).  \label{eqn_sfdp}
\end{equation}

\noindent where $K_{s}$ and $K_r$ are spring and repulsion constants. 
We set them to 1. 
To make this metric scale-invariant, we use a scaling factor $t$ that minimizes \cref{eqn_sfdp}. 
We derive an analytical solution (proof is given in the Appendix \cref{sec_appendix_spr_elec_proof}):
$t = \left(\frac{|V|(|V|-1)}{\sum_{(i,j) \in E}\| \mathbf{x}_i - \mathbf{x}_j \|^3}\right)^{1/3}$.

\mysubsubsection{t-SNE Score:} This metric measures Kullback-Leibler divergence of node distances between the Euclidean layout space and the graph space using t-distributed stochastic neighbor embedding~\cite{Maaten_2008_tSNE}. 
It evaluates how well the graph neighborhoods are preserved in the layout. The t-SNE metric (\cref{sec_appendix_tsne}) is not scale-invariant. 
When computing the optimal 2D projection, to ensure good initialization and a differentiable loss function, we normalize the coordinates using \cref{alpha}. 
However, to ensure a fair comparison across methods, we rescale the coordinates for each layout using a 1-D minimization step before computing the reported metric.

\begin{figure}
    \centering
   \includegraphics[height=0.3\linewidth]{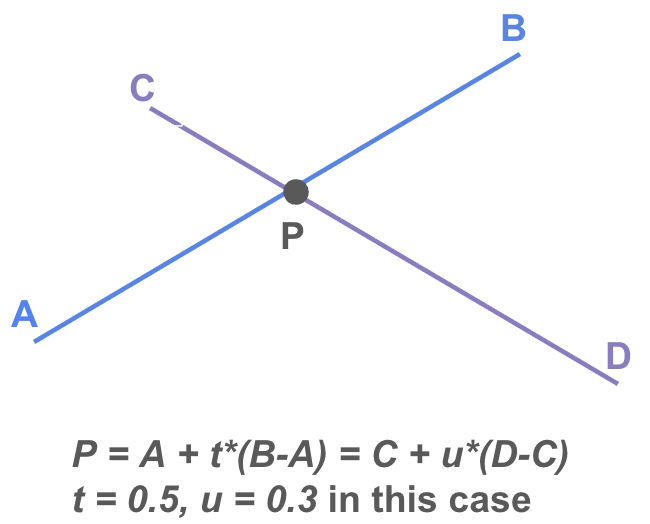} \includegraphics[height=0.3\linewidth]{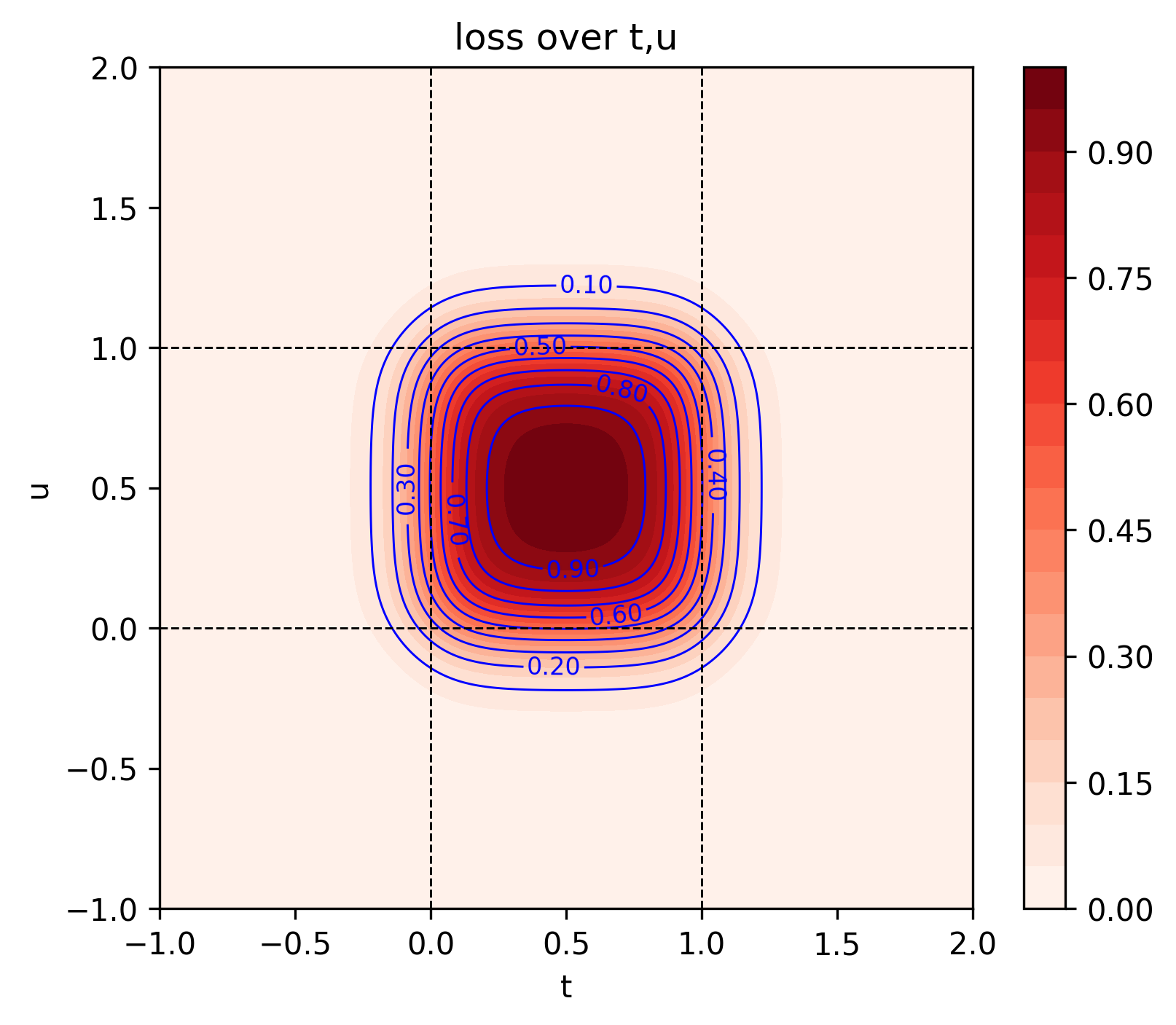}
    \caption{Left: the edge-crossing parameters. Right: contour of the SigmoidX surrogate function for edge crossings.}
    \label{fig:edge_crossing_fun}
\end{figure}

\mysubsubsection{Edge crossings}: This metric counts the number of edge crossings in the layout, indicating visual clutter. Fewer crossings suggest a cleaner, more readable layout.
\begin{equation}
    \text{xing} = \sum_{(i,j),(k,l) \in E, (i,j) \neq (k,l)} \mathbb{I}\left[\text{edge } (i,j) \text{ crosses } (k,l)\right].
\end{equation}
\noindent To apply gradient descent for minimizing edge crossing, a differentiable surrogate function is required.
\sgd~\cite{sgd2} learns a differentiable proxy for edge crossings by training a neural network to predict whether two edges intersect, based on their node coordinates. 
We instead propose a simpler approach. 
Let $t, u$ be the intersection parameters in \cref{fig:edge_crossing_fun}~(left). 
Geometrically, when $t\in [0,1]$, this means that the second edge, or its extension, hits the first edge. 
The parameter $u$ is defined symmetrically to $t$. 
Hence, if the two edges cross, we must have that both $t, u\in [0,1]$. 
But instead of using such a hard, non-differentiable condition to determine whether edges intersect, we use sigmoid-based soft masks:
\begin{equation*}
    M_t = \frac{\sigma(t)\,[1 - \sigma(t - 1)]}{M_{\text{peak}}}, \qquad
    M_u = \frac{\sigma(u)\,[1 - \sigma(u - 1)]}{M_{\text{peak}}},
\end{equation*}
\noindent where $\sigma(t) = \frac{1}{1+e^{-T\cdot t}}$ is the sigmoid function, 
with temperature $T$ controlling the steepness, and \(M_{\text{peak}}\) normalizing the peak value to 1.

The differentiable probability of an edge crossing is then $P(\text{crossing}) = M_t M_u$. 
It approaches 1 when the edges intersect (\(t,u \in [0,1]\)), but quickly and smoothly decreases toward 0 otherwise.
We call this loss function \textit{SigmoidX}.
It is a continuous, differentiable proxy for the binary ``edges intersect’’ test, enabling gradient-based minimization of edge crossings. 
We illustrate this function in~\cref{fig:edge_crossing_fun}. 
Additional details are given in \cref{sec:sigmoidx} of the Appendix. SigmoidX is used to find the optimal projection, but we report the actual edge crossings.

\mysubsubsection{Edge Length Variance:} This metric measures the standard deviation of all edge lengths.
Let \( \ell_{ij} = \|\mathbf{x}_i - \mathbf{x}_j\| \) be the length of edge \( (i, j) \in E \), then $\text{edgelen\_var} = \mathrm{std}(\ell_{ij})/\mathrm{mean}(\ell_{ij})$. 
It captures uniformity in edge representation, affecting aesthetic clarity and interpretability.

\mysubsubsection{Angular Resolution:} This metric measures the minimum angle between incident edges at each node — higher resolution improves visual distinction and reduces ambiguity. 
Typically, it is defined as
$
 \pi - \min_{v \in V} \min_{\substack{(v,u), (v,w) \in E \\ u \neq w}} \arccos{ (\mathbf{x}_u - \mathbf{x}_v, \mathbf{x}_w - \mathbf{x}_v) }.
$
However, this only focuses on the worst case, which has limited expressivity. 
In this paper, we aim to improve the angular resolution across all nodes; hence, we adopt an average measure of how much the minimum angle at each node differs from the ideal angle.
For a node $v$ with degree $d_v \ge 2$, the ideal angular spacing is $\theta_v^{*} = \tfrac{2\pi}{d_v}$. 
Let $\hat{\theta}_v^{\min}$ be the smallest angle between consecutive neighbors of $v$.
We define angular resolution as
\begin{equation*}
    \text{ang\_res}  = \sqrt{\frac{1}{|V'|} \sum_{v \in V'} \left(\theta_v^{*} - \hat{\theta}_v^{\min}\right)^2},
\end{equation*}
where $V' = \{v \mid d_v \ge 2\}$ is the set of nodes with degree over 1.

\section{Experimental Evaluation}
\label{sec:results}

To quantify the potential benefits of \DataFly, we compare the metrics of baseline algorithms with those of the \DataFly\ optimal viewpoints. 
Hereafter, we use \neatoNoSpace($K$) to denote the outcome of laying out a graph in \highD-D; the same notation applies to other algorithms. 
By default, \neato\ refers to \neatoNoSpace(2). 
We denote \OptProj(\highD) as the result of the \DataFly\ optimal viewpoint on the \neato\ \highD-D layout for a specific metric, and write \OptProj\ for \OptProj(10). 
Similarly, \OptProjSfdp\ denotes the optimal projection on the \sfdp\ 10-D layout for a specific metric.

\subsection{Experimental Setup}

\noindent\textbf{Optimization procedure}. To find the optimal projection $P$ in \cref{eqn_optimal}, we initialize $P$ using the first two principal components of the \highD-dimensional layout $X$. At each iteration $t$, we perform a gradient descent update (\cref{eqn_gd}) with a learning rate of $\eta = 0.1$, using the Adam~\cite{AdamW} optimizer. The algorithm tracks the best projection matrix $P$ and the corresponding projected coordinates $X P$ that yield the lowest metric value observed during training. We run 200 epochs, which we found sufficient for convergence in most cases. 
Across multiple runs, we found that \OptProj and \OptProjSfdp produced stable results, likely due to the deterministic initialization of $P$. 
The Graphviz implementations of 
\neato\ and \sfdp\ are also stable, and provide consistent results.

\noindent\textbf{Benchmark Datasets}. 
We evaluate using  two benchmark datasets. 
SuiteSparse14 consists of 14 test graphs 
chosen from the SuiteSparse matrix collection~\cite{Davis_Hu_ufl_2009}. 
They are selected to include graphs of different types (e.g., originating from structural, power network, and scientific problems).
The two exceptions are ``mobius'', which is generated from a $5\times 50$ mesh graph of a M\"obius strip. 
This graph was used as an example in Gajer et al.~\cite{Gajer_Goodrich_Kobourov_2000} to motivate the need for high-dimensional layouts of graphs. 
We also include the $2^{10}$-dimensional hypercube ``hypercube10'', since it can only be realized without distortion in 10D. 
The number of nodes and edges is shown in the first column of 
\cref{tab:combined_suitesparse}.
For disconnected graphs, the largest component is taken.

The second benchmark is the Rome graph collection~\cite{rome_graphs}, comprising 11,531 relatively sparse graphs with 10–100 vertices and 9–158 edges. 
We include this dataset to enable comparison with metric-optimizing
baseline algorithms \sgd~\cite{sgd2} and \smartgd~\cite{wang_2023_smartgd}, which were originally designed for smaller graphs and therefore warrant such a benchmark.  
To present results here, we randomly selected 20 graphs from the benchmark for illustration purposes, which we refer to as Rome20.

\noindent\textbf{Aggregating performance metrics}. To summarize the comparison between \DataFly\ and other algorithms 
over many graphs, we report the \SPC metric~\cite{deepgd}, defined as the average relative difference between the metrics of \DataFly\ and 
the baseline, over the test graphs.
\emph{Smaller values are better}. 
For example, when comparing algorithm A with the baseline algorithm B, a \SPC$=-0.05$ indicates that A is on average 5\% better than B under that metric. Specifically, 
\begin{equation}
    \text{SPC} = \frac{1}{m}\sum_{i=1}^{m}\frac{A_i-B_i}{\max(|A_i|, |B_i|)},
\end{equation}
\noindent where $A_i$ and $B_i$ denote the metrics for the algorithm of interest and the baseline algorithm, 
for test graph $i$, and $m$ is the number of test graphs.

\subsection{Comparing \DataFly\ with Baseline Algorithms\label{sec:compare_baselines}}

Traditional graph layout algorithms, such as force-directed methods and stress majorization~\cite{Eades_1984, Fruchterman_Reingold_1991, neato}, optimize functions directly in low-dimensional spaces (e.g., 2D or 3D), where they are more likely to get trapped in local minima. In contrast, optimizing the same problem in higher dimensions offers more degrees of freedom, helping escape local minima that often become saddle points in higher dimensions, which still admit descent directions~\cite{goodfellow2016deep}~(Section 8.2.3). Given this, a natural question arises: can \DataFly\ achieve lower stress than \neato, lower spring-electrical energy than \sfdp, and improvements across other layout quality metrics?

\begin{table*}[t]
\centering
\scriptsize
\setlength{\tabcolsep}{3.5pt}
\renewcommand{\arraystretch}{0.75}
\caption{Comparison of metrics on SuiteSparse14 for two method pairs:
\neato\ (2D) vs.\ \OptProj\ and \sfdp\ (2D) vs.\ \OptProjSfdp. \emph{Lower is better for all metrics}.
The best value within each method pair and graph/metric combination is highlighted in bold.
The number of nodes/edges is shown in parentheses under the graph names.
SPC values marked with * are statistically significant (Appendix \cref{fig:CI_plots}).}
\label{tab:combined_suitesparse}
\begin{tabular}{llrrrrrr|lrrrrrr}
\toprule
& & \multicolumn{6}{c|}{\neato\ vs.\ \OptProj} & & \multicolumn{6}{c}{\sfdp\ vs.\ \OptProjSfdp} \\
\cmidrule(lr){3-8}\cmidrule(l){10-15}
Graph & mthd
& ang\_r$\downarrow$ & edglen$\downarrow$ & spr\_el$\downarrow$ & stress$\downarrow$ & t-SNE$\downarrow$ & xing$\downarrow$
& mthd
& ang\_r$\downarrow$ & edglen$\downarrow$ & spr\_el$\downarrow$ & stress$\downarrow$ & t-SNE$\downarrow$ & xing$\downarrow$ \\
\midrule

\multirow{2}{*}{\makecell[l]{1138\_bus\\\scriptsize(1138,1458)}}
& \neato    & 1.173 & \textbf{0.296} & -3.988 & \textbf{0.069} & 2.365 & 1406
& \sfdp         & \textbf{1.046} & \textbf{0.550} & \textbf{-4.369} & \textbf{0.092} & \textbf{1.415} & \textbf{511} \\
& \OptProj & \textbf{1.154} & 0.409 & \textbf{-4.212} & 0.076 & \textbf{2.085} & \textbf{1199}
& \OptProjSfdp & 1.105 & 0.569 & -4.322 & 0.098 & 1.584 & 756 \\
\graymidrule

\multirow{2}{*}{\makecell[l]{bcspwr07\\\scriptsize(1612,2106)}}
& \neato    & 1.160 & \textbf{0.291} & -4.465 & \textbf{0.046} & 2.149 & 1572
& \sfdp         & 1.033 & 0.672 & \textbf{-4.998} & 0.088 & \textbf{1.365} & \textbf{678} \\
& \OptProj & \textbf{1.096} & 0.402 & \textbf{-4.664} & 0.062 & \textbf{2.024} & \textbf{1570}
& \OptProjSfdp & \textbf{1.030} & \textbf{0.568} & -4.937 & \textbf{0.084} & 1.526 & 814 \\
\graymidrule

\multirow{2}{*}{\makecell[l]{email\\\scriptsize(1133,5451)}}
& \neato    & 1.191 & 0.515 & \textbf{-2.151} & \textbf{0.133} & 1.369 & 544390
& \sfdp         & 1.207 & 0.546 & \textbf{-2.306} & \textbf{0.142} & \textbf{1.308} & 425417 \\
& \OptProj & \textbf{0.996} & \textbf{0.477} & -2.114 & 0.163 & \textbf{1.336} & \textbf{517851}
& \OptProjSfdp & \textbf{1.097} & \textbf{0.529} & -2.269 & 0.150 & 1.309 & \textbf{399532} \\
\graymidrule

\multirow{2}{*}{\makecell[l]{Erdos991\\\scriptsize(446,1413)}}
& \neato    & 1.277 & 0.439 & \textbf{-2.170} & \textbf{0.109} & 1.451 & 35664
& \sfdp         & 1.260 & 0.502 & \textbf{-2.313} & 0.142 & 1.378 & 32632 \\
& \OptProj & \textbf{1.132} & \textbf{0.409} & -2.157 & 0.129 & \textbf{1.398} & \textbf{29064}
& \OptProjSfdp & \textbf{1.183} & \textbf{0.465} & -2.280 & \textbf{0.131} & \textbf{1.334} & \textbf{27855} \\
\graymidrule

\multirow{2}{*}{\makecell[l]{EX1\\\scriptsize(560,4368)}}
& \neato    & 0.390 & 0.339 & -1.458 & \textbf{0.167} & 1.153 & 487248
& \sfdp         & 0.390 & 0.205 & \textbf{-1.493} & \textbf{0.170} & 1.167 & 492441 \\
& \OptProj & \textbf{0.374} & \textbf{0.201} & \textbf{-1.477} & 0.176 & \textbf{1.123} & \textbf{340132}
& \OptProjSfdp & \textbf{0.367} & \textbf{0.183} & -1.486 & 0.175 & \textbf{1.089} & \textbf{364237} \\
\graymidrule

\multirow{2}{*}{\makecell[l]{football\\\scriptsize(115,613)}}
& \neato    & 0.541 & 0.424 & -1.167 & \textbf{0.128} & 0.539 & 6843
& \sfdp         & 0.539 & 0.519 & \textbf{-1.196} & 0.138 & 0.508 & 5622 \\
& \OptProj & \textbf{0.529} & \textbf{0.399} & \textbf{-1.170} & 0.132 & \textbf{0.518} & \textbf{5183}
& \OptProjSfdp & \textbf{0.526} & \textbf{0.438} & -1.181 & \textbf{0.135} & \textbf{0.507} & \textbf{5334} \\
\graymidrule

\multirow{2}{*}{\makecell[l]{hypercube10\\\scriptsize(1024,5120)}}
& \neato    & 0.595 & 0.341 & -1.935 & \textbf{0.198} & 2.302 & 478232
& \sfdp         & 0.486 & 0.081 & \textbf{-1.980} & \textbf{0.202} & 2.343 & 518024 \\
& \OptProj & \textbf{0.323} & \textbf{0.007} & \textbf{-1.972} & 0.206 & \textbf{1.987} & \textbf{247989}
& \OptProjSfdp & \textbf{0.356} & \textbf{0.040} & -1.970 & 0.206 & \textbf{1.991} & \textbf{264487} \\
\graymidrule

\multirow{2}{*}{\makecell[l]{Journals\\\scriptsize(124,5972)}}
& \neato    & 0.077 & \textbf{0.434} & \textbf{0.210} & \textbf{0.167} & 0.079 & 3747293
& \sfdp         & 0.077 & \textbf{0.446} & \textbf{0.210} & \textbf{0.170} & 0.078 & \textbf{3720645} \\
& \OptProj & \textbf{0.075} & 0.463 & 0.243 & 0.182 & \textbf{0.077} & \textbf{3156318}
& \OptProjSfdp & \textbf{0.074} & 0.462 & 0.237 & 0.182 & \textbf{0.077} & 3741769 \\
\graymidrule

\multirow{2}{*}{\makecell[l]{mobius\\\scriptsize(250,450)}}
& \neato    & 0.639 & 0.187 & -3.535 & \textbf{0.029} & 0.759 & 72
& \sfdp         & 0.819 & 0.524 & \textbf{-3.607} & 0.038 & 0.812 & 37 \\
& \OptProj & \textbf{0.464} & \textbf{0.089} & \textbf{-3.603} & 0.034 & \textbf{0.735} & \textbf{37}
& \OptProjSfdp & \textbf{0.650} & \textbf{0.193} & -3.592 & 0.038 & \textbf{0.730} & \textbf{14} \\
\graymidrule

\multirow{2}{*}{\makecell[l]{qh882\\\scriptsize(1764,3354)}}
& \neato    & 1.277 & 0.421 & -4.177 & \textbf{0.056} & 2.065 & 7590
& \sfdp         & \textbf{1.311} & 0.874 & \textbf{-4.484} & \textbf{0.083} & \textbf{1.399} & \textbf{4337} \\
& \OptProj & \textbf{1.237} & \textbf{0.323} & \textbf{-4.369} & 0.072 & \textbf{1.730} & \textbf{5361}
& \OptProjSfdp & 1.356 & \textbf{0.752} & -4.442 & 0.087 & 1.515 & 5283 \\
\graymidrule

\multirow{2}{*}{\makecell[l]{Si2\\\scriptsize(769,8516)}}
& \neato    & 0.317 & 0.536 & -1.421 & 0.159 & 0.907 & 1745046
& \sfdp         & 0.326 & 0.569 & \textbf{-1.568} & \textbf{0.141} & \textbf{0.825} & 1242149 \\
& \OptProj & \textbf{0.308} & \textbf{0.384} & \textbf{-1.555} & \textbf{0.143} & \textbf{0.828} & \textbf{1150696}
& \OptProjSfdp & \textbf{0.307} & \textbf{0.393} & -1.561 & 0.146 & 0.830 & \textbf{1229816} \\
\graymidrule

\multirow{2}{*}{\makecell[l]{spiral\\\scriptsize(665,7657)}}
& \neato    & 0.533 & 0.467 & -2.464 & \textbf{0.044} & 1.443 & 853028
& \sfdp         & 0.349 & 0.685 & \textbf{-2.857} & \textbf{0.076} & 1.424 & 725528 \\
& \OptProj & \textbf{0.283} & \textbf{0.440} & \textbf{-2.653} & 0.062 & \textbf{1.284} & \textbf{638663}
& \OptProjSfdp & \textbf{0.340} & \textbf{0.618} & -2.815 & 0.079 & \textbf{1.193} & \textbf{689046} \\
\graymidrule

\multirow{2}{*}{\makecell[l]{Trefethen\\\scriptsize(700,5977)}}
& \neato    & 0.353 & 0.616 & -1.481 & \textbf{0.177} & 1.209 & 667259
& \sfdp         & 0.352 & 0.519 & \textbf{-1.512} & 0.180 & 1.239 & 696237 \\
& \OptProj & \textbf{0.346} & \textbf{0.341} & \textbf{-1.498} & 0.181 & \textbf{1.103} & \textbf{457958}
& \OptProjSfdp & \textbf{0.348} & \textbf{0.348} & -1.503 & \textbf{0.179} & \textbf{1.132} & \textbf{519729} \\
\graymidrule

\multirow{2}{*}{\makecell[l]{USpower\\\scriptsize(4941,6594)}}
& \neato    & 1.246 & \textbf{0.372} & -4.829 & \textbf{0.058} & 3.100 & 12364
& \sfdp         & \textbf{1.066} & 0.795 & \textbf{-5.633} & \textbf{0.098} & \textbf{1.712} & \textbf{3614} \\
& \OptProj & \textbf{1.156} & 0.438 & \textbf{-5.177} & 0.073 & \textbf{2.782} & \textbf{12273}
& \OptProjSfdp & 1.129 & \textbf{0.773} & -5.579 & 0.101 & 2.044 & 5207 \\
\midrule

\multicolumn{2}{l}{\sPC$\downarrow$}
& \textbf{-0.128}$^*$ & \textbf{-0.169}$^*$ & \textbf{-0.019} & 0.122$^*$ & \textbf{-0.0744}$^*$ & \textbf{-0.233}$^*$
& 
& \textbf{-0.050}$^*$ & \textbf{-0.178}$^*$ & 0.017$^*$ & 0.015 & \textbf{-0.011} & \textbf{-0.068} \\
\bottomrule
\end{tabular}
\end{table*}

\noindent{\textbf{Comparison with \neato.}} 
Results are summarized in \cref{tab:combined_suitesparse}.
We first note that, as embedding dimensions increase, it is intuitively easier for the Euclidean distance in the layout to approximate the graph-theoretic distance, resulting in lower stress (\cref{eqn_stress}).
As shown in \cref{fig:vis-1-neato} and \cref{tab:stress_vs_dims} in the Appendix, 
the stress decreases as the ambient dimension increases. 
Stress in 10-D is 84.4\% lower on average than in 2D.

\begin{figure}[htbp]
    \centering
    \includegraphics[width=\linewidth]{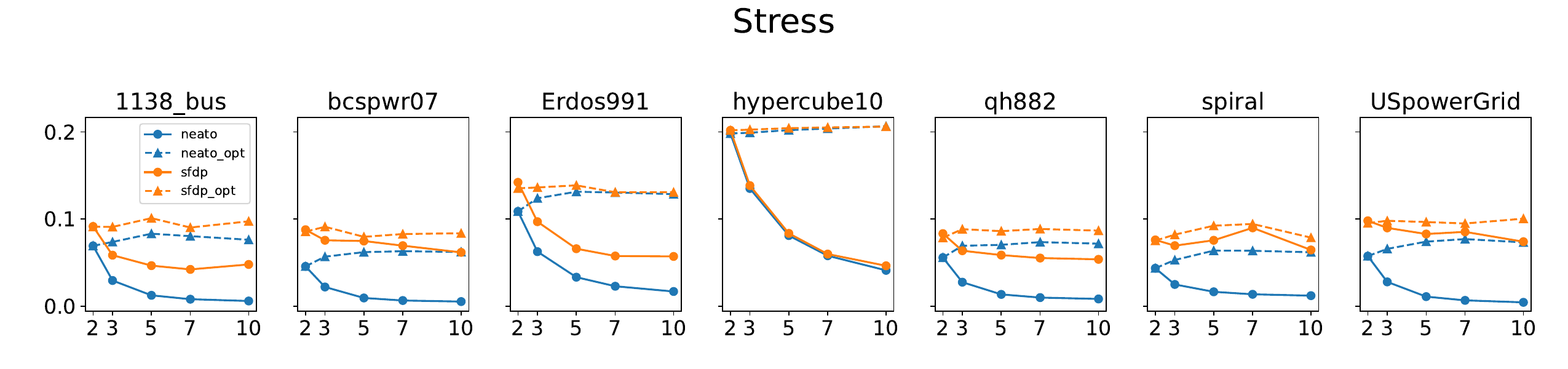}
    \includegraphics[width=\linewidth]{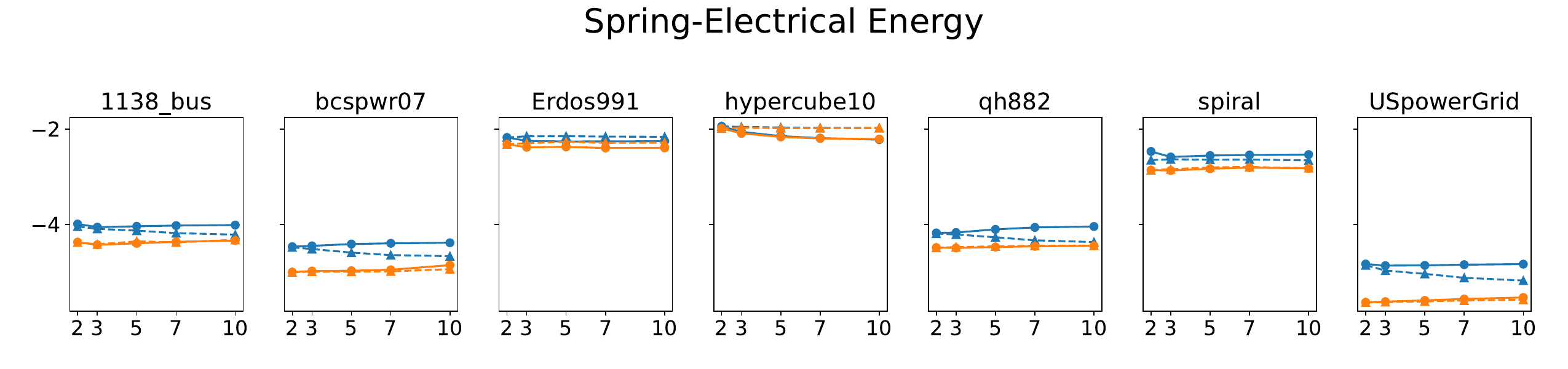}
    \includegraphics[width=\linewidth]{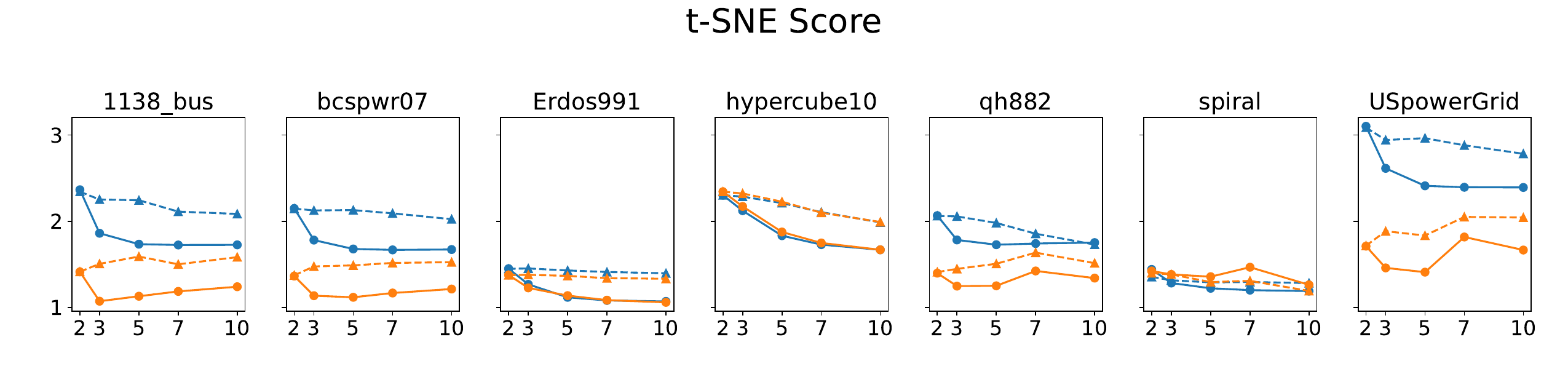}    
    \caption{
    How metrics change when embedded in higher dimensions: solid lines show values in the ambient dimension, dotted lines show values after optimal projection. Stress improves markedly with added dimensions, but this also makes low-stress projections harder to obtain. Spring-electrical energy gains little from extra dimensions. For the t-SNE score, higher dimensions often increase the metric, while projections from them typically outperform direct 2D optimization.
    }
    \label{fig:vis-1-neato}
\end{figure}

However, does this lower stress in high dimensions result in \DataFly\ achieving lower stress than a 2D optimized \neato? 
As shown in 
\cref{tab:combined_suitesparse}~(left),
this can be true in some cases, e.g., for ``Si2'', \OptProj\  achieved a stress of 0.143 vs 0.159 for \neato. 
Overall, the SPC aggregate metric shows that stress for \OptProj\ is 12.2\% higher (worse) than for \neato. This suggests that while \neato\ can occasionally get trapped in a local optimum in 2D, this is infrequent — \neato\ generally achieves lower stress in 2D.

When considering other metrics, the advantage of \DataFly becomes clearer. \OptProj\ achieves 12.8\% better angular resolution, 16.9\% lower edge-length variance, 7.44\% better t-SNE score, and 23.3\% fewer edge crossings. Its performance on spring-electrical energy is on par with direct \neato 2D layouts.

\Cref{fig:CI_plots} (top) in the Appendix shows paired-bootstrap ($n=5000$) 95\% confidence intervals (CIs) for the SPC metric across evaluation measures. A CI entirely below 0 indicates \OptProj\ significantly outperforms \neato. Under this view, \OptProj\ clearly outperforms in angular resolution, edge length variance, t-SNE score, and edge crossings. SPC values for these metrics in \cref{tab:combined_suitesparse} are marked with *, and the same convention is used for subsequent results.

\noindent{\textbf{Comparison with \sfdp.}} The spring-electrical energy, which \sfdp optimizes, decreases as the layout dimension increases, as shown in \cref{fig:vis-1-neato} and \cref{tab:sfdp_vs_dims} in the Appendix.
However, the reduction is not as drastic as we saw with stress and \neato (\cref{tab:stress_vs_dims}). Overall, spring-electrical energy is 10.6\% lower in 10-D than in 2D. In particular, when the graphs originate from 2D applications (e.g., 1138\_bus and USpowerGrid), higher dimensions do not help at all, while for other graphs (e.g., email, EX1), higher dimensions improve spring-electrical energy, though not significantly. This seems to indicate that an optimal layout for this loss function can be realized well
in a lower dimension. 
One way to assess whether a high-dimensional embedding can ``live'' comfortably in a lower-dimensional subspace is to examine the variance explained by PCA~\cite{hotelling1933_PCA} when projecting to that lower dimension. The results in the Appendix (\cref{tab:variance_explained}) support this: at the same target dimension, the PCA-explained variance for \sfdp\ is consistently higher than for \neato. For example, projecting 10D embeddings to 2D via PCA retains 75.4\% of the original variance for \sfdp, compared with 70.2\% for \neato.

The optimal projection \OptProjSfdp\ achieves similar (within 2\%) spring-electrical energy, stress, and t-SNE score as \sfdp; see \cref{tab:combined_suitesparse}~(right).
However, \OptProjSfdp\ performs better on angular resolution, edge length variance, and edge crossings.

\newcommand{\final}{}

\ifdefined\final
    \begin{figure}[htbp]
        \centering
        \includegraphics[width=0.5\textwidth]{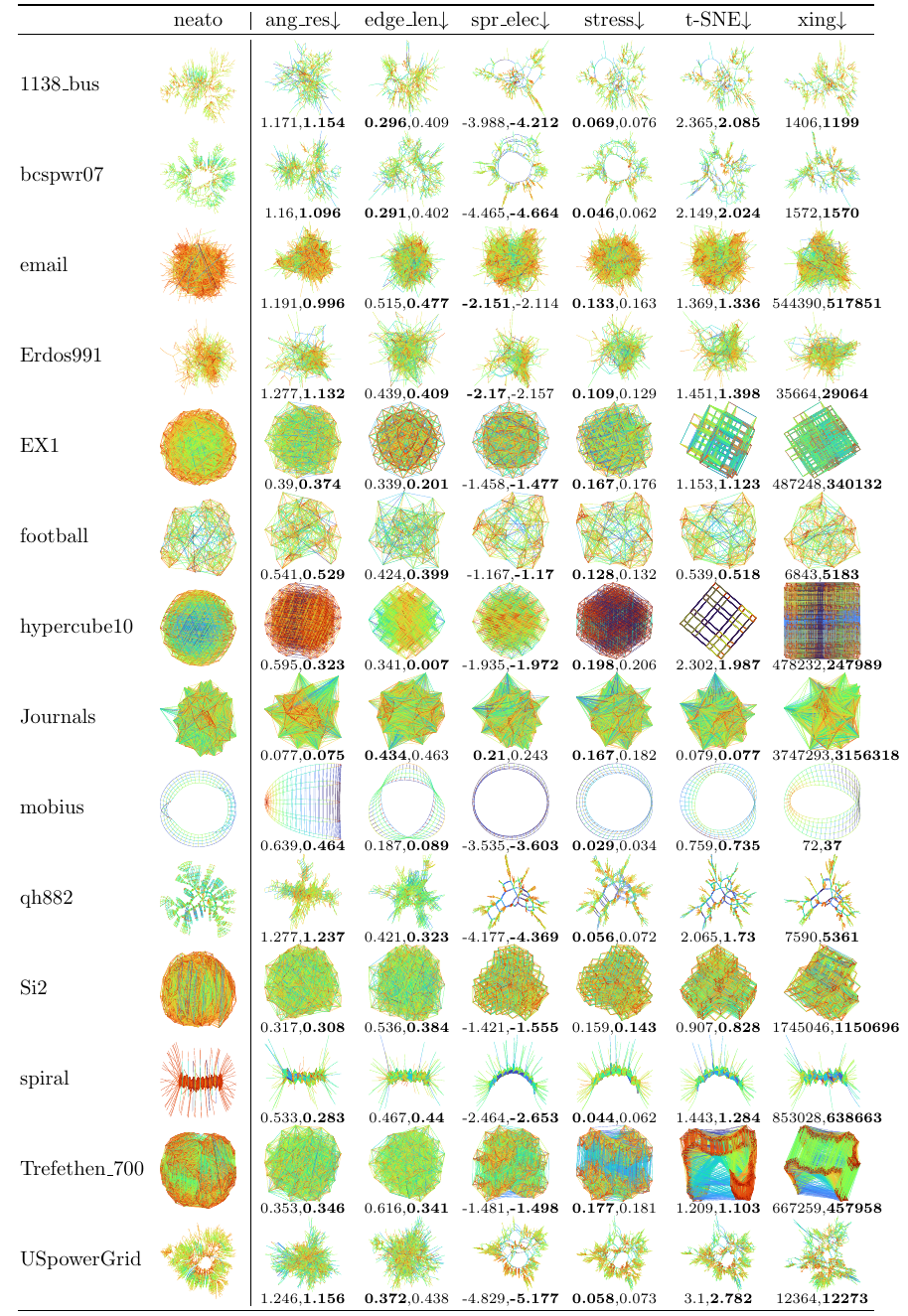}

        \caption{Optimal projections of the 10D \neato layout. Numbers represent (\neato, \OptProj) metrics. \textbf{Bold} = better.  
        Edges colored by length: {\color{red}red} = short, {\color{blue}blue} = long.\label{tab:fly_neato_dim10_vis}}
        
    \end{figure}
\else

\begin{figure}[htbp]
    \centering
    \includegraphics[width=0.5\textwidth]{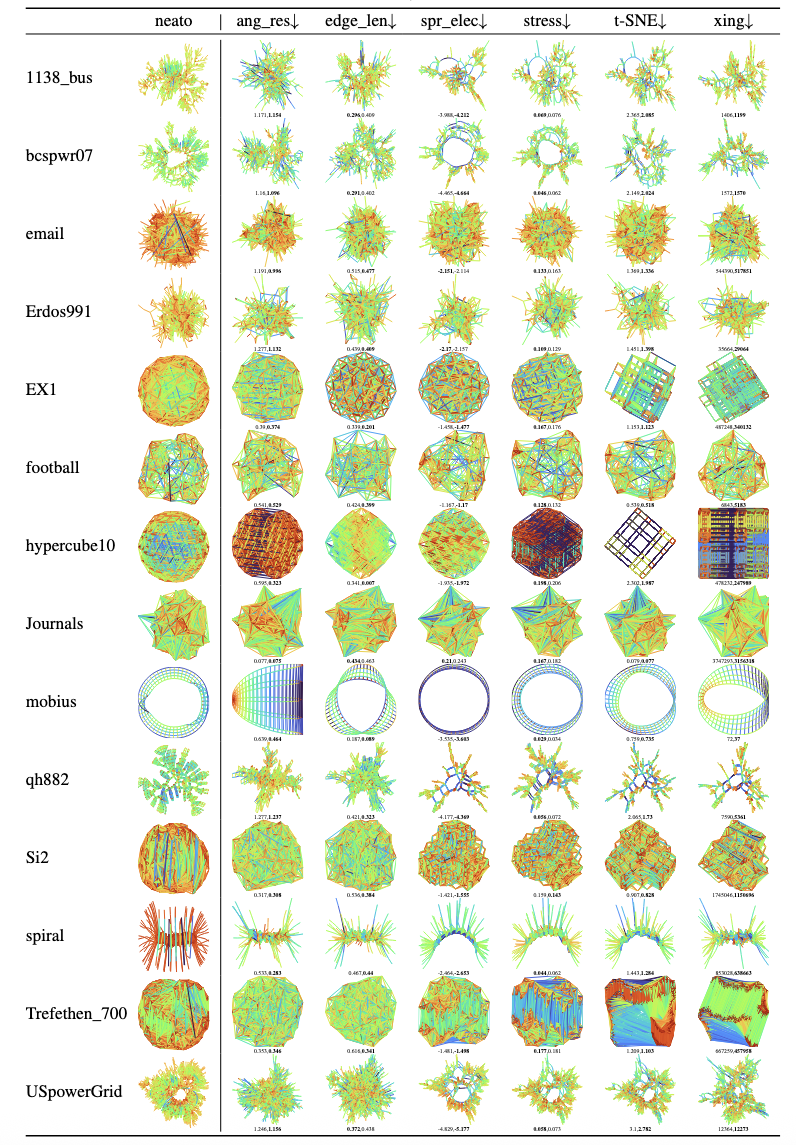}
    \caption{Optimal projections of the 10D \neato layout. Numbers show \neato and \OptProj metrics. \textbf{Bold} = better. Edges are colored by length. Red = short, blue = long.}
    \label{tab:fly_neato_dim10_vis}
\end{figure}

\fi

\subsection{Visualizations with \DataFly\label{sec:visual_comparison}}

It is informative to compare the optimal projection viewpoints identified by \DataFly\ with those from the corresponding baseline algorithms.
\cref{tab:fly_neato_dim10_vis} presents visualizations from \OptProj\ alongside those from \neato. \DataFly\ sometimes reveals compelling graph structures that differ markedly from those produced by \neato. For example, for the ``Si2'' and ``EX1'' graphs, the \DataFly\ visualization that optimizes edge crossings highlights a lattice structure that is less apparent in the \neato\ layout. Similarly, for ``mobius'', the \DataFly\ viewpoint that minimizes edge crossings is much smoother and has half as many edge crossings as \neato\ (37 vs. 72). In ``Trefethen\_700'', \DataFly\ produces a clearer, more organized layout that reveals linear and rectangular patterns, while the \neato\ layout appears more entangled, with increased edge crossings and visual clutter. Additional visual comparisons between \neato, \sfdp, and \DataFly\ are given in the Appendix (\cref{tab:fly_neato_dim10_vis_rome20_a}, \cref{tab:fly_neato_dim10_vis_rome20_b} and \cref{tab:sfdp_vis}), which further confirm the effectiveness of \DataFly\ in revealing graph structures that may be obscured in direct 2D layouts. 
These findings provide strong support for \textbf{RQ1}.

\subsection{Comparing \DataFly with Other Algorithms}

We observe that embedding graphs into high dimensions and then finding the optimal 2D projection can produce layouts with substantially better metric scores than those generated by baseline 2D algorithms (e.g., \neato or \sfdp). To understand how effective this approach is compared with directly optimizing the metrics, we compare \DataFly with \sgd~\cite{sgd2} and \smartgd~\cite{wang_2023_smartgd} in minimizing edge crossings, and with \sgd\ on angular resolution and edge length variance. We were unable to obtain comparable results for \sgd and \smartgd\ across other metrics from the authors of these algorithms.
\sgd\ is designed to optimize several graph aesthetic metrics directly using gradient descent on relatively small graphs. \smartgd\ was also trained on the Rome collection of small graphs to optimize specific metrics using a generative adversarial network~\cite{rgan}. 
Therefore, we first used the Rome20 dataset for comparison. 

The results are given in \cref{tab_rome_xing}, with CIs analyzed in the Appendix \cref{fig:additional-ci1}. For minimizing edge crossings, the last row of the table shows that \OptProj\  achieves 22.1\% fewer crossings on average than \sgd, while \smartgd\ is 4.4\% better than \sgd.  In terms of CPU time, \OptProj\ took only 10 seconds, \smartgd\ takes 1 second, while the \sgd\ algorithm took over 1280 seconds. This also indicates that the gradient for crossing minimization in \sgd, which involves a neural network model that approximates the crossings, requires significant computational effort.
The best performing algorithm, Vertex Movement (VM)~\cite{xing-heuristic}, takes 760 seconds and is 60.3\% better than \sgd. 
While VM achieves low crossing numbers, it produces drawings that do not optimize other aesthetic criteria; see the example drawing of grafo7023 in \cref{fig:VM_example}, versus the corresponding drawing produced by \DataFly\ in \cref{tab:fly_neato_dim10_vis_rome20_b} (row 5, last column).

In terms of angular resolution, \OptProj\  is 62\% better than \sgd\ and again much faster, requiring only a few seconds vs 218 seconds for \sgd. When looking at edge length variance, \sgd reduced it to almost zero and performed better than \DataFly. This is not a surprise, since even though edge length variance should be very low in high dimensions, linear projection is unlikely to preserve it due to the different orientations of edges in high dimensions. However, \sgd\ required 148 seconds, while \OptProj completed in 10 seconds.

\begin{table}
\centering
\caption{Comparison on Rome20 graphs. 
    We used \sgd as the benchmark. 
    *=statistically significant (Appendix \cref{fig:additional-ci1}).}
\label{tab_rome_xing}
\setlength{\tabcolsep}{0.5pt}
\begin{tabular}{l|rrrr|rr|rr}
  \toprule
    &  \multicolumn{4}{c|}{xing$\downarrow$} &  \multicolumn{2}{c|}{ang\_res$\downarrow$} &  \multicolumn{2}{c}{edge\_len$\downarrow$} \\ 
grafo id & \OptProj & \sgd & \smartgd & VM & \OptProj & \sgd & \OptProj & \sgd \\
\midrule
2815.35 & 2 & 8 & 2 & \textbf{1} & \textbf{0.700} & 2.158 & 0.097 & \textbf{0.012} \\
10379.95 & 82 & 80 & 82 & \textbf{40} & \textbf{0.773} & 1.881 & \textbf{0.282} & 0.388 \\
9033.80 & 54 & 54 & 49 & \textbf{30} & \textbf{0.778} & 1.681 & \textbf{0.279} & 0.306 \\
11674.33 & 3 & 5 & 6 & \textbf{1} & \textbf{0.615} & 1.997 & 0.122 & \textbf{0.000} \\
847.22 & 4 & 16 & 5 & \textbf{0} & \textbf{0.622} & 1.808 & 0.051 & \textbf{0.000} \\
762.27 & 3 & 5 & 3 & \textbf{1} & \textbf{0.568} & 1.980 & 0.157 & \textbf{0.000} \\
11412.32 & 1 & \textbf{0} & 1 & \textbf{0} & \textbf{0.696} & 2.171 & 0.116 & \textbf{0.000} \\
1583.66 & 12 & 18 & 15 & \textbf{9} & \textbf{0.704} & 2.085 & 0.245 & \textbf{0.206} \\
3587.36 & 7 & 8 & 7 & \textbf{3} & \textbf{0.868} & 1.848 & 0.173 & \textbf{0.013} \\
8296.86 & 66 & 50 & 79 & \textbf{42} & \textbf{0.771} & 1.798 & \textbf{0.240} & 0.302 \\
11543.34 & \textbf{0} & 7 & 4 & \textbf{0} & \textbf{0.753} & 2.114 & 0.174 & \textbf{0.021} \\
7023.45 & 8 & 16 & 12 & \textbf{2} & \textbf{0.718} & 2.034 & 0.194 & \textbf{0.095} \\
9592.72 & 13 & 24 & 18 & \textbf{3} & \textbf{0.692} & 2.102 & 0.204 & \textbf{0.174} \\
3138.60 & 31 & \textbf{18} & 42 & 20 & \textbf{0.889} & 2.161 & \textbf{0.226} & 0.294 \\
2955.38 & 6 & 10 & 16 & \textbf{5} & \textbf{0.754} & 1.968 & 0.217 & \textbf{0.039} \\
10480.95 & 59 & 81 & 63 & \textbf{30} & \textbf{0.835} & 1.934 & \textbf{0.296} & 0.307 \\
2747.16 & \textbf{0} & \textbf{0} & \textbf{0} & \textbf{0} & \textbf{0.635} & 2.038 & 0.046 & \textbf{0.000} \\
5780.48 & 25 & 30 & 33 & \textbf{13} & \textbf{0.886} & 1.703 & 0.203 & \textbf{0.102} \\
11311.40 & \textbf{4} & 12 & 6 & 5 & \textbf{0.696} & 2.097 & 0.216 & \textbf{0.143} \\
10451.95 & 96 & 64 & 116 & \textbf{50} & \textbf{0.868} & 1.805 & \textbf{0.278} & 0.337 \\
\midrule
SPC            & -0.221${}^*$       & 0           & -0.044       & \textbf{-0.603}${}^*$ & \textbf{-0.620}${}^*$ & 0 & \textbf{0.456}${}^*$ & 0 \\
\bottomrule
\end{tabular}
\end{table}

\subsection{Additional comparison on edge crossing minimization\label{sec:crossing_rome}}

As seen in \cref{tab:combined_suitesparse},
edge-crossings for both \OptProjSfdp\ and \sfdp\ 
are generally lower than for \neato\ and \OptProj.
We formally compare these four algorithms together with \smartgd\ and \sgd\ on crossing minimization of the SuiteSparse14 dataset in \cref{tab:sfdp_xing}. 
Indeed, on average, \sfdp\ has 25.9\% fewer crossings than \neato, while \OptProjSfdp\ has 33.9\% fewer crossings. 
This is surprising given that in the graph drawing community, minimizing stress (which \neato\ does) and minimizing crossings were known to be highly associated with user preference~\cite{Purchase_1997, Tim-user-study}. 

We observe that \smartgd\ and \sgd\ have substantially higher edge crossings (70\% and 54.5\% worse than \neato), while the two \DataFly\ algorithms achieve improvements of 23.3\% and 35\% over \neato.
This confirms that neither \smartgd\ nor \sgd\ works well for large graphs. 
Comparing the runtime on these large graphs, \OptProj and \OptProjSfdp take 10,470 and 9,800 seconds, respectively. 
\sgd\ is the slowest, taking 16,500 seconds. 
\OptProj\ and \OptProjSfdp\ are only moderately faster because, although their optimization involves only 20  variables, compared with $2|V|$ variables for \sgd, that advantage is offset by the high cost of computing the edge-crossings loss required for these algorithms. 
Inference with \smartgd\ is fast on these large graphs, taking only 315 seconds, reflecting its neural network–based, non-iterative nature; however, training the model demands substantial data and computational resources.

\begin{table}
\center
\caption{Comparing edge-crossings for various algorithms on SuiteSparse14. *=statistically significant (Appendix \cref{fig:additional-ci2})}
\label{tab:sfdp_xing}
\setlength{\tabcolsep}{1pt}
\begin{tabular}{lrrrrrr}
\toprule
 & \neato & \OptProj & \sfdp & \OptProjSfdp & \smartgd\ & \sgd \\
\midrule
1138\_bus      & 1406   & 1199         & \textbf{511} & 756     & 12061         & 15431 \\
bcspwr07       & 1572   & 1570         & \textbf{678} & 814     & 22176         & 47795 \\
email          & 544390 & 517851       & 425417       & \textbf{399532} & 2852817 & 1234370 \\
Erdos991       & 35664  & 29064        & 32632        & \textbf{27855}  & 138411  & 47165 \\
EX1            & 487248 & \textbf{340132} & 492441    & 364237  & 1862924       & 741591 \\
football       & 6843   & \textbf{5183} & 5622        & 5334    & 9353          & 5578 \\
hypercube10    & 478232 & \textbf{247989} & 518024    & 264487  & 2821132       & 1111129 \\
Journals       & 3747293 & 3156318     & 3720645      & 3741769 & 3607080       & \textbf{2772694} \\
mobius         & 72     & 37           & 37           & \textbf{14}     & 10704   & 544 \\
qh882          & 7590   & 5361         & \textbf{4337} & 5283    & 47892        & 203344 \\
Si2            & 1745046 & \textbf{1150696} & 1242149  & 1229816 & 4181606      & 3347111 \\
spiral         & 853028 & \textbf{638663} & 725528    & 689046  & 2338855       & 2006179 \\
Trefethen      & 667259 & \textbf{457958} & 696237    & 519729  & 1878613       & 1723377 \\
USpower        & 12364  & 12273        & \textbf{3614} & 5207    & 201292       & 1254507 \\
\midrule
SPC            & 0      & -0.233${}^*$       & -0.259${}^*$        & \textbf{-0.339}${}^*$  & 0.700${}^*$    & 0.545${}^*$  \\
\bottomrule
\end{tabular}
\end{table}

\begin{table}[]
\setlength{\tabcolsep}{3pt}
    \centering
    \caption{SPC for edge crossings relative to \neato\ over all Rome Graphs. *=statistically significant (Appendix \cref{fig:additional-ci2})}
    \label{tab:rome_full_xing}
\begin{tabular}{l|rrrrrr}
\toprule
& \neato & \OptProj & \sfdp & \OptProjSfdp & \smartgd & \sgd\\
\midrule
SPC & 0 & \textbf{-0.268}${}^*$  & -0.0338${}^*$  & -0.215${}^*$  & -0.0393${}^*$  & -0.0477${}^*$ \\
\bottomrule
\end{tabular}
\end{table}

We also ran the edge-crossing minimization algorithm \textit{Vertex Movement}~(VM)~\cite{xing-heuristic} on these large graphs, but it did not finish within 24 hours for 9 of the 14 instances, confirming its scalability limitations.

Finally, for a comprehensive evaluation of edge-crossing minimization on small graphs, we compared all the algorithms on the complete Rome graph collection in \cref{tab:rome_full_xing}. 
On this full set of 11,531 graphs, \OptProj\ and \OptProjSfdp\ have 26.8\% and 21.5\% lower edge crossings than \neato, respectively. Both substantially outperform \sgd\ and \smartgd, although the latter two do surpass \neato.
Regarding the runtime, \OptProj\ and \OptProjSfdp\ each take about 3300 seconds for all graphs. 
The inference of \smartgd\ is also very fast on these small graphs, taking just 615 seconds. 
The geometric heuristic VM is slower, taking 122,000 seconds; furthermore, \emph{it failed on 12\% of the graphs due to floating-point errors.} 
Therefore, we do not include VM in the table. 
\sgd is the slowest, requiring 558,000 seconds, almost 170 times longer than \OptProj and \OptProjSfdp.

\subsection{Findings for \textbf{RQ1}}

When comparing \DataFly\ with the two baseline algorithms \neato\ and \sfdp\ across eight different evaluation metrics, \DataFly\ outperformed them on six metrics. When comparing it with the two state-of-the-art algorithms \smartgd\ and \sgd, which are designed to optimize specific metrics, we found that \DataFly\ is highly effective at minimizing edge crossings and improving angular resolution. While VM achieves the absolute lowest number of edge crossings, it is designed solely for that objective and has very high computational complexity. VM does not produce an aesthetic layout and is not suitable for large graphs.
These observations support an affirmative answer to \textbf{RQ1}.

\section{\DataFly Exploration Tool}
\label{sec:system}

The \DataFly\ application is a web-based platform for interactively exploring high-dimensional graph embeddings through different viewpoints.
Having established that these viewpoints are quantitatively strong (\cref{sec:results}) and qualitatively informative (\cref{tab:fly_neato_dim10_vis}), we now describe the system that exposes them to users.
Through interactive 2D visualizations, \DataFly\ reveals patterns and structures that may be hidden in a static layout by allowing users to navigate among related projections. In addition to optimal projections, \DataFly\ supports PCA projections, which are particularly well-suited for interactive exploration.
We describe the PCA projections next, then the system's main functionality, and finally a usability study.

\subsection{PCA viewpoints and interactive rotation\label{sec:PCA}}

Given the \HighD-dimensional node-embeddings matrix $X \in \mathbb{R}^{|V|\times K}$, we compute its principal components through Principal Component Analysis (PCA)~\cite{hotelling1933_PCA}. 
PCA yields \HighD\ orthogonal directions in the embedding space, ranked by decreasing variance of the projected points, that is, the spread of nodes in the projection.
The first principal component captures the most variance, the second the next most, and so on.
Any pair of principal components can be used as the axes of a 2D projection, yielding up to
$\binom{K}{2} = K(K-1)/2$ possible 2D viewpoints.
In the tool, these viewpoints provide a structured set of interpretable linear projections that users can browse interactively.

To support informed exploration, we compute quality metrics such as stress and edge crossings for each candidate viewpoint and display them as a line chart at the bottom of the UI (\cref{fig:system_showcase}).
The user can then navigate to the view that optimizes a given metric by clicking either the corresponding point on the line chart or its thumbnail.
Since computing these metrics for every possible viewpoint is expensive, we first rank all principal-component pairs by their explained variance and retain only the top 50.
Beyond the practical limitation that the interface can display only a limited number of viewpoints along the $x$-axis, this filtering is motivated by the observation that low-variance projections tend to cluster nodes tightly, leading to visual occlusion.
Empirically, the most insightful views typically correspond to component pairs that capture relatively high variance (i.e., greater spatial spread).

\subsection{User Interface and Functionality} 

\begin{figure*}[htbp]
    \vspace{-0.2cm}
    \centering
    \includegraphics[width=0.97\textwidth]{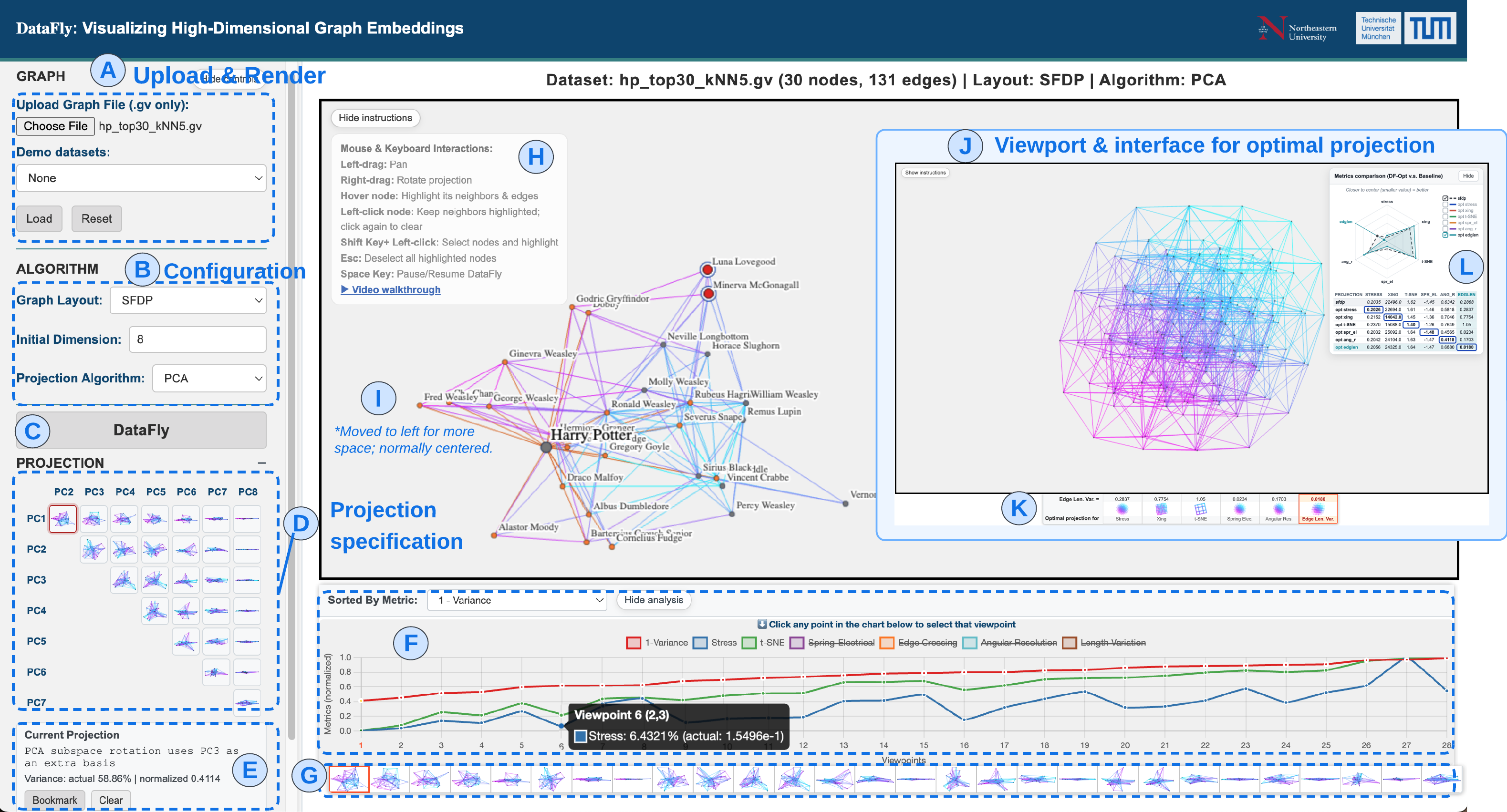}
    \caption{\DataFly\ interface and functionality overview. Main: Rotated PCA view of Harry Potter co-occurrence graph. Inset: Optimal Projection view of hypercube 8D.}
    \label{fig:system_showcase}
    \vspace{-0.1cm}
\end{figure*}

\Cref{fig:system_showcase} shows the \DataFly\ visualization and interactive exploration system, with components labeled in the figure (e.g., \circmark{A}).  
Users begin by uploading a 
file or selecting a built-in demo dataset \circmark{A}. The system processes the data and renders the visualization \circmark{I}.  
Users can configure the layout algorithm (\neato, \sfdp, \pivotmds, or \spectral), the initial embedding dimension (2–10), and the projection method (PCA or optimal projection) via \circmark{B}.  

As the user interacts with the system, edges are colored by BFS by default to help preserve the mental map; neighboring nodes and edges can be highlighted and remain so until the user deselects them. Mouse and keyboard based zooming, panning and other interactions are supported with instructions in an
information panel \circmark{H}.

\mysubsubsection{Interaction modes.} Three modes are provided.
The first is the \emph{optimal-projection} mode: when the ``Projection Algorithm'' in \circmark{B} is set to ``optimal projection,'' \DataFly\ presents the layout that optimizes stress as the main display~\circmark{J}  (depicted inset in the figure for demonstration), along with six thumbnails \circmark{K} corresponding to optimal projections for stress, edge crossings, t-SNE, spring-electrical energy, angular resolution, and edge-length variance, respectively.
The metric values achieved by all of the six layouts are displayed through a table in component \circmark{L} in the optimal projection interface~\circmark{J}. An interactive radar chart that provides comparison of layout quality across all metric values is also in \circmark{L}, letting users visually compare each layout against the baseline (2D \neato, \sfdp, \pivotmds, or \spectral).
More details are shown in \cref{fig:metrics_comparison_component}, and further discussions of the interactive features are  in \cref{sec:appendix_improvements} .

The PCA mode displays thumbnails \circmark{G} as previews of 2D projections
from PCA component pairs 
sorted by the selected metric (default: highest to lowest variance). The toggle list in \circmark{F} supports re-sorting by a different metric. Users can browse thumbnails to spot patterns and click to transition to the selected viewpoint.
Likewise, users can use the ``Projection'' matrix with thumbnails in \circmark{D} to 
manually select PCA viewpoints.
From the panel \circmark{E}, users can view projection details, bookmark rotated projections, and quickly switch to saved views.

Thirdly, the system also provides an automatic cinematic mode, allowing users to initiate transition animations through all viewpoints by clicking the ``DataFly'' button \circmark{C}, and to pause or resume at any time by pressing the space key.

\mysubsubsection{Interactive rotation.} Users can drag the view to perform a local rotation within a 3D exploration subspace. 
For PCA projections, this subspace is spanned by the two principal components of the current viewpoint and the remaining component with the highest variance; in optimal projection mode, it is formed by the optimized projection plane and an additional orthogonal direction. In both cases, the resulting viewpoint remains a linear projection, preserving interpretability while letting users perturb a view, extract structure from motion, mitigate node occlusion, and inspect nearby projections without losing continuity.

Together, these functionalities and interface elements are intended to support three complementary tasks: overview, selection, and refinement. Thumbnails in \circmark{D}, \circmark{G} and \circmark{K} provide an overview of all viewpoints at once, and users can switch easily to a projection by clicking on a thumbnail; the line chart section \circmark{F} supports quantitative comparison by quality metrics and selection of views that optimize a chosen criterion; and direct interactions (e.g., zooming, panning, interactive rotation) support local refinement near a promising view. 
Animated transitions and a consistent edge-coloring scheme, together with node highlighting, help preserve correspondence between successive projections, so users can compare related views as transformations of the same embedding rather than as unrelated layouts. 

\mysubsubsection{Scalability.} 
\DataFly\ leverages GPU-accelerated rendering and some scalable layout algorithms. 
As shown in Table~\ref{tab:system-scalability-runtime}, it scales well to moderate-sized graphs (thousands of nodes). For large graphs, it remains usable, but the bottleneck shifts to high-dimensional layout computation. More discussion about scalability can be found in  \cref{sec:scalability}. 

\begin{table}[t]
\centering
\caption{Runtime scalability of the DataFly backend. Backend total includes graph loading, layout computation, coordinate extraction, and response preparation; layout only measures the high-dimensional layout computation.}
\label{tab:system-scalability-runtime}
\scriptsize
\setlength{\tabcolsep}{4pt}
\begin{tabular}{rrrrrr}
\toprule
\multirow{2}{*}{$|V|$} & \multirow{2}{*}{$|E|$} & \multicolumn{2}{c}{PivotMDS (s)} & \multicolumn{2}{c}{\textsc{sfdp} (s)} \\
\cmidrule(lr){3-4}\cmidrule(lr){5-6}
 & & Backend & Layout & Backend & Layout \\
\midrule
492 & 1,417 & 0.12 & 0.06 & 0.64 & 0.54 \\
1,133 & 5,451 & 0.59 & 0.16 & 2.37 & 1.94 \\
4,941 & 6,594 & 0.96 & 0.41 & 4.57 & 3.97 \\
9,941 & 35,780 & 3.19 & 1.49 & 73.7 & 57.7 \\
49,989 & 224,509 & 19.3 & 8.97 & 196.6 & 183.9 \\
169,422 & 554,926 & 55.3 & 29.1 & 876.9 & 821.4 \\
\bottomrule
\end{tabular}
\end{table}

\section{Usability Study and Results}
\label{sec:case-study-results} 

To validate the practical utility of \DataFly{} in helping users explore high-dimensional graph layouts (\textbf{RQ2}), we conducted a qualitative usability study with four participants via remote video conferencing using a screen-sharing and think-aloud protocol. Each 40--50 minute session comprised four phases: a pre-interview on the participant's expertise, a tutorial video, hands-on exploration of a few datasets, and a closing Likert-scale questionnaire. 
Participants used the system to perform unguided exploration, identify nodes of interest, trace neighbors, switch projections, and compare viewpoints across metrics while verbalizing their observations. 
The study was conducted in accordance with the Technical University of Munich ethics guidelines (TUM Ethikkommission) for non-medical research.
\mysubsubsection{Participants.} 
We recruited four participants: two visualization experts (E1, E2), PhD students who regularly work with graph data using tools (NetworkX, Gephi). The other participants were non-experts (N1, N2) with computer science backgrounds but no prior experience with graph embeddings or dimensionality reduction. This setup enabled us to validate \DataFly's accessibility to a broader audience. Datasets matched participant backgrounds: 
non-experts explored the Last.fm and Game of Thrones networks, experts additionally explored 
subsets of a VIS co-authorship, and a hypercube graph.
The participants had no prior exposure to \DataFly. The 
hypercube graph is shown in \cref{fig:system_showcase}, and 
the remaining graphs can be found in the Appendix (\cref{sec:appendix_improvements}).
Informed consent was obtained from all participants prior to participation.

\mysubsubsection{Expert feedback.}
Both experts found the tool visually appealing and intuitive. 
E1 noted that \emph{``in one layout a node looked like a neighbor, but in the other layout didn't.''} They particularly valued the interactive rotation for resolving occlusion, finding that it revealed 
connections between clusters.
E1 also observed
that PCA viewpoints supported open-ended exploration while optimal projections could serve presentation: \emph{``I like the duality of being able to explore, but also look at a layout that maybe is nice to present in a communication setting.''} On the hypercube dataset, they noted that higher-dimensional embeddings produced markedly different projections, whereas the traditional 2D layout made all views appear noisy and similar --- \emph{``in 2D, they all kind of look the same. It looks noisy.''} 

E2 began by comparing layout algorithms and exploring the animated transitions. They found the transitions helpful for building intuition about the high-dimensional structure, though initially confused by the apparent rotation direction of the animations.
They praised the metric chart for validating projection quality, finding it helpful to \emph{``compare the values for the others as well.''} 
On the music dataset, they remarked that thumbnails were effective for previewing and skipping uninformative projections.

Each expert proposed improvements: E1 suggested ego-network highlighting and edge tracking through transitions (both now implemented), while E2 recommended supporting additional graph formats.

\mysubsubsection{Non-expert feedback.} N1 started with the Last.fm network, 
noting its small-world structure. They went on to identify coherent genre clusters, pointing out a tight grouping of nu-metal and alternative metal bands, and used the rotation feature to position the cluster as visually salient. 

N2 began with the Game of Thrones network, and made immediate comparisons with other 3D viewpoint editors, 
stating that controls felt both natural and intuitive. Using the projection thumbnails to see different viewpoints, they noted the separation between the Westeros and Essos storylines was visually salient across nearly all projections. 
They also used the bookmark feature to save interesting viewpoints. 

Both non-expert participants found similar insights in the Game of Thrones graph: both expressed surprise at the large number of connections of Robert Baratheon, and the 
small number of connections for many Stark family characters. Each used rotation and alternate projections to ``convince'' themselves of these observations, with N2 noting they would not have believed this without the visualization.

\mysubsubsection{Likert ratings.}
The post-session ratings (1--5) are evaluated across five aspects: \textit{Intuitive} (how intuitive the tool is to use), \textit{Would Use} (how likely would you use the tool for work or personal interest), \textit{Appeal} (overall visual appeal of the tool), \textit{Suitable} (how suitable the tool is for answering questions about the data), and \textit{Interaction} (overall effectiveness of the tool's user interactions and instructions). Exact values are found in the Appendix (\cref{sec:appendix_improvements}).

The tool received consistently high marks for visual appeal (median 4.5) and user interaction quality (median 4.5). Intuitiveness ratings were high among experts and N1 (all rated 4) but lower for N2 (rated 2), who struggled with interpreting the axes and viewpoints. Suitability scores were consistently high (median 4). Likelihood of use for work or personal interest varied (range 2--4).

\mysubsubsection{Results and findings summary.} 
The usability study validates that \DataFly{} helps users relate high-dimensional embeddings to recognizable structural patterns. Interactive rotation was repeatedly cited as essential for resolving occlusion and revealing otherwise hidden connections. Participants also valued the duality between exploratory and communicative uses: while PCA viewpoints supported open-ended exploration, the optimal projection mode was favored for producing aesthetic, presentation-ready layouts. Quantitatively, participants gave consistently high ratings for visual appeal, interaction quality, and task suitability. Based on participant feedback, we have iteratively improved the tool and outlined further updates in the Appendix (\cref{sec:appendix_improvements}).

\mysubsubsection{Discussion on RQ2.}
Taken together, the qualitative examples presented earlier (e.g., \cref{tab:fly_neato_dim10_vis}), which demonstrate \DataFly{}’s ability to reveal graph structures not visible in direct 2D layouts, along with the findings of this usability study, provide qualitative evidence supporting an affirmative answer to \textbf{RQ2}.

\section{Discussion and Limitations}
\label{sec:limit}

While \DataFly currently supports viewpoints via linear projection, either optimized for one of the six metrics or allowing users to explore through the sequence of PCA projections, one
might also consider non-linear projection techniques such as Multidimensional Scaling~\cite{Kruskal_MDS_1964}, t-SNE~\cite{Maaten_2008_tSNE}, or UMAP~\cite{mcinnes2018_umap}.
These methods can also produce a 2D representation of a high-dimensional graph layout, 
but they are less suitable for the interactive exploration scenario considered here.
First, they produce only a single projection for a fixed parameter setting, so exploring alternative views requires rerunning the method, potentially yielding layouts that differ significantly. 
Second, due to non-linearity, there is no meaningful relationship between two different embeddings. 
In this sense, non-linear projections are an abstract representation of high-dimensional data. 
\DataFly allows users to explore, in a principled way, several viewpoints of a high-dimensional graph embedding using interpretable linear projection techniques.

Several improvements can be made. Firstly, while DataFly's scalability benefits from optimizing only the $2K$ parameters of a projection matrix rather than all $2|V|$ node positions, several quality metrics remain computationally expensive as the graph grows. One clear avenue for improvement is incorporating the plane sweep algorithm~\cite{bentley1979_edge_crossing_algorithms} to speed up edge crossing computation.
Secondly, due to Graphviz limitations, our experiments were restricted to layouts up to 10 dimensions, except for some preliminary results in \cref{sec_appendix_30D}; future work will investigate systematically whether higher-dimensional layouts provide additional benefits. Finally, the projection matrix is initialized using the first two principal components of the high-dimensional embedding, ensuring repeatability. A natural extension would be to explore multiple initializations to improve the likelihood of reaching a global optimum.

\section{Conclusions}

\label{sec:conc}

This paper proposes an approach to graph visualization that embeds graphs in high-dimensional space and systematically searches for informative 2D viewpoints, moving beyond the conventional paradigm of producing a single static 2D layout. 

Our experimental results show that optimal projections derived from high-dimensional embeddings can produce superior visualizations, with metrics (e.g., edge crossings) outperforming 2D layout algorithms such as \sgd\ and \smartgd\, explicitly designed to minimize them. \DataFly also reveals structural patterns not apparent in widely used 2D layouts such as \neato\ and \sfdp. We further introduce SigmoidX, an effective differentiable loss function for edge crossings that is simpler than the neural network–based surrogate used in \sgd, while achieving significantly fewer crossings than both \sgd\ and \smartgd.

To make these high-dimensional embeddings accessible to users, we propose an interactive tool \DataFly for exploring high-dimensional graph embeddings and discovering insights. The system enables users to find projections that optimize specific metrics or explore a set of PCA-based viewpoints, accompanied by a chart displaying quality metrics such as stress, t-SNE score, and edge crossings. Users can interactively explore the high-dimensional layout landscape by ``flying'' from one viewpoint to another. Our usability study with experts and non-experts provides encouraging evidence that \DataFly{} helps users discover structural patterns that remain hidden in conventional 2D layouts.

\clearpage
\section*{Acknowledgments}
The authors thank Pan Xu for her input during the initial exploration of this project.

\nocite{
  cook1995grand,
  bertini2006visual,
  gove2019gragnostics,
  behrisch2017magnostics,
  henry2007nodetrix
}
\bibliography{ref}

\clearpage

\section*{Appendix}

\appendix

\renewcommand{\thefigure}{A\arabic{figure}}
\setcounter{figure}{0}

\renewcommand{\thesection}{A\arabic{section}}
\setcounter{section}{0}

\renewcommand{\thetable}{A\arabic{table}}
\setcounter{table}{0}

\renewcommand{\theequation}{A\arabic{equation}}
\setcounter{equation}{0}

\section{Differentiable surrogate function for edge crossing}
\label{sec:sigmoidx}

We define a differentiable approximation for detecting whether two edges intersect as follows.

Let the two edges be
\[
e_1: p + t r, \quad e_2: q + u s, \quad t, u \in [0,1],
\]
where \(p, q \in \mathbb{R}^2\) are the start points, and \(r, s\) are the direction vectors. The endpoints of these two edges are $p+r$ and $q+s$.

The intersection parameters are computed as
\[
t = \frac{(q - p) \times s}{r \times s}, \qquad
u = \frac{(q - p) \times r}{r \times s},
\]
where ``\(\times\)'' denotes the 2D cross product. Geometrically, when $t\in [0,1]$, it means that $e_2$, or its extension, hits $e_1$. 
Similarly, $u\in [0,1]$ means that $e_1$, or its extension, hits $e_2$. 
Hence, if the two edges cross, we must have $t, u\in [0,1]$.

Instead of using a hard condition \((t,u \in [0,1])\) to decide whether the edges intersect, which is not differentiable, we use sigmoid-based soft masks:
\[
M_t = \frac{\sigma_T(t)\,[1 - \sigma_T(t - 1)]}{M_{\text{peak}}}, \qquad
M_u = \frac{\sigma_T(u)\,[1 - \sigma_T(u - 1)]}{M_{\text{peak}}},
\]
where \(\sigma_T(u) = 1/(1+e^{-T u})\) is the sigmoid function with temperature $T$ (we use $T=10$), and $M_{\text{peak}} = \sigma_T(0.5)\,[1 - \sigma_T(-0.5)]$ normalizes the peak value.

The differentiable edge-crossing probability is
\[
P(\text{crossing}) = M_t M_u,
\]
which approaches 1 when the edges intersect (\(t,u \in [0,1]\)) and smoothly decreases toward 0 otherwise. We call this loss function \textit{SigmoidX}.
It is a continuous, differentiable proxy for the binary ``edges intersect’’ test, enabling gradient-based optimization to minimize edge crossings. 
We also experimented with other surrogate functions, including a pyramid function over the unit square, but found that \textit{SigmoidX} performs better.

In practice, we add a small constant $\epsilon=10^{-12}$ to the cross-product denominator when computing the intersection parameters. This prevents numerical instability when two projected edges are nearly parallel. We also exclude all edge pairs that share a vertex; thus, incident edges are not counted as crossings and do not contribute to the \textit{SigmoidX} loss.

\section{t-SNE metric formula}
\label{sec_appendix_tsne}

\begin{equation}    
\text{tsne} = \sum_{\substack{i,j\\i \neq j}} p_{ij}\log\frac{p_{ij}}{q_{ij}},
\label{equ:tsne-loss}
\end{equation}
where
\begin{equation*}    
    p_{ij} = \frac{p_{j|i}+p_{i|j}}{2N}, 
    p_{j|i} = \frac{\exp(-\frac{d_{ij}^2}{2\sigma_i^2})}{\sum_{\substack{k\\k\neq i}}\exp(-\frac{d_{ik}^2}{2\sigma_i^2})},
    q_{ij} = \frac{(1+||\mathbf{x}_i-\mathbf{x}_j||^2)^{-1} }{\sum_{\substack{k,l\\k \neq l}}(1+||\mathbf{x}_k-\mathbf{x}_l||^2)^{-1}}.
\end{equation*}

\section{Scale-invariant spring-electrical energy}
\label{sec_appendix_spr_elec_proof}

This metric treats edges as springs and nodes as electrically repelling particles and measures the energy of the system:
\begin{equation*}
\text{se\_eng}(x) = \sum_{(i,j) \in E} \frac{\| \mathbf{x}_i - \mathbf{x}_j \|^3}{3K_{s}}  
- \sum_{i \ne j} K_{r}^2\ln(\|\mathbf{x}_i - \mathbf{x}_j\|).
\end{equation*}
where $K_{s}$ and $K_r$ are spring and repulsion constants. We set them to 1. To make this metric scale-invariant, we can find a scaling factor $t$ such that $\text{se\_eng}(t x)$ is minimized. If we denote the first term in the energy as $A/3$ (for attractive spring energy) and the second as $R$ (repulsive electrical energy), then we want to minimize

\begin{equation*}
\text{se\_eng}(t x) = \frac{A}{3} t^3 - R - \ln(t)*M
\end{equation*}

\noindent where $M = |V|(|V|-1)$. Taking the derivative and setting it to zero, the optimum is achieved when $t = (\frac{M}{A})^{1/3}$. The final energy is 
\begin{align*} 
\text{se\_eng}(x)
&=\frac{M}{3} + \frac{M}{3}\ln\frac{A}{M} - R \\
&= \frac{|V|(|V|-1)}{3} + \frac{|V|(|V|-1)}{3}\ln\frac{\sum_{(i,j) \in E} \|\mathbf{x}_i-\mathbf{x}_j\|^3}{|V|(|V|-1)} \\
&\ \ \ \ - \sum_{i \ne j} \ln(\|\mathbf{x}_i - \mathbf{x}_j\|)
\end{align*}

\section{Additional results for \neato}
\label{sec_appendix_additional_res_neato}

Tables~\ref{tab:fly_neato_dim10_vis_rome20_a}-\ref{tab:fly_neato_dim10_vis_rome20_b} show visualizations of graphs in the Rome20 dataset, comparing \neato\ with \DataFly's optimal projection \OptProj\ across various metrics. 
The corresponding metric values for \neato and \OptProj\ are shown below each visualization. 
Because \OptProj\ is a projection of a high-dimensional \neato\ layout, its drawings remain visually similar to those of \neato, yet apart from stress and spring-electrical loss, \OptProj\ typically achieves better metric values. For example, for grafo11543, \OptProj\ finds a planar layout, whereas \neato\ has four edge crossings.

Table~\ref{tab:stress_vs_dims} presents the stress achieved by \neato\  when optimizing directly in various dimensions from 2D to 10D. As expected, stress is significantly lower in 10D for all graphs.

\begin{table*}[htbp]
\setlength{\tabcolsep}{4pt}
\centering
\renewcommand{\arraystretch}{3}
\caption{Optimal projections of the 10D Neato layout of the Rome20 dataset (part 1/2). Numbers show Neato vs. \OptProj metrics. \textbf{Bold} = better. Edges colored by length: {\color{red}red} = short, {\color{blue}blue} = long.}
\label{tab:fly_neato_dim10_vis_rome20_a}
\begin{tabular}{lc|cccccc}
\toprule
& neato & ang\_res$\downarrow$  & edge\_len$\downarrow$  & spr\_elec$\downarrow$  & stress$\downarrow$  & tsne$\downarrow$  & xing$\downarrow$  \\
\midrule
grafo10379.95 & \parbox{1.7cm}{\centering \includegraphics[height=1.5cm]{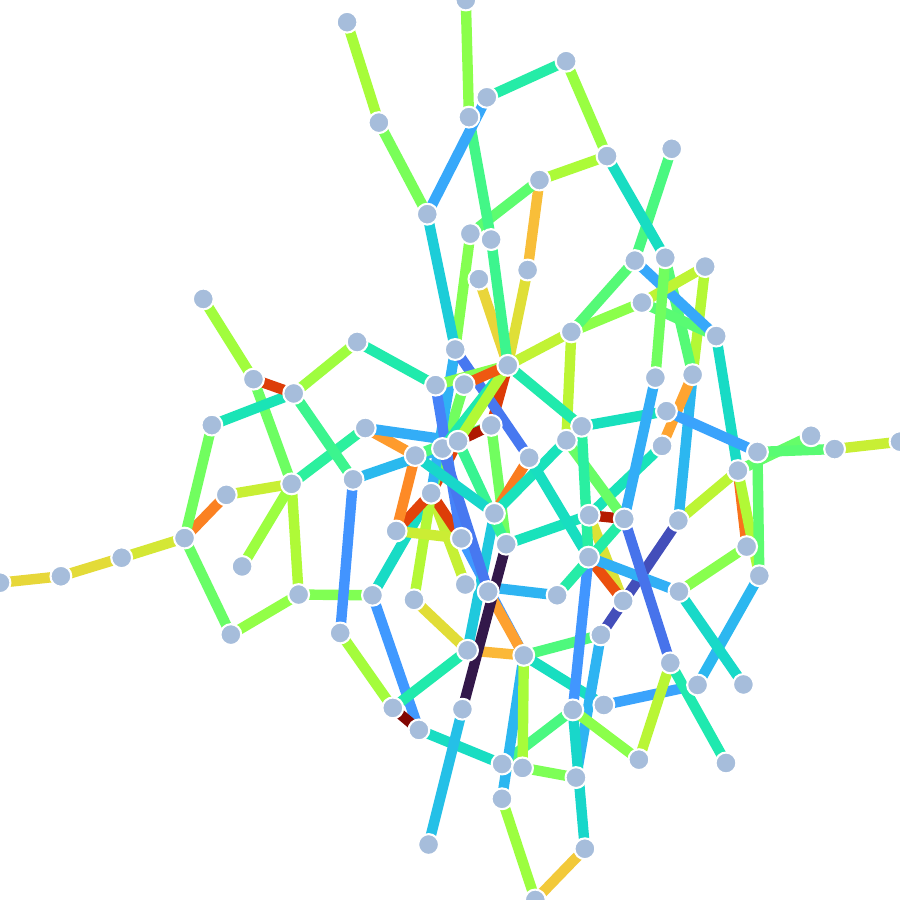} \\ \vspace{-0.0cm} \fontsize{7pt}{0pt}\selectfont } & \parbox{1.7cm}{\centering \includegraphics[height=1.5cm]{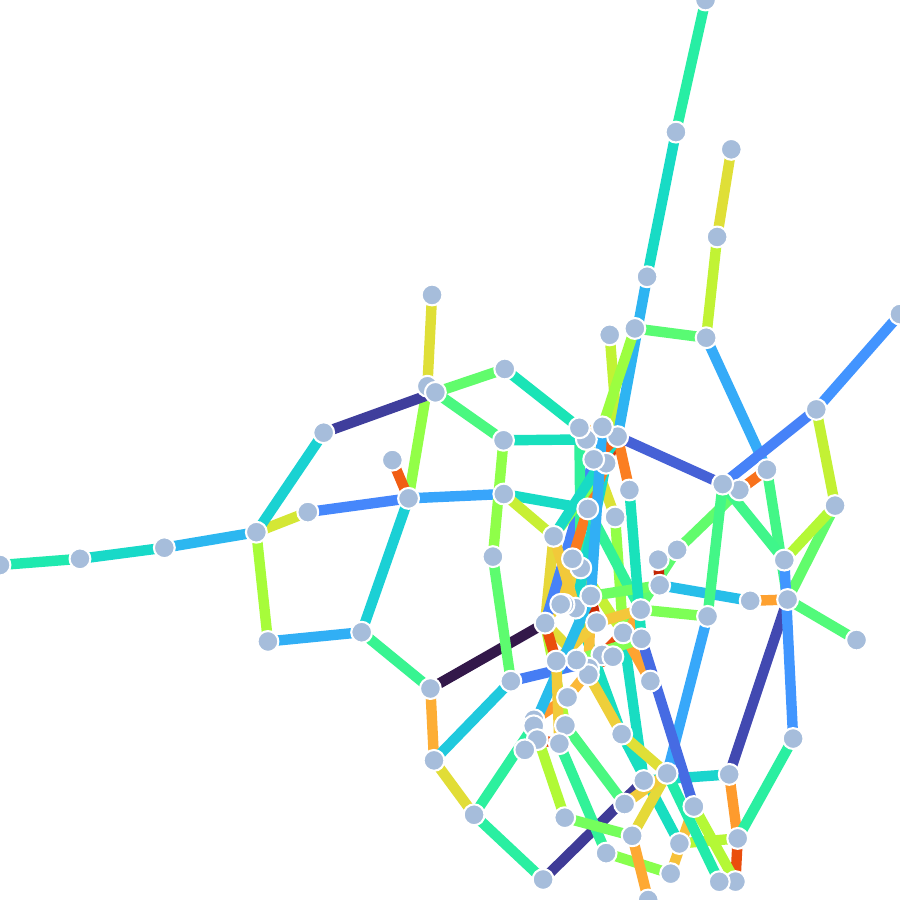} \\ \vspace{-0.0cm} \fontsize{7pt}{0pt}\selectfont 1.036,\textbf{0.773}} & \parbox{1.7cm}{\centering \includegraphics[height=1.5cm]{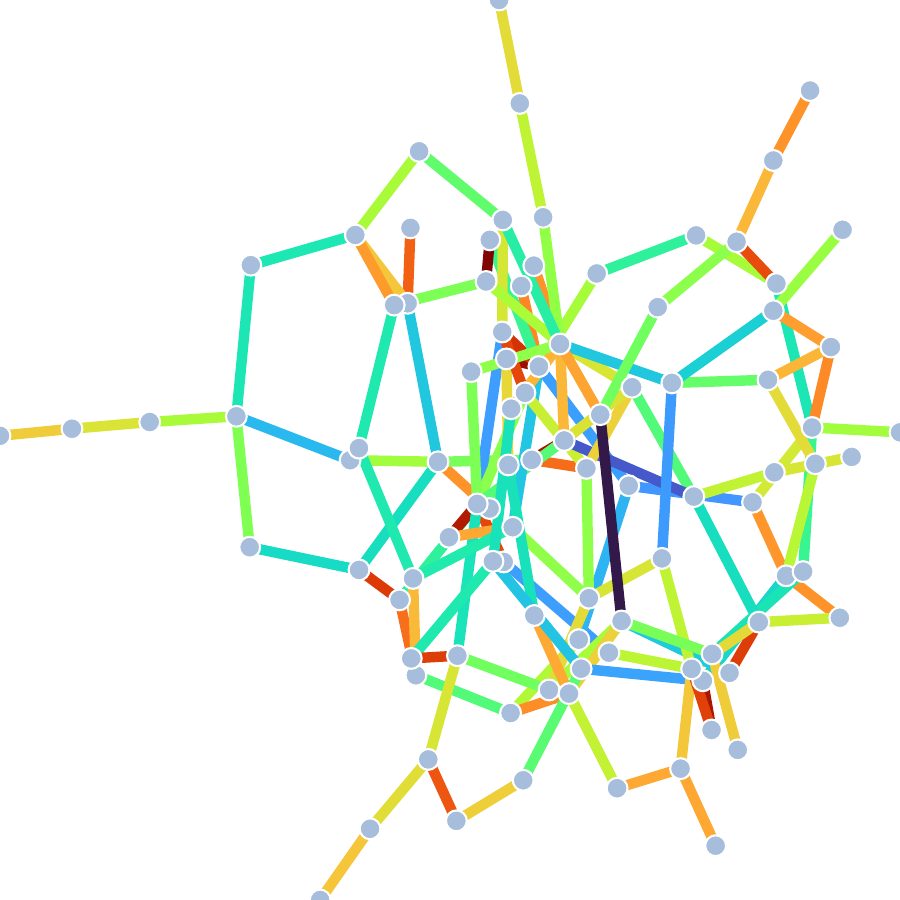} \\ \vspace{-0.0cm} \fontsize{7pt}{0pt}\selectfont \textbf{0.214},0.282} & \parbox{1.7cm}{\centering \includegraphics[height=1.5cm]{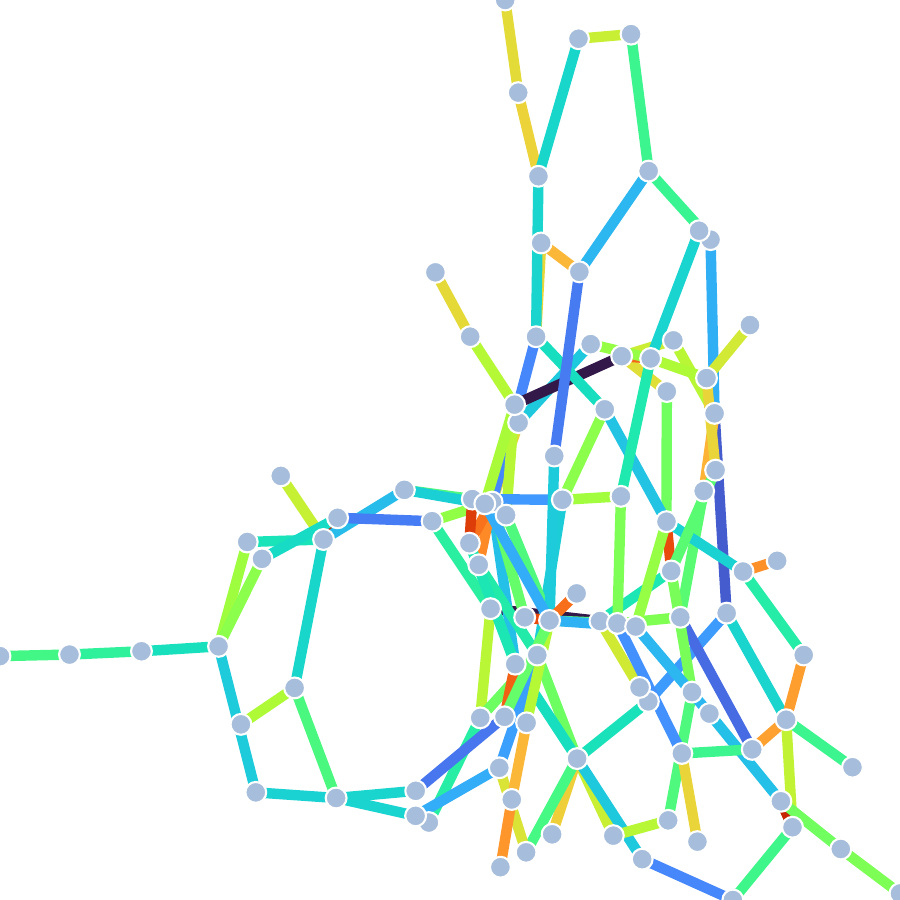} \\ \vspace{-0.0cm} \fontsize{7pt}{0pt}\selectfont -2.139,\textbf{-2.16}} & \parbox{1.7cm}{\centering \includegraphics[height=1.5cm]{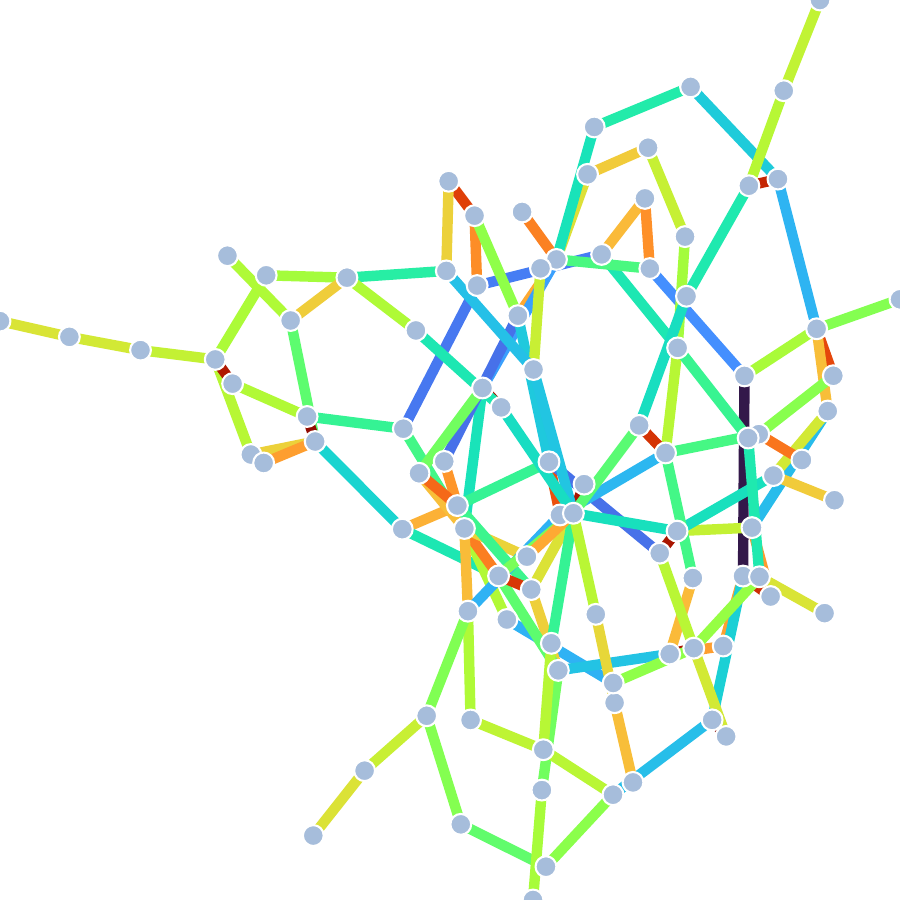} \\ \vspace{-0.0cm} \fontsize{7pt}{0pt}\selectfont \textbf{0.085},0.087} & \parbox{1.7cm}{\centering \includegraphics[height=1.5cm]{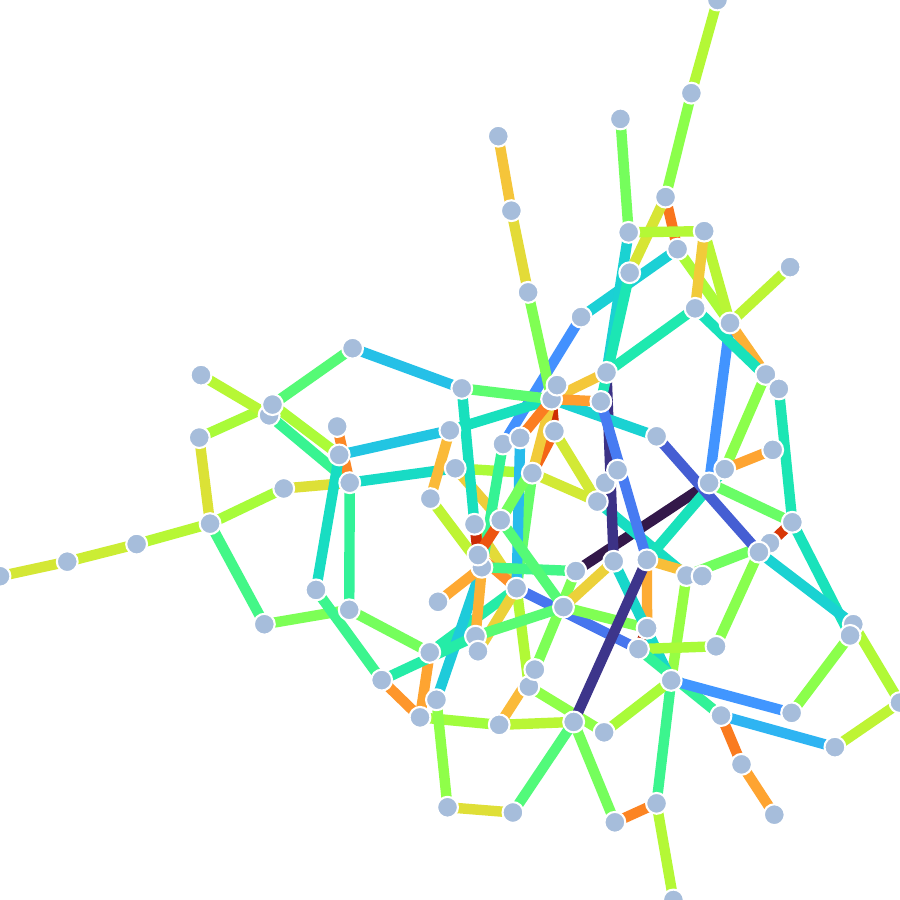} \\ \vspace{-0.0cm} \fontsize{7pt}{0pt}\selectfont 1.335,\textbf{1.28}} & \parbox{1.7cm}{\centering \includegraphics[height=1.5cm]{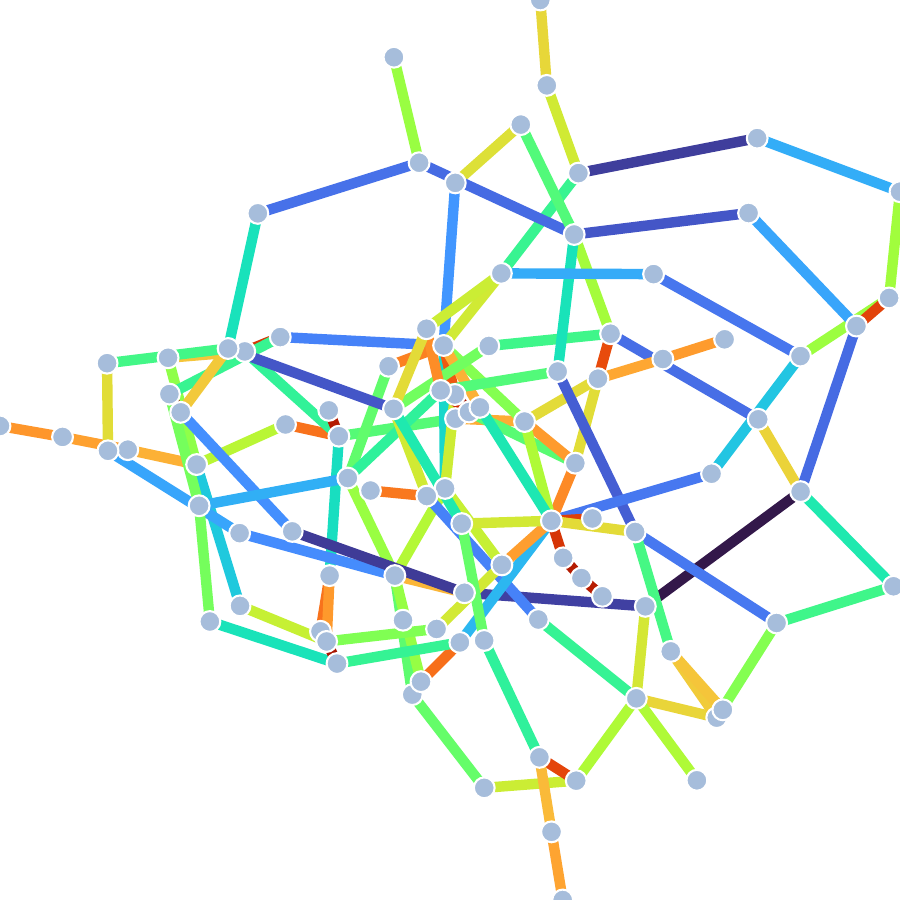} \\ \vspace{-0.0cm} \fontsize{7pt}{0pt}\selectfont 99,\textbf{82}} \\
grafo10451.95 & \parbox{1.7cm}{\centering \includegraphics[height=1.5cm]{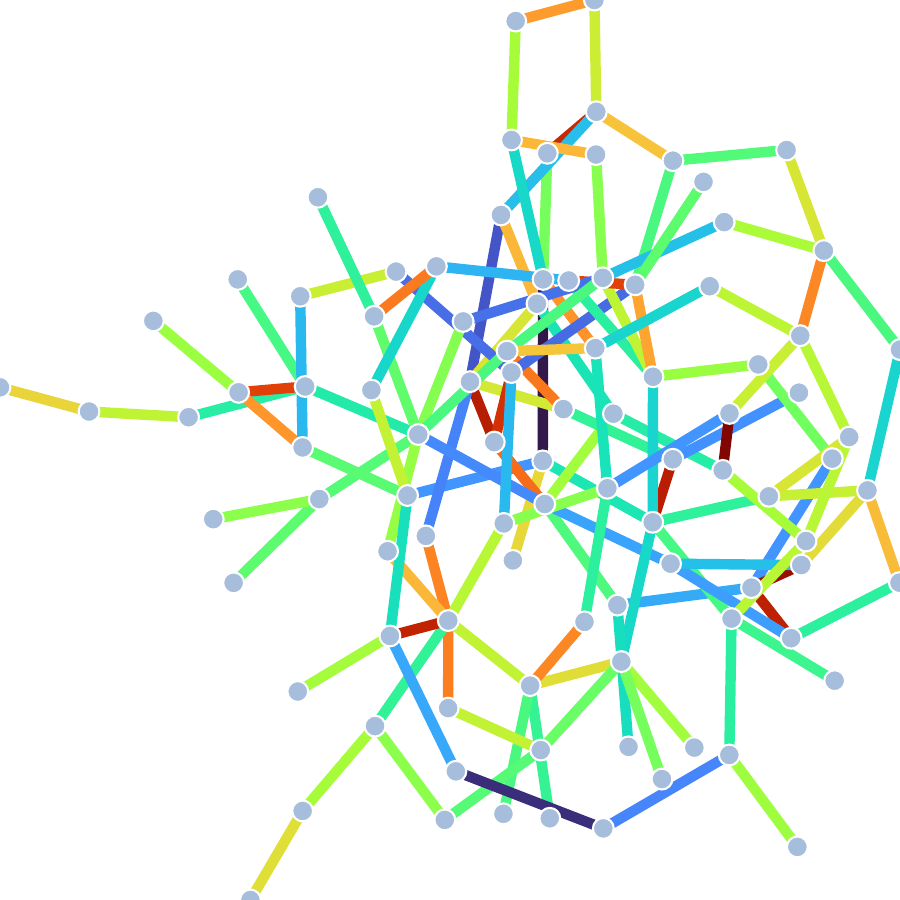} \\ \vspace{-0.0cm} \fontsize{7pt}{0pt}\selectfont } & \parbox{1.7cm}{\centering \includegraphics[height=1.5cm]{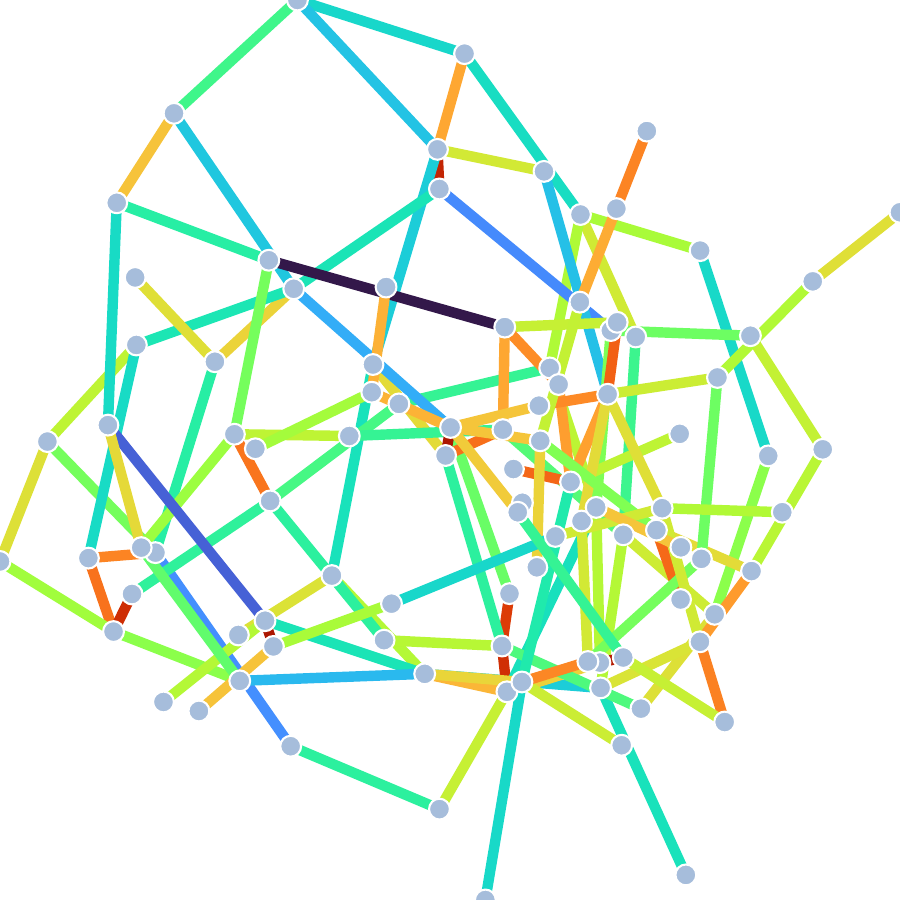} \\ \vspace{-0.0cm} \fontsize{7pt}{0pt}\selectfont 1.046,\textbf{0.868}} & \parbox{1.7cm}{\centering \includegraphics[height=1.5cm]{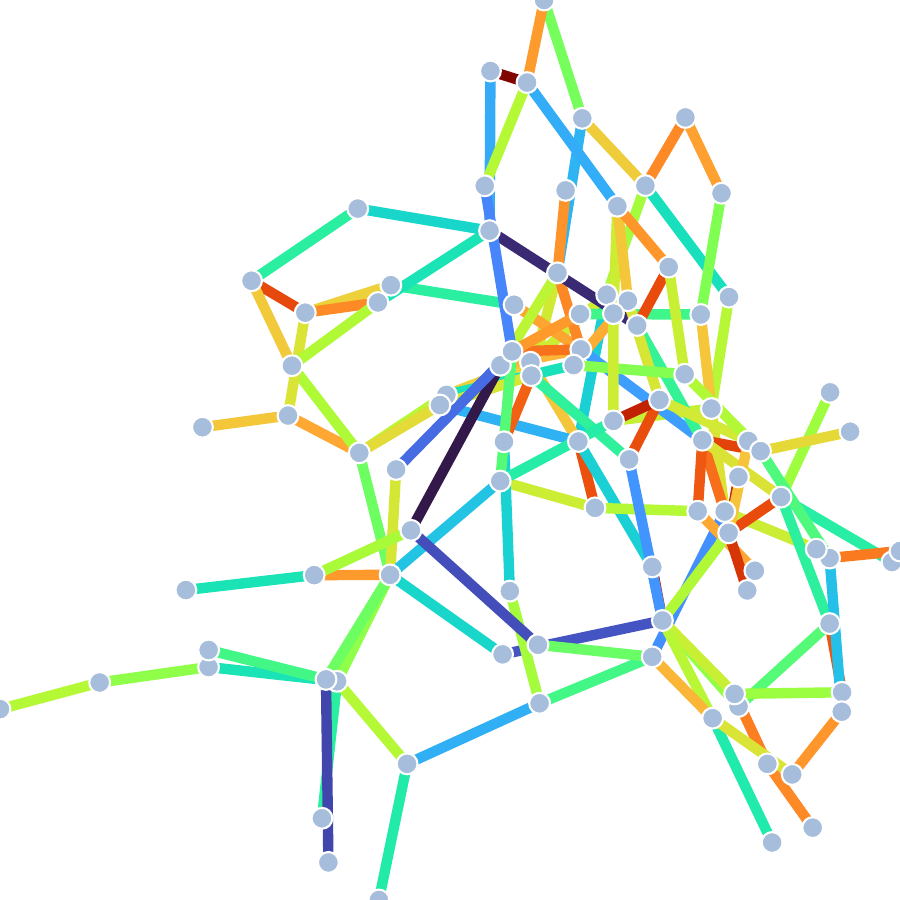} \\ \vspace{-0.0cm} \fontsize{7pt}{0pt}\selectfont \textbf{0.212},0.278} & \parbox{1.7cm}{\centering \includegraphics[height=1.5cm]{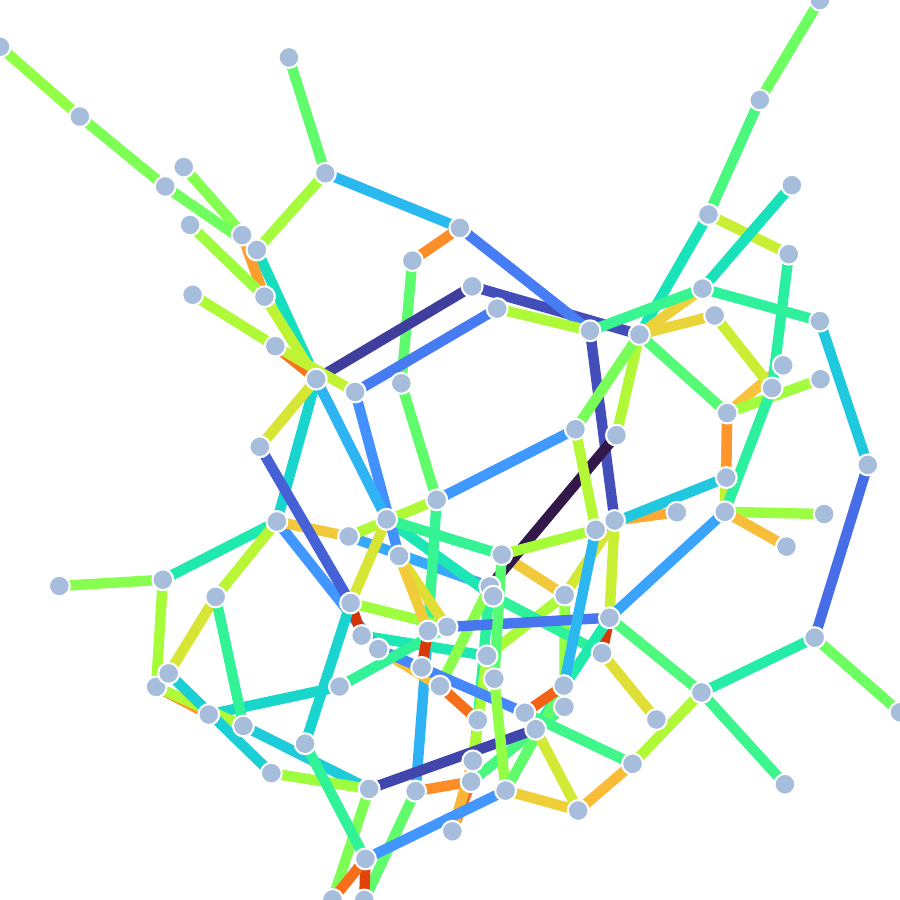} \\ \vspace{-0.0cm} \fontsize{7pt}{0pt}\selectfont -2.032,\textbf{-2.063}} & \parbox{1.7cm}{\centering \includegraphics[height=1.5cm]{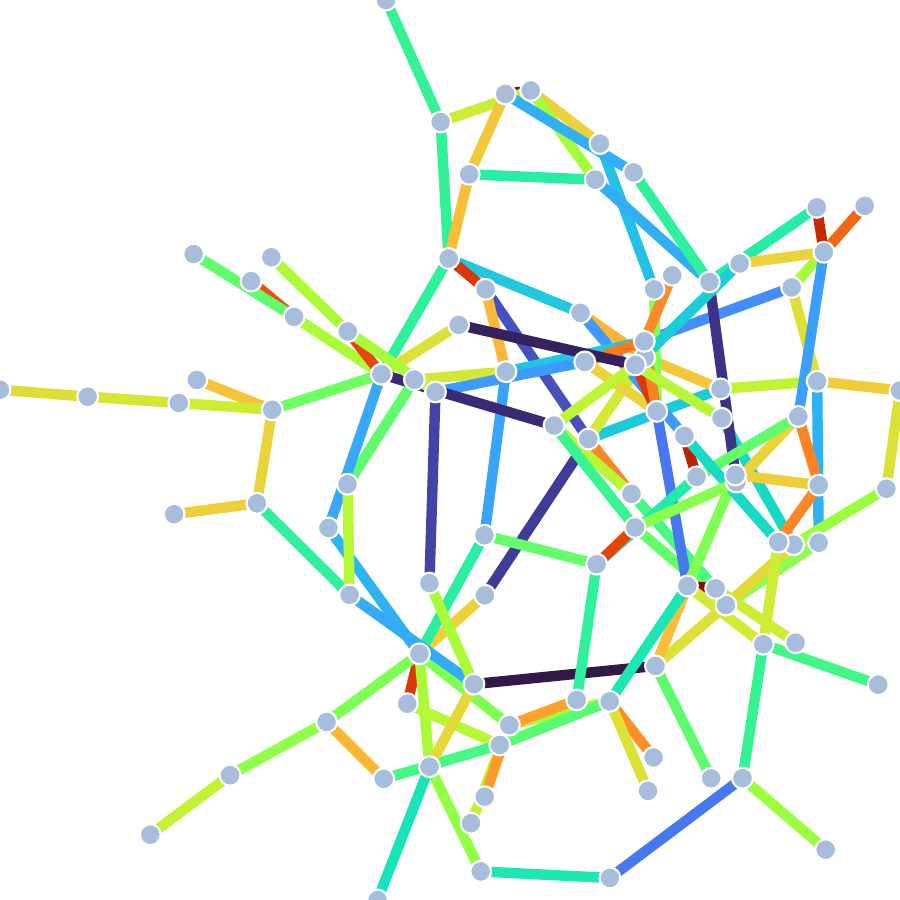} \\ \vspace{-0.0cm} \fontsize{7pt}{0pt}\selectfont 0.103,\textbf{0.1}} & \parbox{1.7cm}{\centering \includegraphics[height=1.5cm]{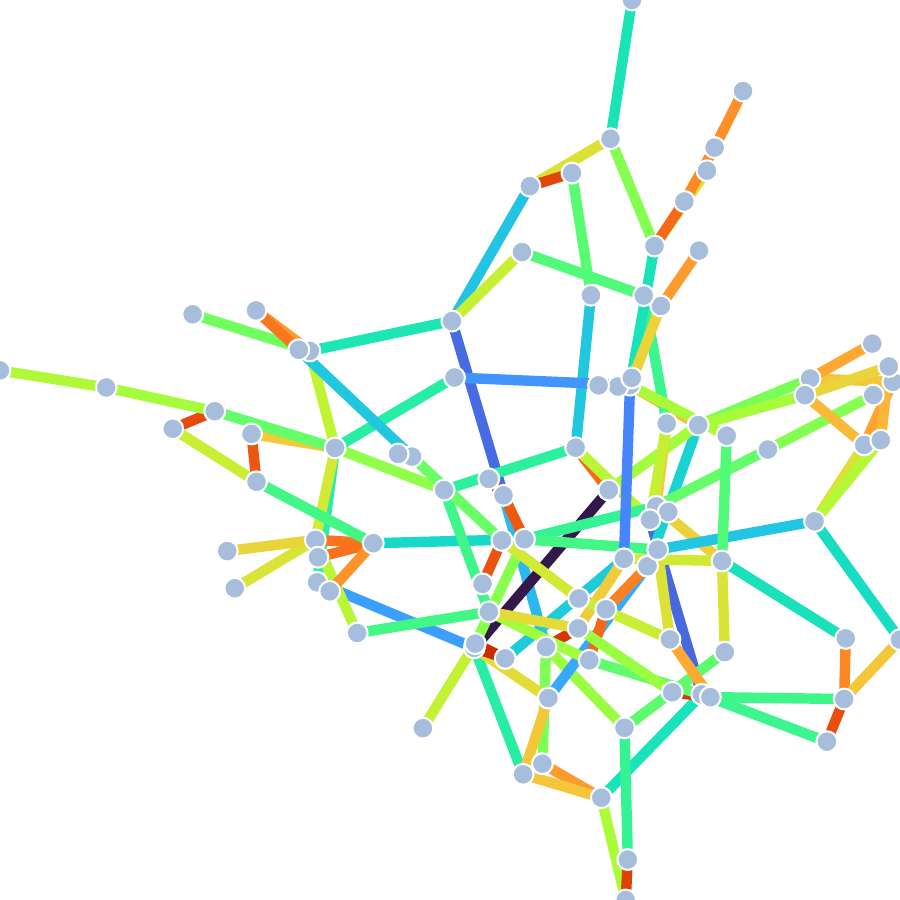} \\ \vspace{-0.0cm} \fontsize{7pt}{0pt}\selectfont 1.444,\textbf{1.307}} & \parbox{1.7cm}{\centering \includegraphics[height=1.5cm]{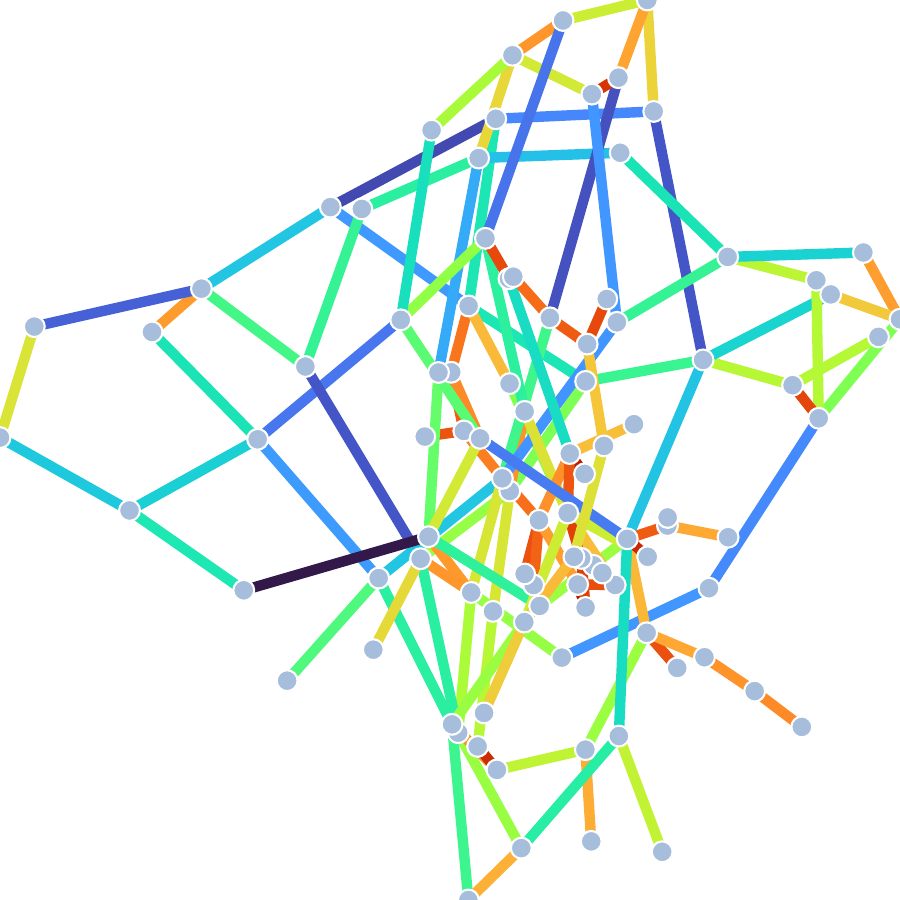} \\ \vspace{-0.0cm} \fontsize{7pt}{0pt}\selectfont 117,\textbf{96}} \\
grafo10480.95 & \parbox{1.7cm}{\centering \includegraphics[height=1.5cm]{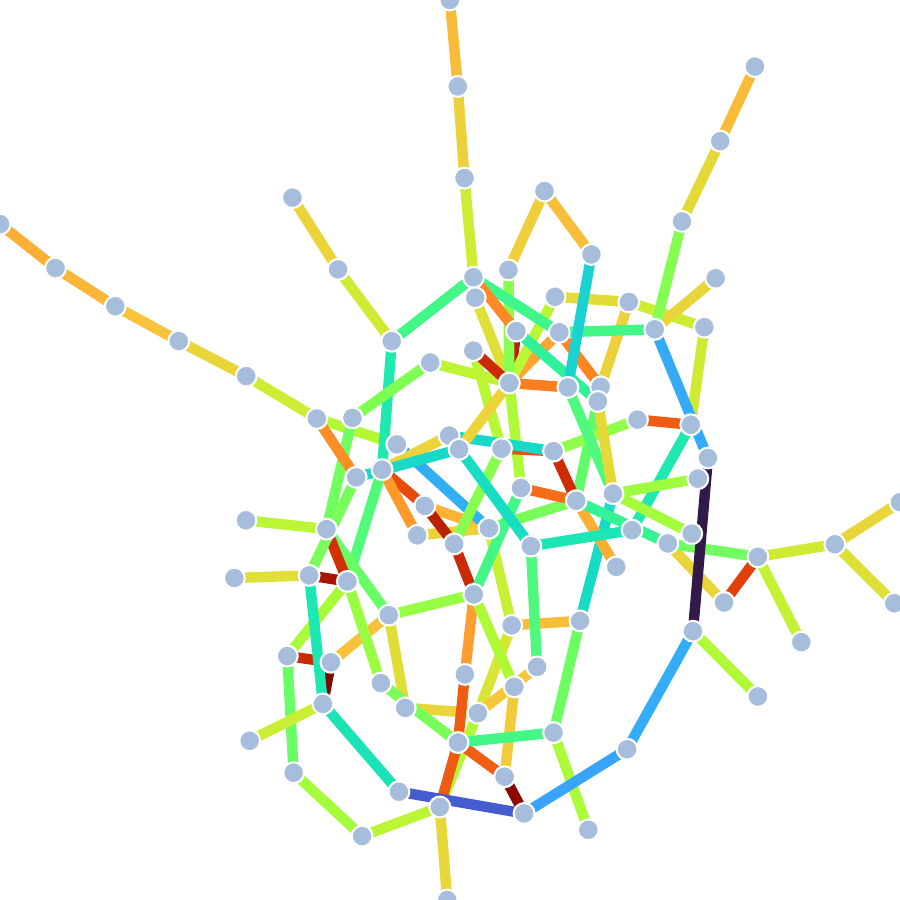} \\ \vspace{-0.0cm} \fontsize{7pt}{0pt}\selectfont } & \parbox{1.7cm}{\centering \includegraphics[height=1.5cm]{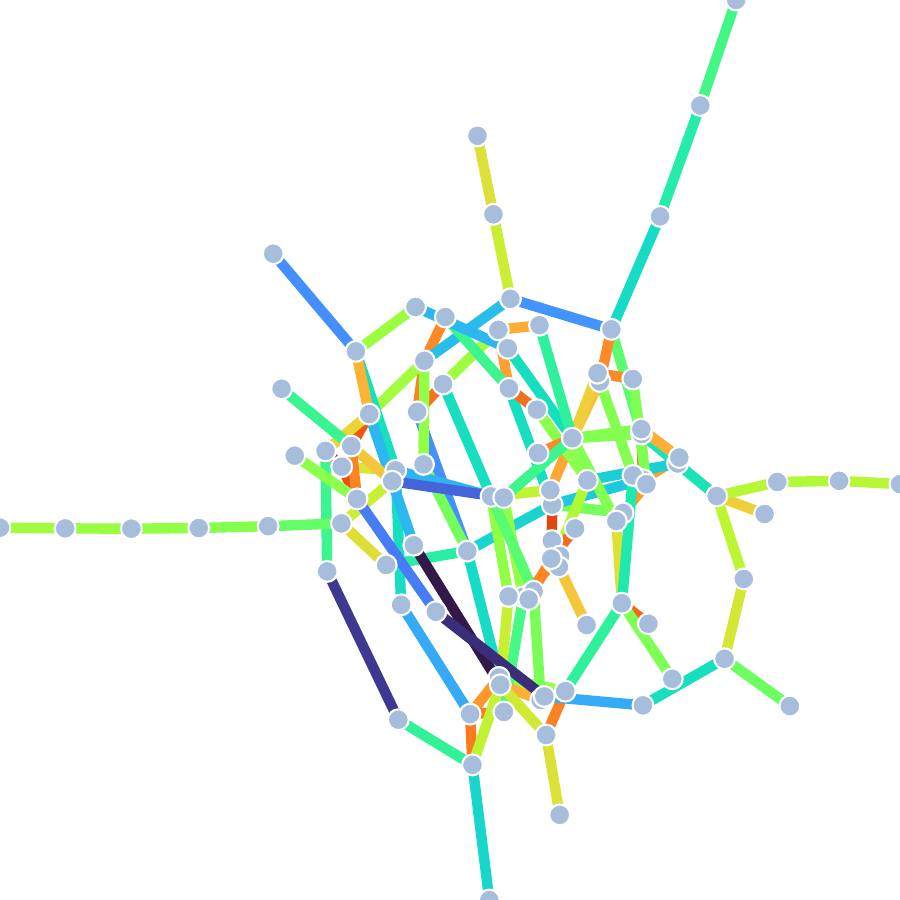} \\ \vspace{-0.0cm} \fontsize{7pt}{0pt}\selectfont 0.893,\textbf{0.835}} & \parbox{1.7cm}{\centering \includegraphics[height=1.5cm]{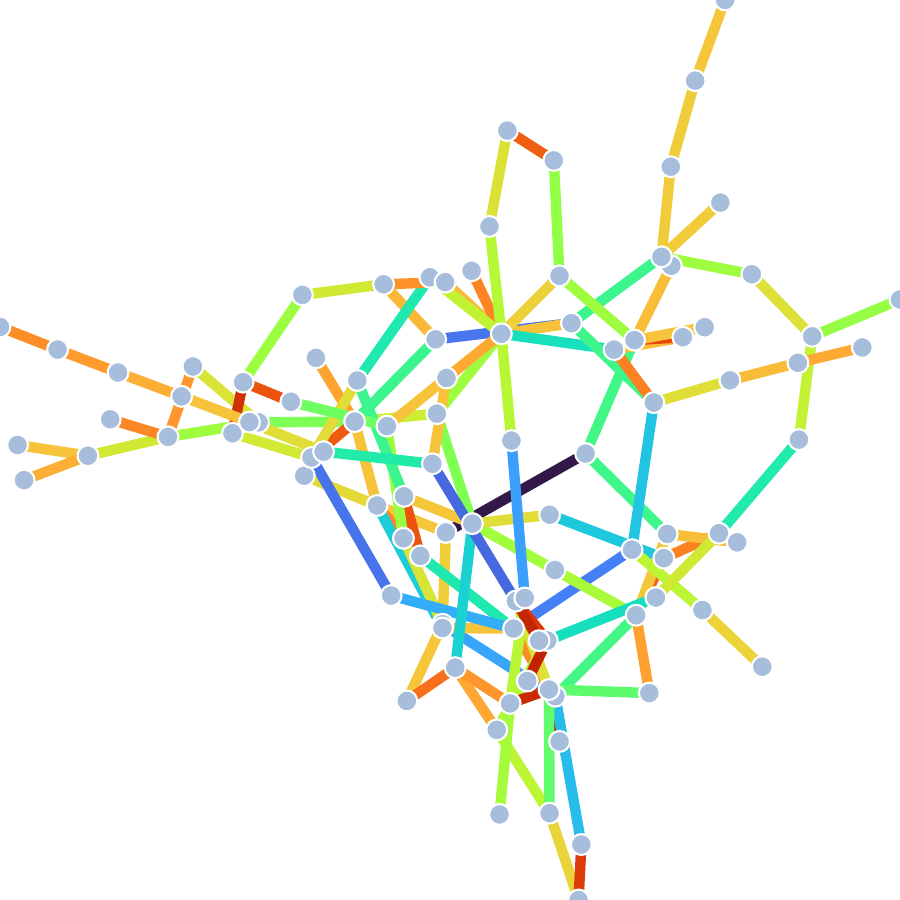} \\ \vspace{-0.0cm} \fontsize{7pt}{0pt}\selectfont \textbf{0.245},0.296} & \parbox{1.7cm}{\centering \includegraphics[height=1.5cm]{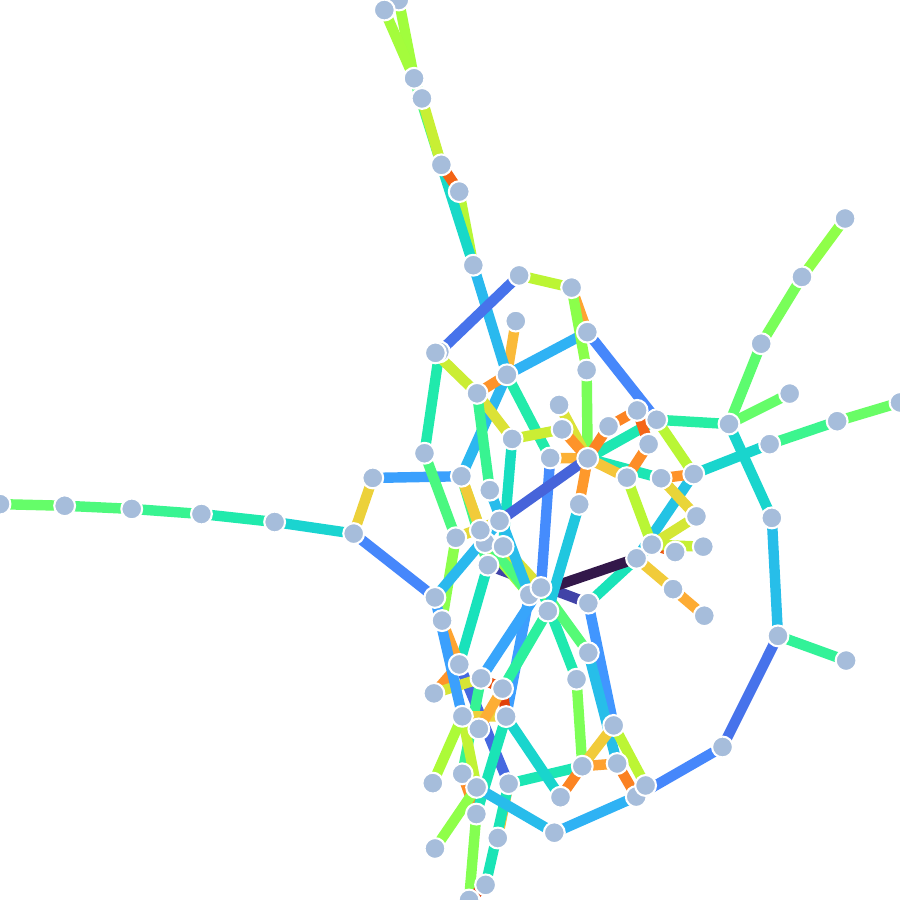} \\ \vspace{-0.0cm} \fontsize{7pt}{0pt}\selectfont -2.191,\textbf{-2.244}} & \parbox{1.7cm}{\centering \includegraphics[height=1.5cm]{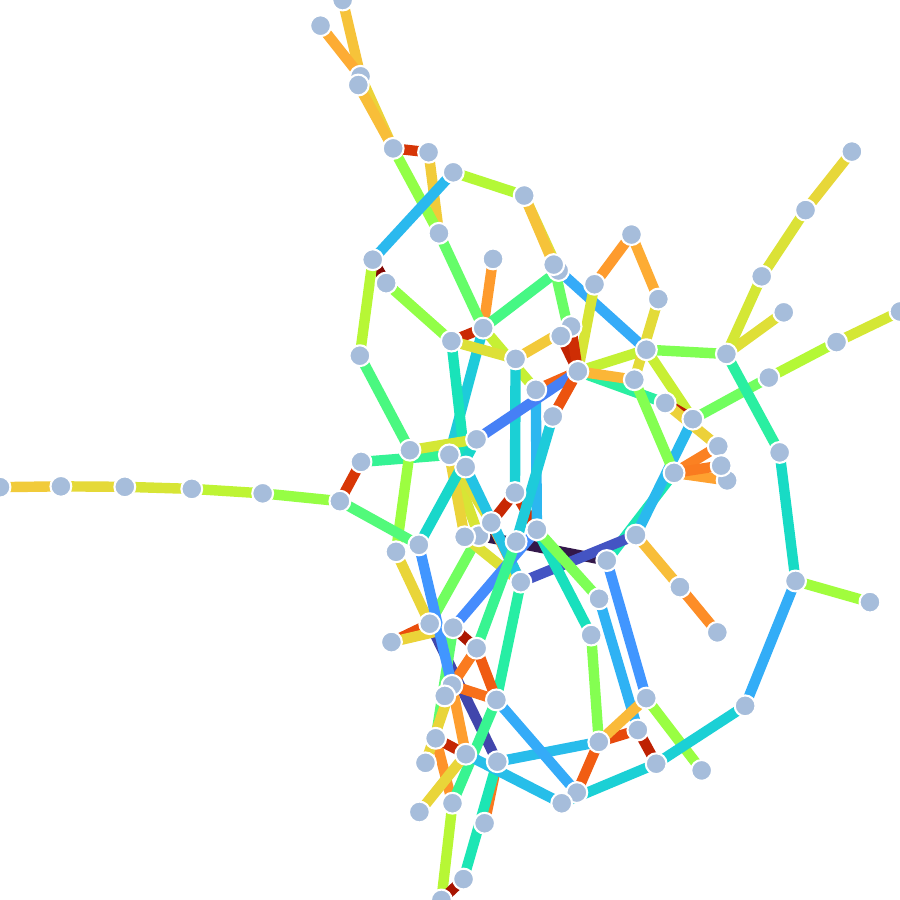} \\ \vspace{-0.0cm} \fontsize{7pt}{0pt}\selectfont 0.091,\textbf{0.088}} & \parbox{1.7cm}{\centering \includegraphics[height=1.5cm]{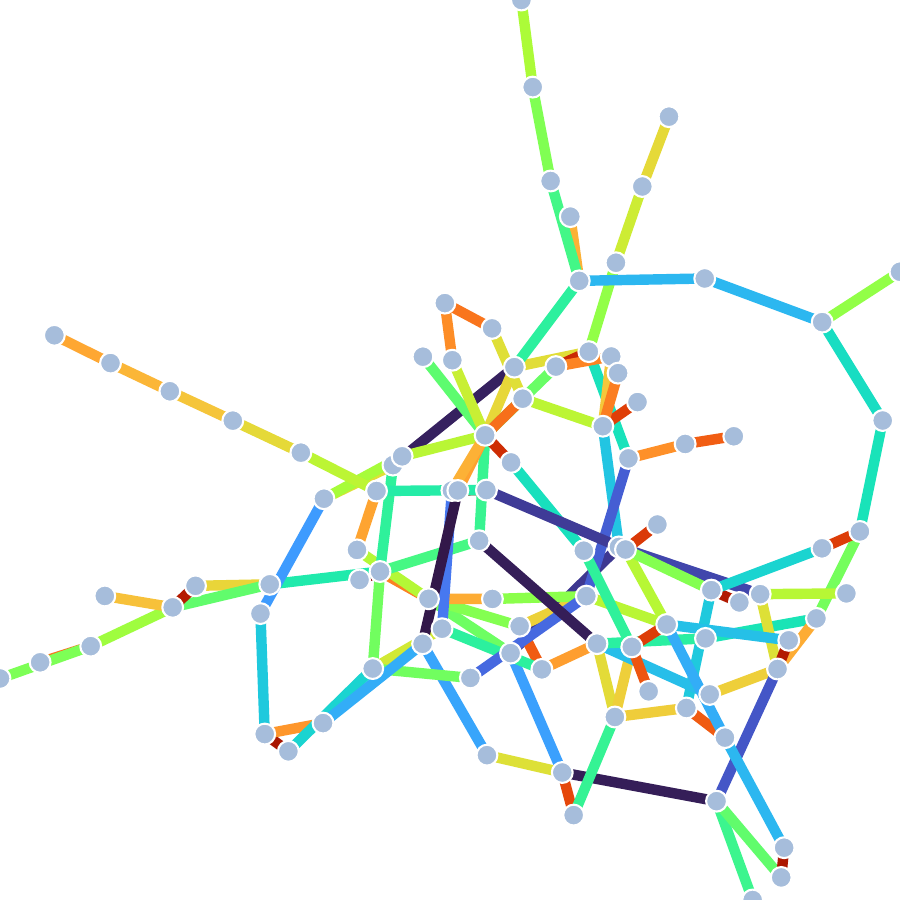} \\ \vspace{-0.0cm} \fontsize{7pt}{0pt}\selectfont 1.381,\textbf{1.288}} & \parbox{1.7cm}{\centering \includegraphics[height=1.5cm]{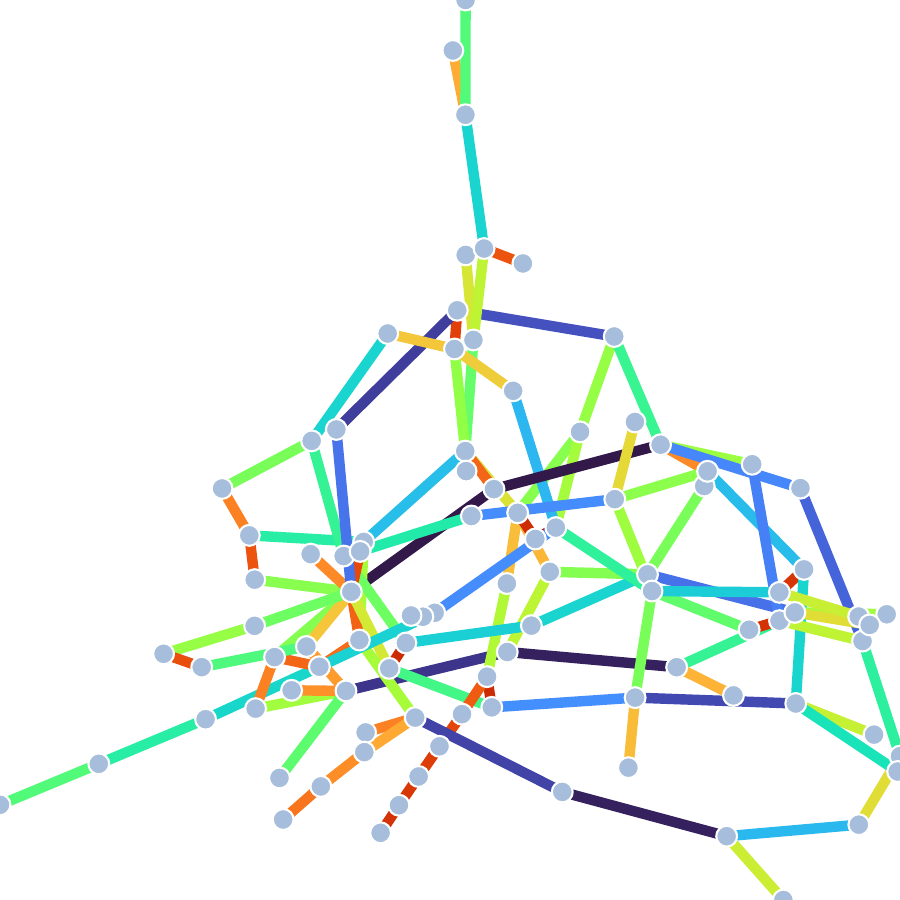} \\ \vspace{-0.0cm} \fontsize{7pt}{0pt}\selectfont 70,\textbf{59}} \\
grafo11311.40 & \parbox{1.7cm}{\centering \includegraphics[height=1.5cm]{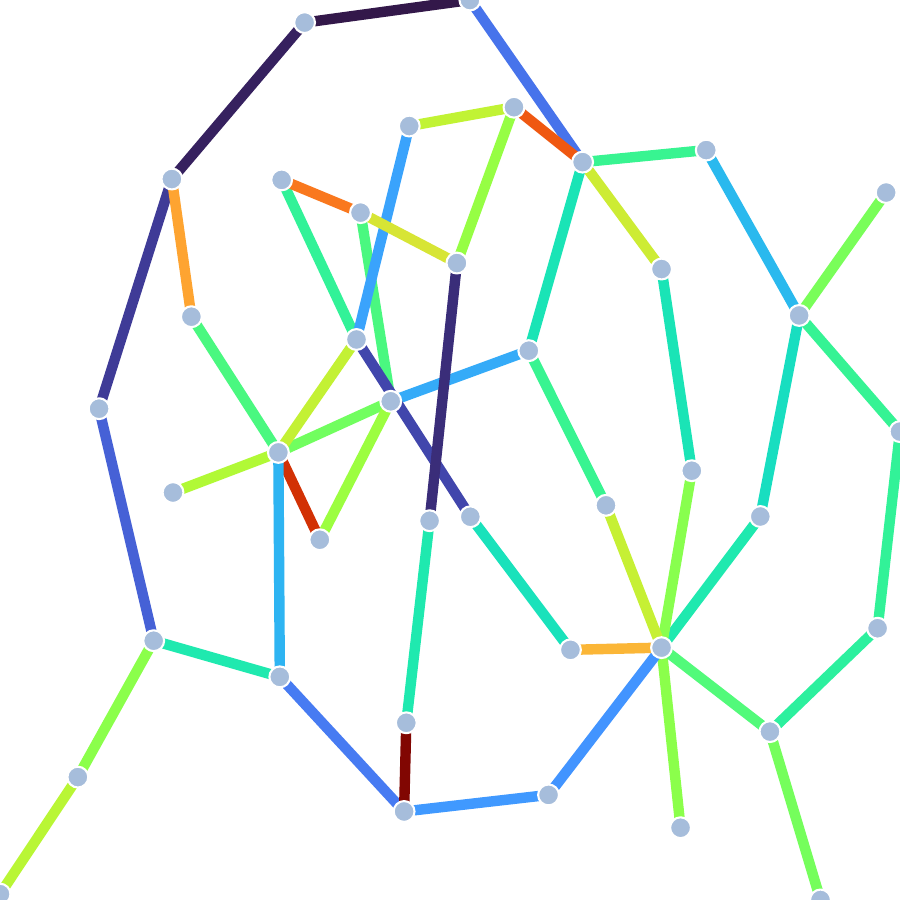} \\ \vspace{-0.0cm} \fontsize{7pt}{0pt}\selectfont } & \parbox{1.7cm}{\centering \includegraphics[height=1.5cm]{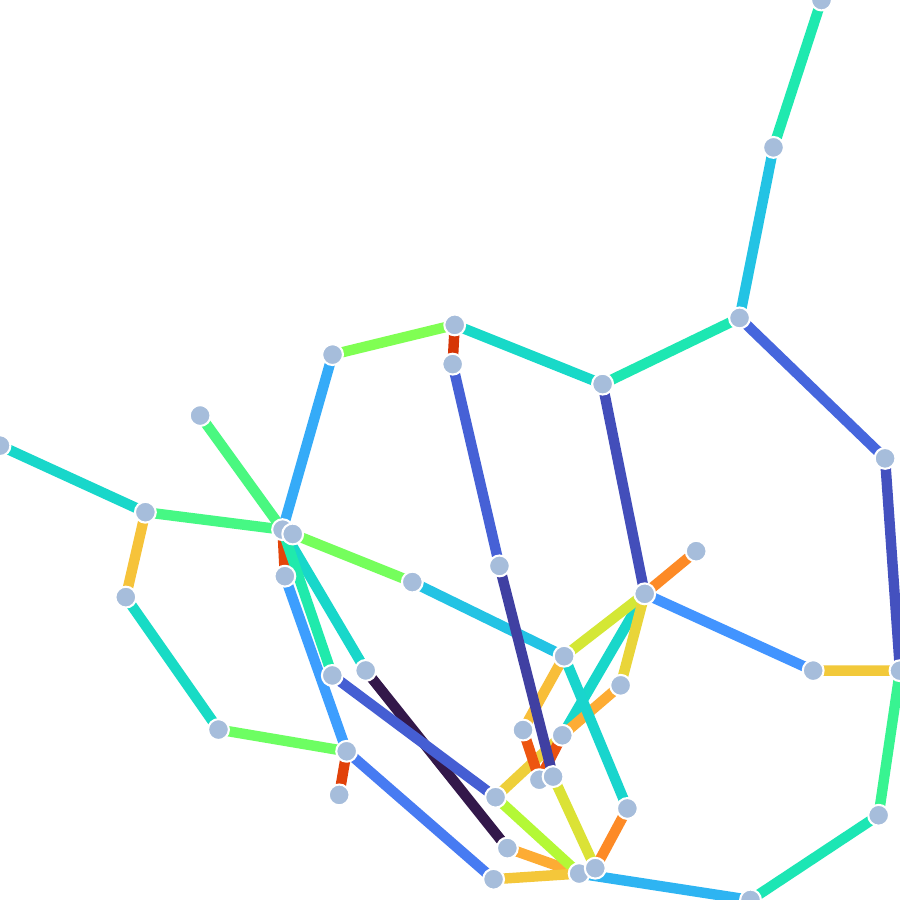} \\ \vspace{-0.0cm} \fontsize{7pt}{0pt}\selectfont 0.87,\textbf{0.696}} & \parbox{1.7cm}{\centering \includegraphics[height=1.5cm]{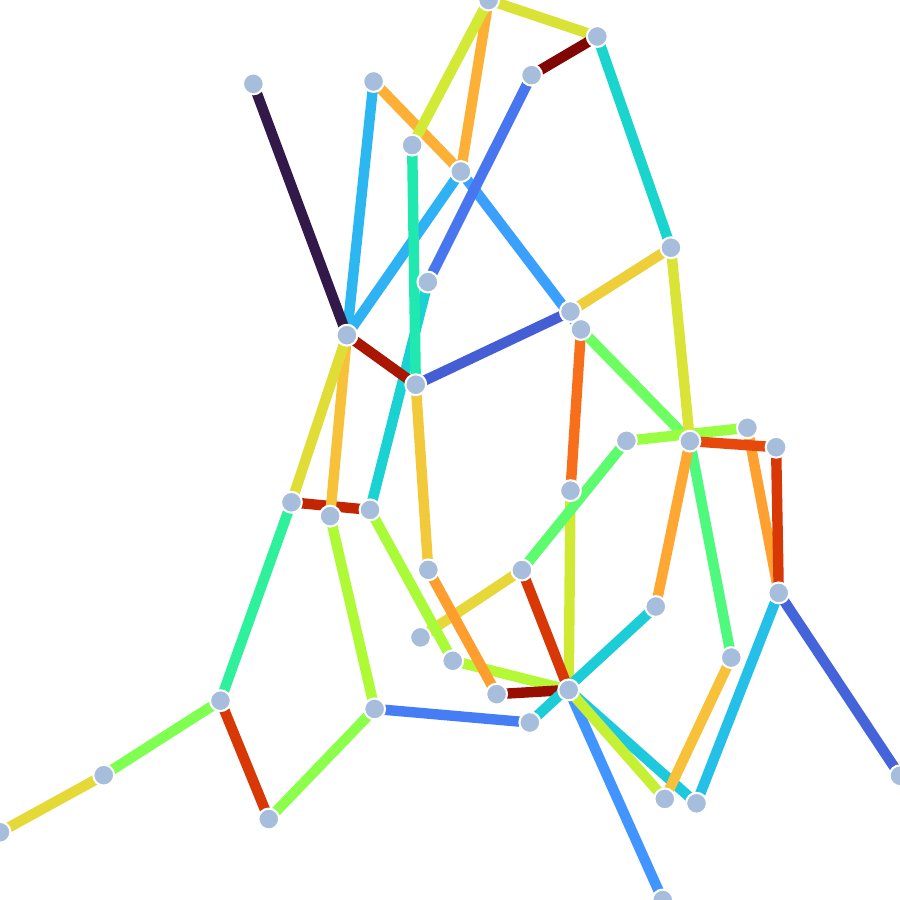} \\ \vspace{-0.0cm} \fontsize{7pt}{0pt}\selectfont \textbf{0.196},0.216} & \parbox{1.7cm}{\centering \includegraphics[height=1.5cm]{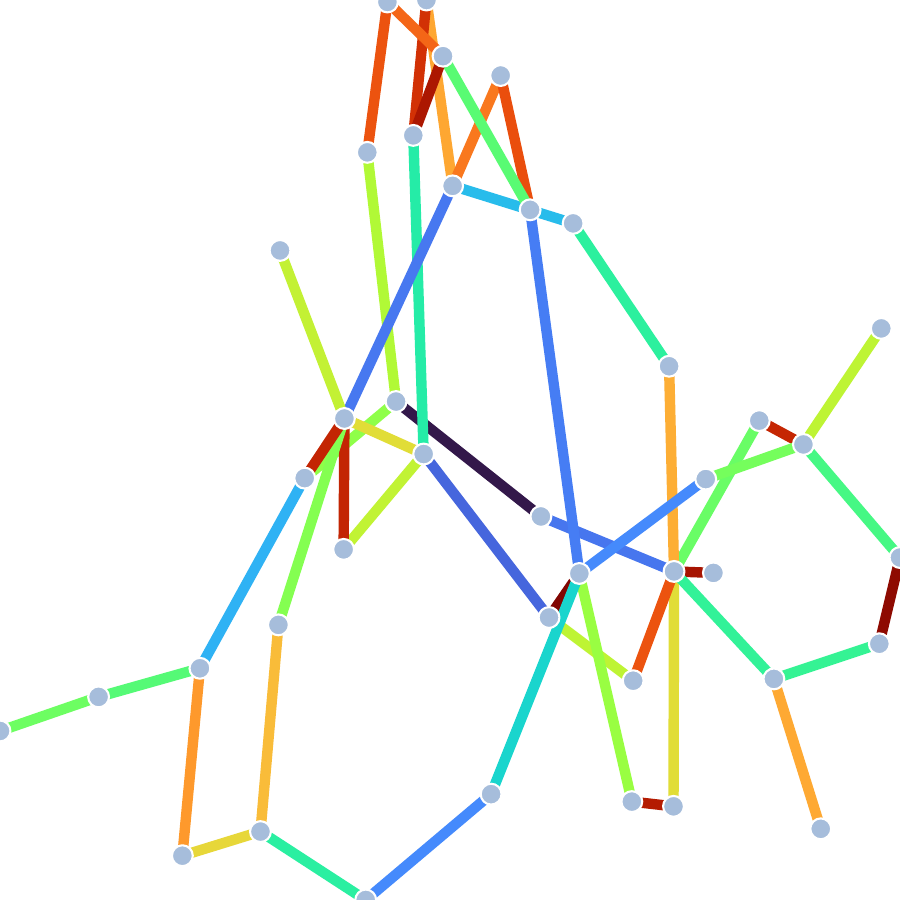} \\ \vspace{-0.0cm} \fontsize{7pt}{0pt}\selectfont -1.726,\textbf{-1.727}} & \parbox{1.7cm}{\centering \includegraphics[height=1.5cm]{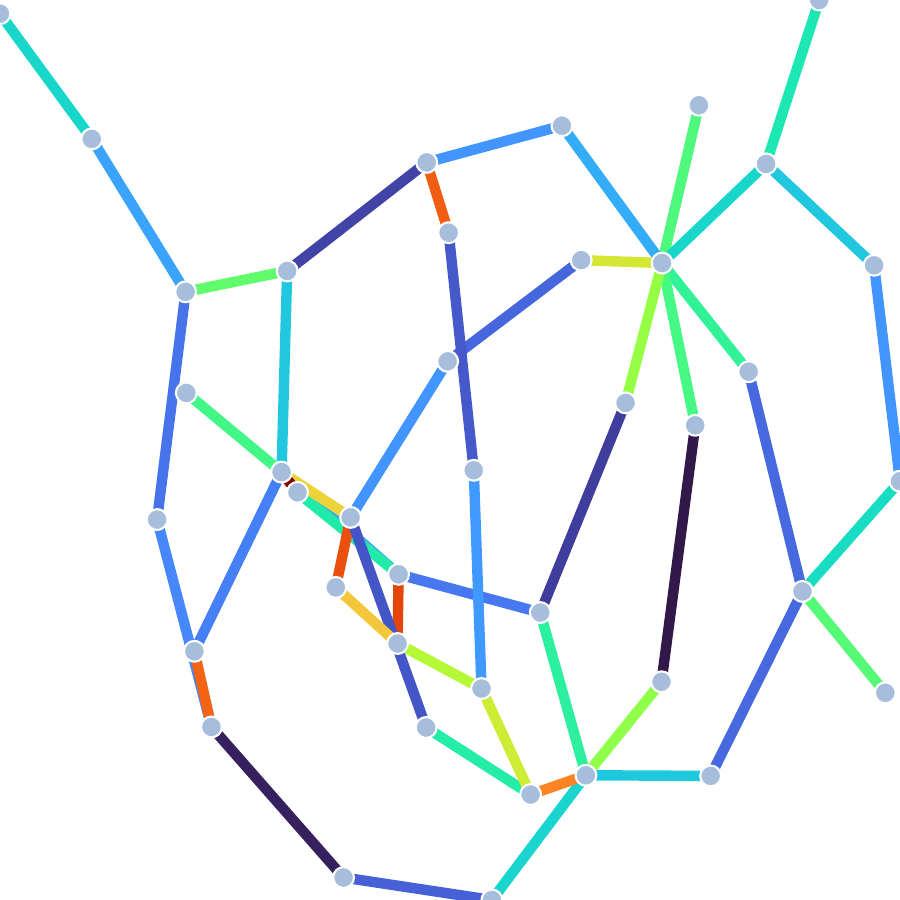} \\ \vspace{-0.0cm} \fontsize{7pt}{0pt}\selectfont \textbf{0.066},0.073} & \parbox{1.7cm}{\centering \includegraphics[height=1.5cm]{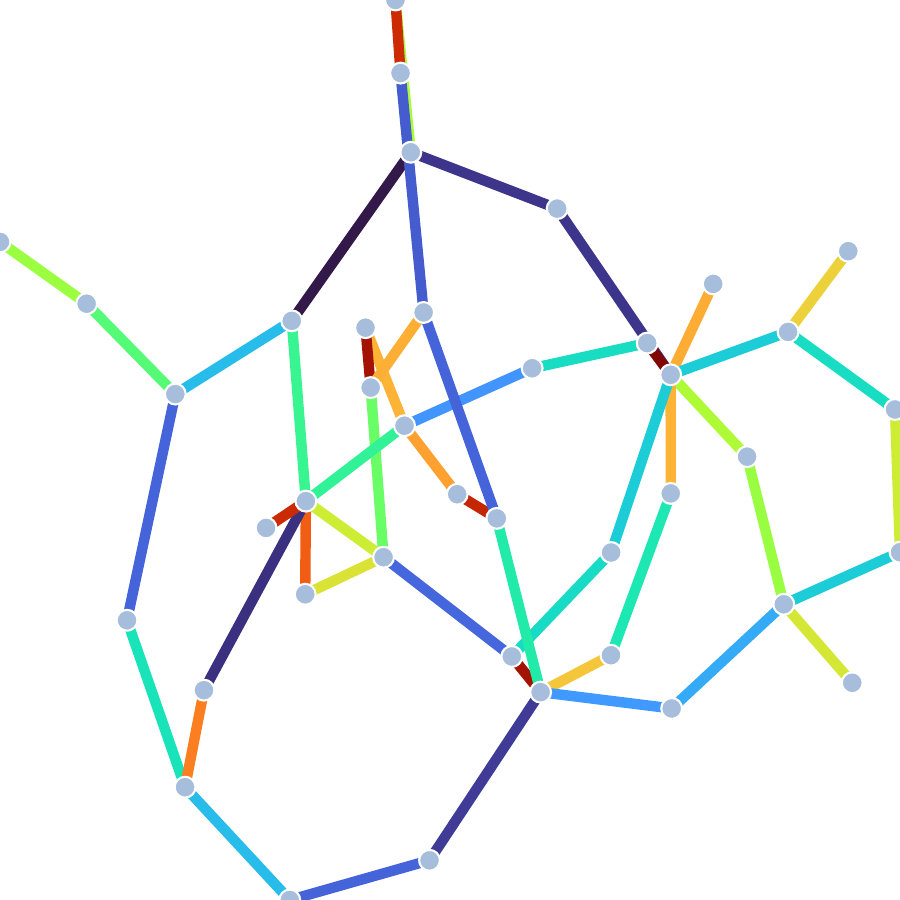} \\ \vspace{-0.0cm} \fontsize{7pt}{0pt}\selectfont 0.784,\textbf{0.716}} & \parbox{1.7cm}{\centering \includegraphics[height=1.5cm]{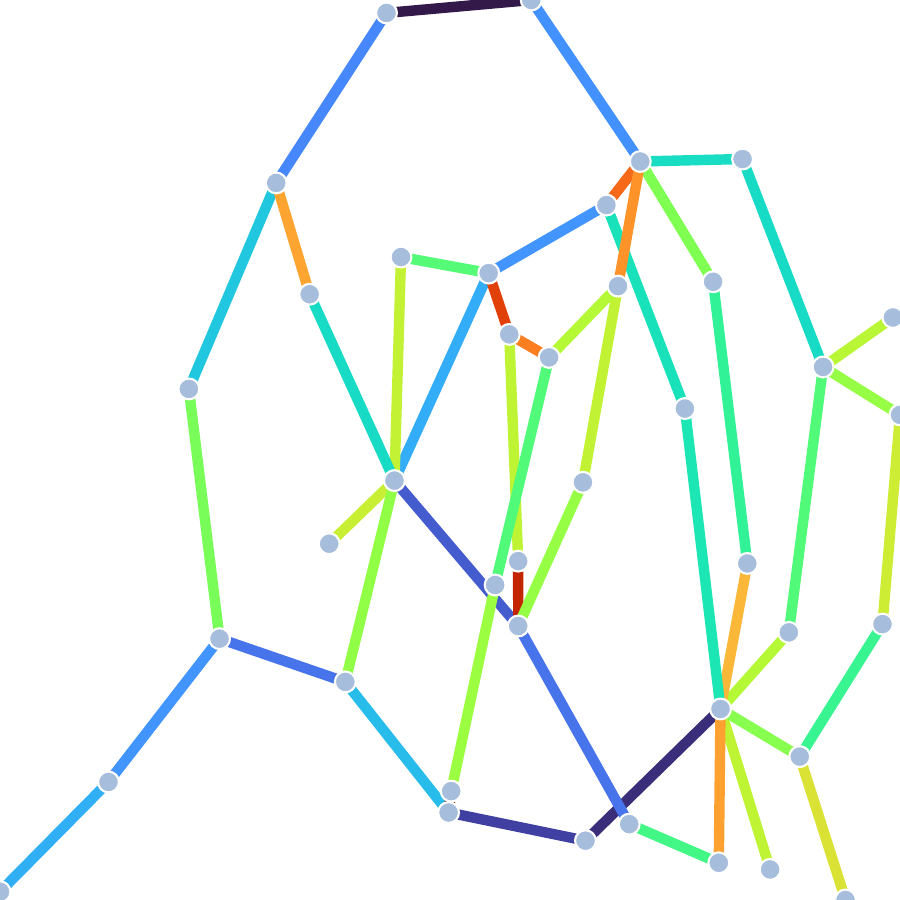} \\ \vspace{-0.0cm} \fontsize{7pt}{0pt}\selectfont 6,\textbf{4}} \\
grafo11412.32 & \parbox{1.7cm}{\centering \includegraphics[height=1.5cm]{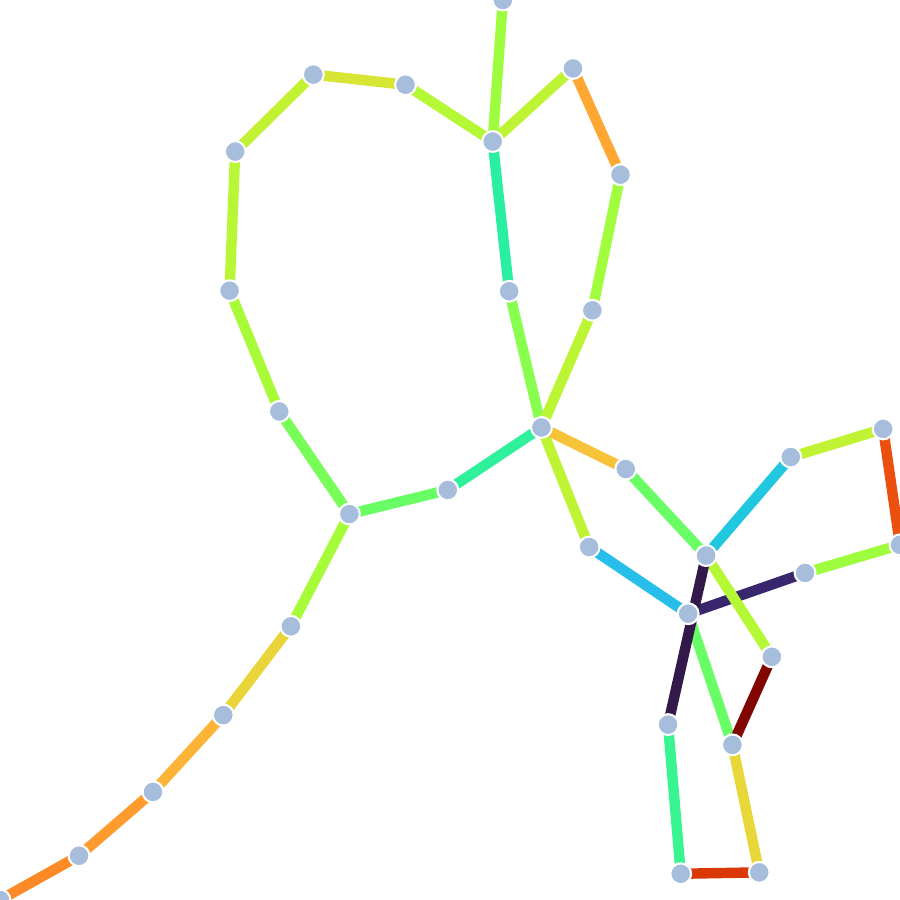} \\ \vspace{-0.0cm} \fontsize{7pt}{0pt}\selectfont } & \parbox{1.7cm}{\centering \includegraphics[height=1.5cm]{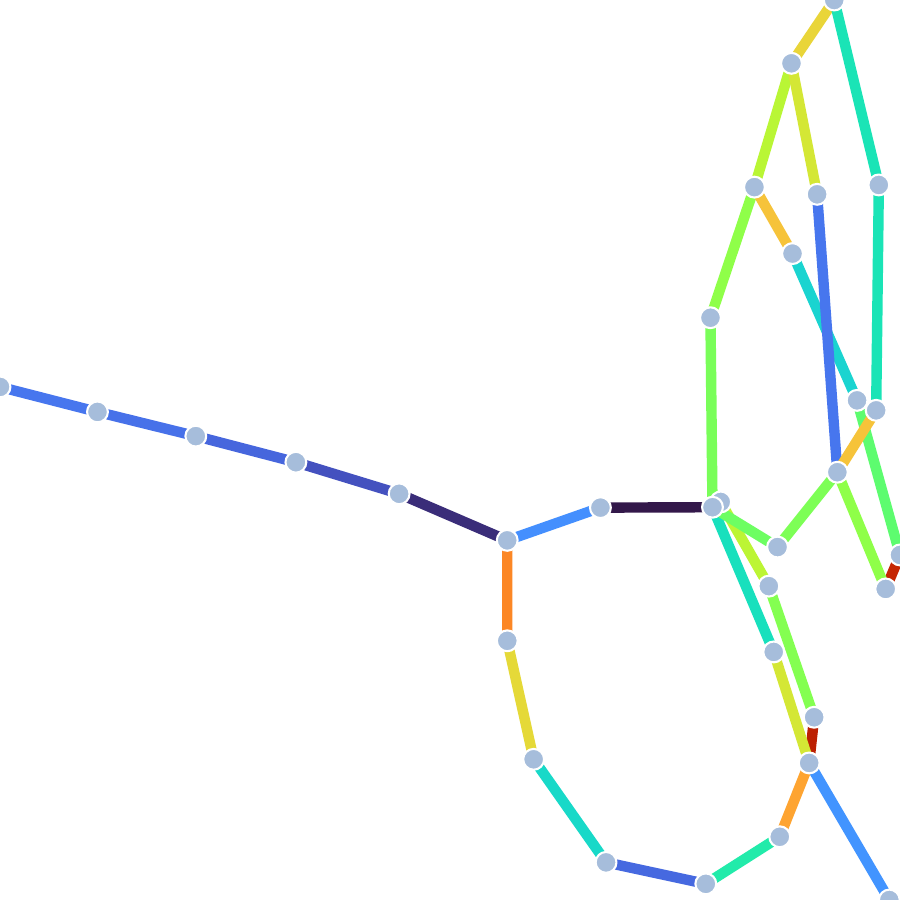} \\ \vspace{-0.0cm} \fontsize{7pt}{0pt}\selectfont 0.842,\textbf{0.696}} & \parbox{1.7cm}{\centering \includegraphics[height=1.5cm]{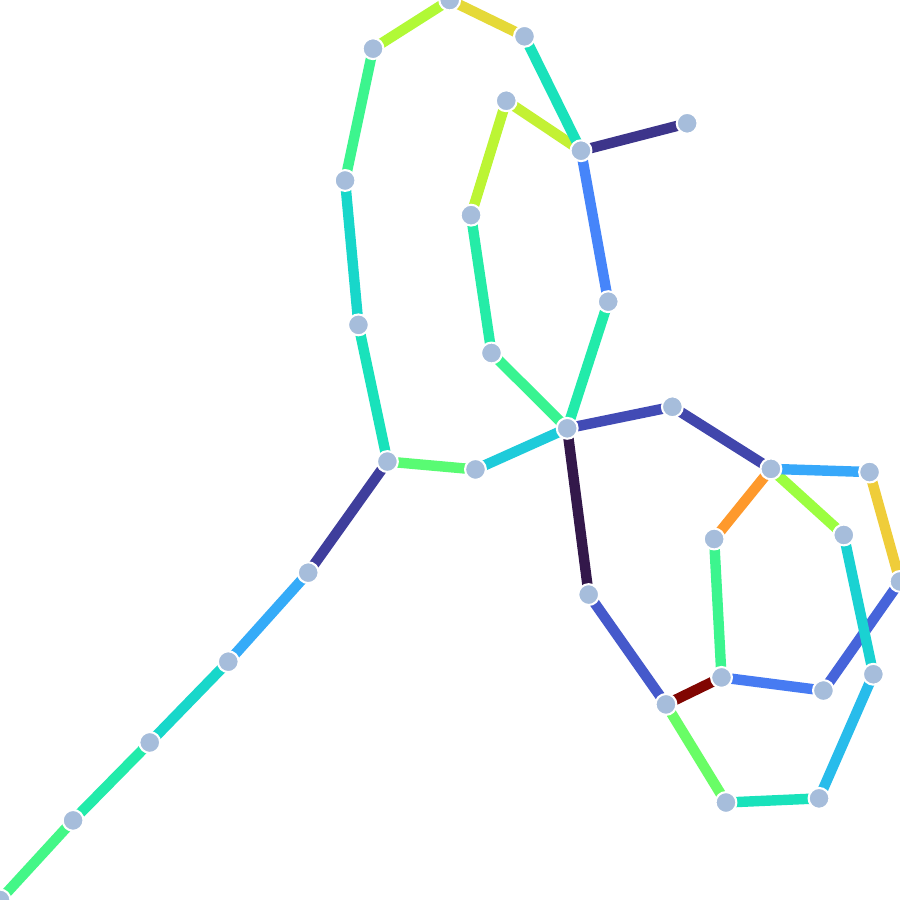} \\ \vspace{-0.0cm} \fontsize{7pt}{0pt}\selectfont \textbf{0.101},0.116} & \parbox{1.7cm}{\centering \includegraphics[height=1.5cm]{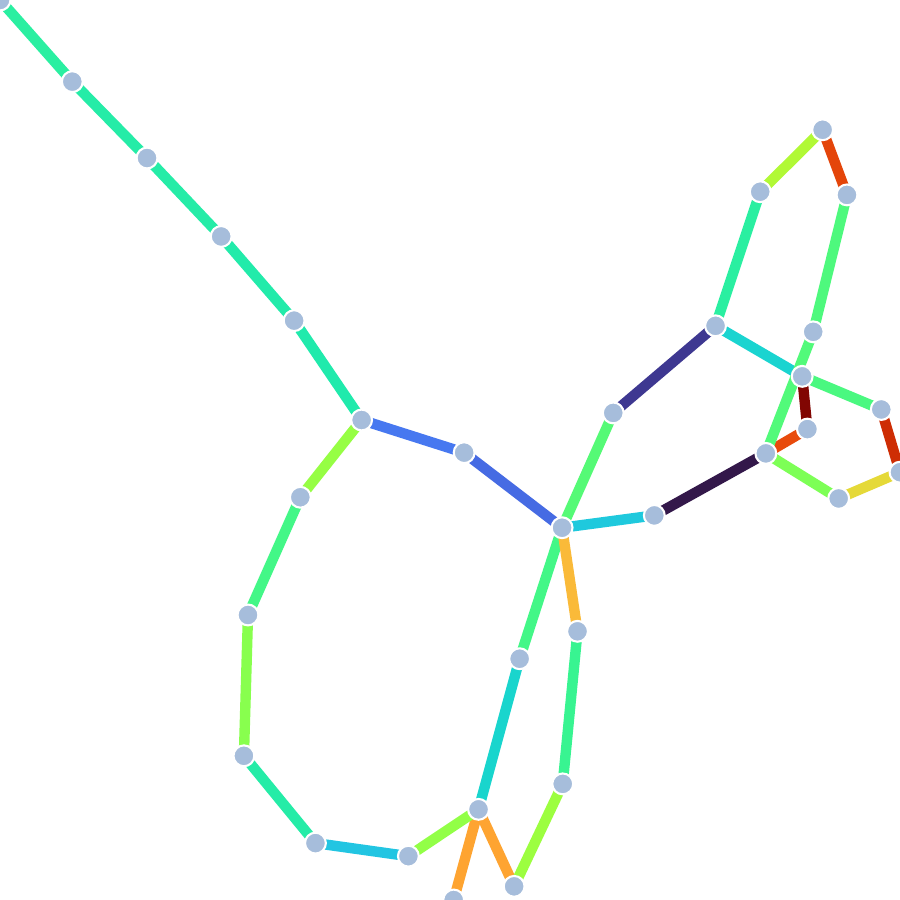} \\ \vspace{-0.0cm} \fontsize{7pt}{0pt}\selectfont -1.979,\textbf{-2.035}} & \parbox{1.7cm}{\centering \includegraphics[height=1.5cm]{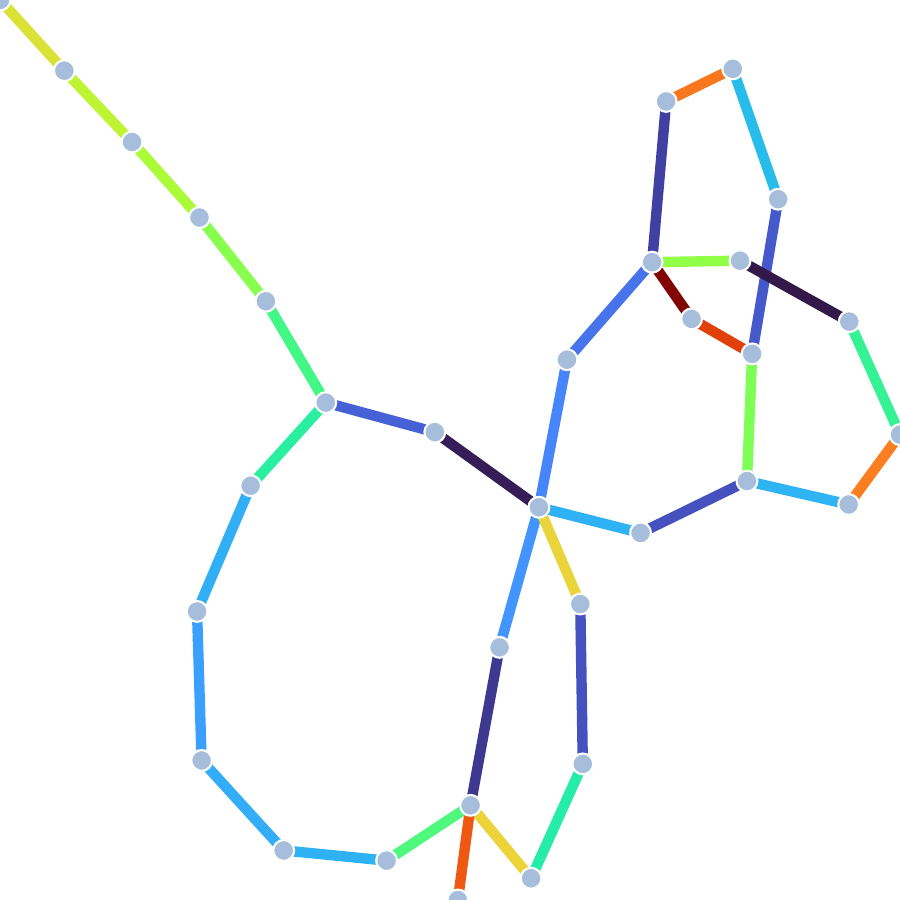} \\ \vspace{-0.0cm} \fontsize{7pt}{0pt}\selectfont \textbf{0.032},0.033} & \parbox{1.7cm}{\centering \includegraphics[height=1.5cm]{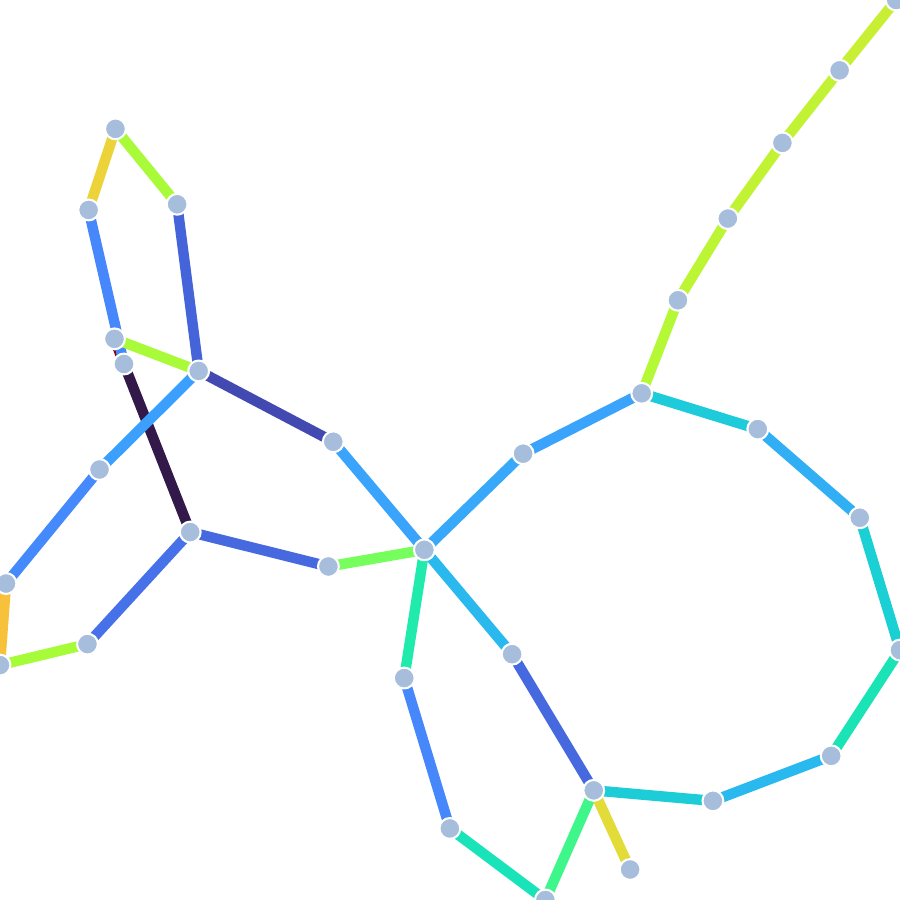} \\ \vspace{-0.0cm} \fontsize{7pt}{0pt}\selectfont 0.569,\textbf{0.512}} & \parbox{1.7cm}{\centering \includegraphics[height=1.5cm]{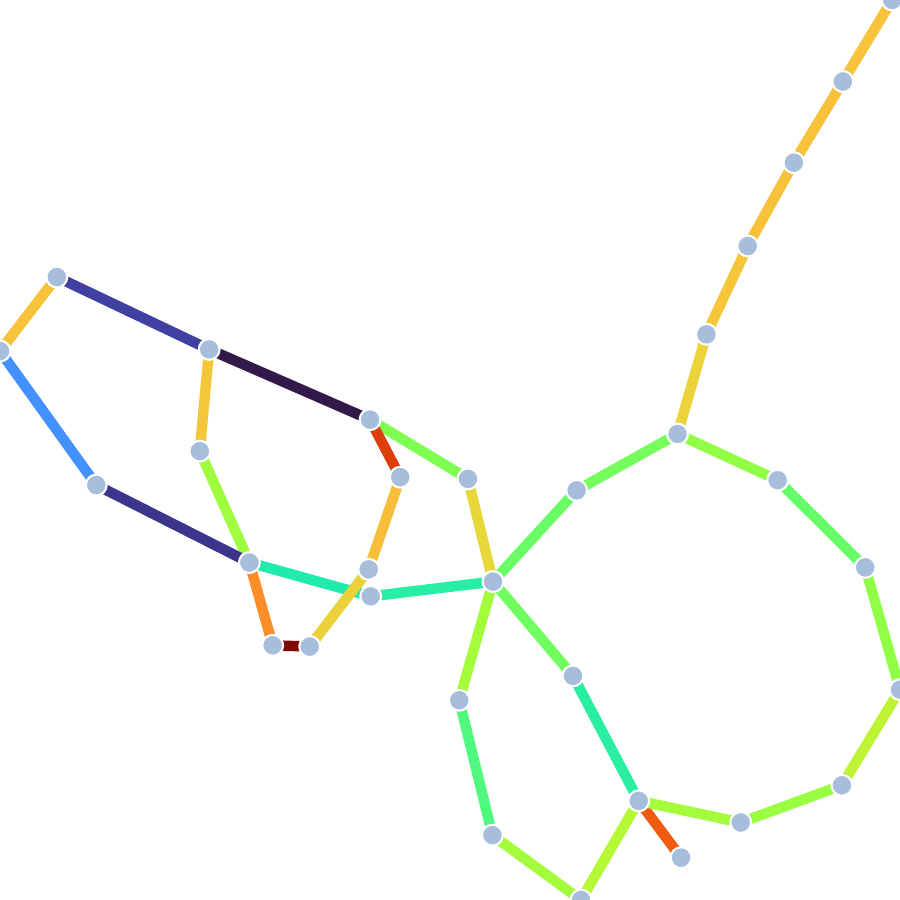} \\ \vspace{-0.0cm} \fontsize{7pt}{0pt}\selectfont 3,\textbf{1}} \\
grafo11543.34 & \parbox{1.7cm}{\centering \includegraphics[height=1.5cm]{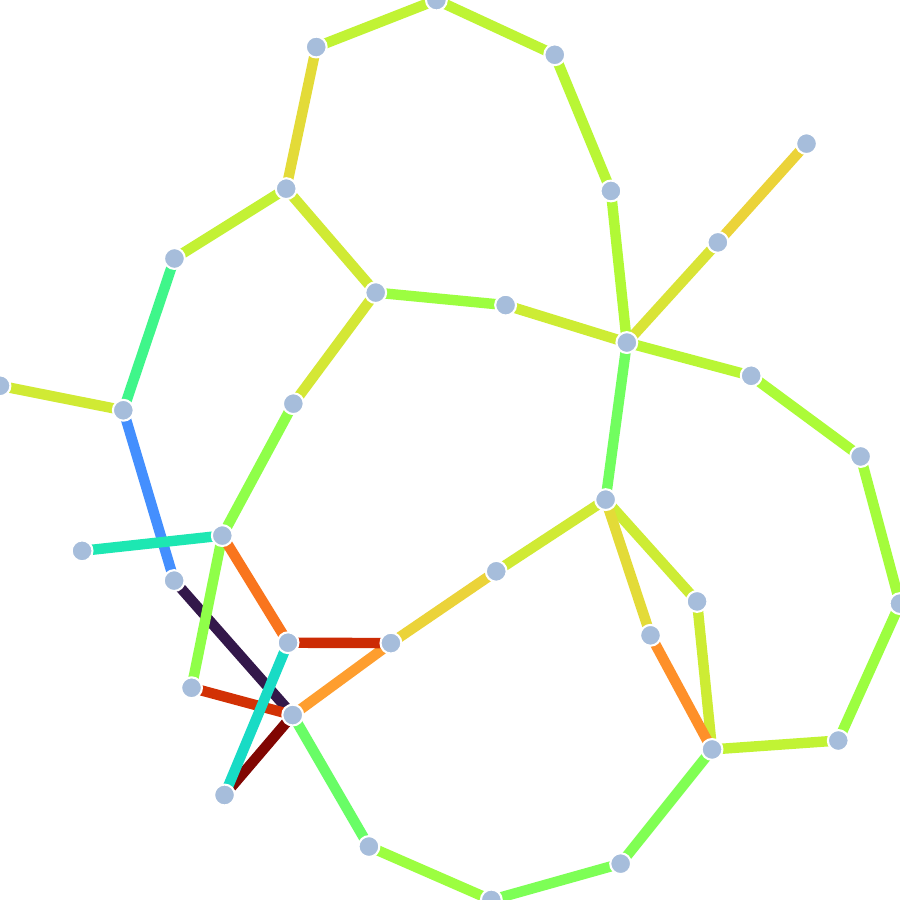} \\ \vspace{-0.0cm} \fontsize{7pt}{0pt}\selectfont } & \parbox{1.7cm}{\centering \includegraphics[height=1.5cm]{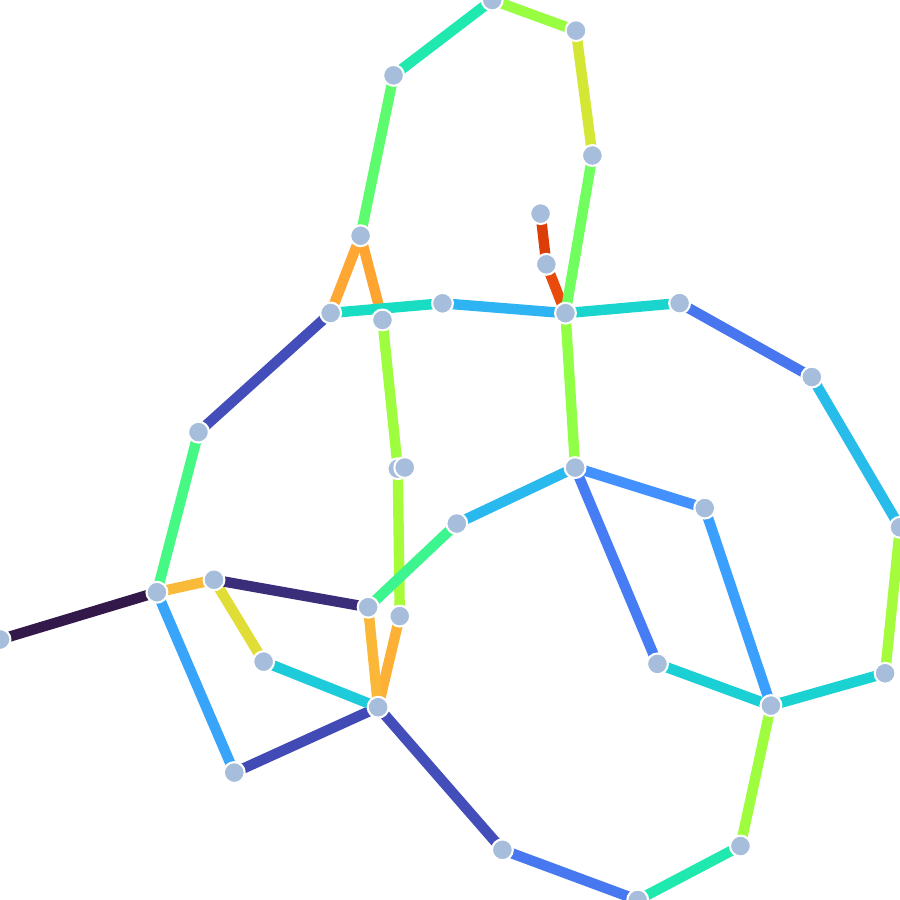} \\ \vspace{-0.0cm} \fontsize{7pt}{0pt}\selectfont 0.872,\textbf{0.753}} & \parbox{1.7cm}{\centering \includegraphics[height=1.5cm]{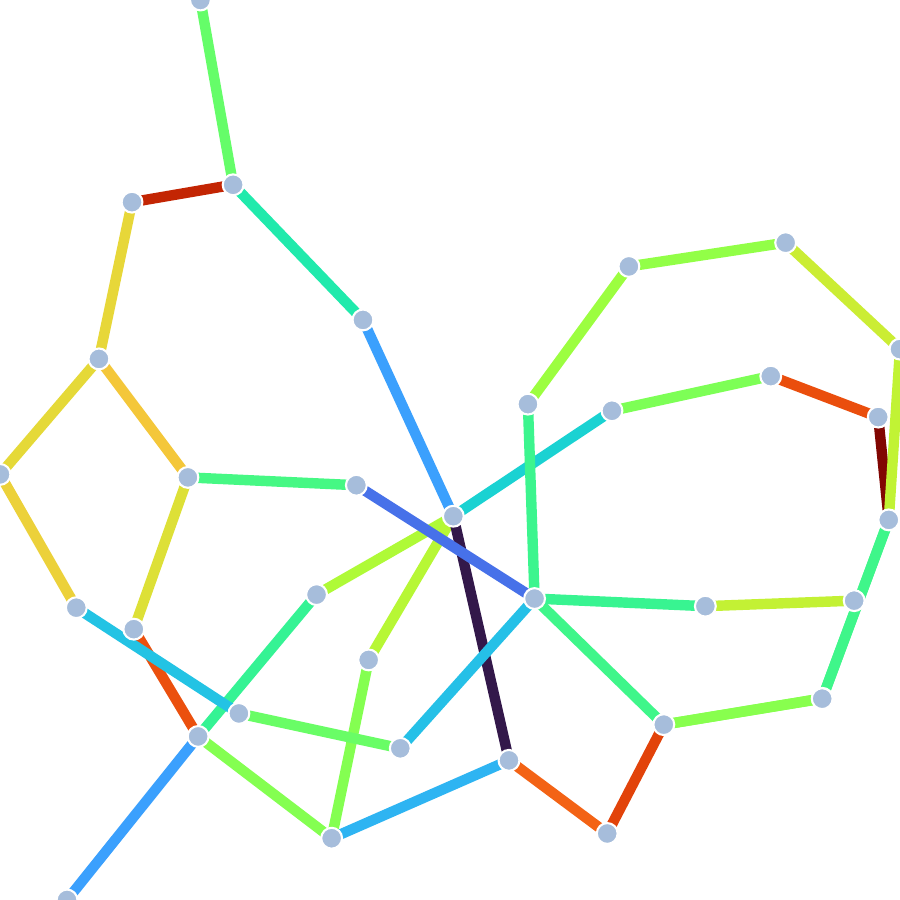} \\ \vspace{-0.0cm} \fontsize{7pt}{0pt}\selectfont \textbf{0.092},0.174} & \parbox{1.7cm}{\centering \includegraphics[height=1.5cm]{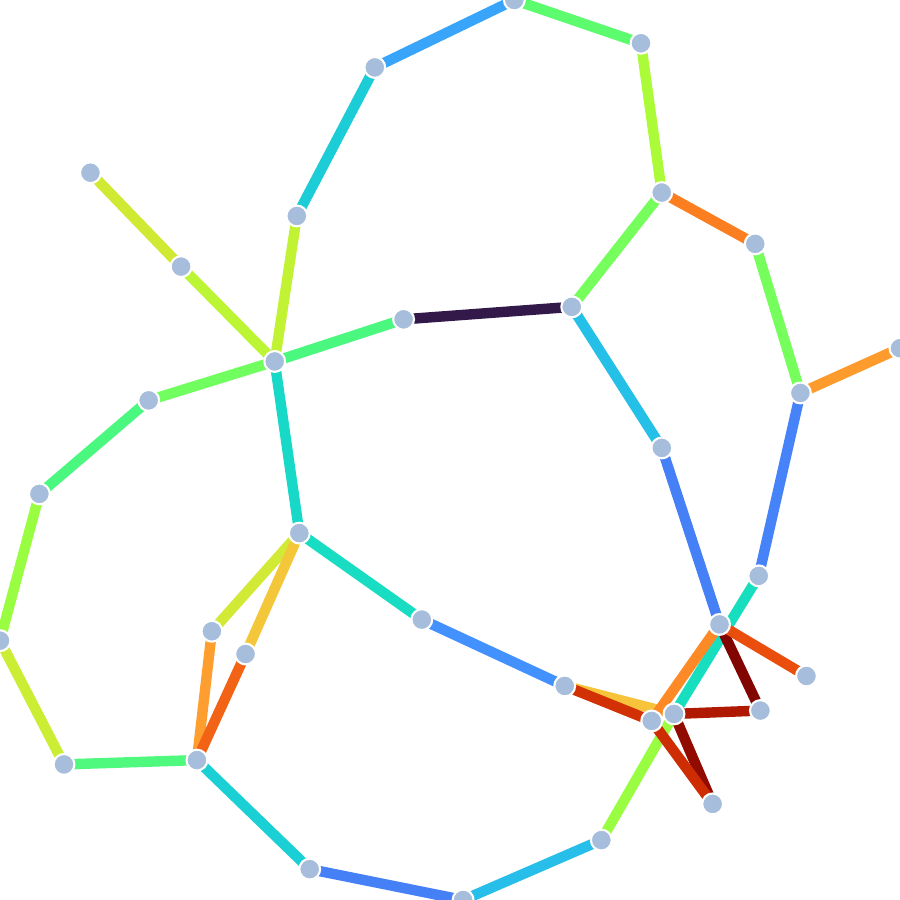} \\ \vspace{-0.0cm} \fontsize{7pt}{0pt}\selectfont -1.829,\textbf{-1.839}} & \parbox{1.7cm}{\centering \includegraphics[height=1.5cm]{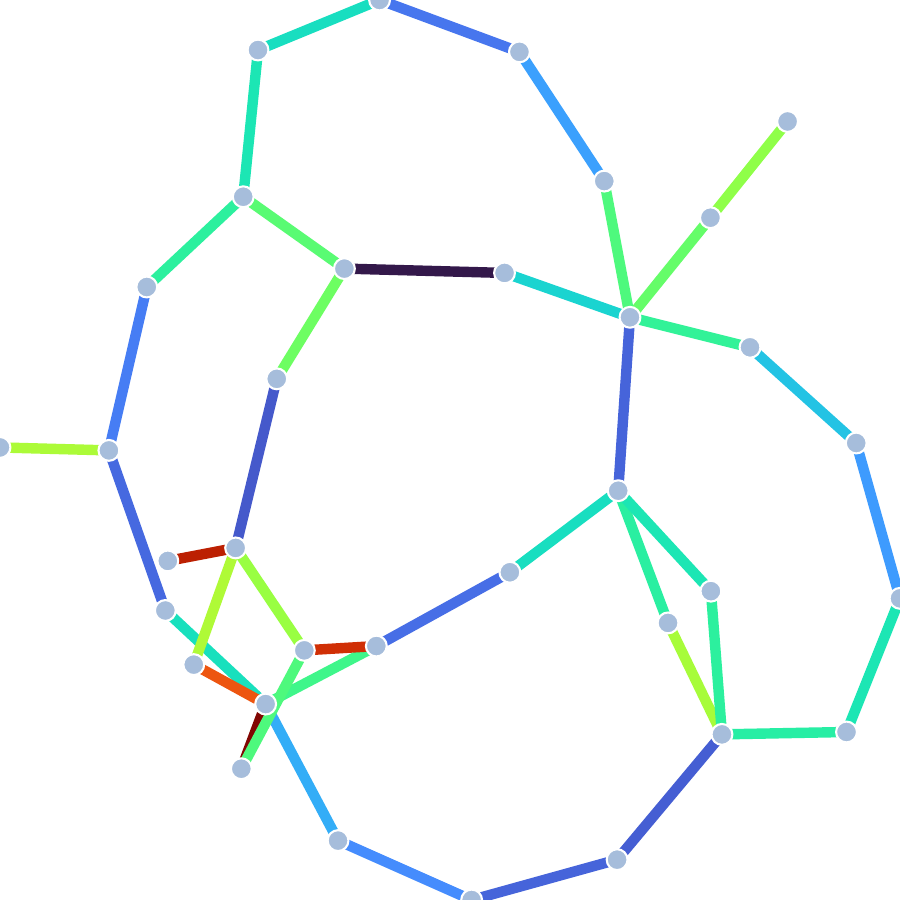} \\ \vspace{-0.0cm} \fontsize{7pt}{0pt}\selectfont \textbf{0.036},0.042} & \parbox{1.7cm}{\centering \includegraphics[height=1.5cm]{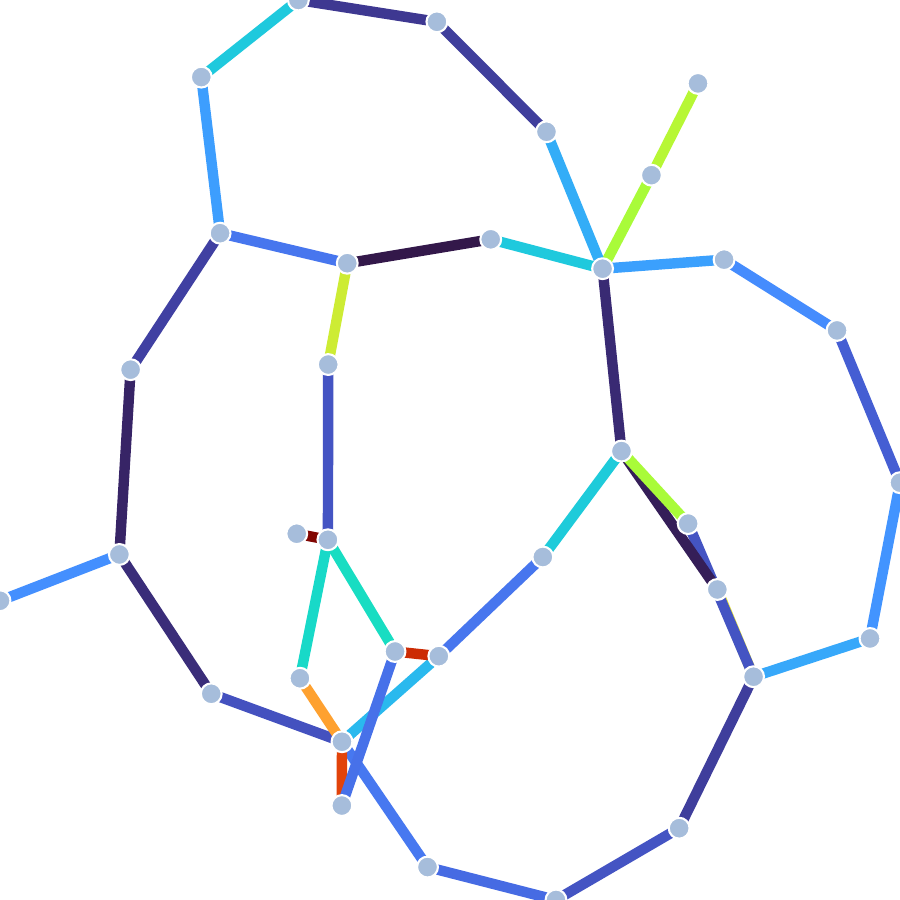} \\ \vspace{-0.0cm} \fontsize{7pt}{0pt}\selectfont 0.625,\textbf{0.6}} & \parbox{1.7cm}{\centering \includegraphics[height=1.5cm]{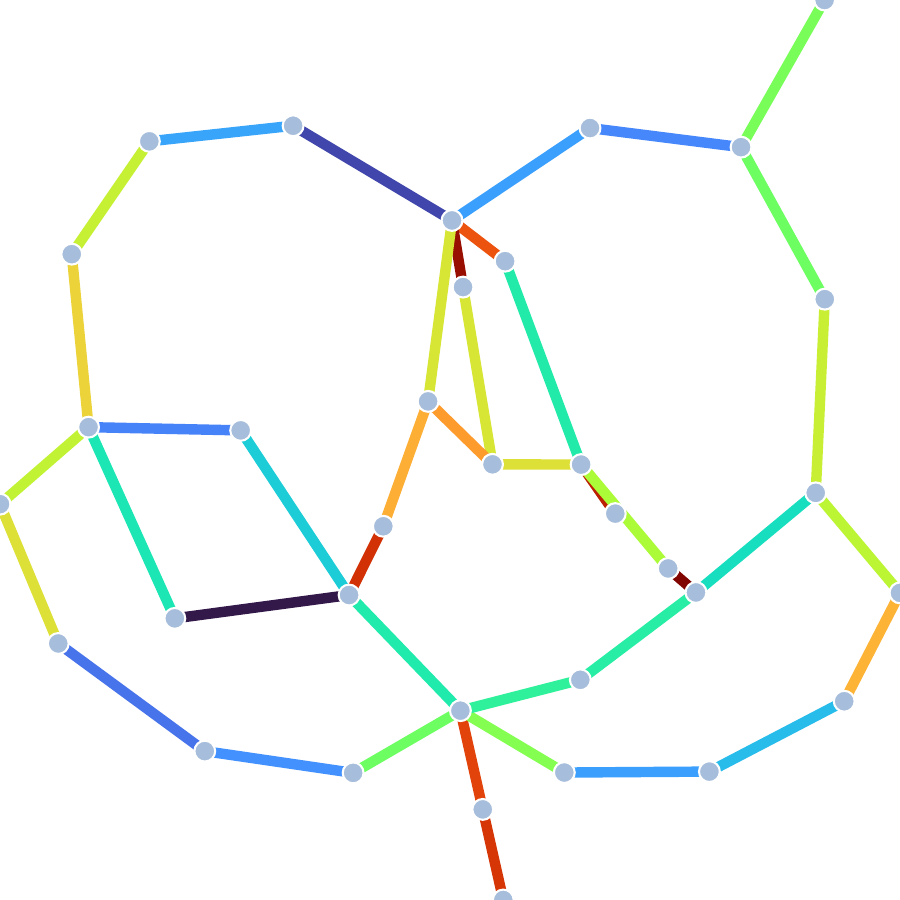} \\ \vspace{-0.0cm} \fontsize{7pt}{0pt}\selectfont 4,\textbf{0}} \\
grafo11674.33 & \parbox{1.7cm}{\centering \includegraphics[height=1.5cm]{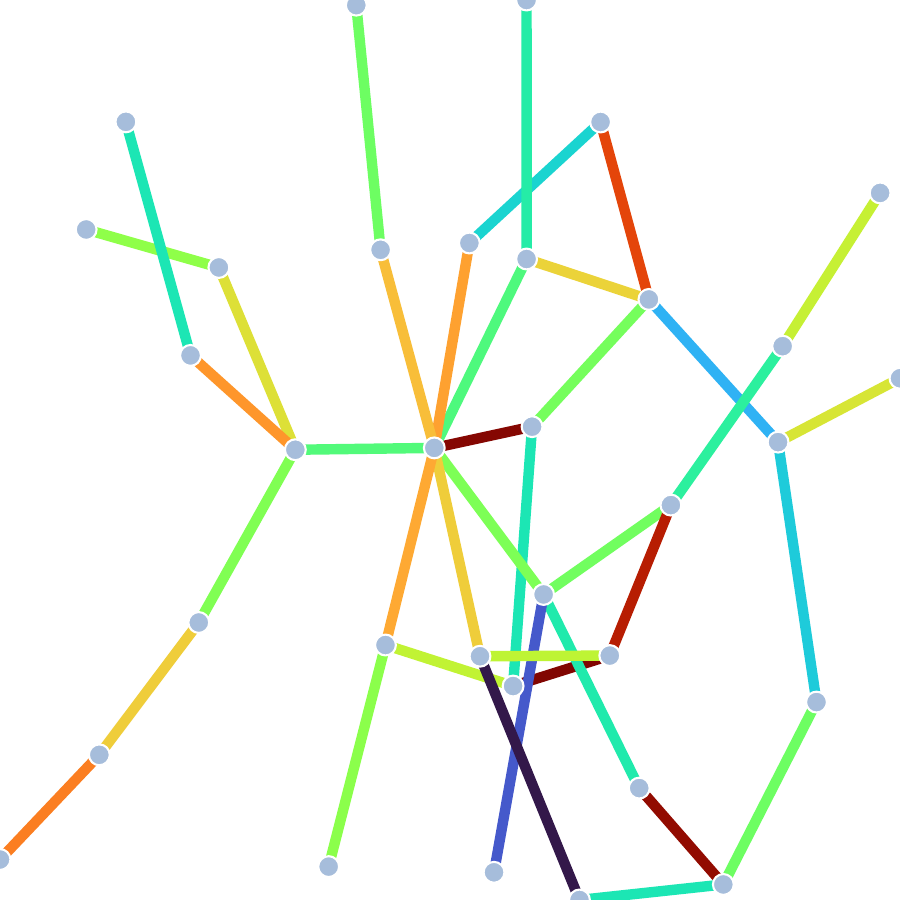} \\ \vspace{-0.0cm} \fontsize{7pt}{0pt}\selectfont } & \parbox{1.7cm}{\centering \includegraphics[height=1.5cm]{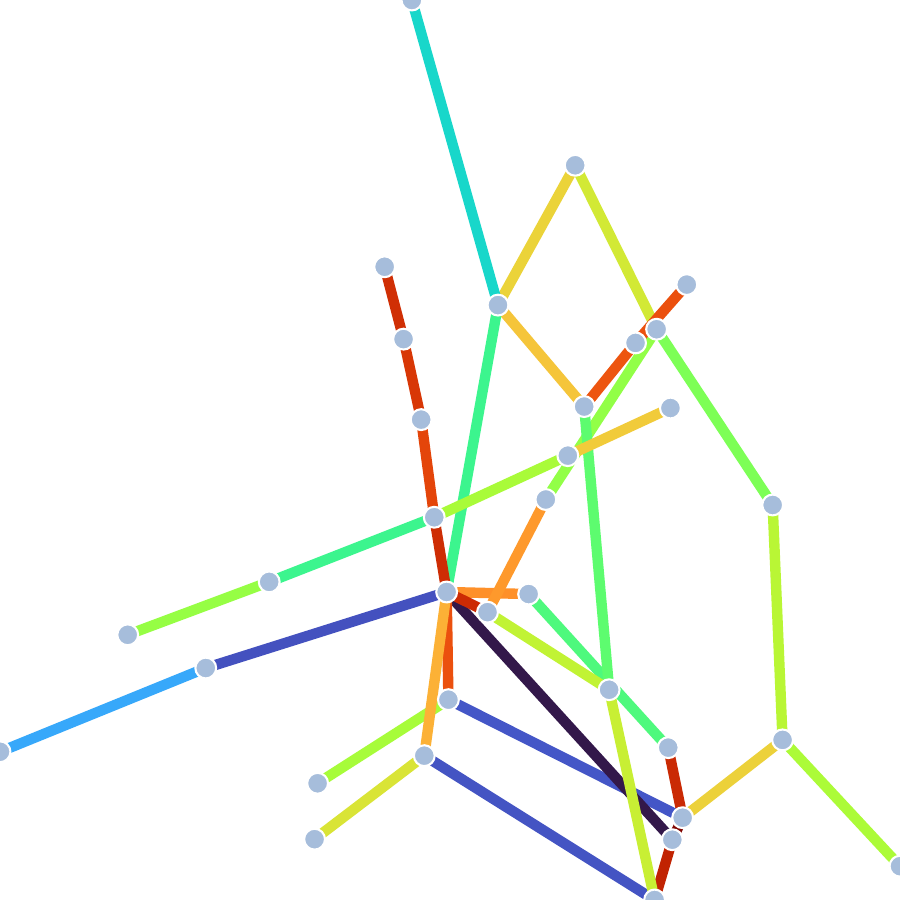} \\ \vspace{-0.0cm} \fontsize{7pt}{0pt}\selectfont 0.924,\textbf{0.615}} & \parbox{1.7cm}{\centering \includegraphics[height=1.5cm]{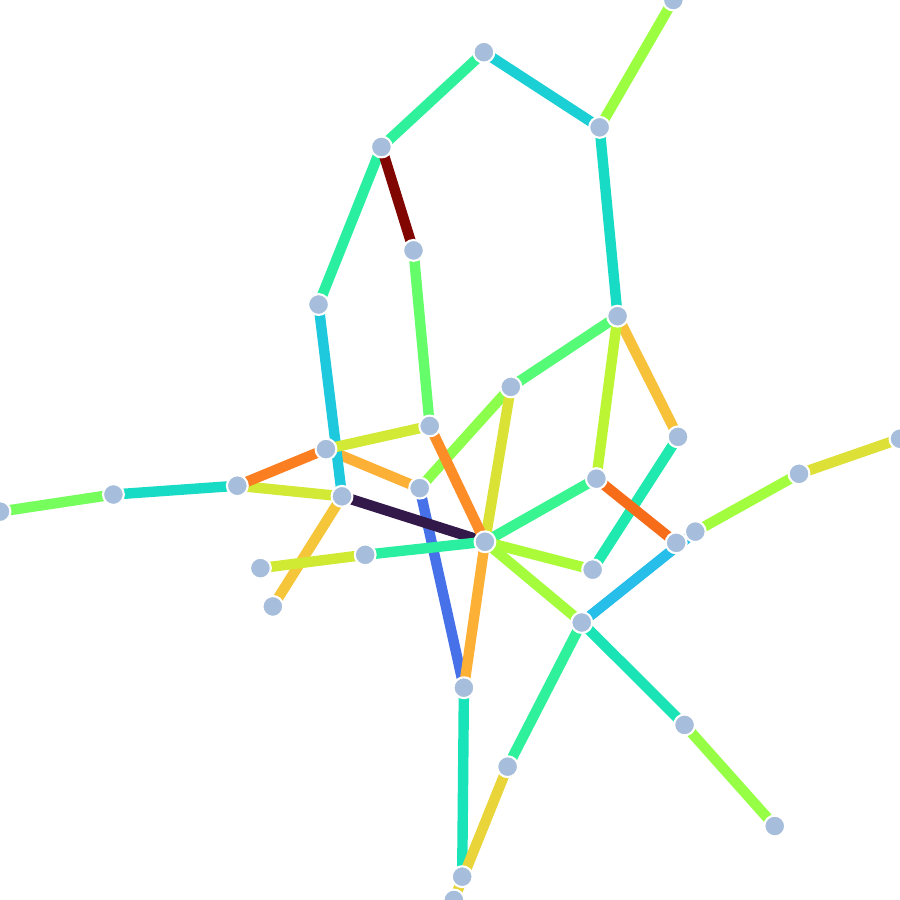} \\ \vspace{-0.0cm} \fontsize{7pt}{0pt}\selectfont \textbf{0.121},0.122} & \parbox{1.7cm}{\centering \includegraphics[height=1.5cm]{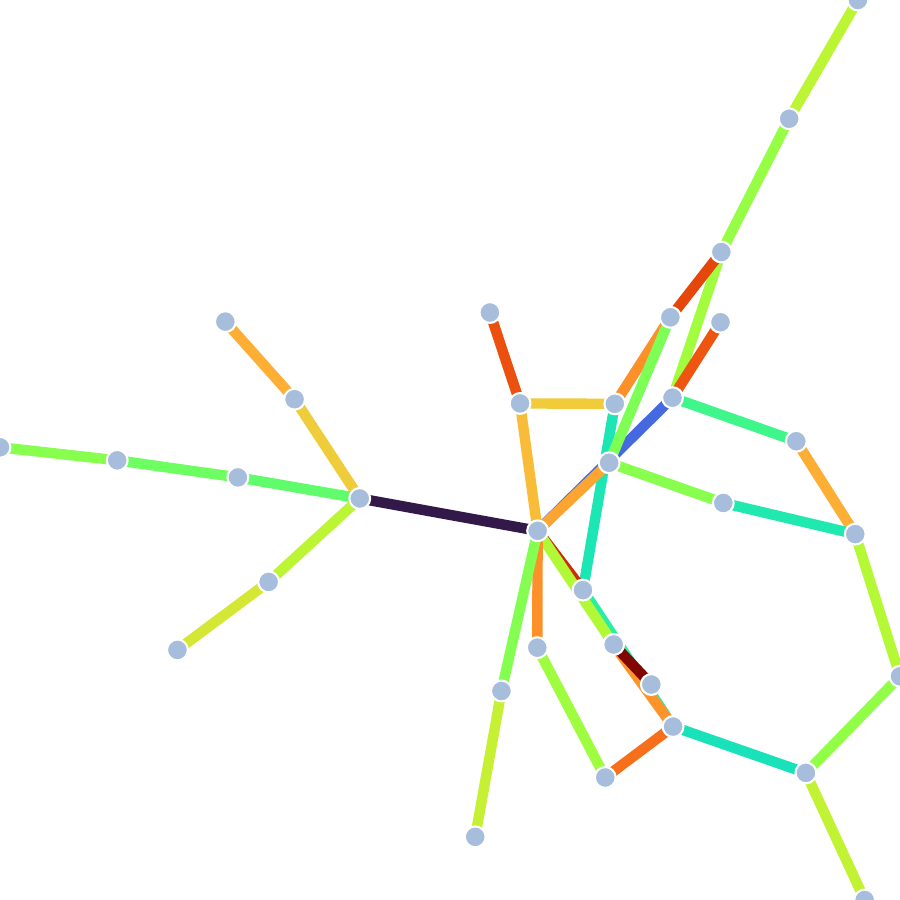} \\ \vspace{-0.0cm} \fontsize{7pt}{0pt}\selectfont -1.611,\textbf{-1.713}} & \parbox{1.7cm}{\centering \includegraphics[height=1.5cm]{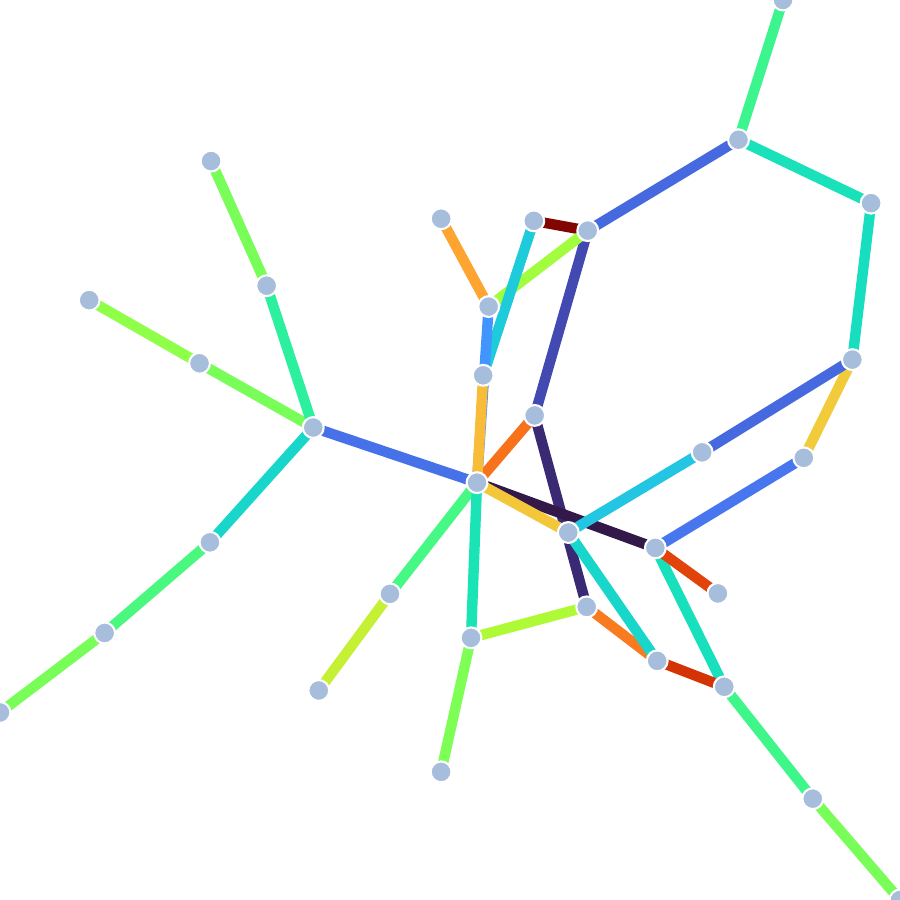} \\ \vspace{-0.0cm} \fontsize{7pt}{0pt}\selectfont 0.074,\textbf{0.061}} & \parbox{1.7cm}{\centering \includegraphics[height=1.5cm]{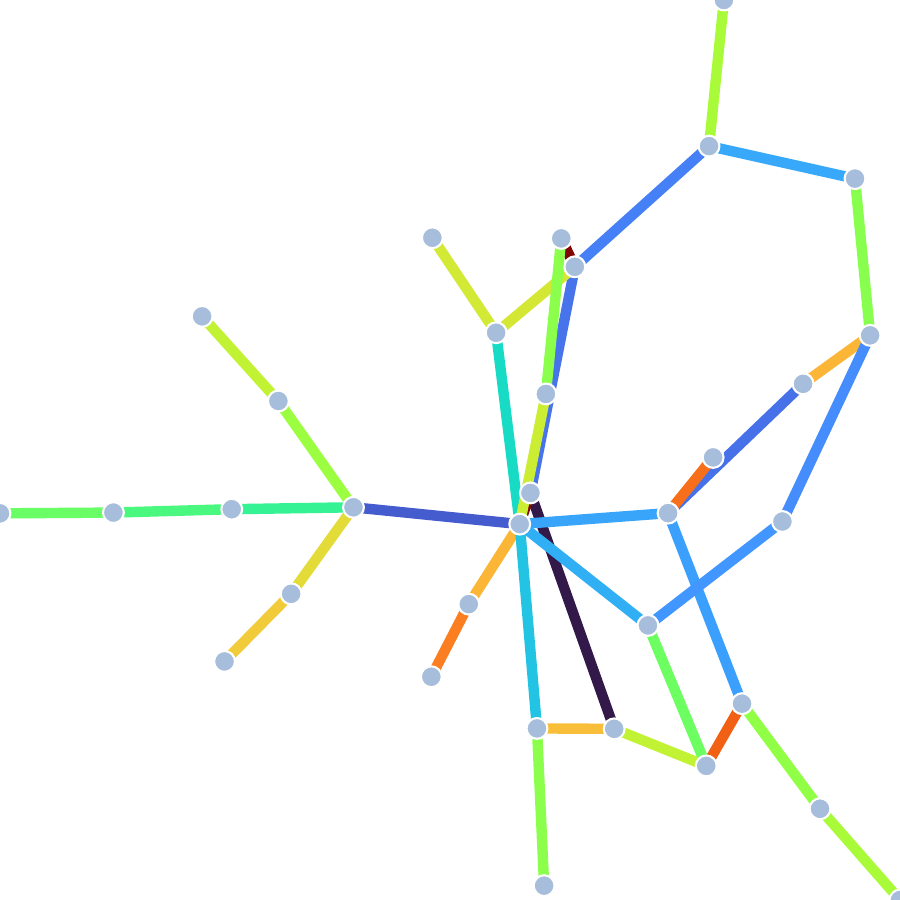} \\ \vspace{-0.0cm} \fontsize{7pt}{0pt}\selectfont 0.893,\textbf{0.66}} & \parbox{1.7cm}{\centering \includegraphics[height=1.5cm]{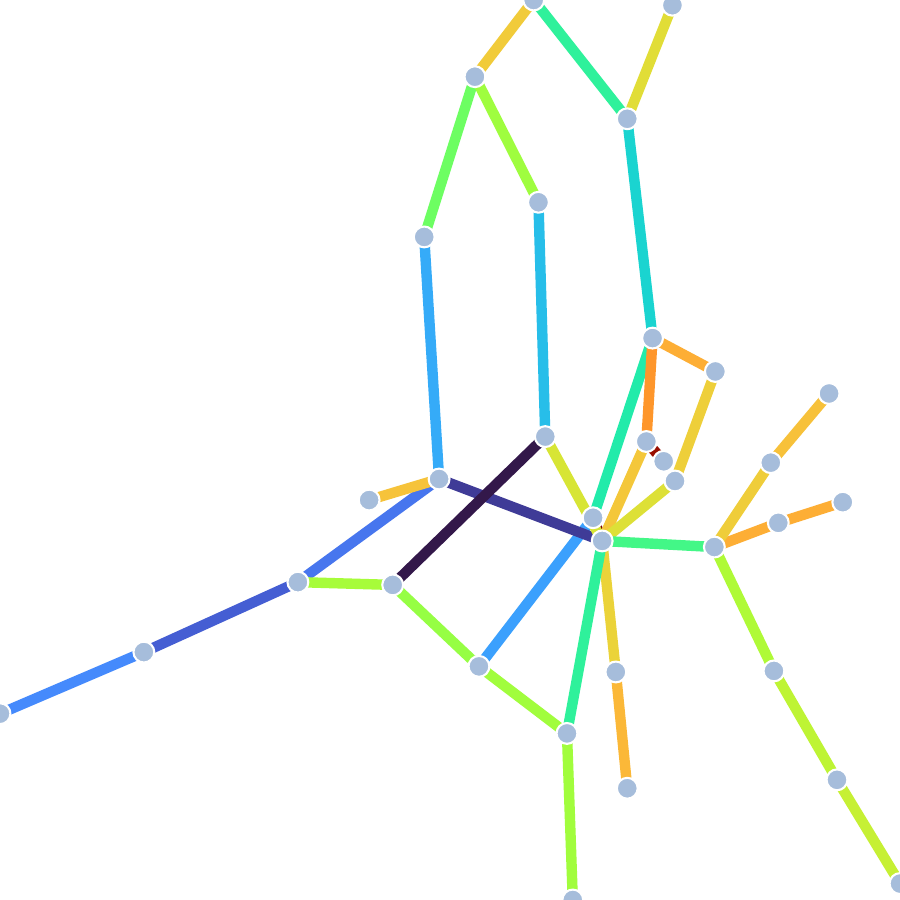} \\ \vspace{-0.0cm} \fontsize{7pt}{0pt}\selectfont 11,\textbf{3}} \\
grafo1583.66 & \parbox{1.7cm}{\centering \includegraphics[height=1.5cm]{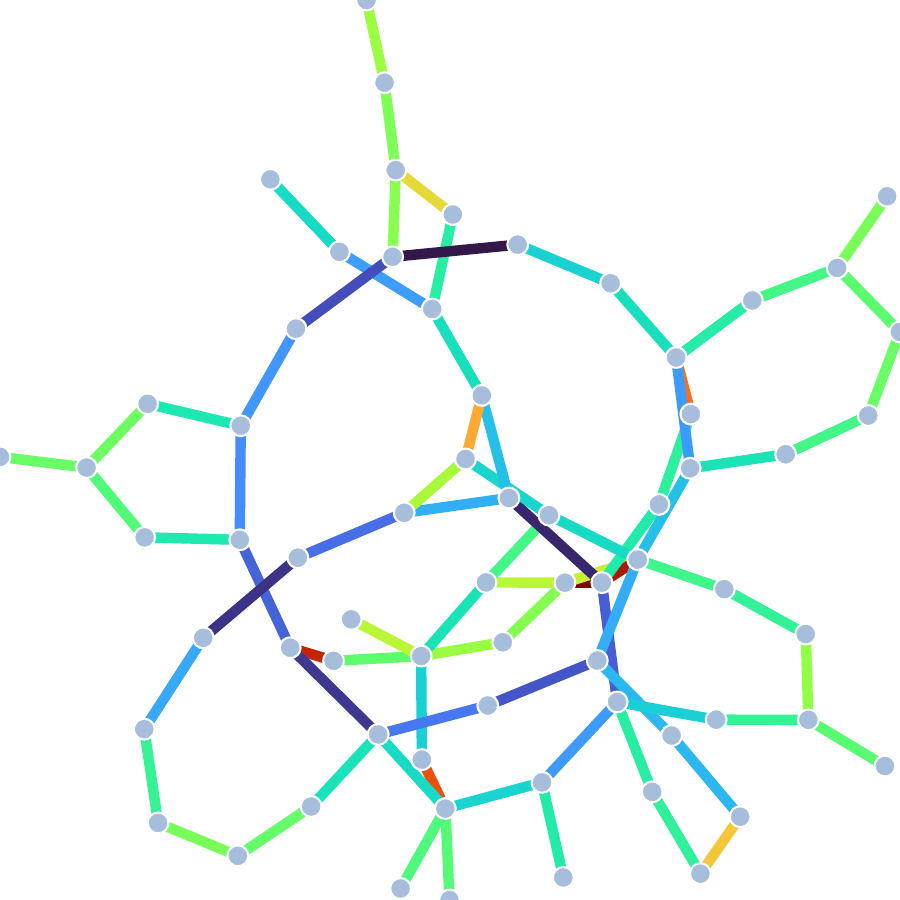} \\ \vspace{-0.0cm} \fontsize{7pt}{0pt}\selectfont } & \parbox{1.7cm}{\centering \includegraphics[height=1.5cm]{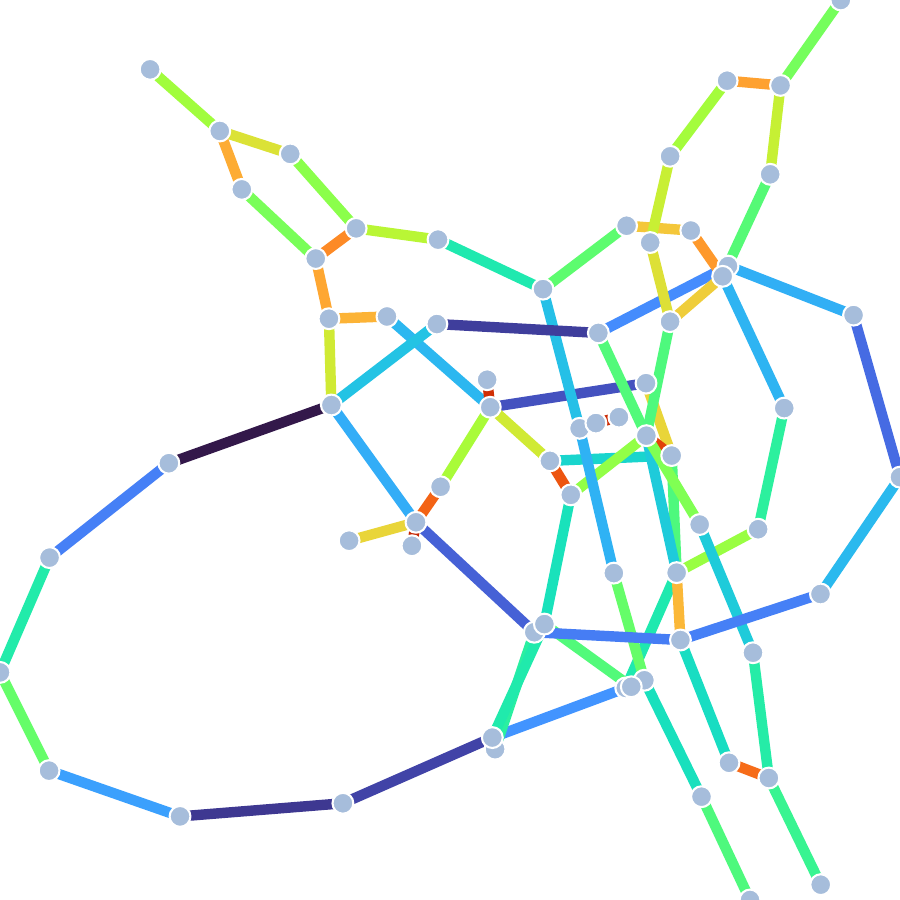} \\ \vspace{-0.0cm} \fontsize{7pt}{0pt}\selectfont 0.815,\textbf{0.704}} & \parbox{1.7cm}{\centering \includegraphics[height=1.5cm]{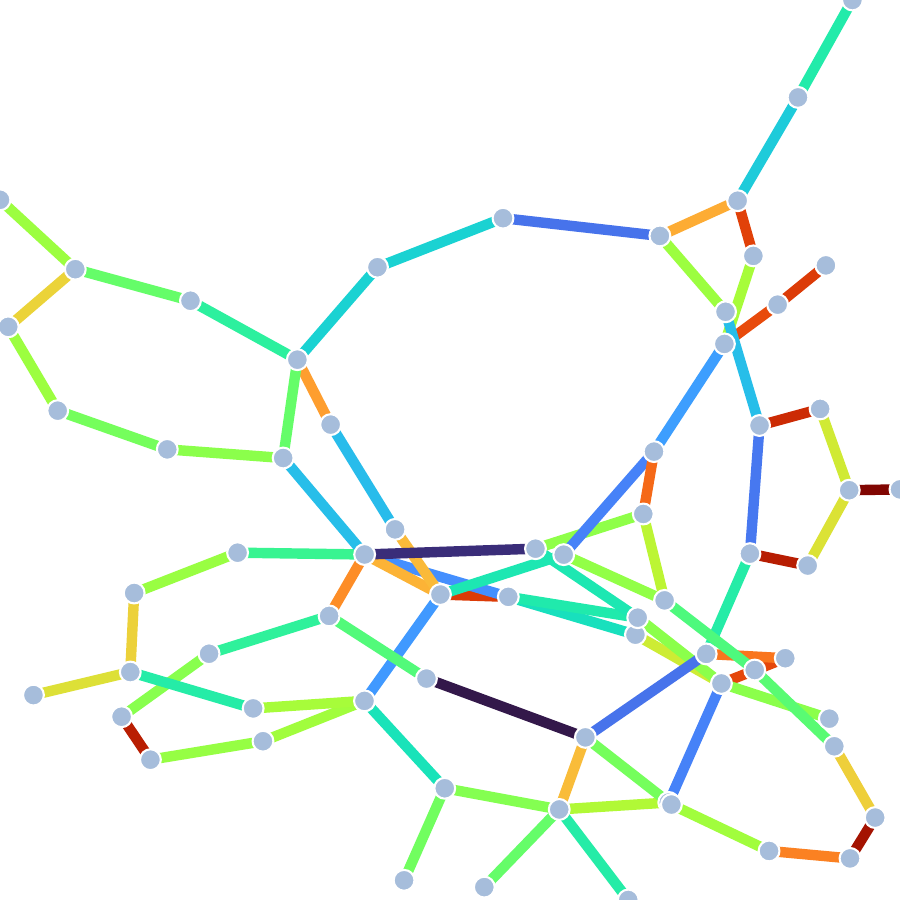} \\ \vspace{-0.0cm} \fontsize{7pt}{0pt}\selectfont \textbf{0.186},0.245} & \parbox{1.7cm}{\centering \includegraphics[height=1.5cm]{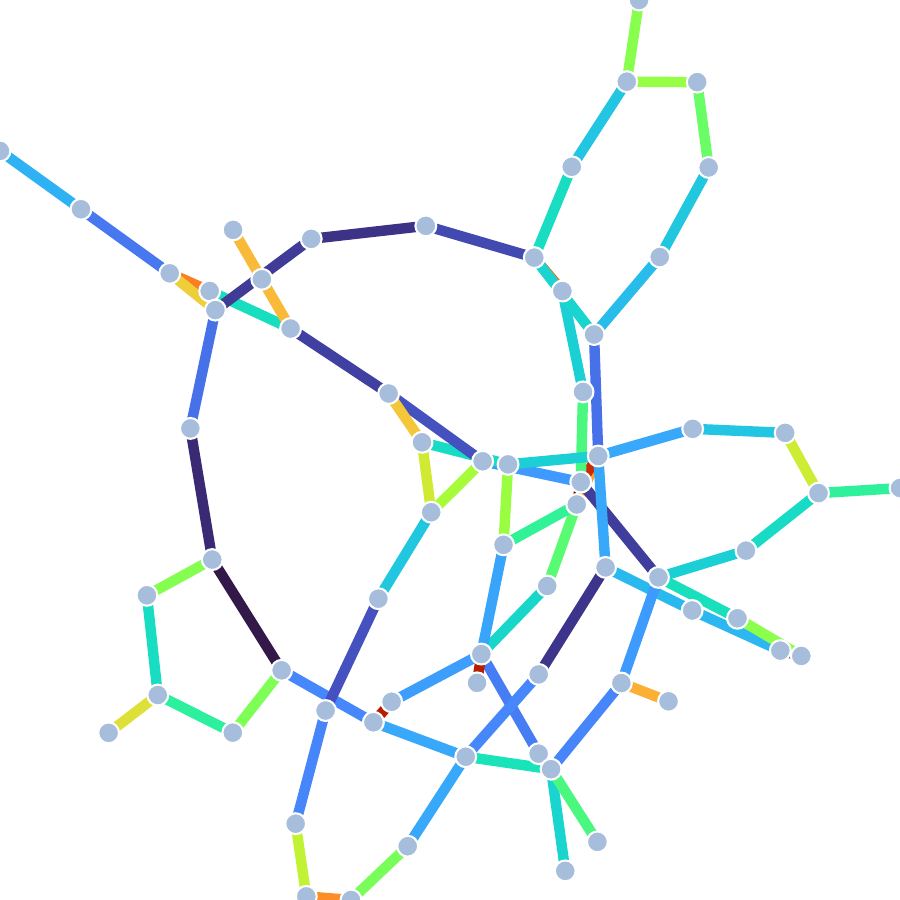} \\ \vspace{-0.0cm} \fontsize{7pt}{0pt}\selectfont \textbf{-2.253},-2.252} & \parbox{1.7cm}{\centering \includegraphics[height=1.5cm]{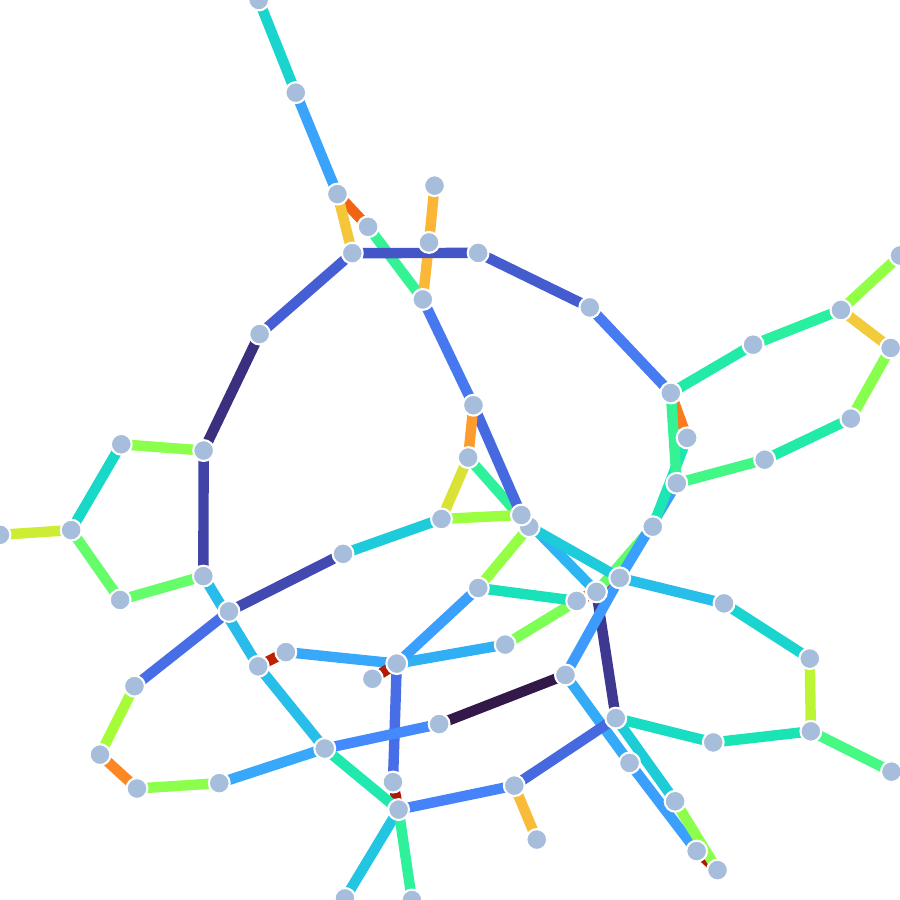} \\ \vspace{-0.0cm} \fontsize{7pt}{0pt}\selectfont \textbf{0.06},0.066} & \parbox{1.7cm}{\centering \includegraphics[height=1.5cm]{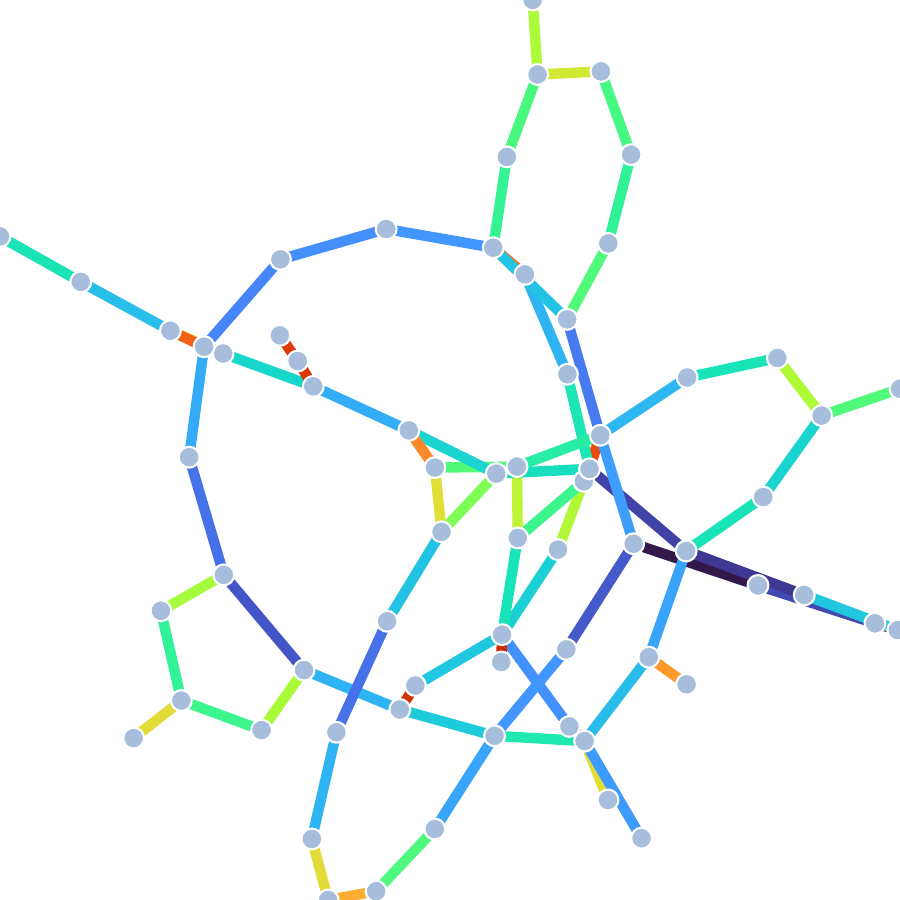} \\ \vspace{-0.0cm} \fontsize{7pt}{0pt}\selectfont 0.964,\textbf{0.927}} & \parbox{1.7cm}{\centering \includegraphics[height=1.5cm]{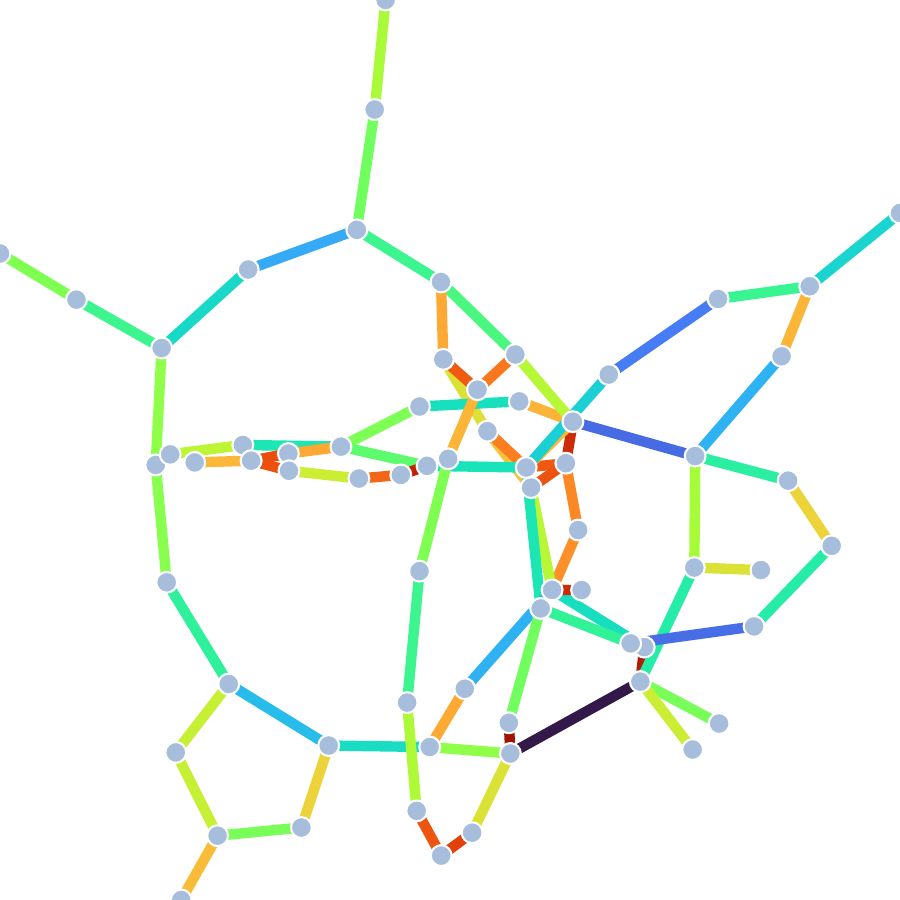} \\ \vspace{-0.0cm} \fontsize{7pt}{0pt}\selectfont 13,\textbf{12}} \\
grafo2747.16 & \parbox{1.7cm}{\centering \includegraphics[height=1.5cm]{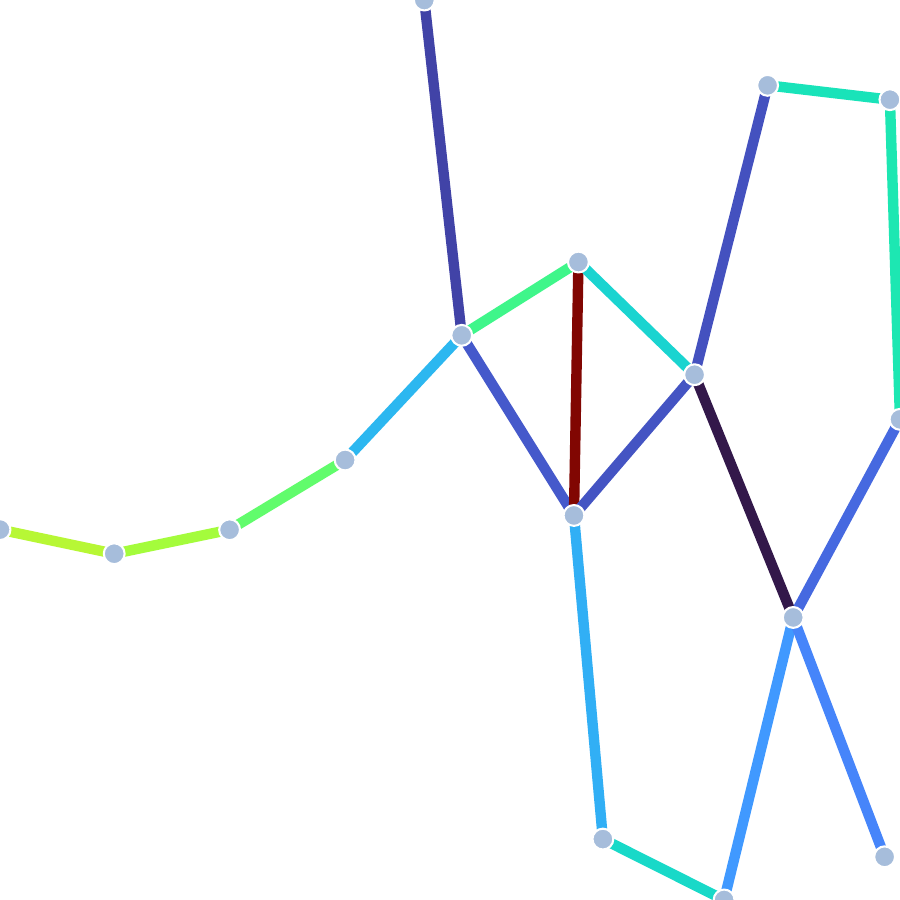} \\ \vspace{-0.0cm} \fontsize{7pt}{0pt}\selectfont } & \parbox{1.7cm}{\centering \includegraphics[height=1.5cm]{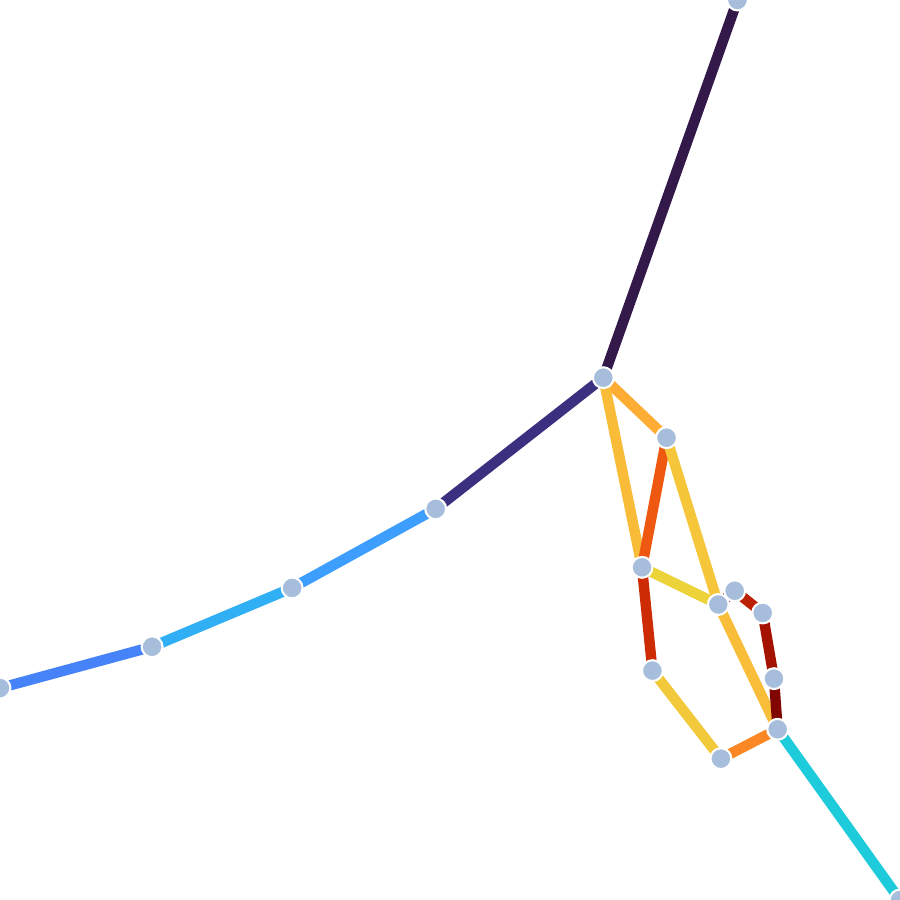} \\ \vspace{-0.0cm} \fontsize{7pt}{0pt}\selectfont 0.845,\textbf{0.635}} & \parbox{1.7cm}{\centering \includegraphics[height=1.5cm]{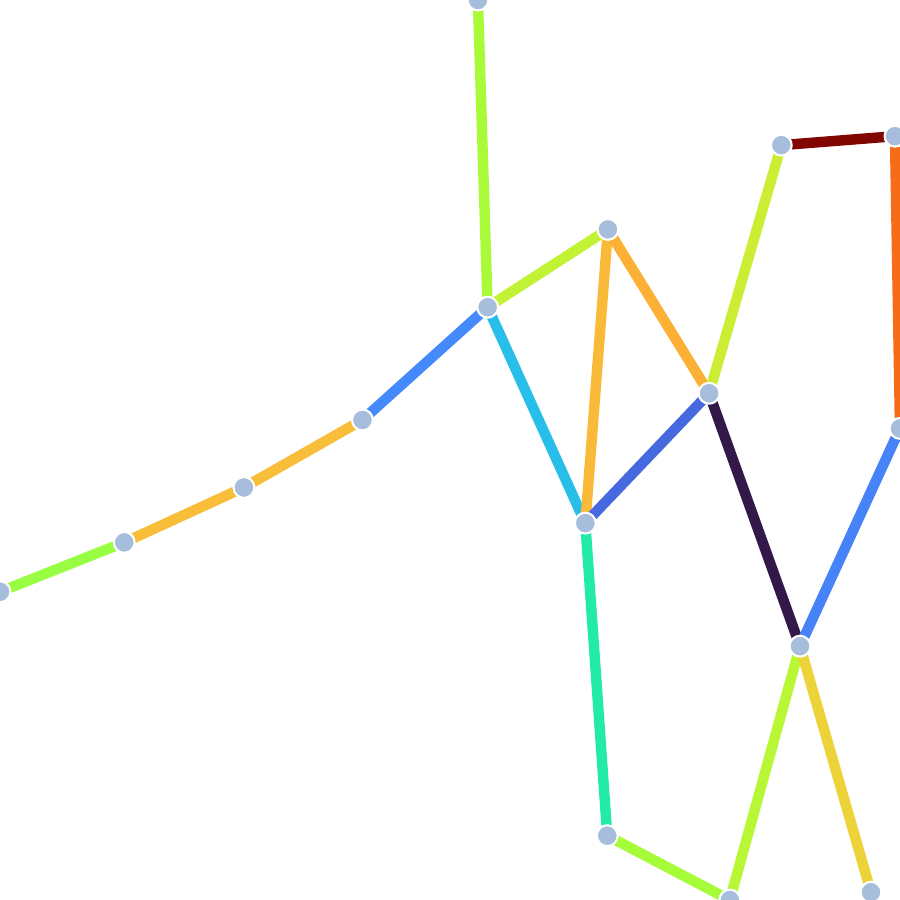} \\ \vspace{-0.0cm} \fontsize{7pt}{0pt}\selectfont 0.07,\textbf{0.046}} & \parbox{1.7cm}{\centering \includegraphics[height=1.5cm]{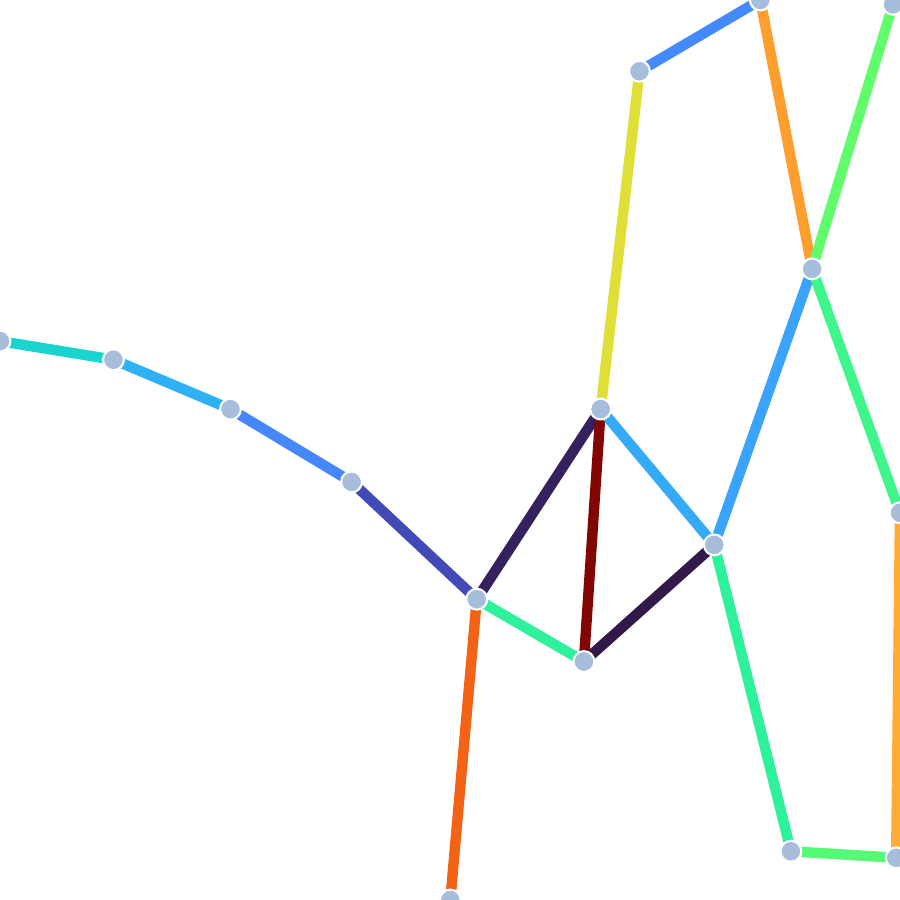} \\ \vspace{-0.0cm} \fontsize{7pt}{0pt}\selectfont -1.438,\textbf{-1.472}} & \parbox{1.7cm}{\centering \includegraphics[height=1.5cm]{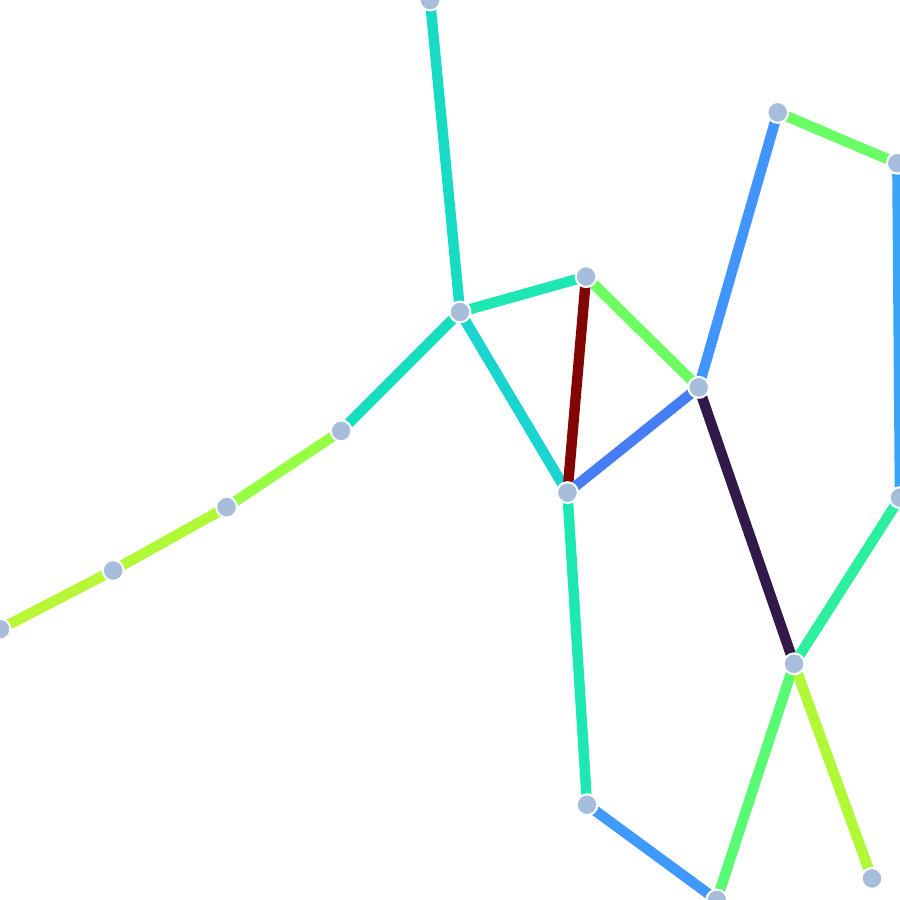} \\ \vspace{-0.0cm} \fontsize{7pt}{0pt}\selectfont \textbf{0.01},0.01} & \parbox{1.7cm}{\centering \includegraphics[height=1.5cm]{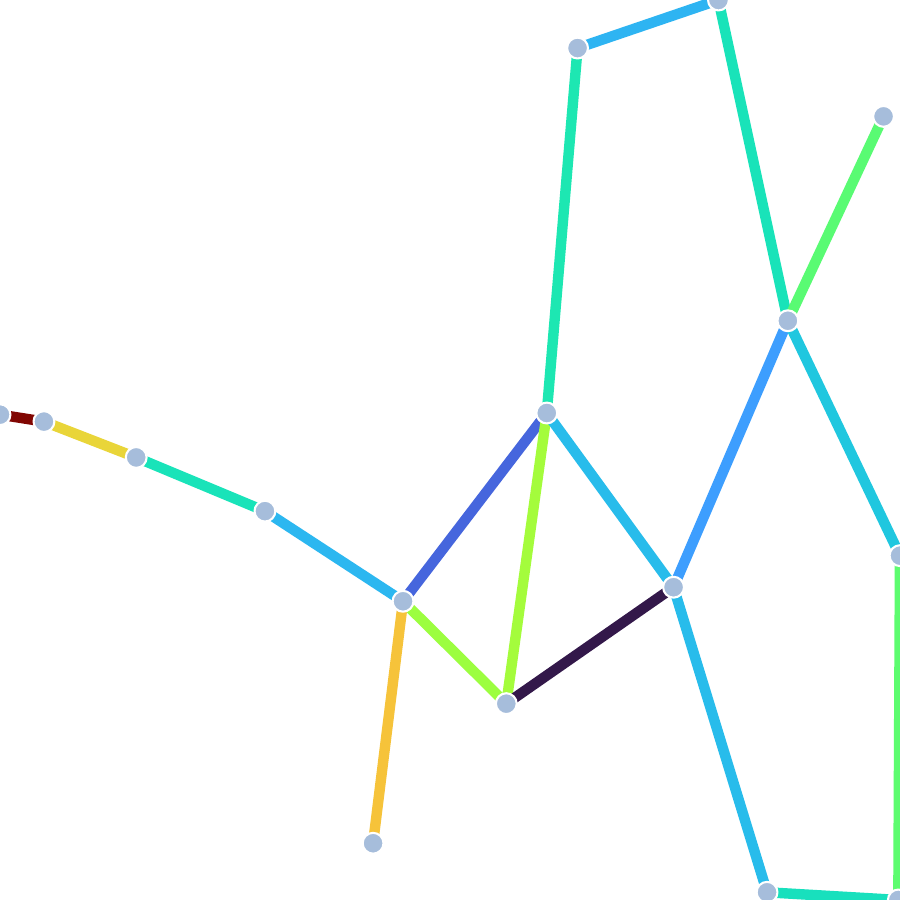} \\ \vspace{-0.0cm} \fontsize{7pt}{0pt}\selectfont 0.204,\textbf{0.184}} & \parbox{1.7cm}{\centering \includegraphics[height=1.5cm]{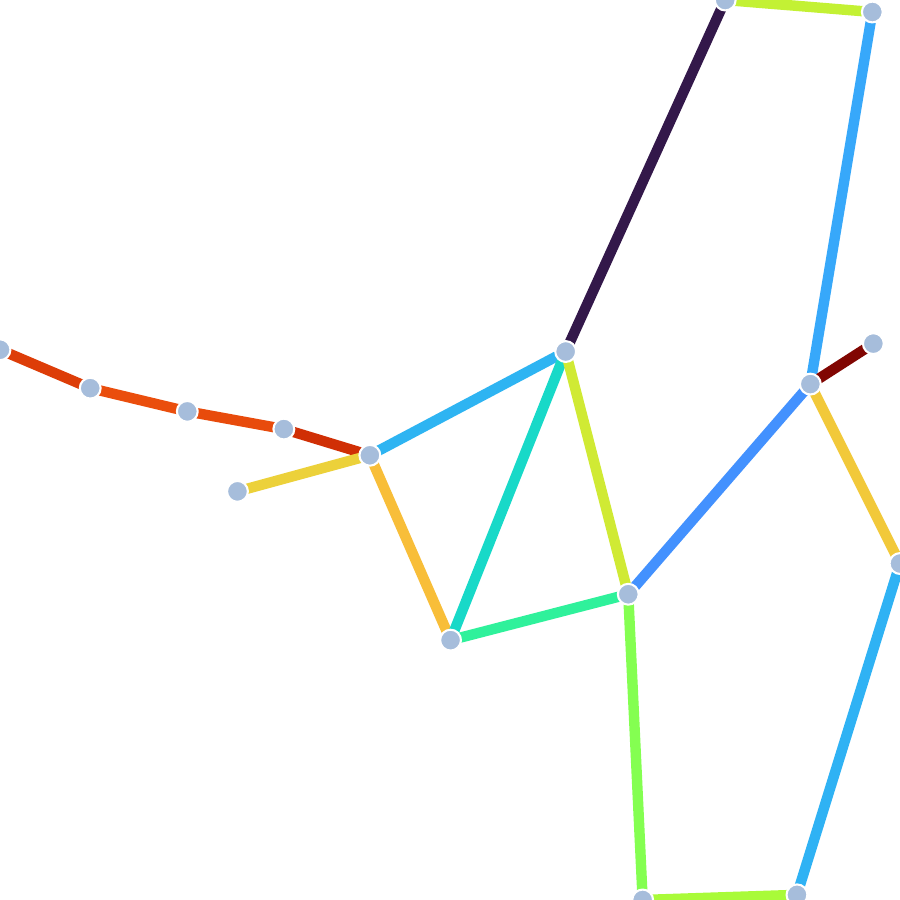} \\ \vspace{-0.0cm} \fontsize{7pt}{0pt}\selectfont 0,0} \\
grafo2815.35 & \parbox{1.7cm}{\centering \includegraphics[height=1.5cm]{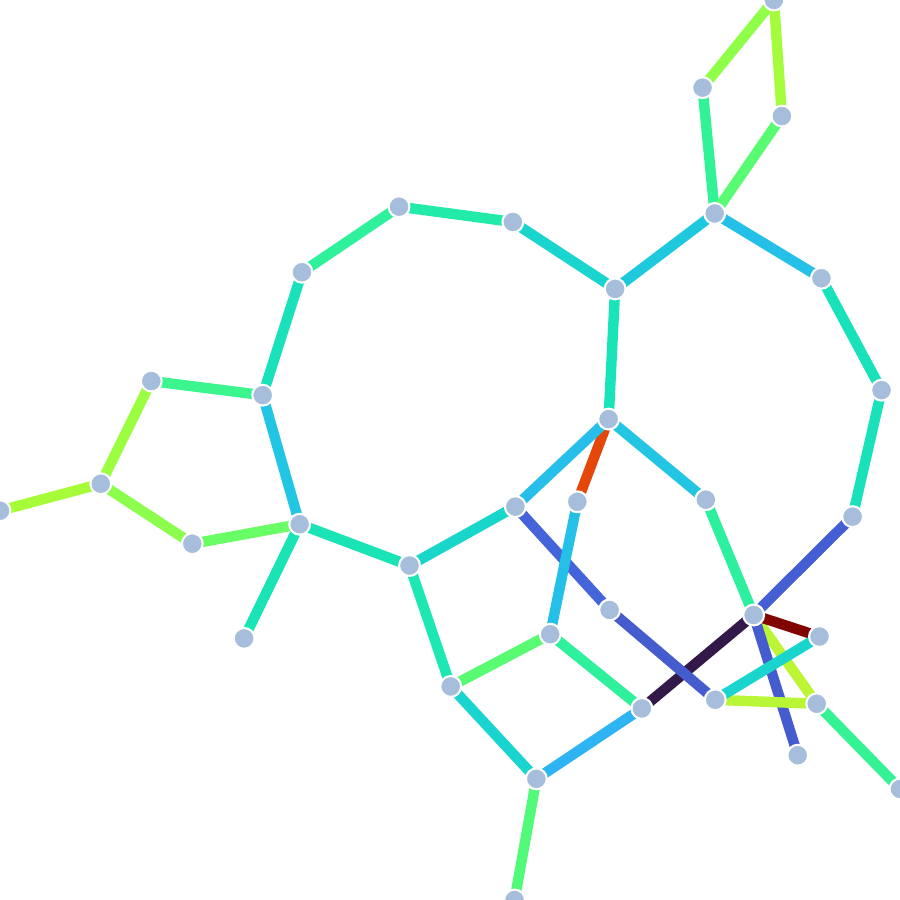} \\ \vspace{-0.0cm} \fontsize{7pt}{0pt}\selectfont } & \parbox{1.7cm}{\centering \includegraphics[height=1.5cm]{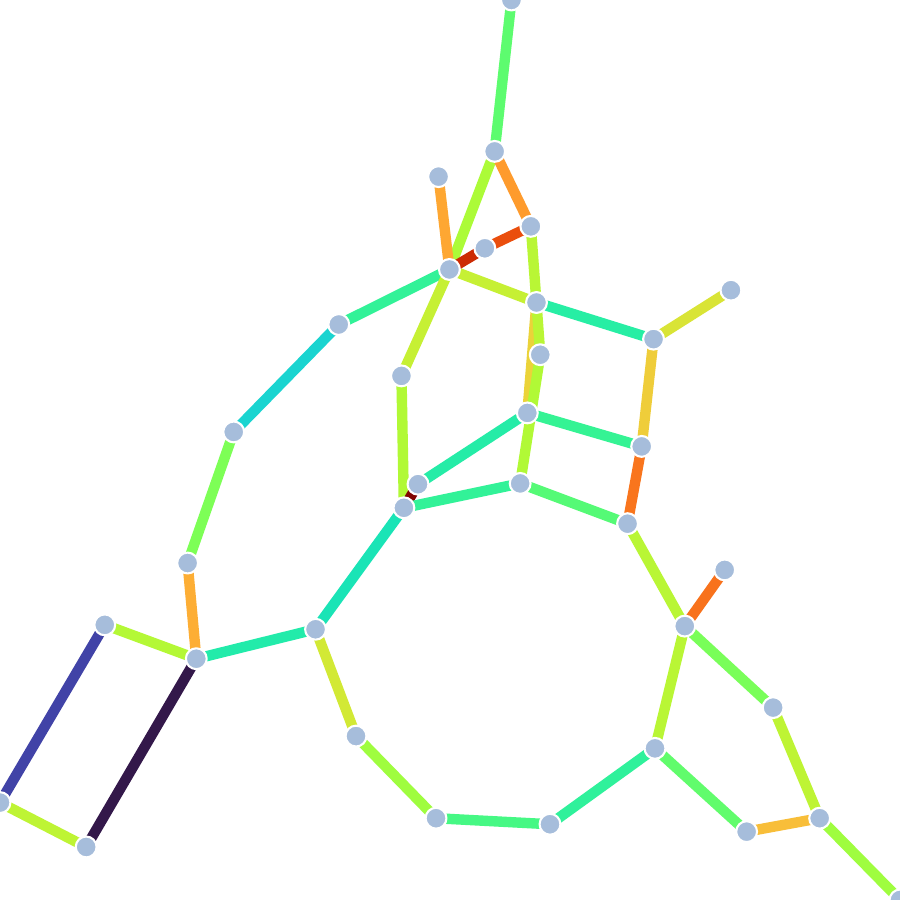} \\ \vspace{-0.0cm} \fontsize{7pt}{0pt}\selectfont 0.934,\textbf{0.7}} & \parbox{1.7cm}{\centering \includegraphics[height=1.5cm]{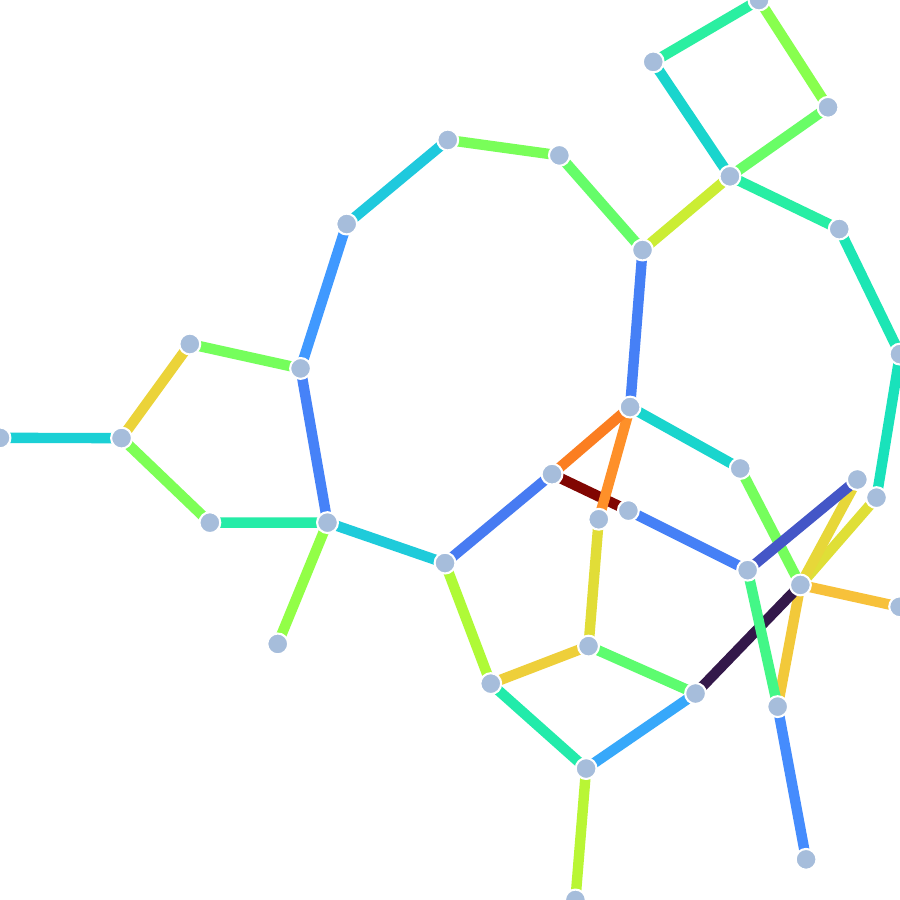} \\ \vspace{-0.0cm} \fontsize{7pt}{0pt}\selectfont 0.108,\textbf{0.097}} & \parbox{1.7cm}{\centering \includegraphics[height=1.5cm]{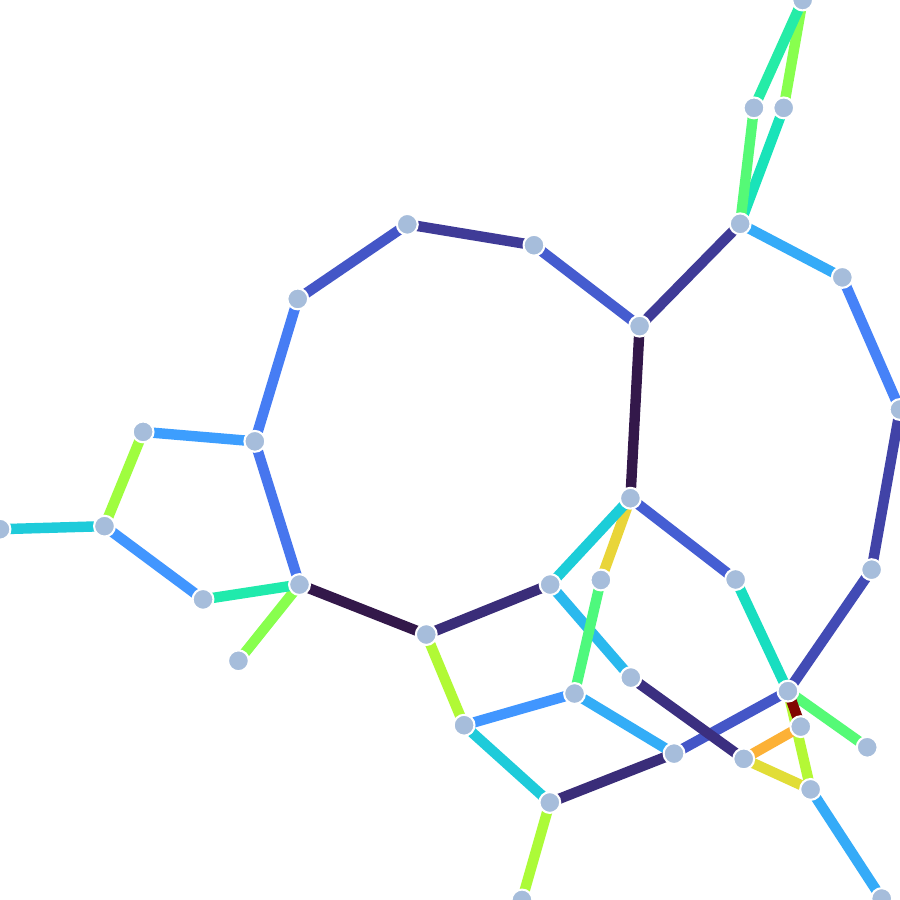} \\ \vspace{-0.0cm} \fontsize{7pt}{0pt}\selectfont -1.872,\textbf{-1.896}} & \parbox{1.7cm}{\centering \includegraphics[height=1.5cm]{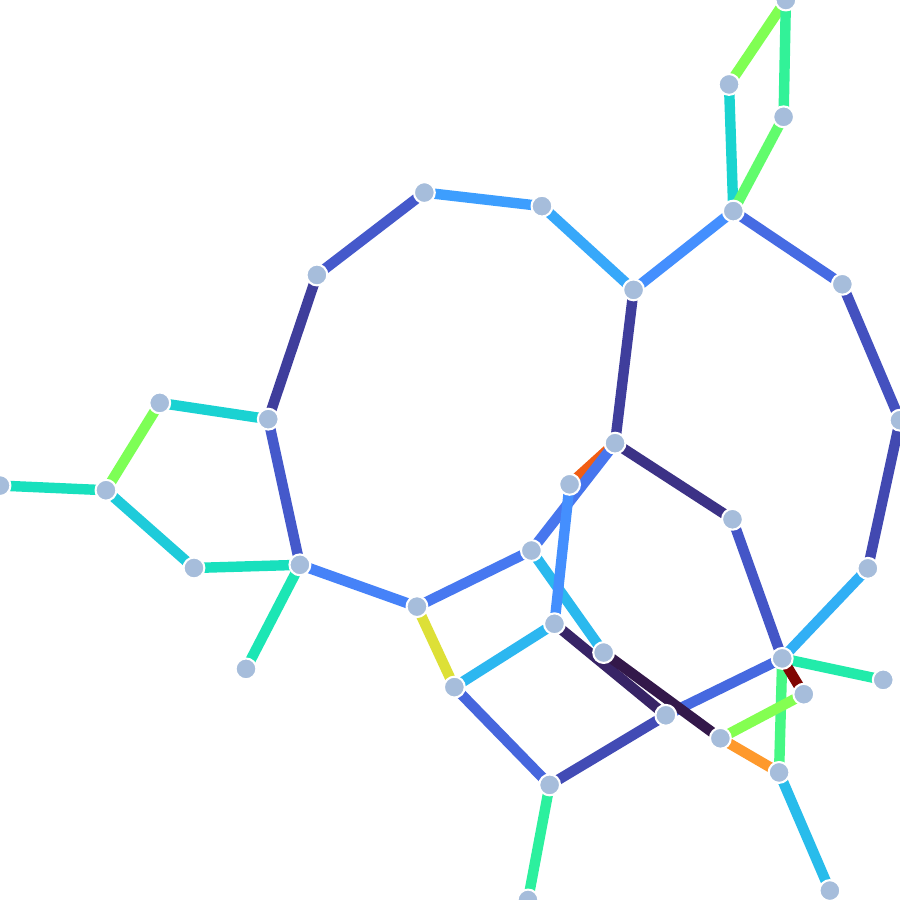} \\ \vspace{-0.0cm} \fontsize{7pt}{0pt}\selectfont \textbf{0.032},0.033} & \parbox{1.7cm}{\centering \includegraphics[height=1.5cm]{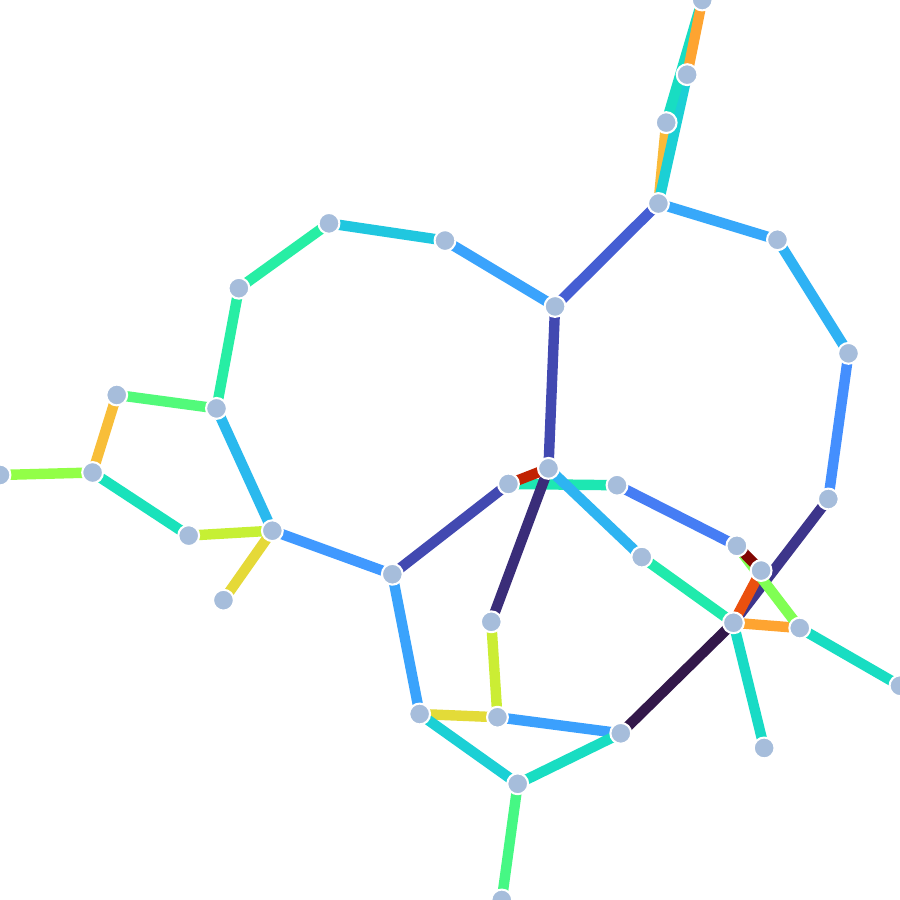} \\ \vspace{-0.0cm} \fontsize{7pt}{0pt}\selectfont 0.603,\textbf{0.582}} & \parbox{1.7cm}{\centering \includegraphics[height=1.5cm]{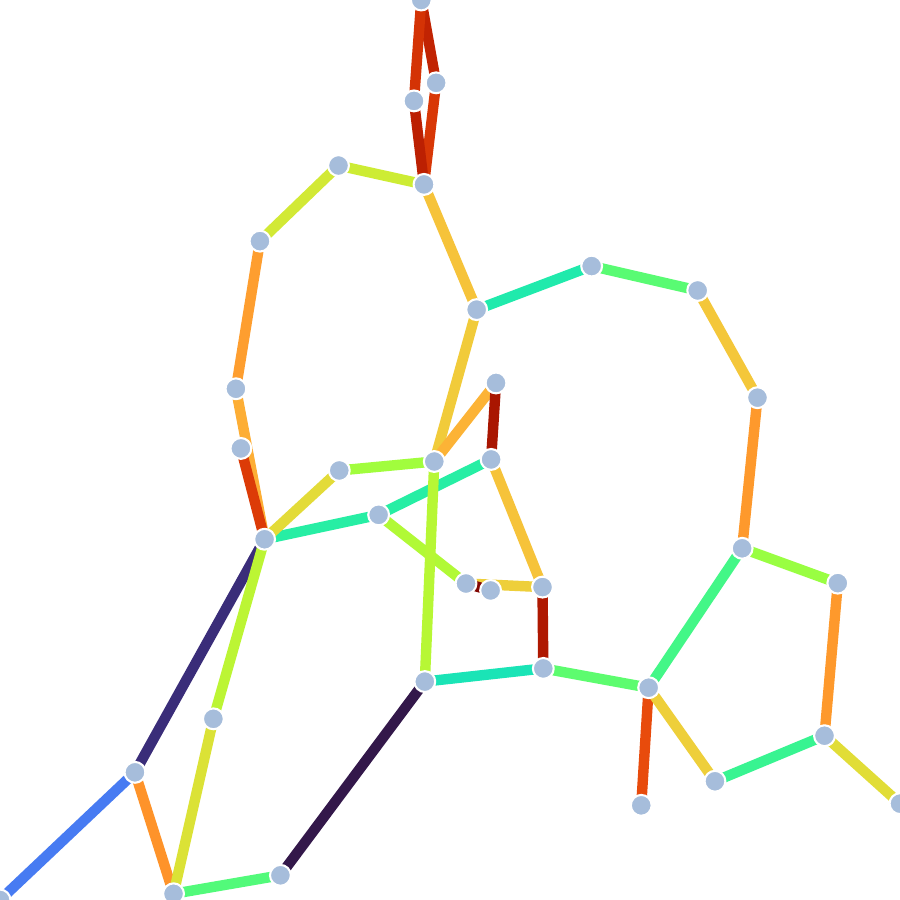} \\ \vspace{-0.0cm} \fontsize{7pt}{0pt}\selectfont 5,\textbf{2}} \\
\bottomrule
\end{tabular}
\end{table*}

\begin{table*}[htbp]
\setlength{\tabcolsep}{4pt}
\centering
\renewcommand{\arraystretch}{3}
\caption{Optimal projections of the 10D Neato layout of the Rome20 dataset (part 2/2). Numbers show Neato vs. \OptProj metrics. \textbf{Bold} = better.
Edges colored by length: {\color{red}red} = short, {\color{blue}blue} = long.}
\label{tab:fly_neato_dim10_vis_rome20_b}
\begin{tabular}{lc|cccccc}
\toprule
& neato & ang\_res$\downarrow$  & edge\_len$\downarrow$  & spr\_elec$\downarrow$  & stress$\downarrow$  & tsne$\downarrow$  & xing$\downarrow$  \\
\midrule
grafo2955.38 & \parbox{1.7cm}{\centering \includegraphics[height=1.5cm]{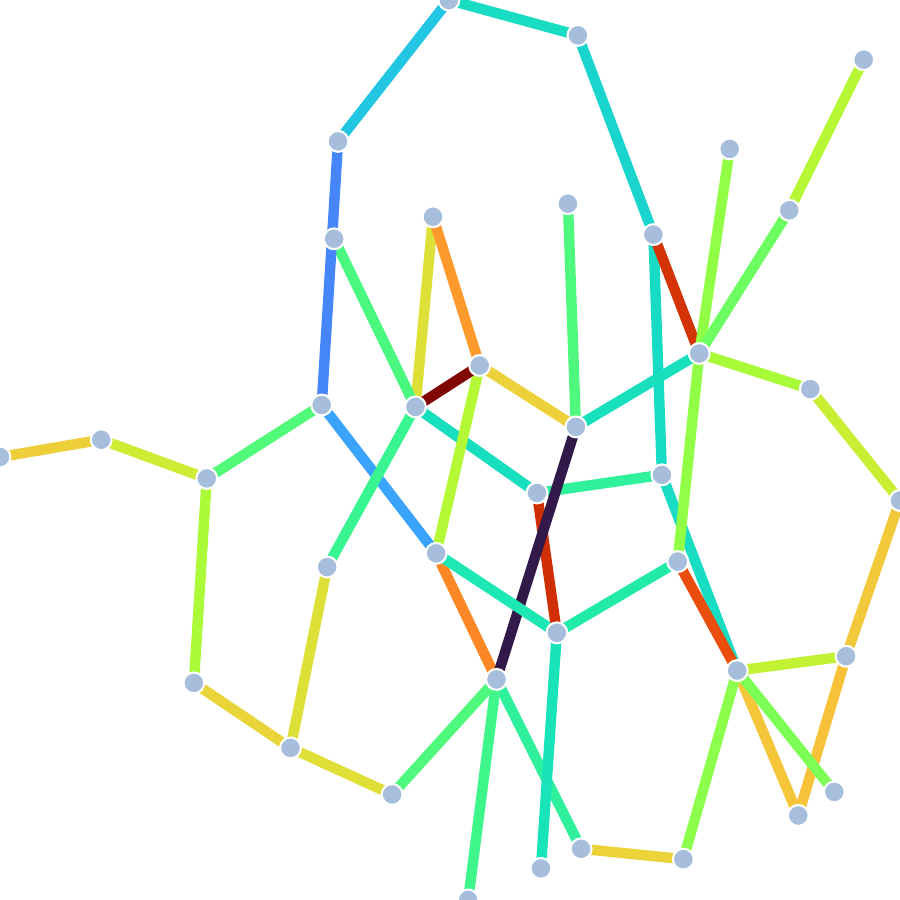} \\ \vspace{-0.0cm} \fontsize{7pt}{0pt}\selectfont } & \parbox{1.7cm}{\centering \includegraphics[height=1.5cm]{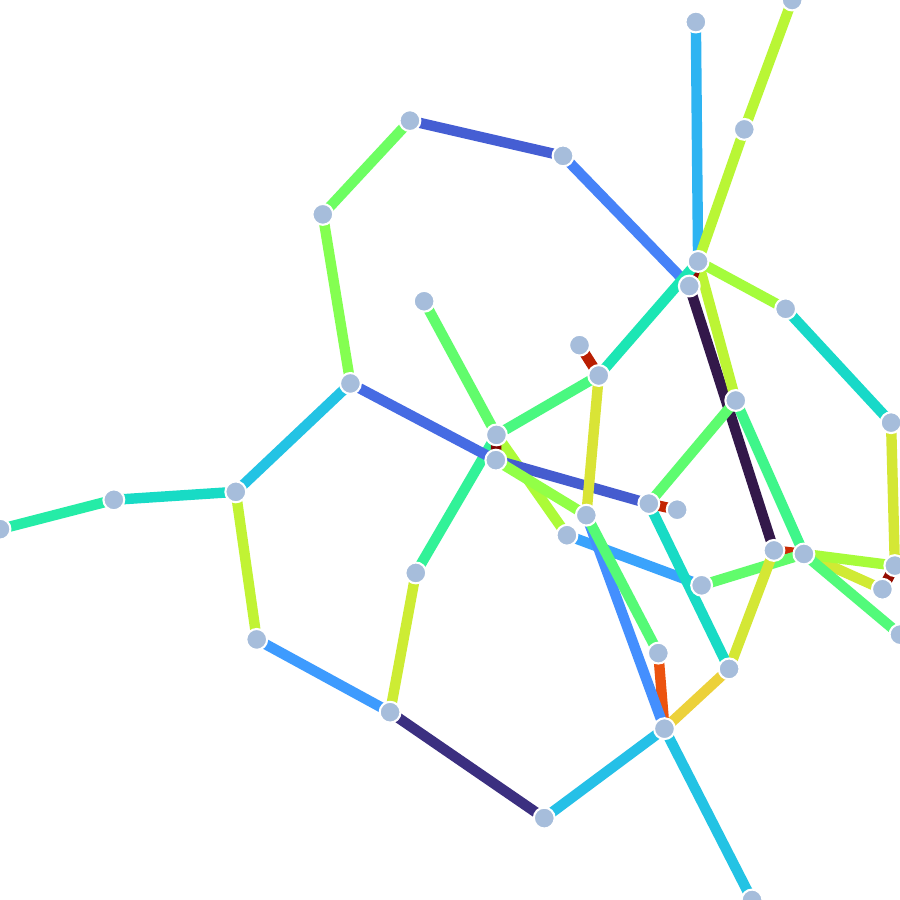} \\ \vspace{-0.0cm} \fontsize{7pt}{0pt}\selectfont 0.96,\textbf{0.754}} & \parbox{1.7cm}{\centering \includegraphics[height=1.5cm]{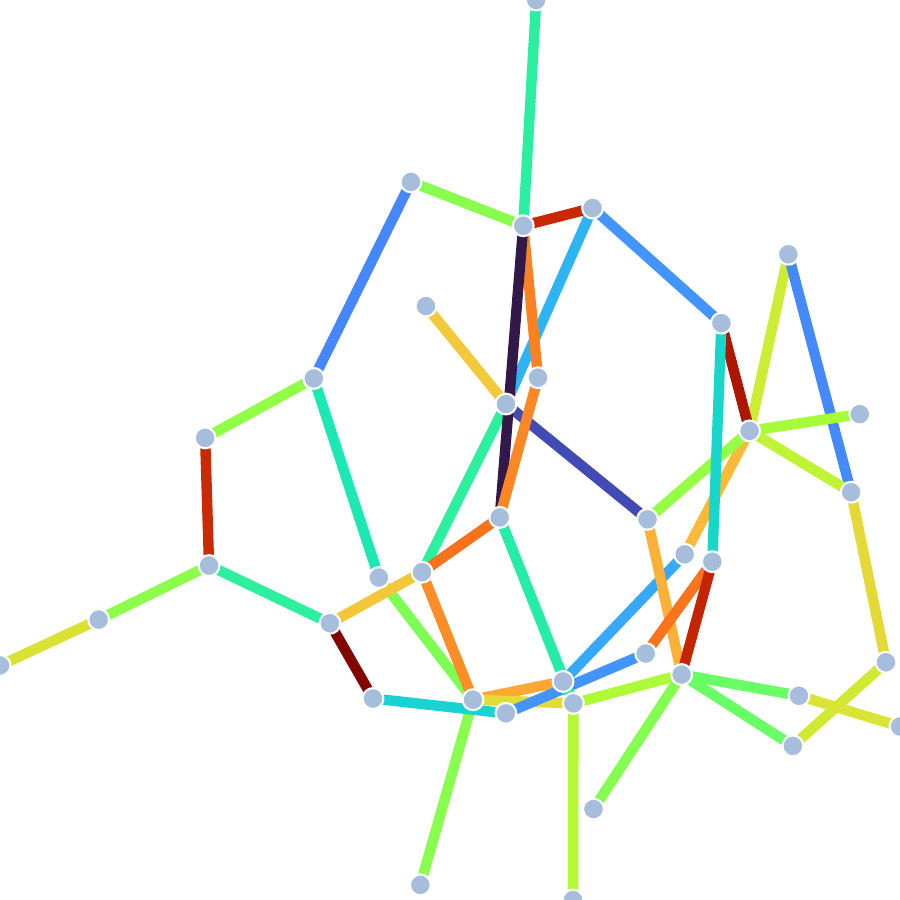} \\ \vspace{-0.0cm} \fontsize{7pt}{0pt}\selectfont \textbf{0.153},0.217} & \parbox{1.7cm}{\centering \includegraphics[height=1.5cm]{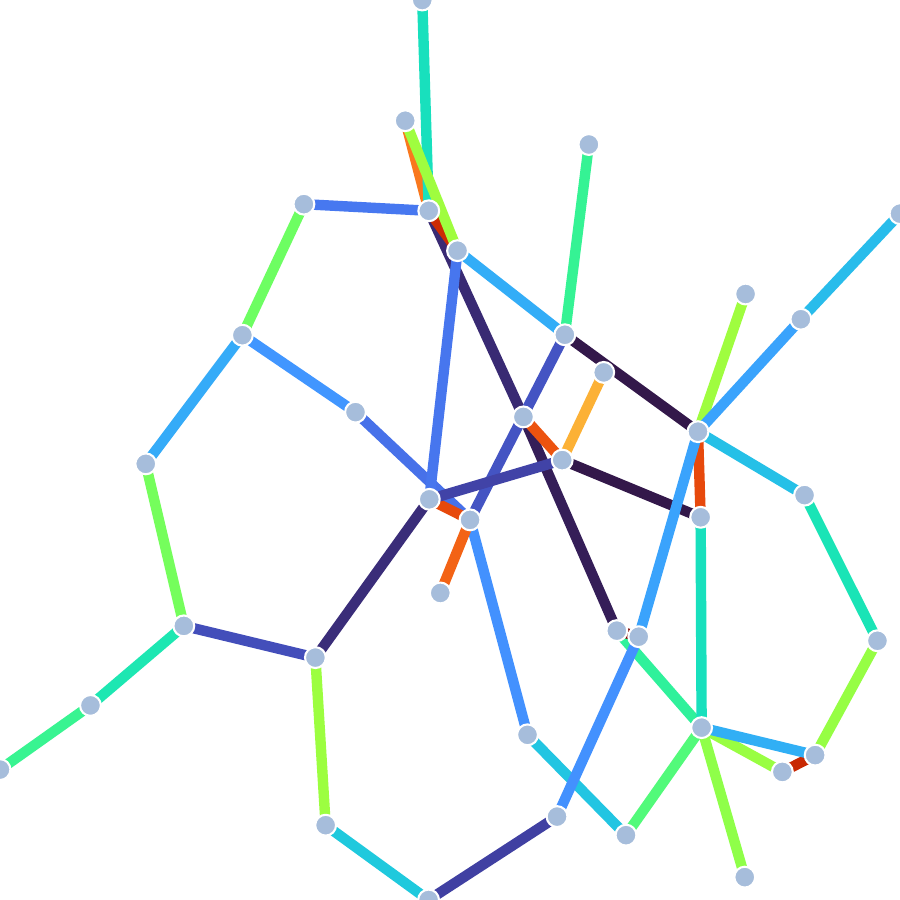} \\ \vspace{-0.0cm} \fontsize{7pt}{0pt}\selectfont -1.692,\textbf{-1.7}} & \parbox{1.7cm}{\centering \includegraphics[height=1.5cm]{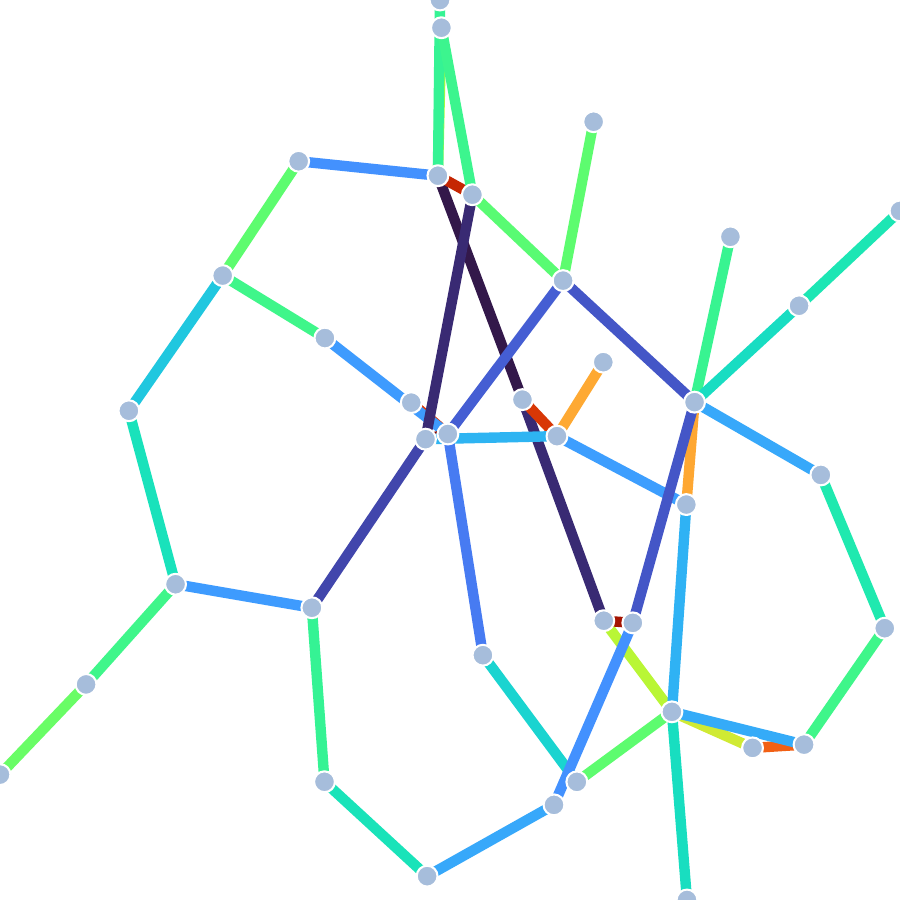} \\ \vspace{-0.0cm} \fontsize{7pt}{0pt}\selectfont \textbf{0.063},0.07} & \parbox{1.7cm}{\centering \includegraphics[height=1.5cm]{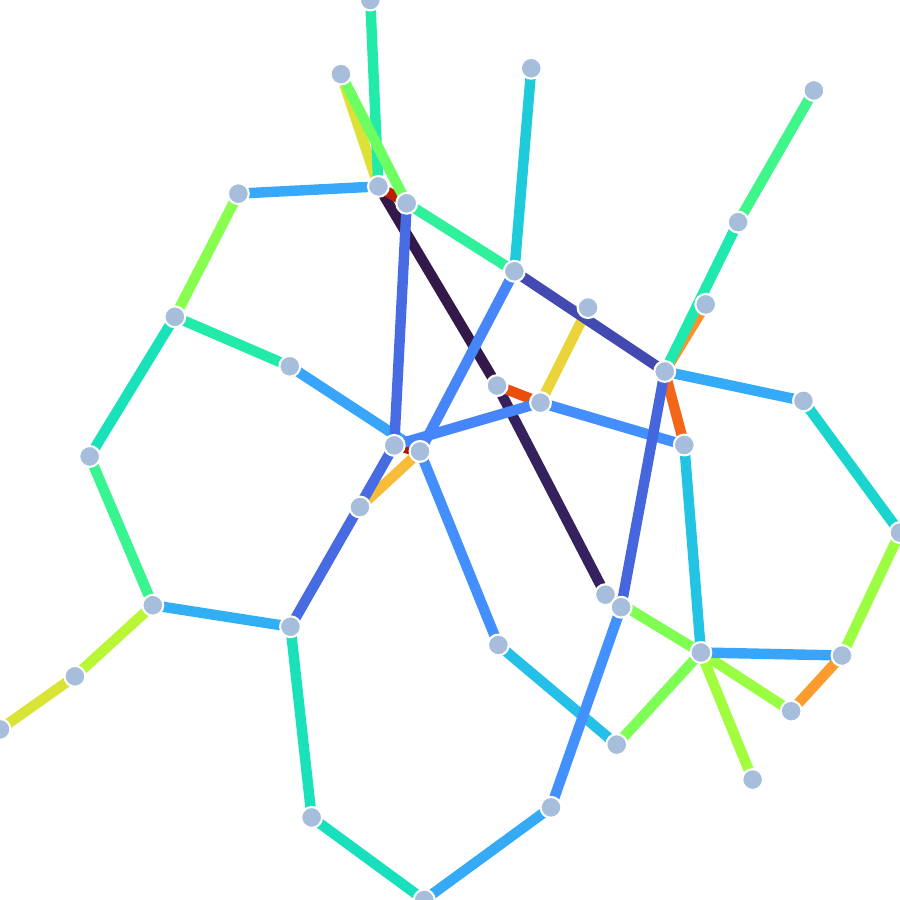} \\ \vspace{-0.0cm} \fontsize{7pt}{0pt}\selectfont 0.78,\textbf{0.708}} & \parbox{1.7cm}{\centering \includegraphics[height=1.5cm]{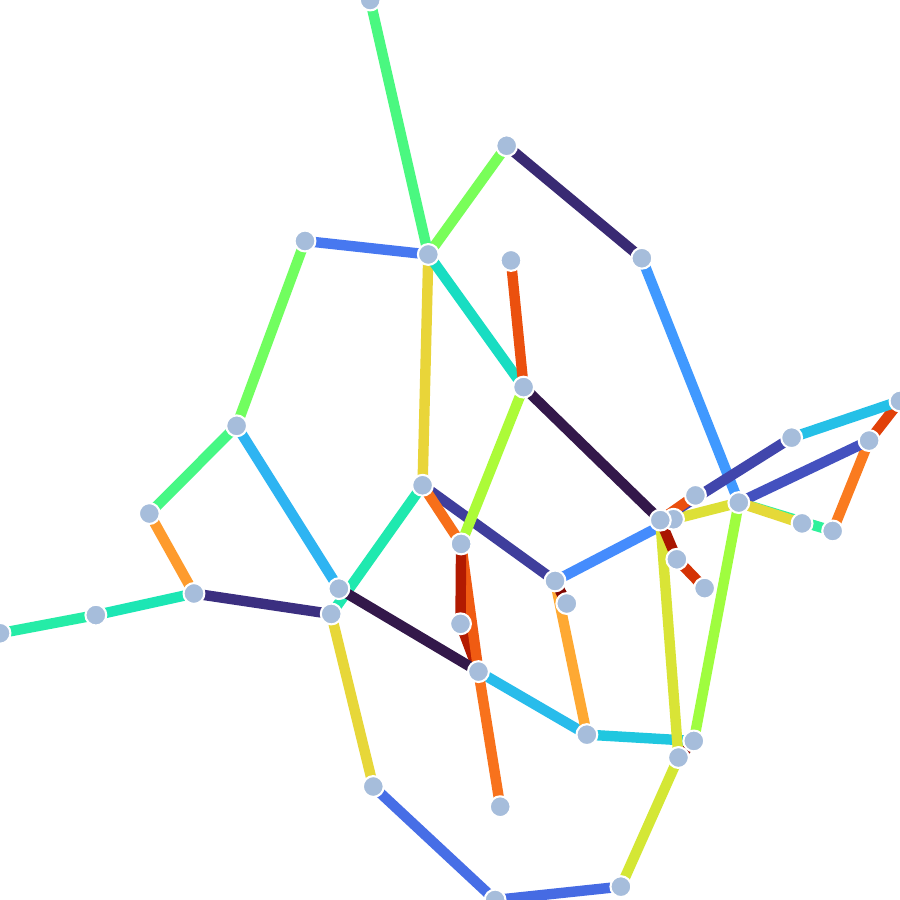} \\ \vspace{-0.0cm} \fontsize{7pt}{0pt}\selectfont 9,\textbf{6}} \\
grafo3138.60 & \parbox{1.7cm}{\centering \includegraphics[height=1.5cm]{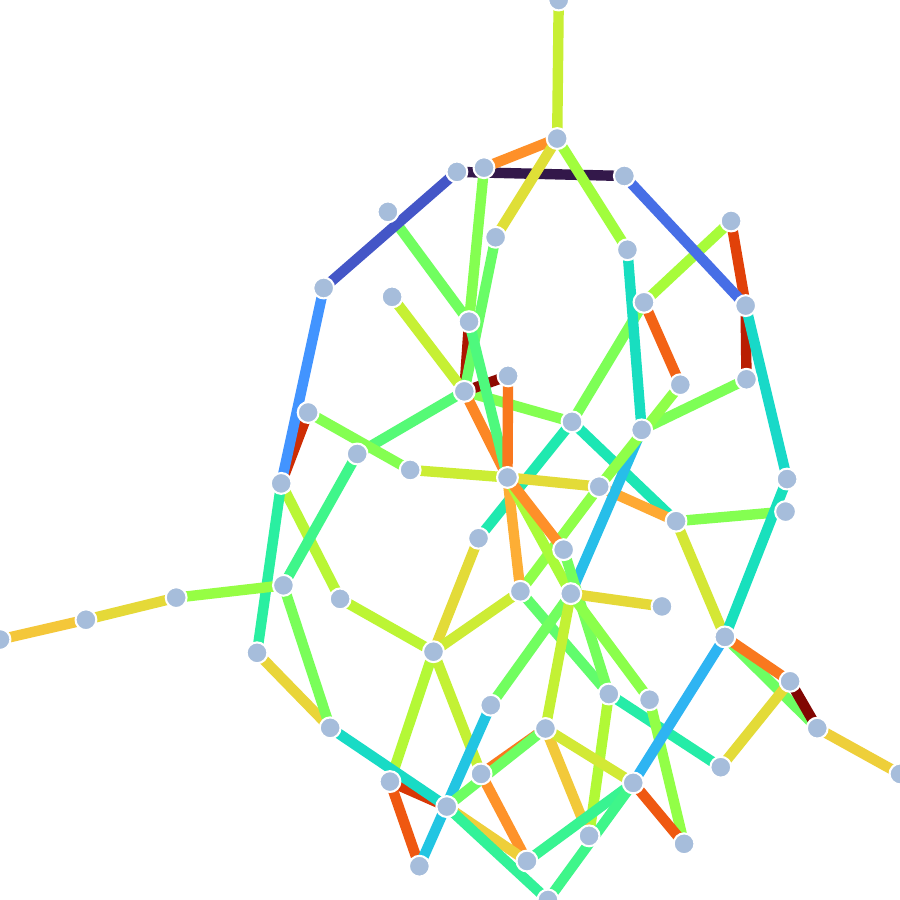} \\ \vspace{-0.0cm} \fontsize{7pt}{0pt}\selectfont } & \parbox{1.7cm}{\centering \includegraphics[height=1.5cm]{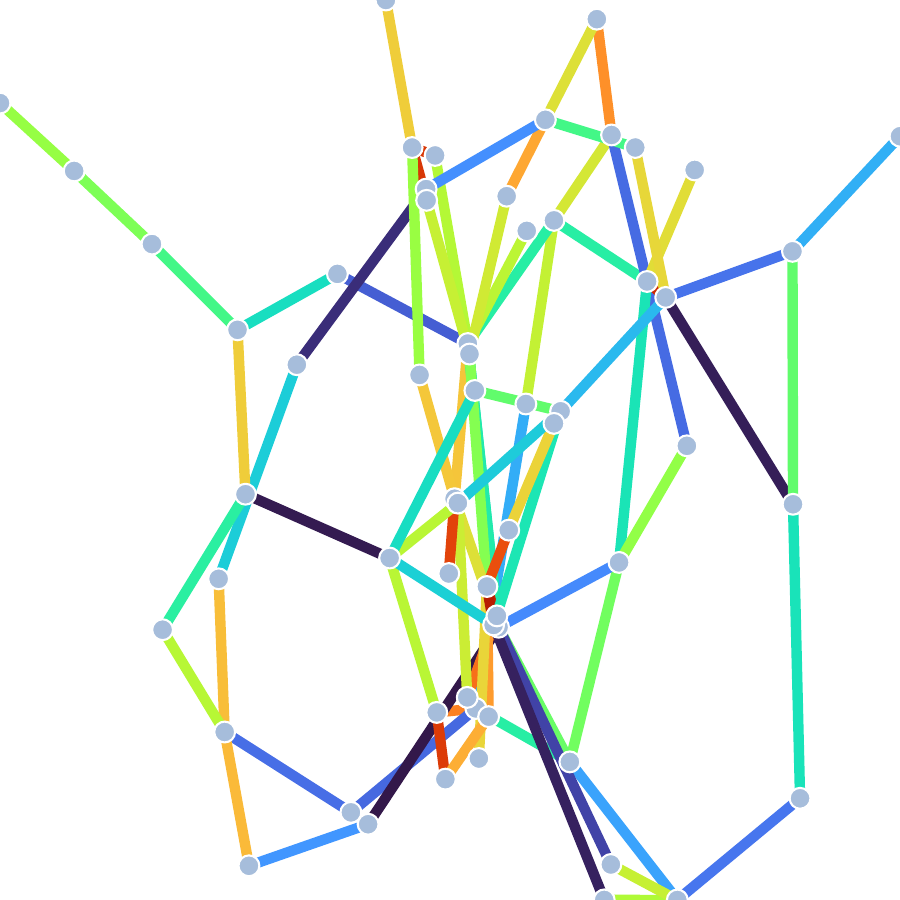} \\ \vspace{-0.0cm} \fontsize{7pt}{0pt}\selectfont 1.137,\textbf{0.889}} & \parbox{1.7cm}{\centering \includegraphics[height=1.5cm]{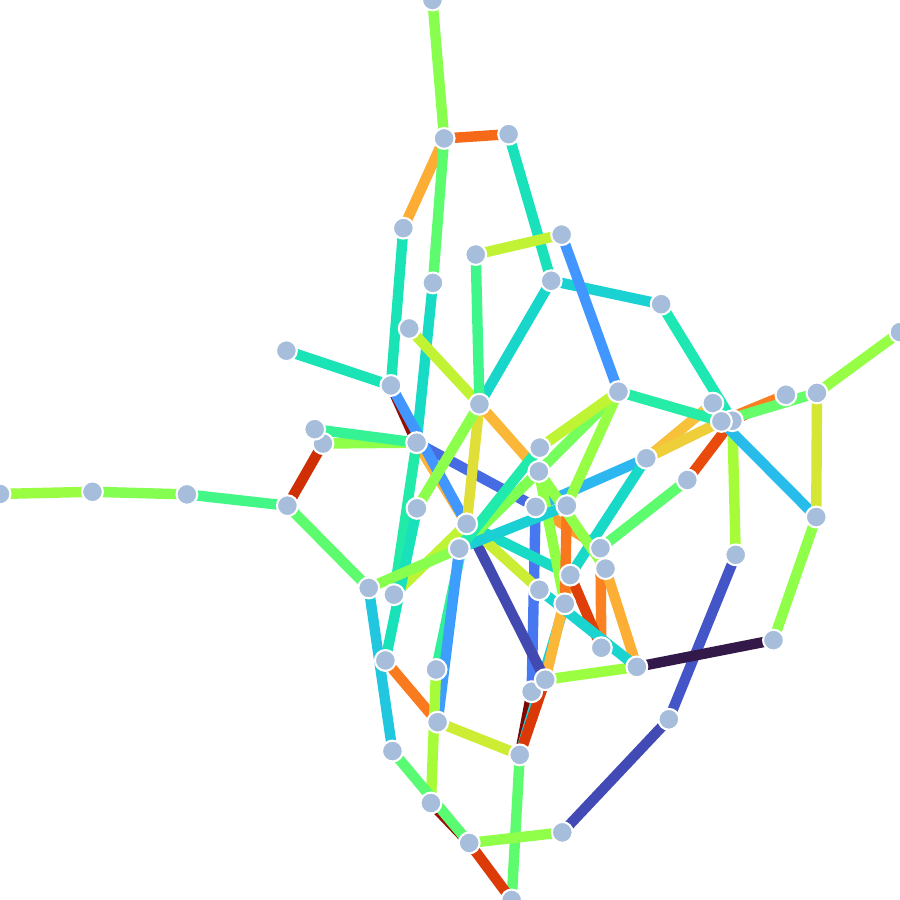} \\ \vspace{-0.0cm} \fontsize{7pt}{0pt}\selectfont 0.244,\textbf{0.226}} & \parbox{1.7cm}{\centering \includegraphics[height=1.5cm]{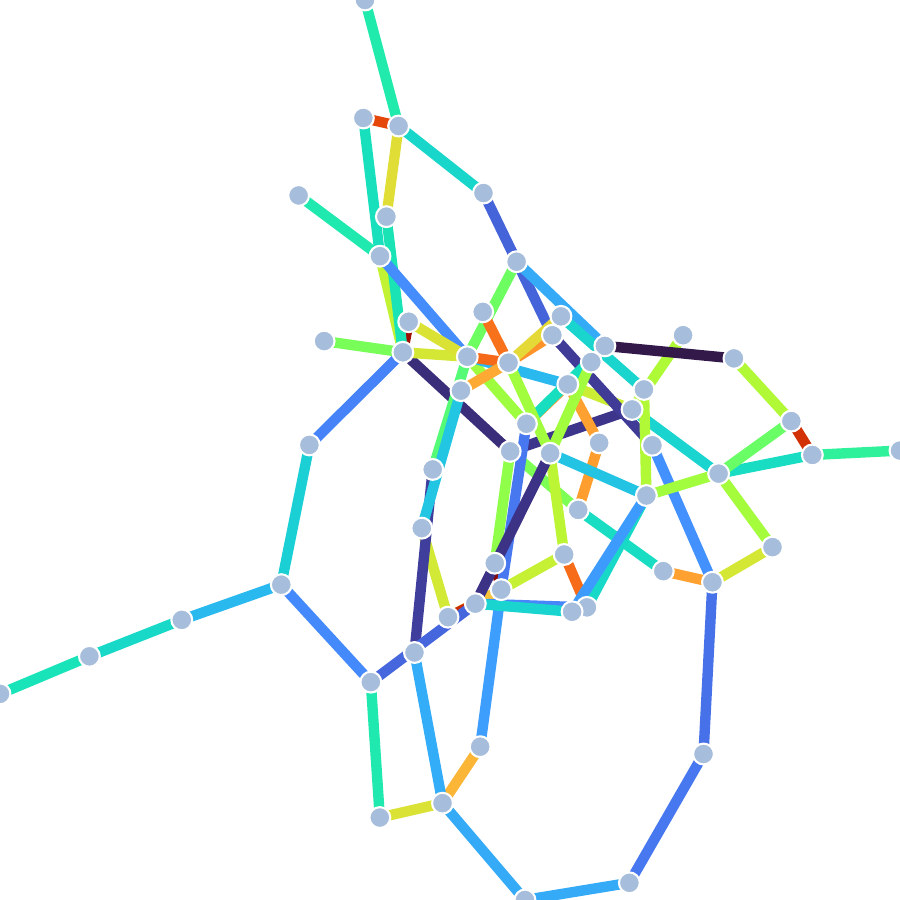} \\ \vspace{-0.0cm} \fontsize{7pt}{0pt}\selectfont \textbf{-1.809},-1.791} & \parbox{1.7cm}{\centering \includegraphics[height=1.5cm]{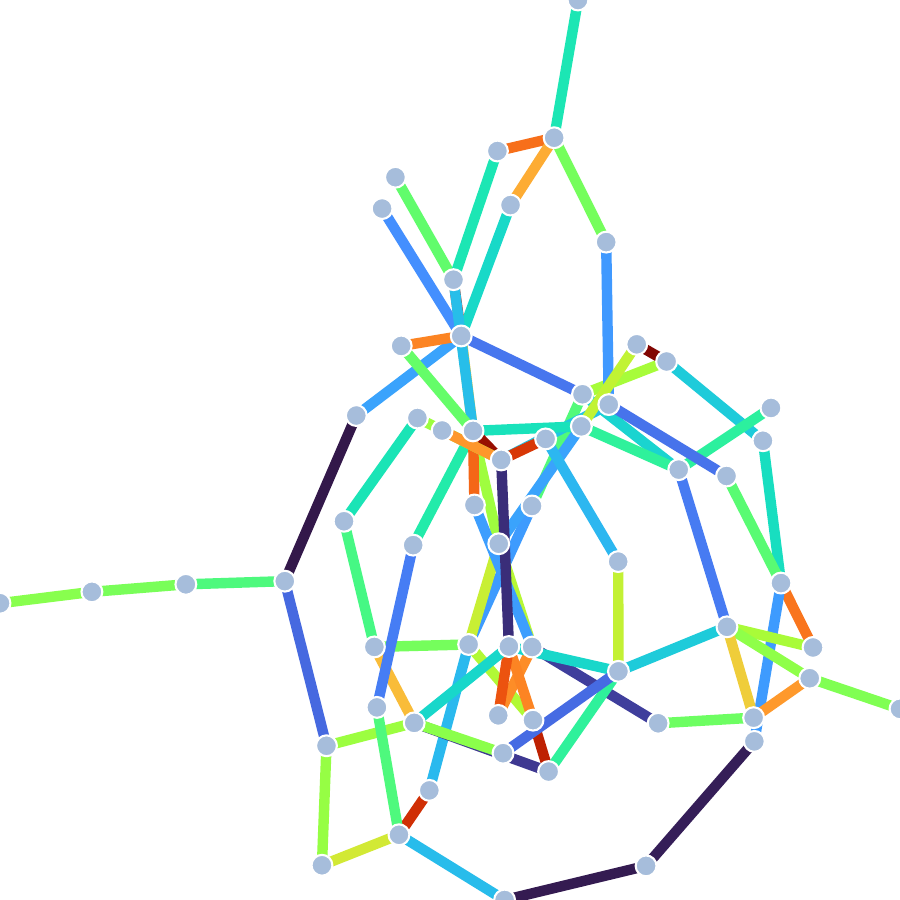} \\ \vspace{-0.0cm} \fontsize{7pt}{0pt}\selectfont \textbf{0.089},0.095} & \parbox{1.7cm}{\centering \includegraphics[height=1.5cm]{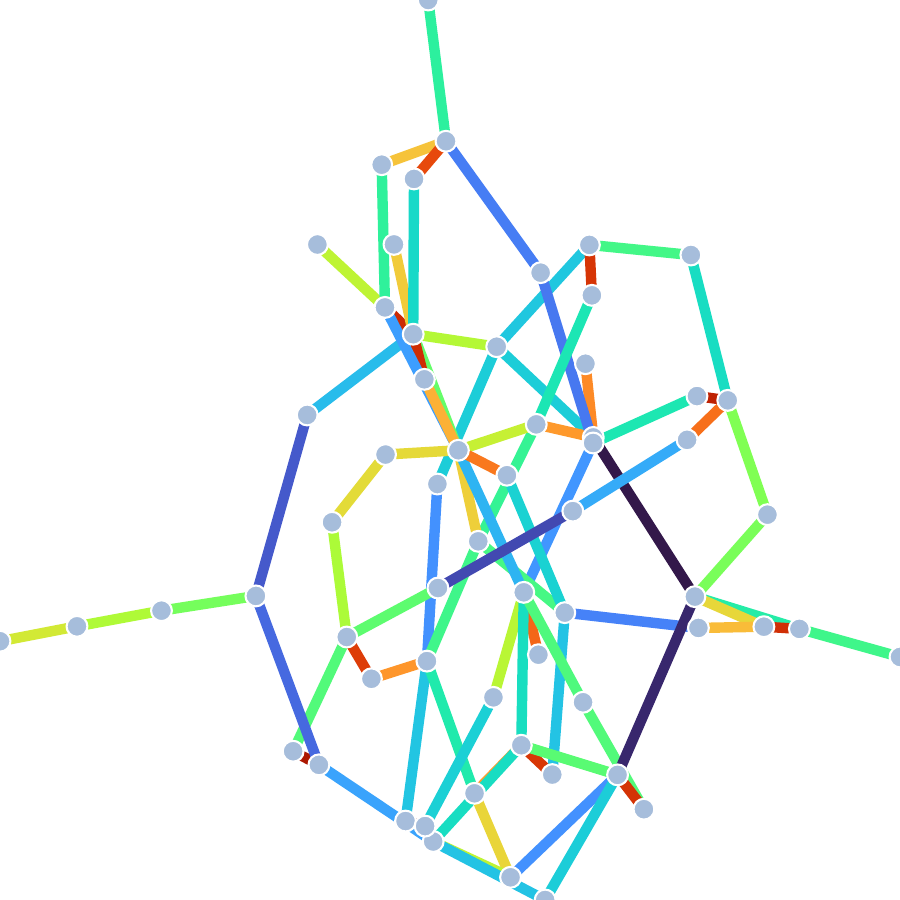} \\ \vspace{-0.0cm} \fontsize{7pt}{0pt}\selectfont 1.054,\textbf{1.0}} & \parbox{1.7cm}{\centering \includegraphics[height=1.5cm]{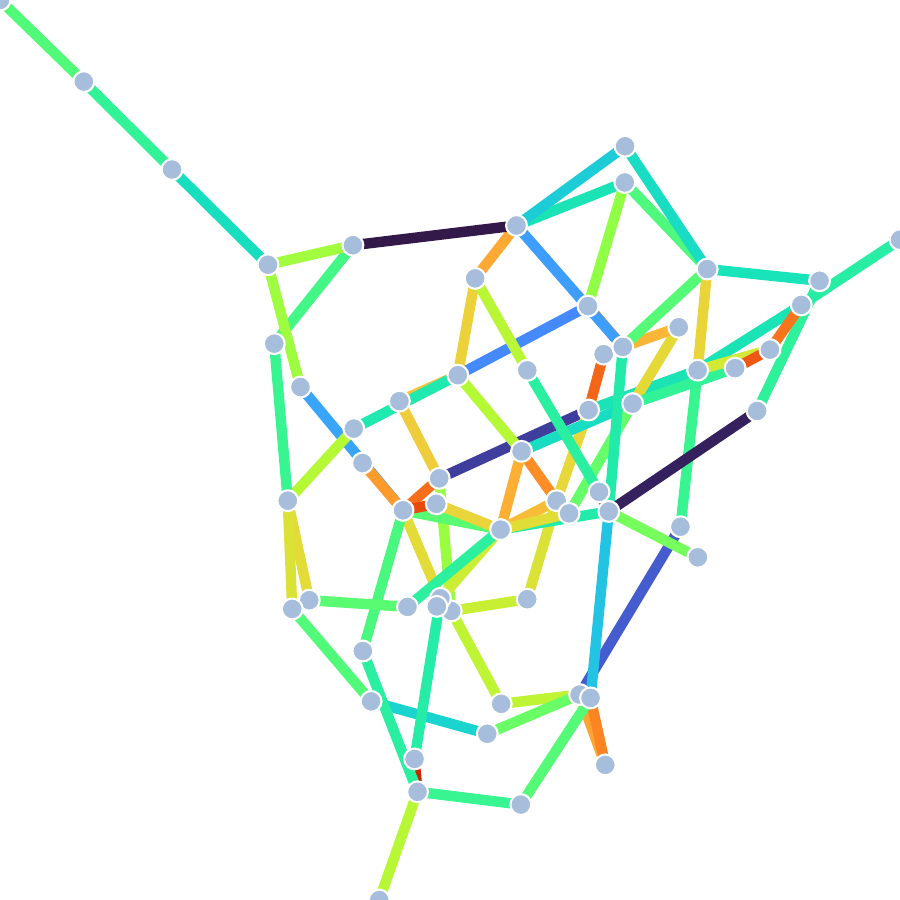} \\ \vspace{-0.0cm} \fontsize{7pt}{0pt}\selectfont 42,\textbf{31}} \\
grafo3587.36 & \parbox{1.7cm}{\centering \includegraphics[height=1.5cm]{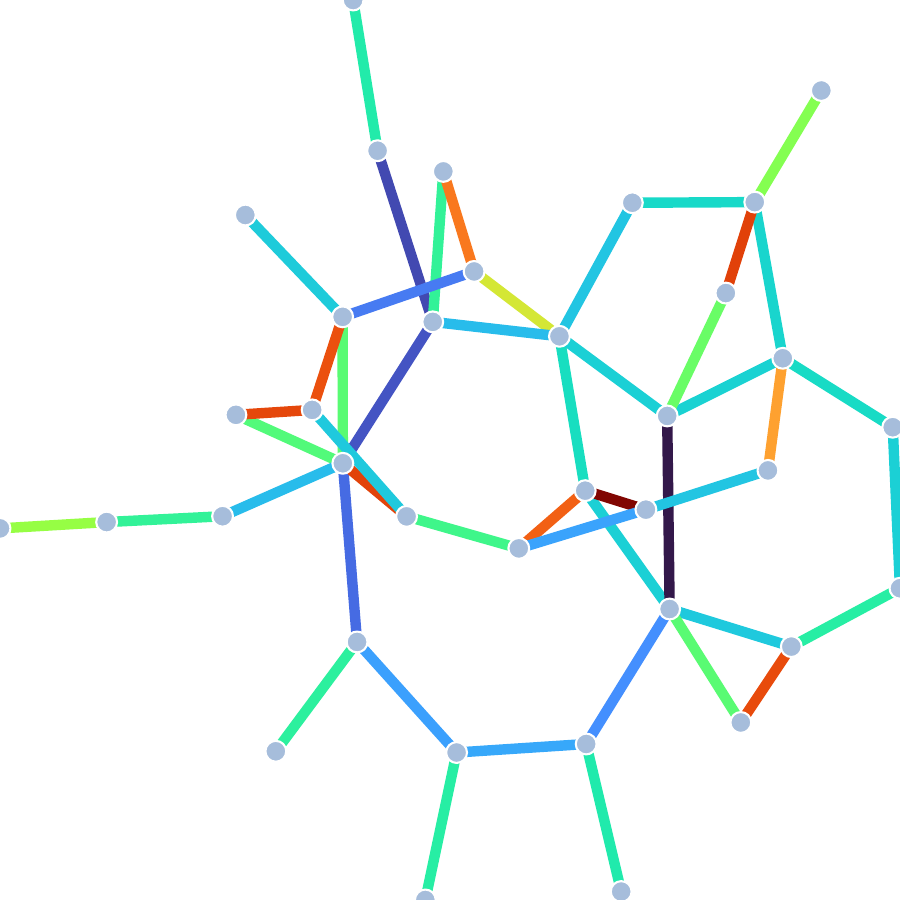} \\ \vspace{-0.0cm} \fontsize{7pt}{0pt}\selectfont } & \parbox{1.7cm}{\centering \includegraphics[height=1.5cm]{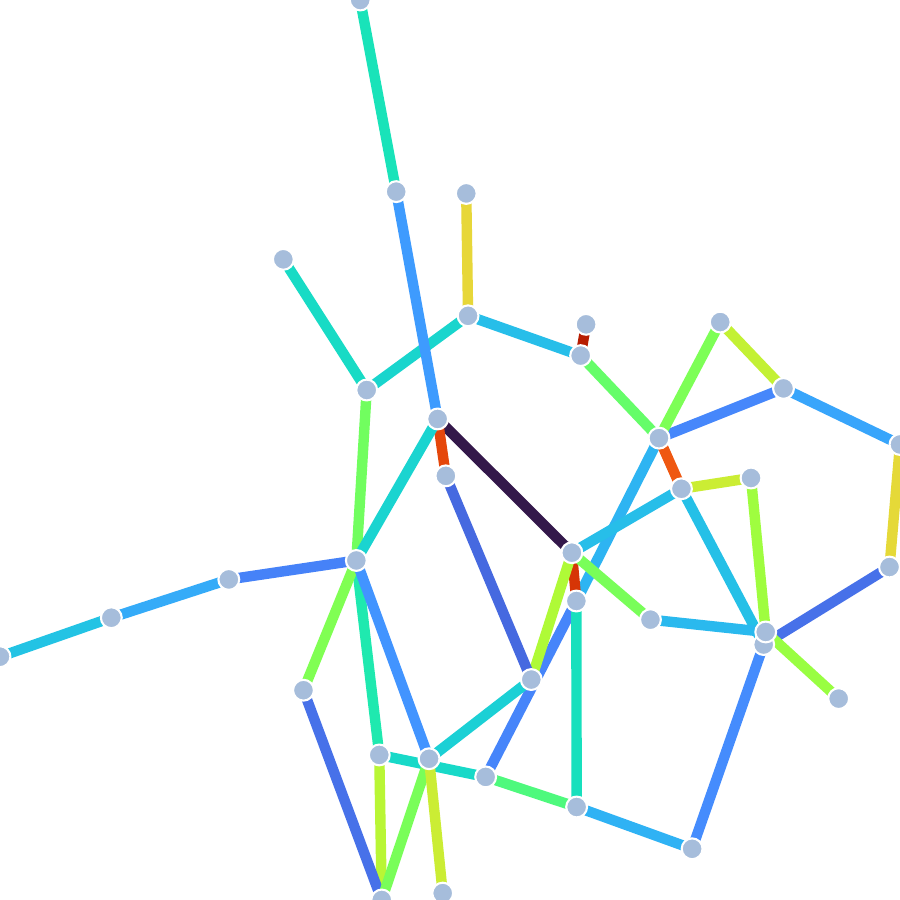} \\ \vspace{-0.0cm} \fontsize{7pt}{0pt}\selectfont 1.186,\textbf{0.868}} & \parbox{1.7cm}{\centering \includegraphics[height=1.5cm]{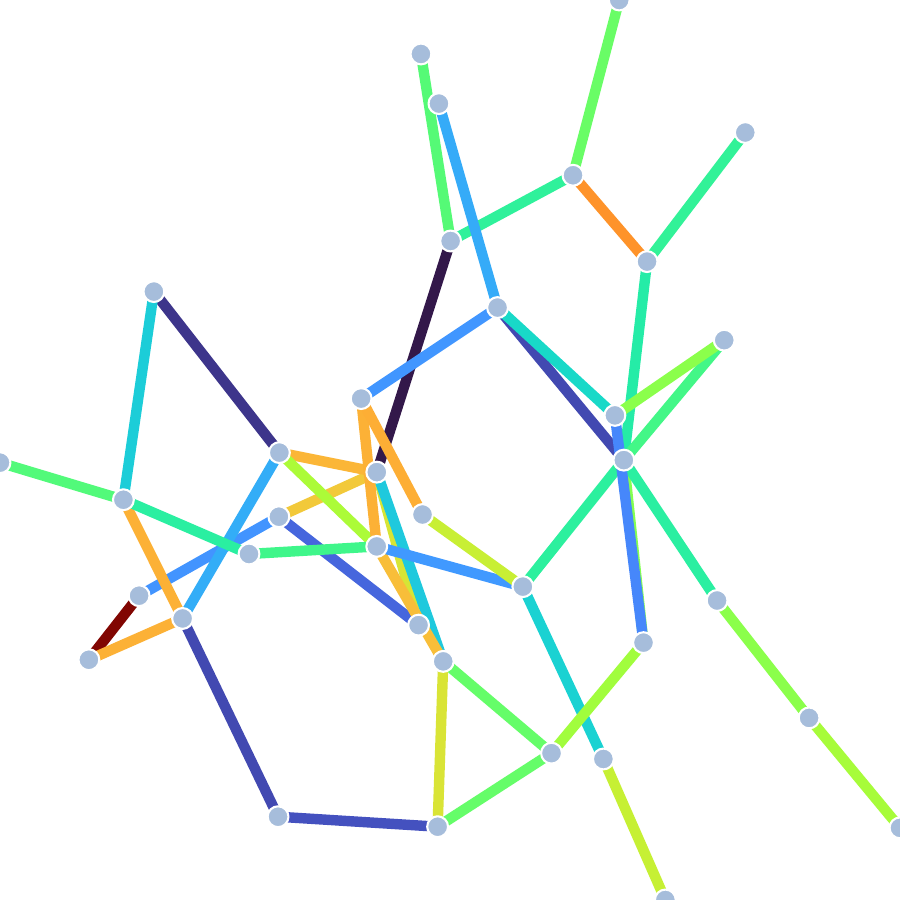} \\ \vspace{-0.0cm} \fontsize{7pt}{0pt}\selectfont 0.176,\textbf{0.173}} & \parbox{1.7cm}{\centering \includegraphics[height=1.5cm]{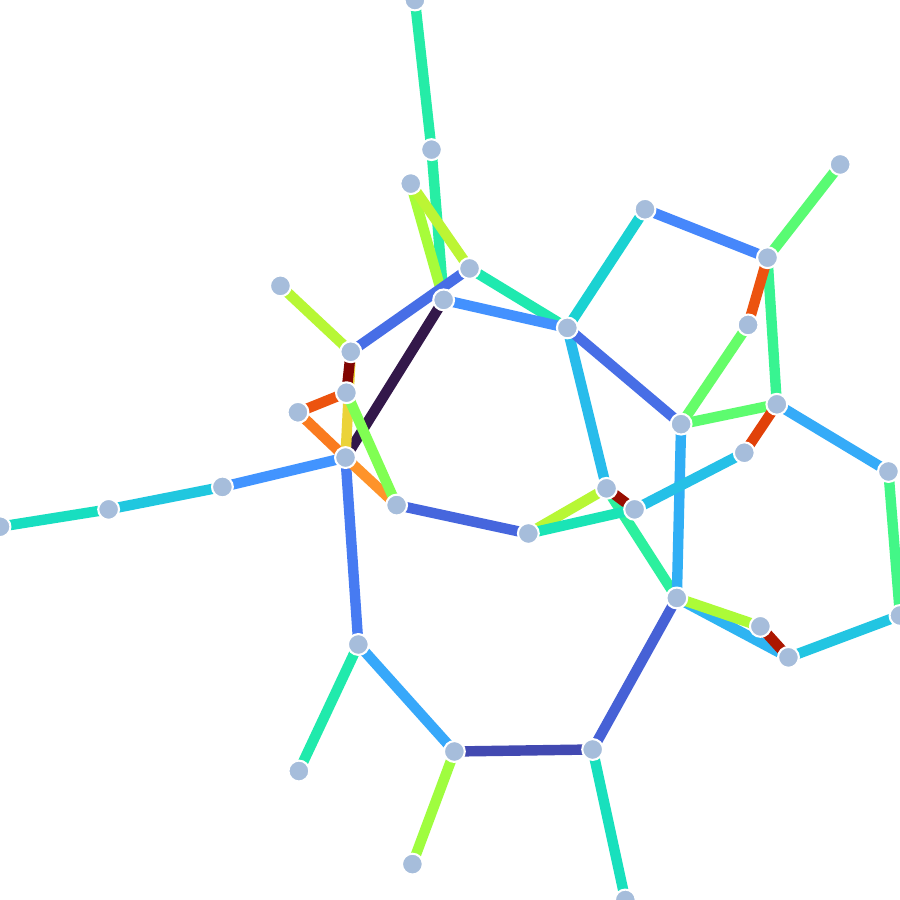} \\ \vspace{-0.0cm} \fontsize{7pt}{0pt}\selectfont -1.748,\textbf{-1.755}} & \parbox{1.7cm}{\centering \includegraphics[height=1.5cm]{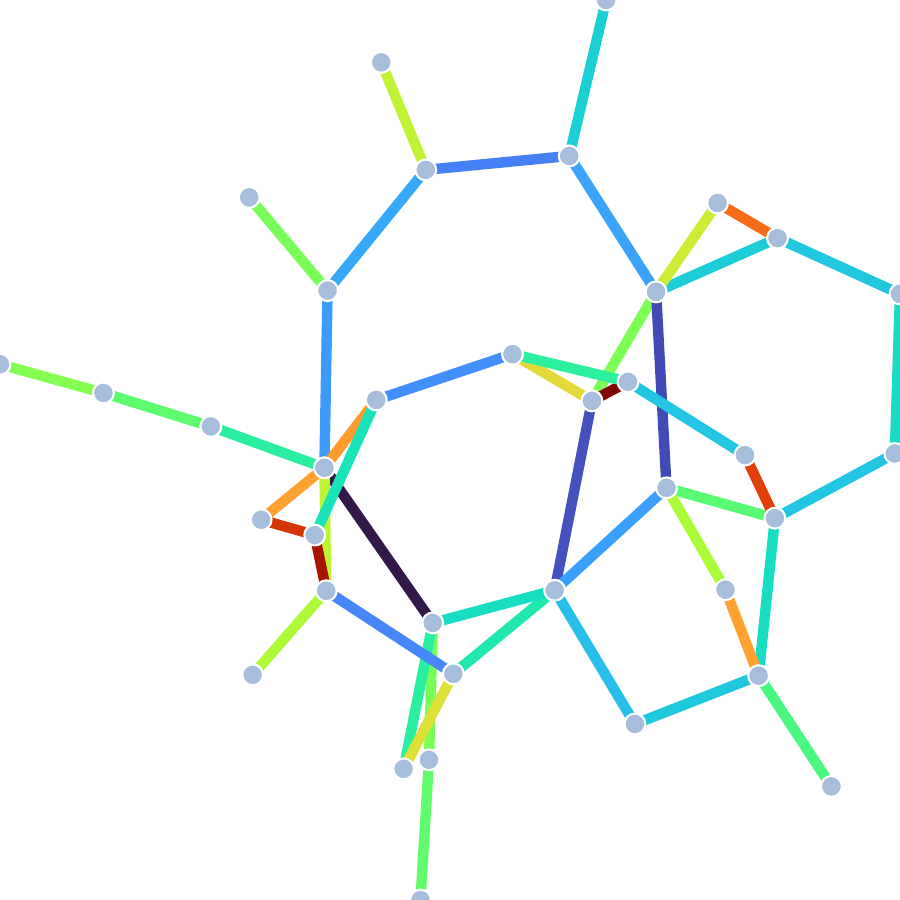} \\ \vspace{-0.0cm} \fontsize{7pt}{0pt}\selectfont \textbf{0.041},0.046} & \parbox{1.7cm}{\centering \includegraphics[height=1.5cm]{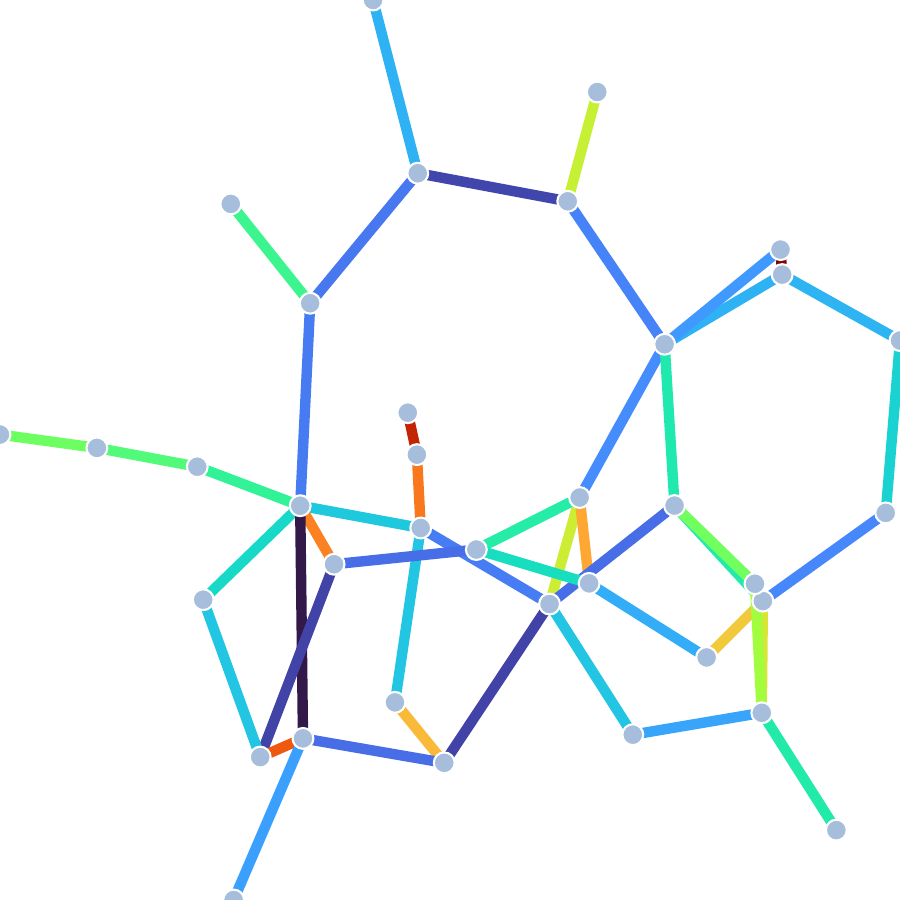} \\ \vspace{-0.0cm} \fontsize{7pt}{0pt}\selectfont \textbf{0.621},0.669} & \parbox{1.7cm}{\centering \includegraphics[height=1.5cm]{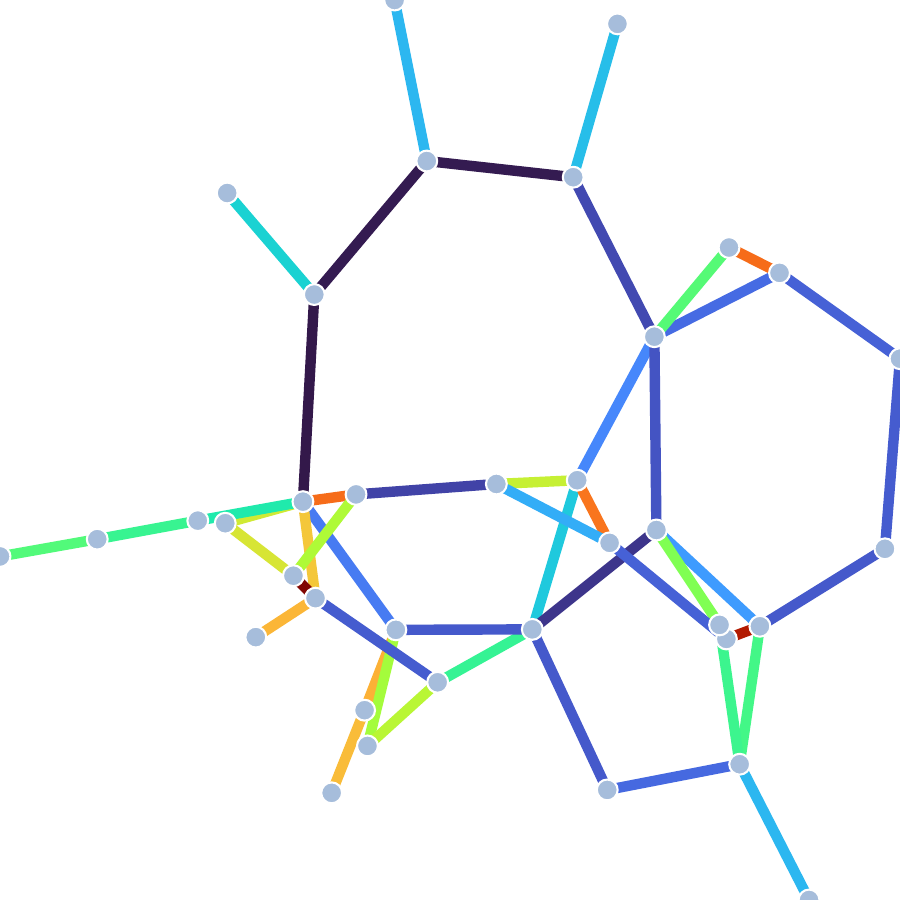} \\ \vspace{-0.0cm} \fontsize{7pt}{0pt}\selectfont \textbf{6},7} \\
grafo5780.48 & \parbox{1.7cm}{\centering \includegraphics[height=1.5cm]{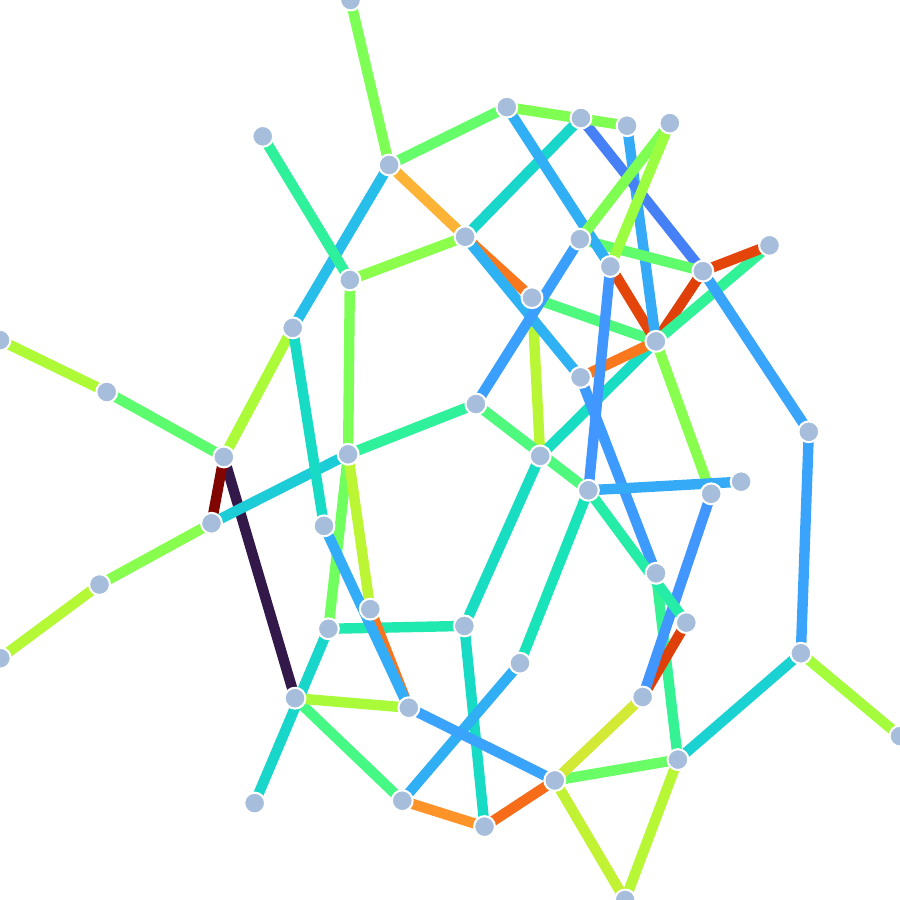} \\ \vspace{-0.0cm} \fontsize{7pt}{0pt}\selectfont } & \parbox{1.7cm}{\centering \includegraphics[height=1.5cm]{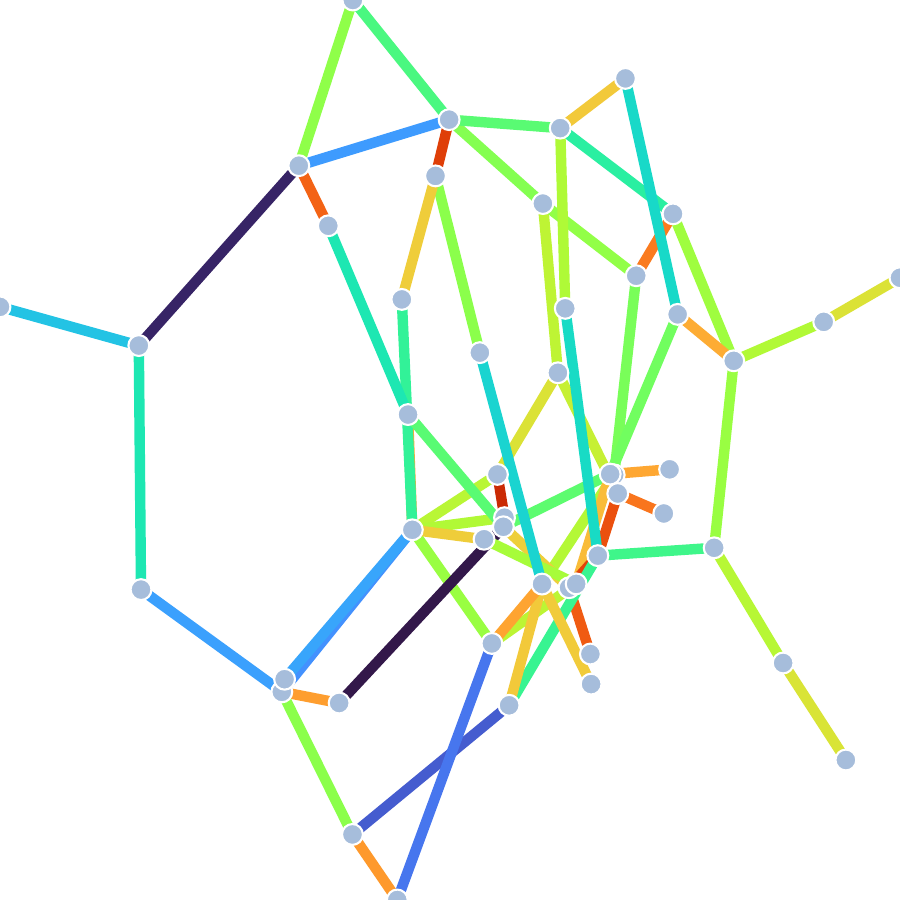} \\ \vspace{-0.0cm} \fontsize{7pt}{0pt}\selectfont 1.143,\textbf{0.886}} & \parbox{1.7cm}{\centering \includegraphics[height=1.5cm]{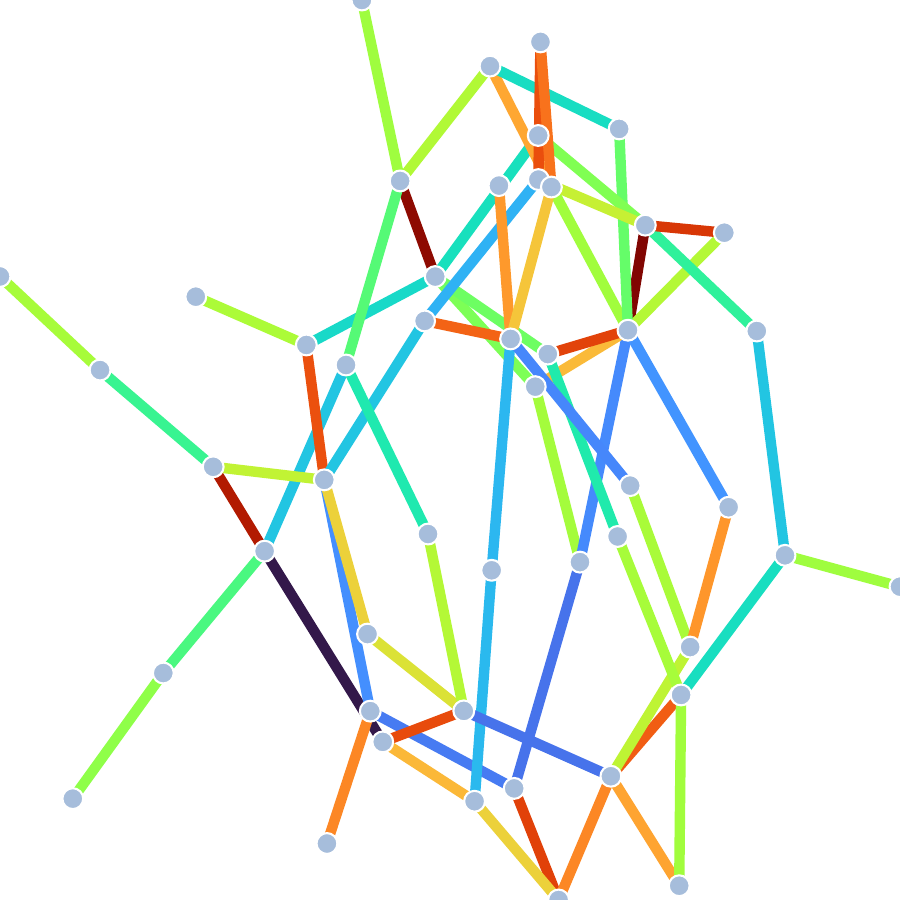} \\ \vspace{-0.0cm} \fontsize{7pt}{0pt}\selectfont 0.221,\textbf{0.203}} & \parbox{1.7cm}{\centering \includegraphics[height=1.5cm]{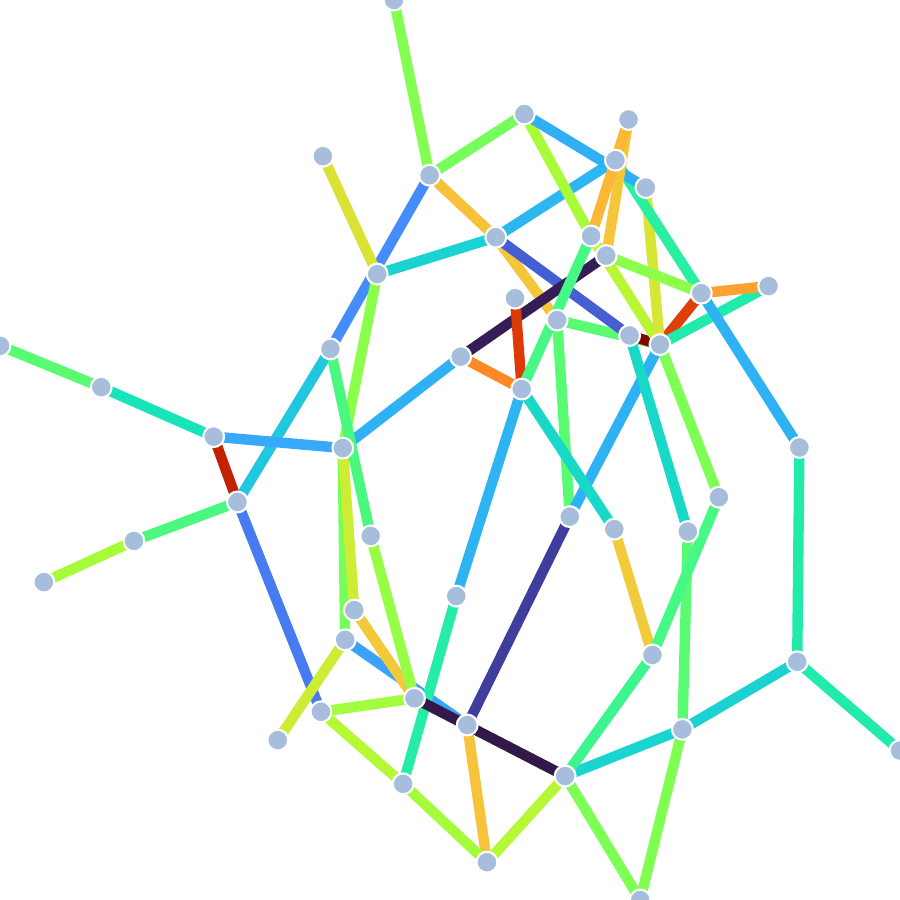} \\ \vspace{-0.0cm} \fontsize{7pt}{0pt}\selectfont -1.657,\textbf{-1.672}} & \parbox{1.7cm}{\centering \includegraphics[height=1.5cm]{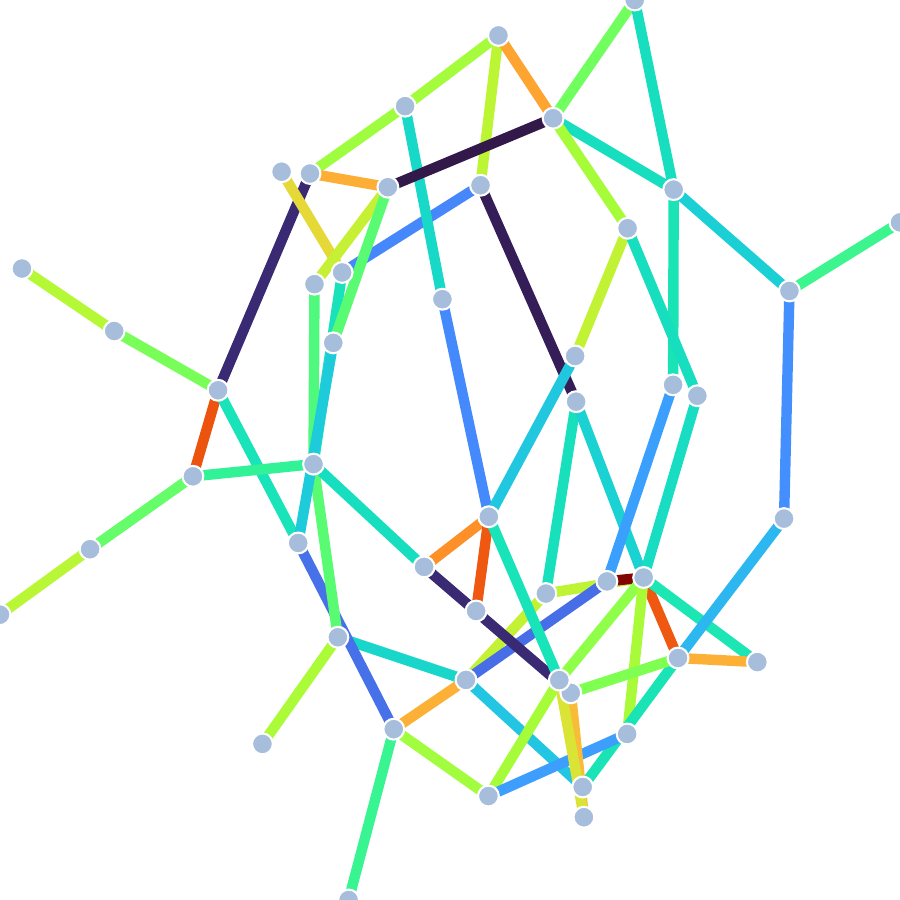} \\ \vspace{-0.0cm} \fontsize{7pt}{0pt}\selectfont \textbf{0.086},0.089} & \parbox{1.7cm}{\centering \includegraphics[height=1.5cm]{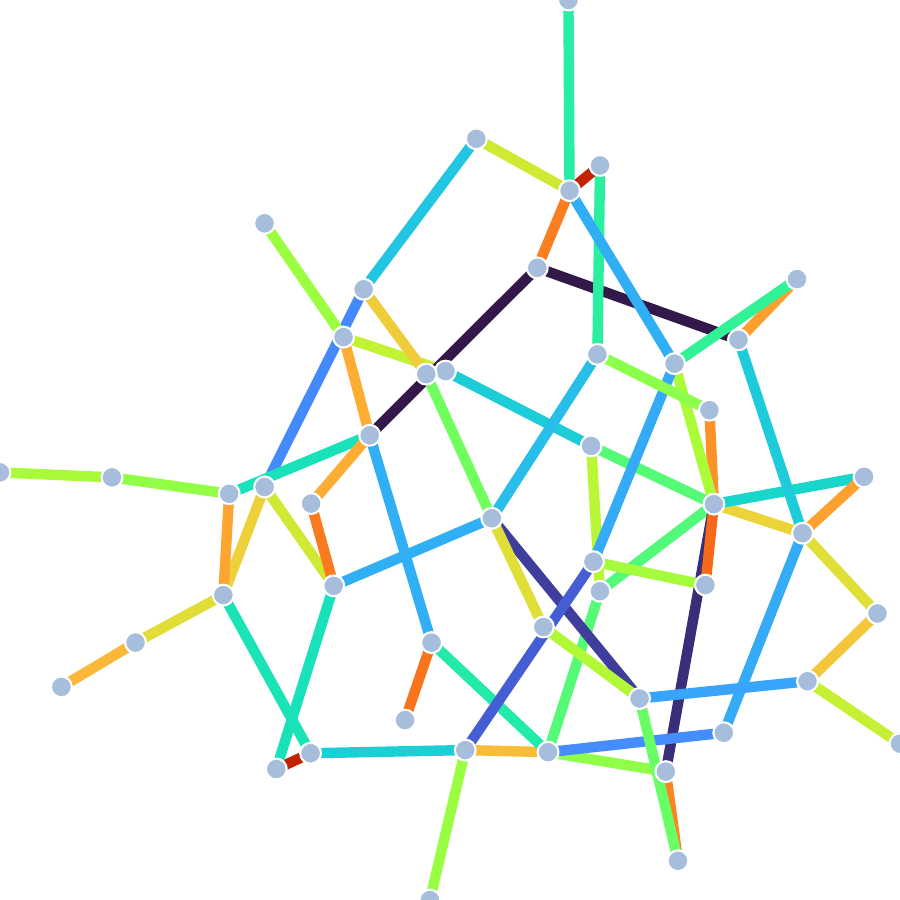} \\ \vspace{-0.0cm} \fontsize{7pt}{0pt}\selectfont 0.972,\textbf{0.892}} & \parbox{1.7cm}{\centering \includegraphics[height=1.5cm]{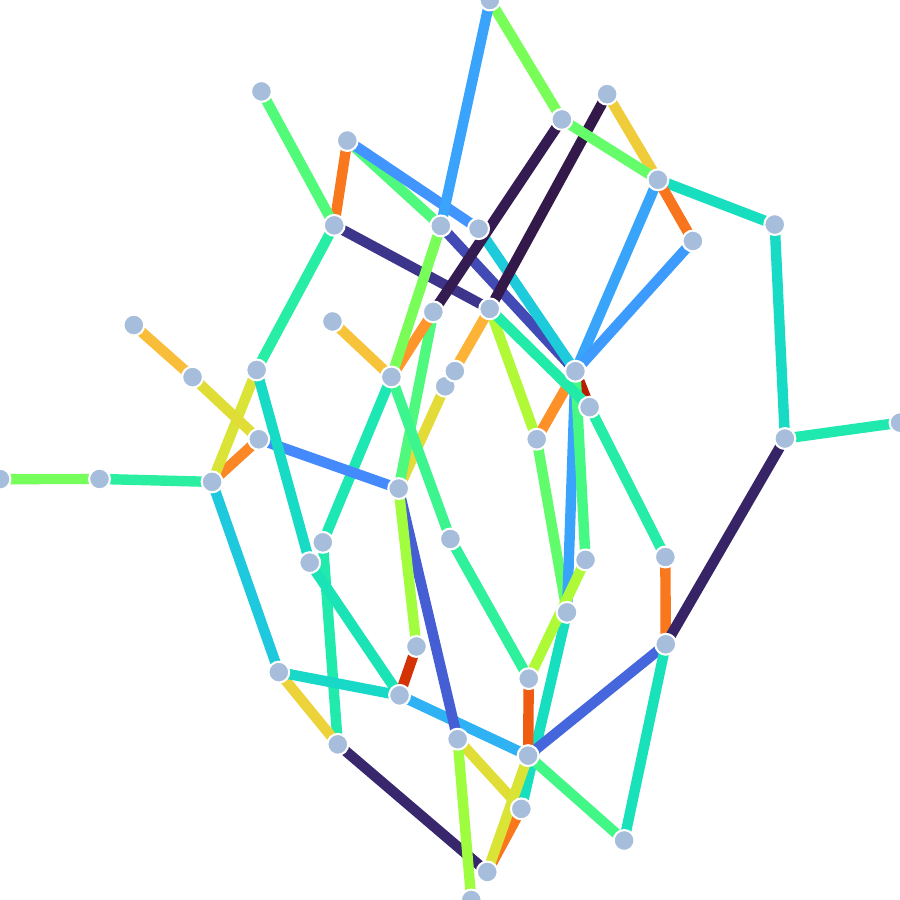} \\ \vspace{-0.0cm} \fontsize{7pt}{0pt}\selectfont 41,\textbf{25}} \\
grafo7023.45 & \parbox{1.7cm}{\centering \includegraphics[height=1.5cm]{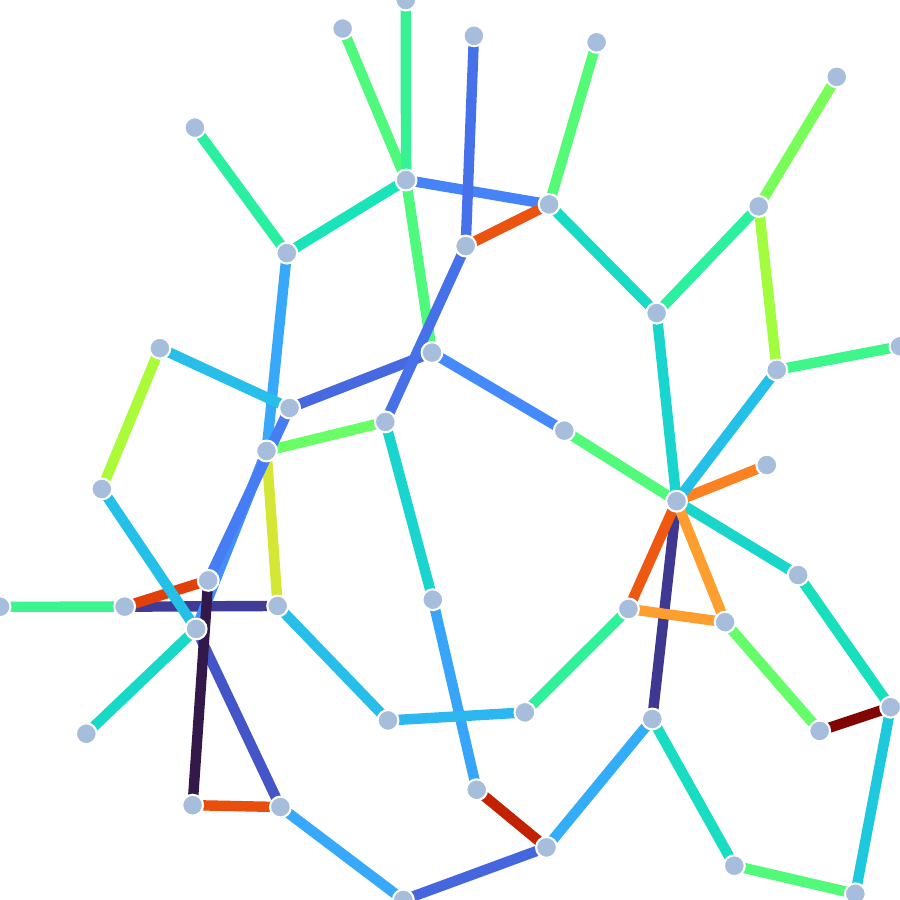} \\ \vspace{-0.0cm} \fontsize{7pt}{0pt}\selectfont } & \parbox{1.7cm}{\centering \includegraphics[height=1.5cm]{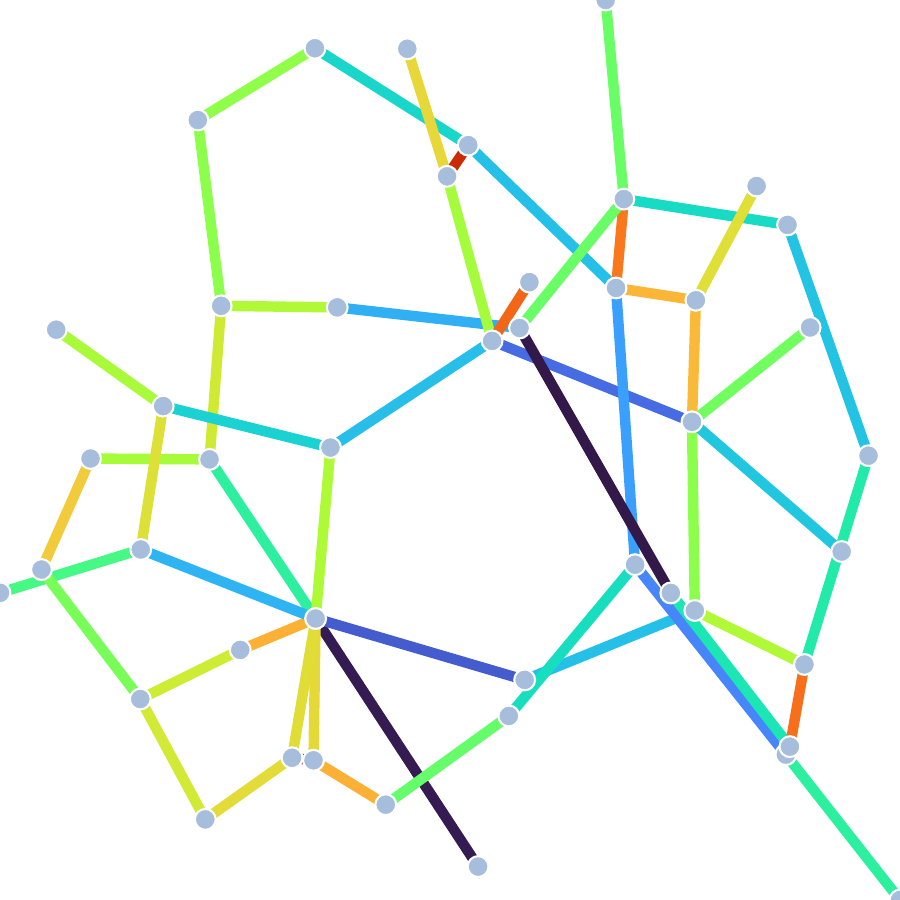} \\ \vspace{-0.0cm} \fontsize{7pt}{0pt}\selectfont 0.913,\textbf{0.718}} & \parbox{1.7cm}{\centering \includegraphics[height=1.5cm]{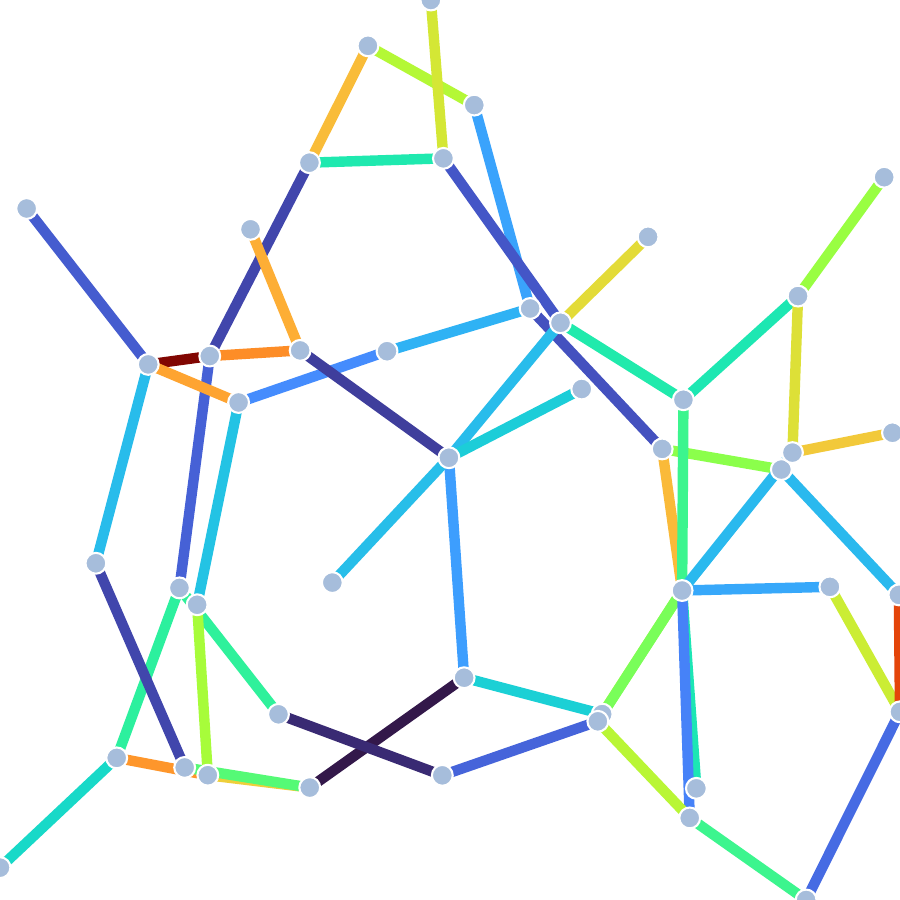} \\ \vspace{-0.0cm} \fontsize{7pt}{0pt}\selectfont \textbf{0.156},0.194} & \parbox{1.7cm}{\centering \includegraphics[height=1.5cm]{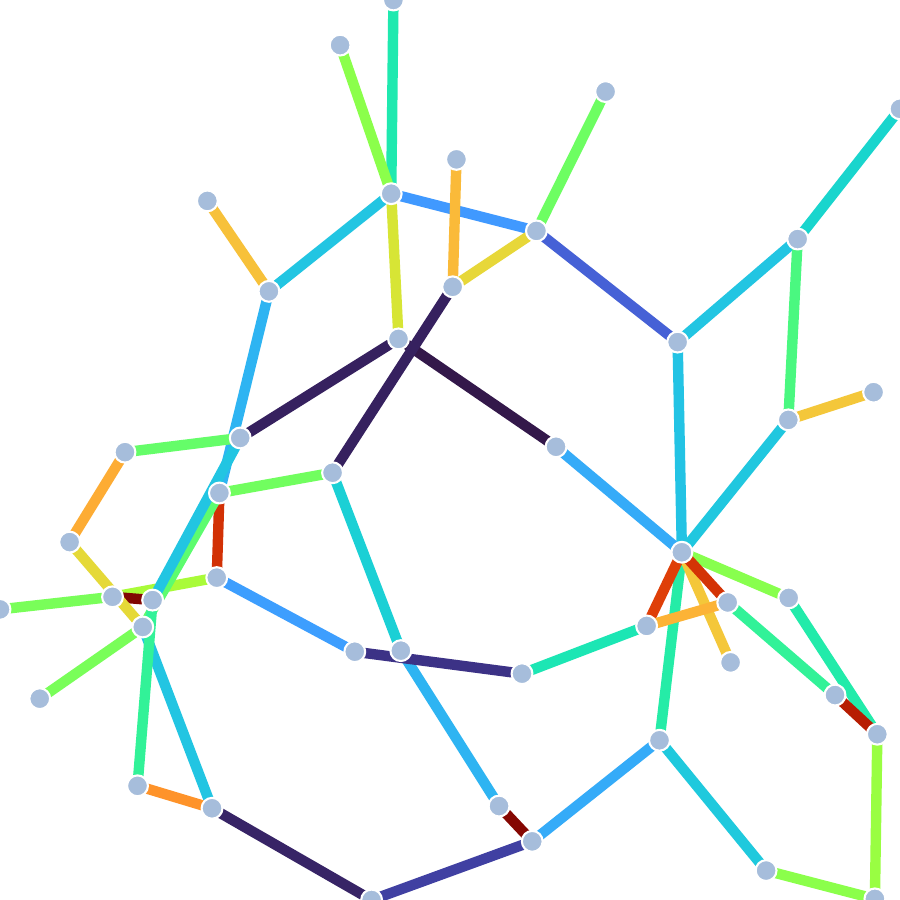} \\ \vspace{-0.0cm} \fontsize{7pt}{0pt}\selectfont -1.836,\textbf{-1.84}} & \parbox{1.7cm}{\centering \includegraphics[height=1.5cm]{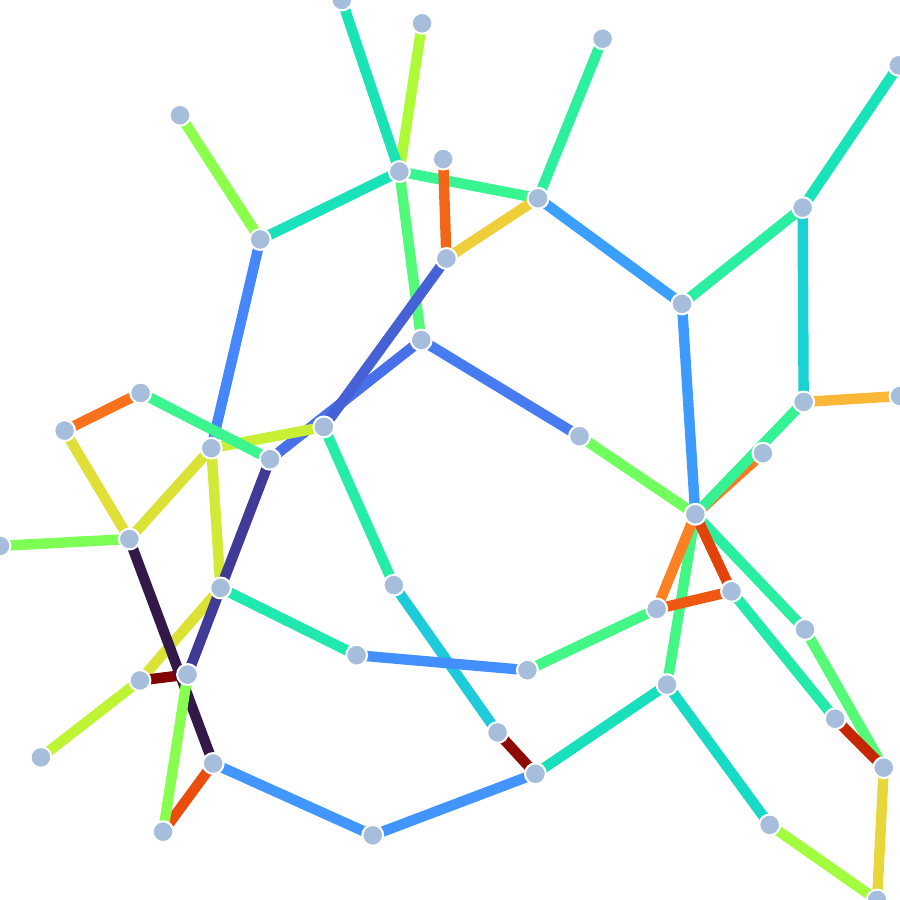} \\ \vspace{-0.0cm} \fontsize{7pt}{0pt}\selectfont \textbf{0.063},0.069} & \parbox{1.7cm}{\centering \includegraphics[height=1.5cm]{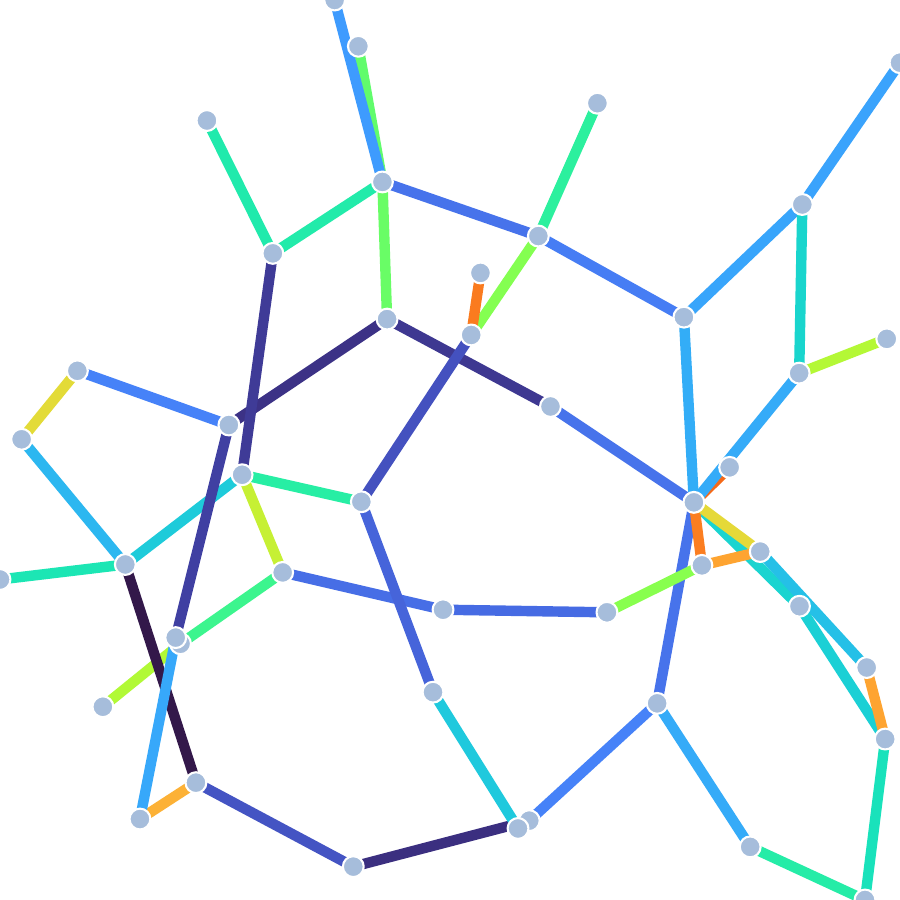} \\ \vspace{-0.0cm} \fontsize{7pt}{0pt}\selectfont 0.886,\textbf{0.816}} & \parbox{1.7cm}{\centering \includegraphics[height=1.5cm]{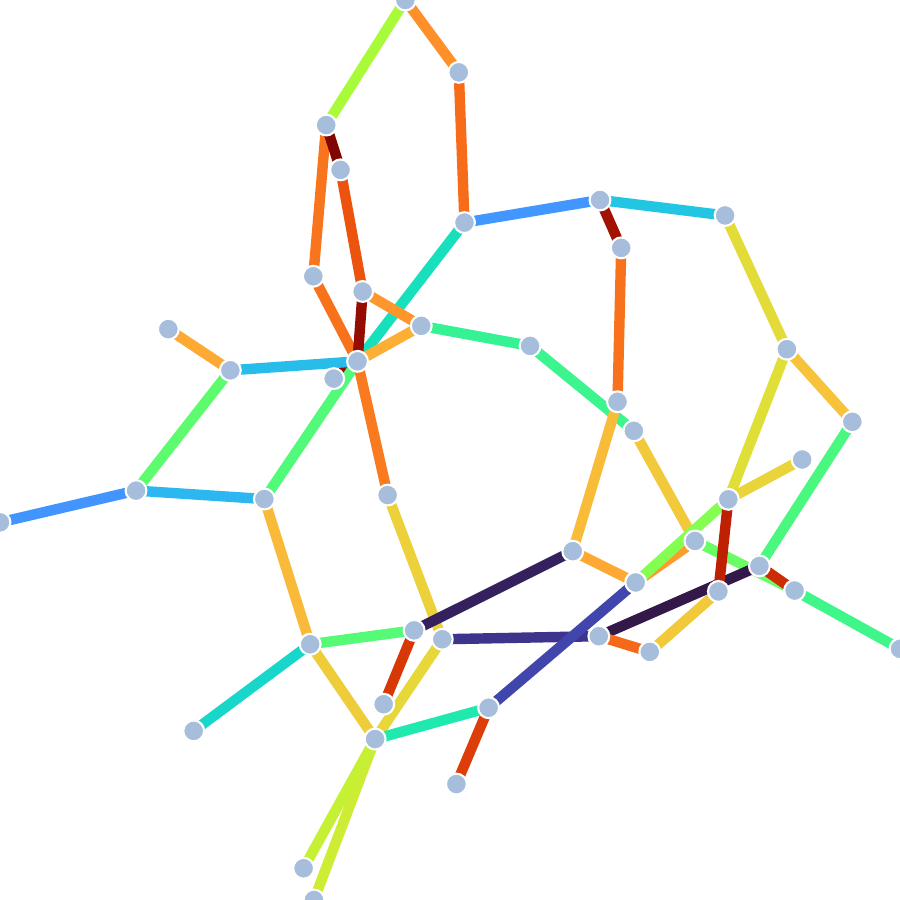} \\ \vspace{-0.0cm} \fontsize{7pt}{0pt}\selectfont 15,\textbf{8}} \\
grafo762.27 & \parbox{1.7cm}{\centering \includegraphics[height=1.5cm]{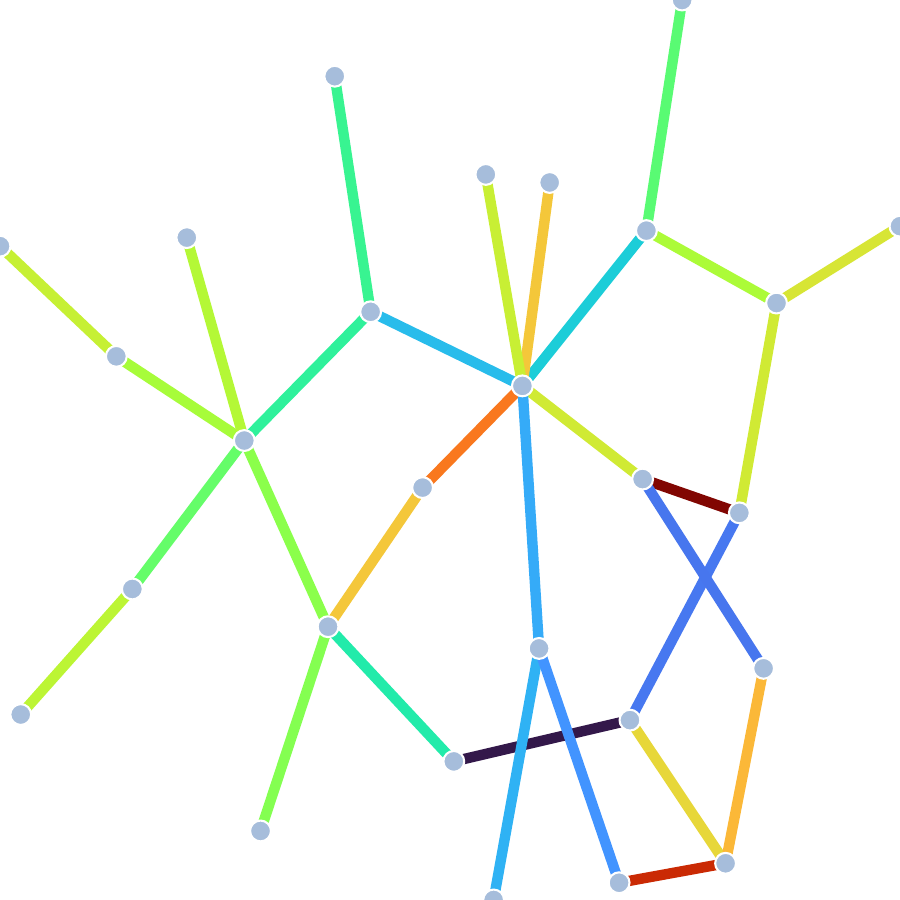} \\ \vspace{-0.0cm} \fontsize{7pt}{0pt}\selectfont } & \parbox{1.7cm}{\centering \includegraphics[height=1.5cm]{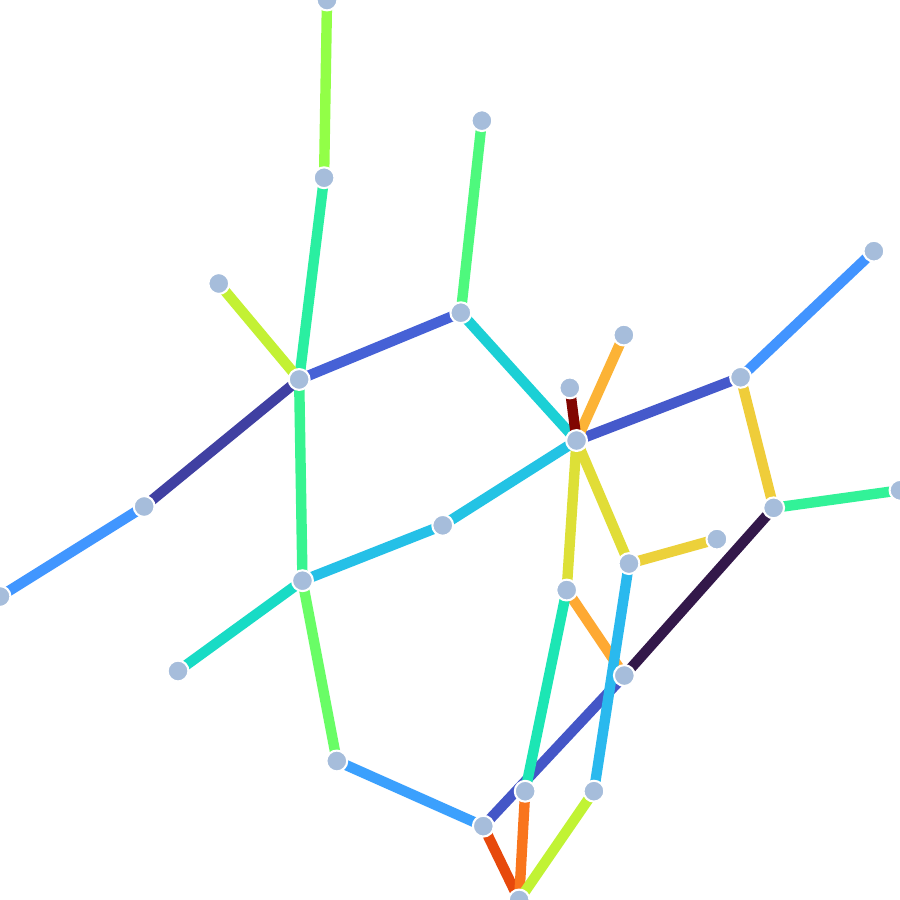} \\ \vspace{-0.0cm} \fontsize{7pt}{0pt}\selectfont 0.807,\textbf{0.568}} & \parbox{1.7cm}{\centering \includegraphics[height=1.5cm]{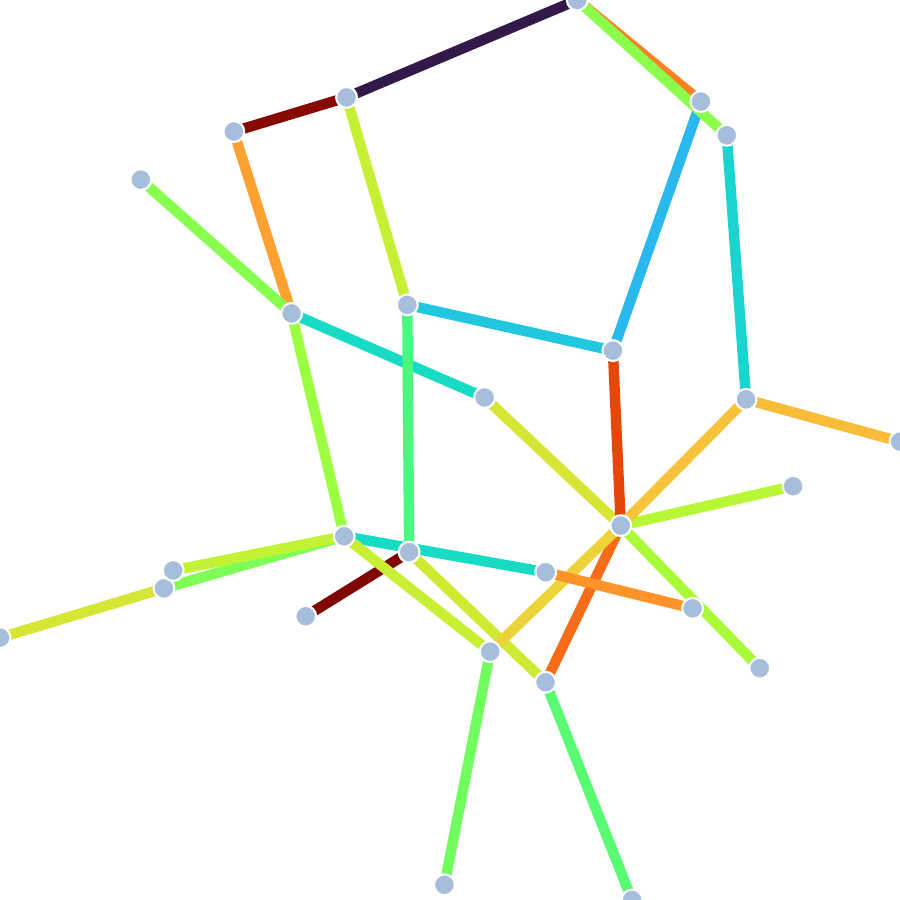} \\ \vspace{-0.0cm} \fontsize{7pt}{0pt}\selectfont \textbf{0.123},0.157} & \parbox{1.7cm}{\centering \includegraphics[height=1.5cm]{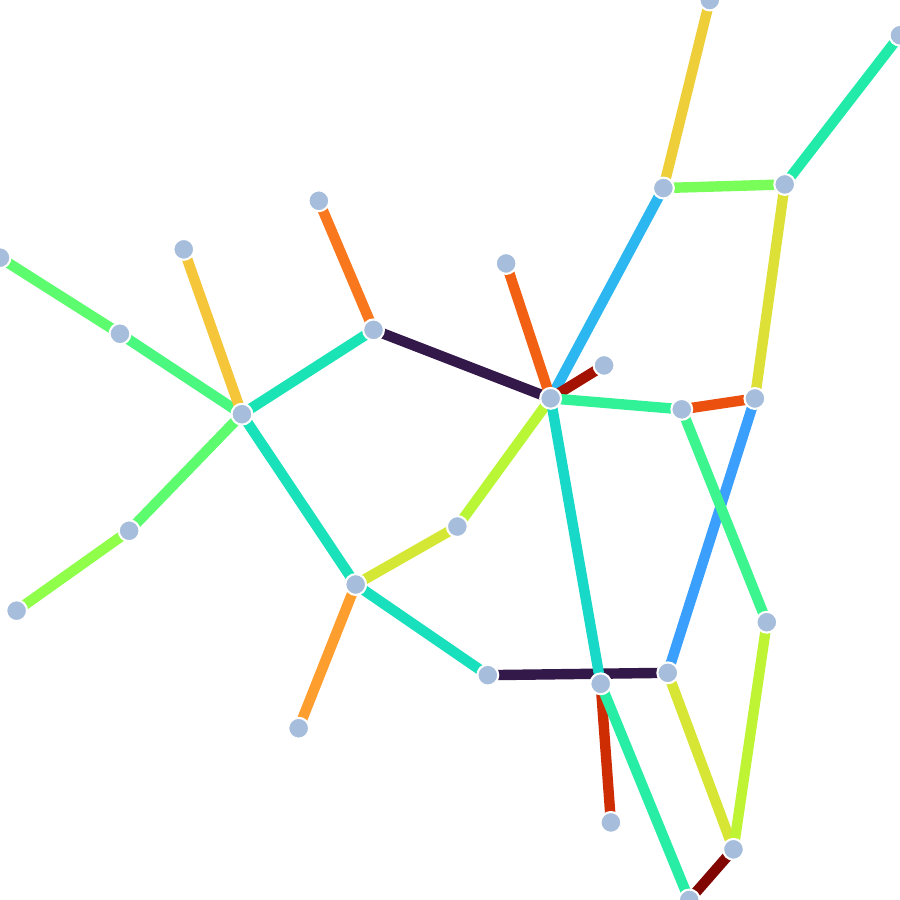} \\ \vspace{-0.0cm} \fontsize{7pt}{0pt}\selectfont -1.544,\textbf{-1.567}} & \parbox{1.7cm}{\centering \includegraphics[height=1.5cm]{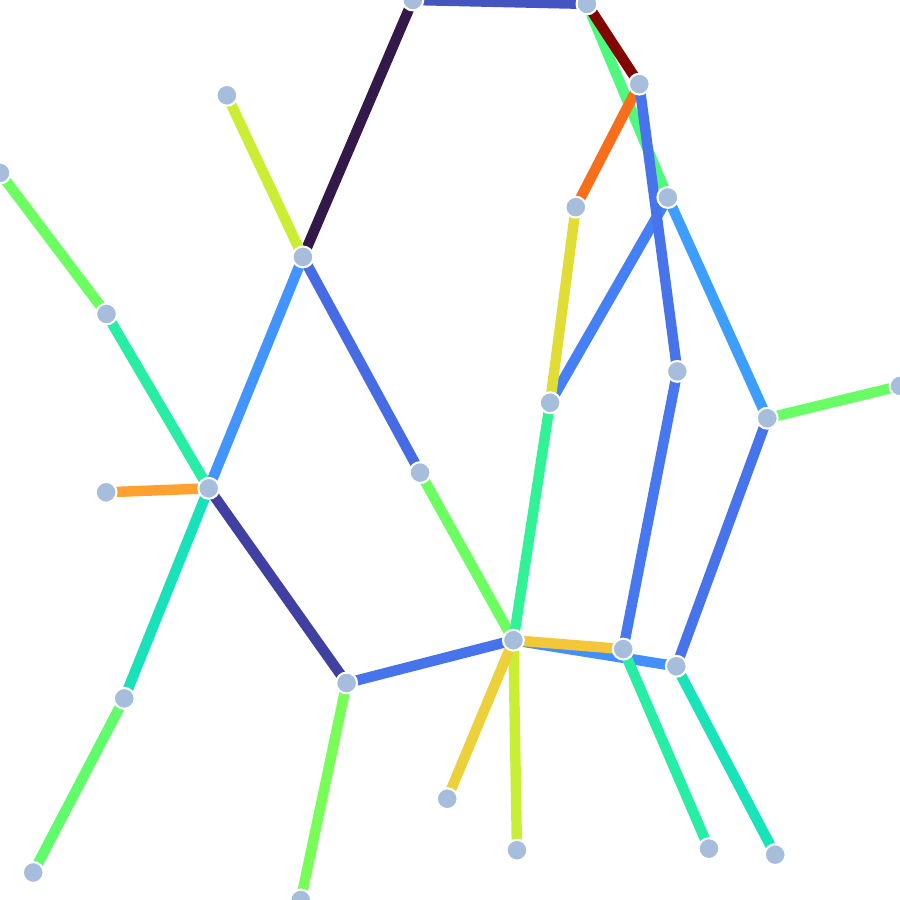} \\ \vspace{-0.0cm} \fontsize{7pt}{0pt}\selectfont \textbf{0.052},0.057} & \parbox{1.7cm}{\centering \includegraphics[height=1.5cm]{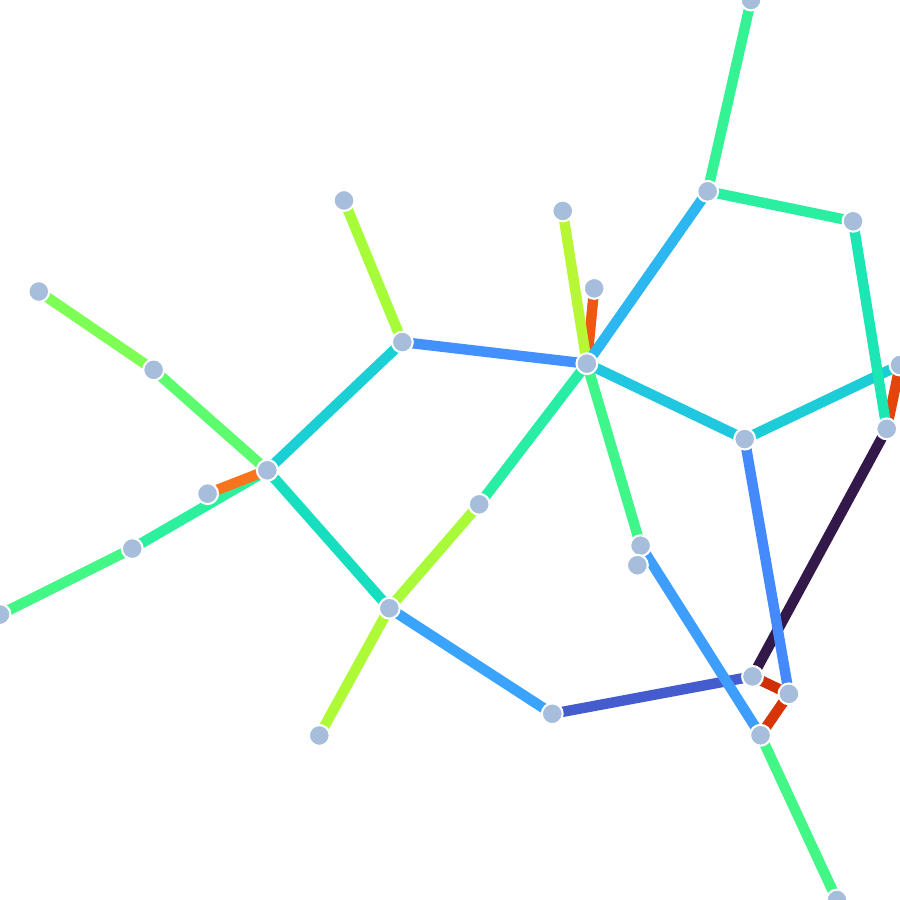} \\ \vspace{-0.0cm} \fontsize{7pt}{0pt}\selectfont 0.579,\textbf{0.554}} & \parbox{1.7cm}{\centering \includegraphics[height=1.5cm]{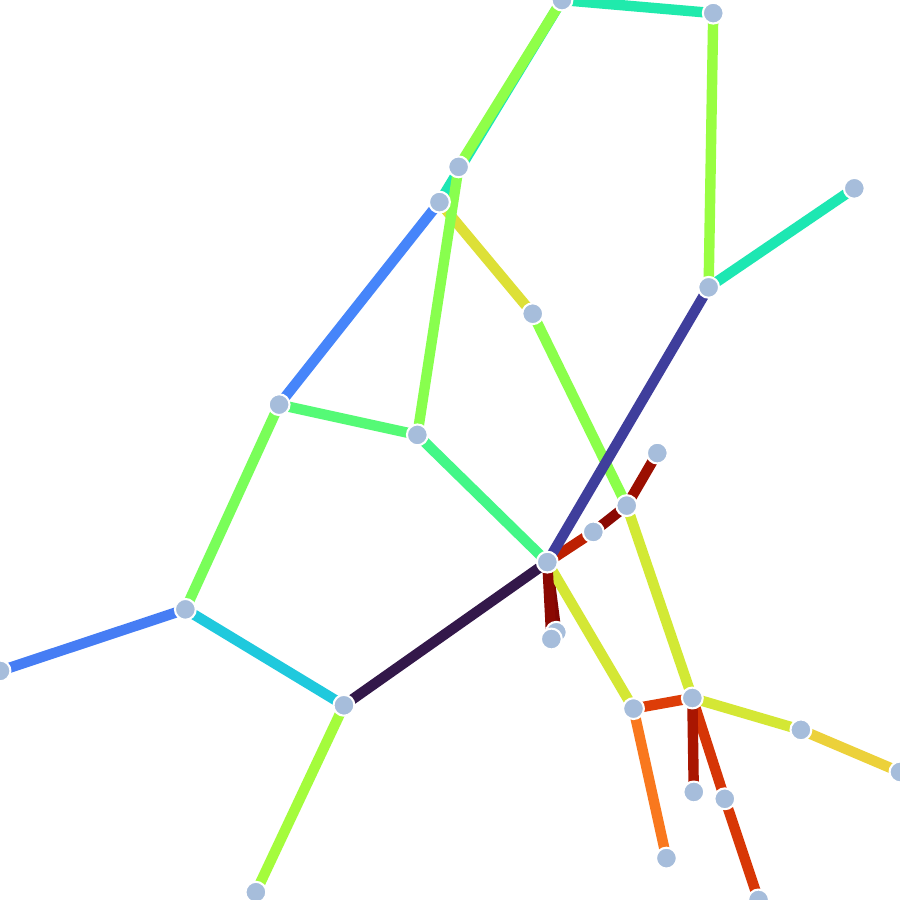} \\ \vspace{-0.0cm} \fontsize{7pt}{0pt}\selectfont 3,3} \\
grafo8296.86 & \parbox{1.7cm}{\centering \includegraphics[height=1.5cm]{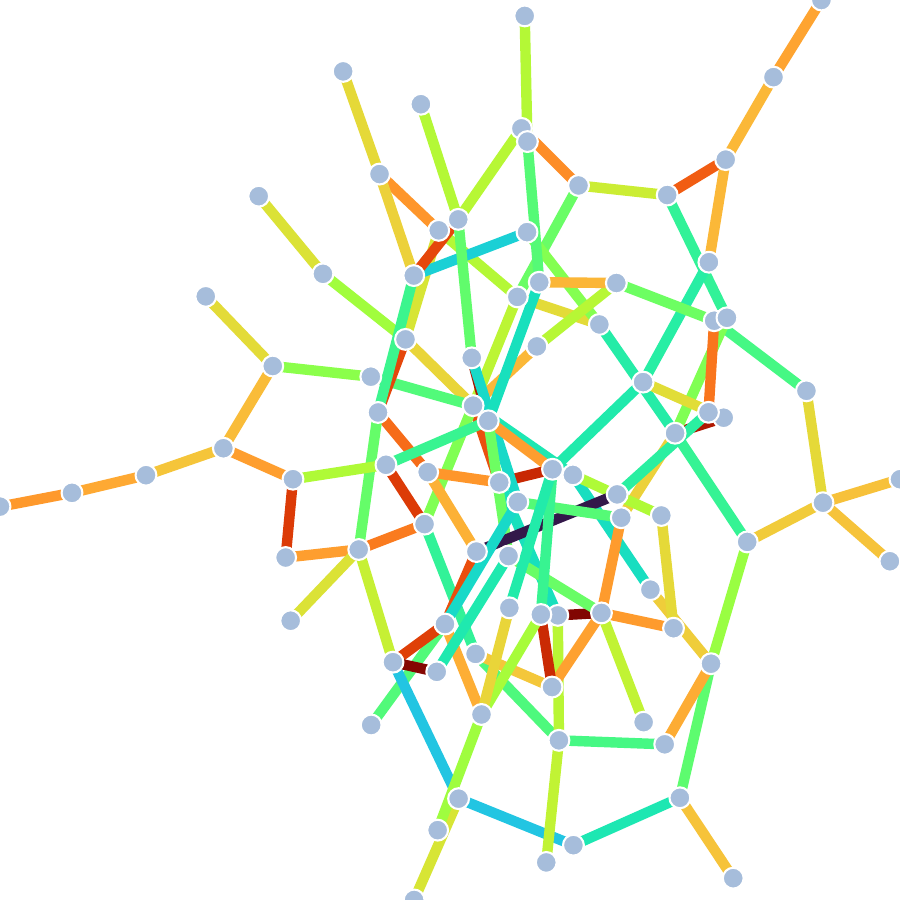} \\ \vspace{-0.0cm} \fontsize{7pt}{0pt}\selectfont } & \parbox{1.7cm}{\centering \includegraphics[height=1.5cm]{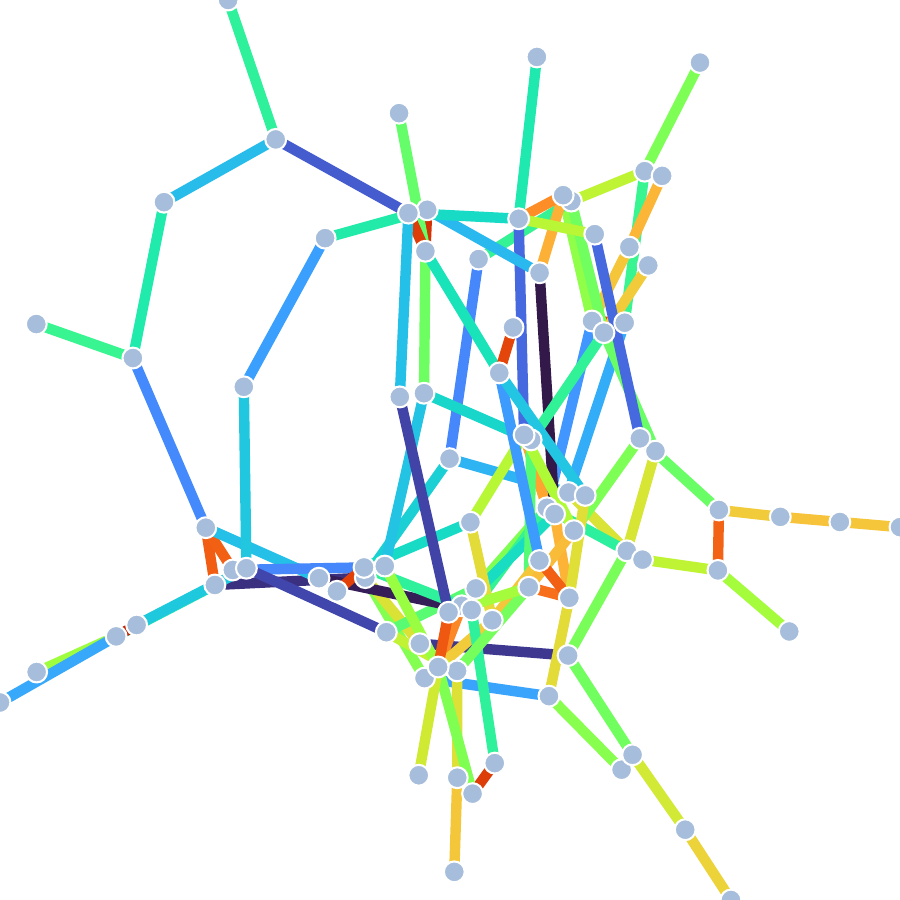} \\ \vspace{-0.0cm} \fontsize{7pt}{0pt}\selectfont 0.964,\textbf{0.771}} & \parbox{1.7cm}{\centering \includegraphics[height=1.5cm]{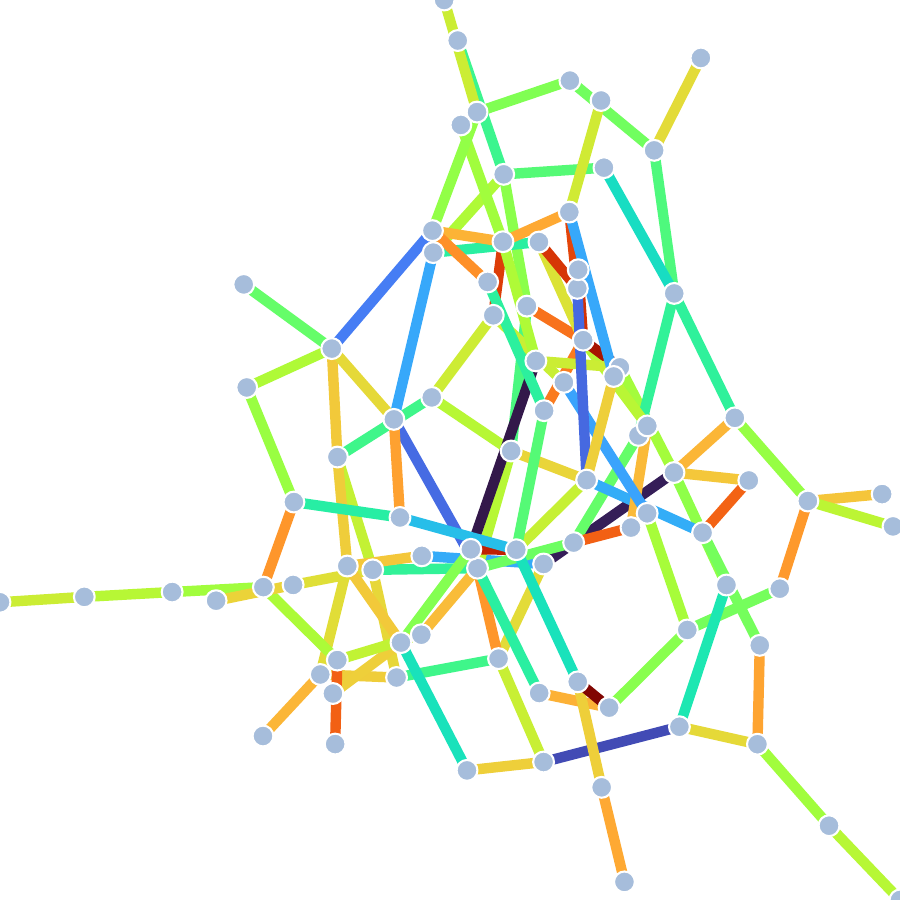} \\ \vspace{-0.0cm} \fontsize{7pt}{0pt}\selectfont \textbf{0.214},0.24} & \parbox{1.7cm}{\centering \includegraphics[height=1.5cm]{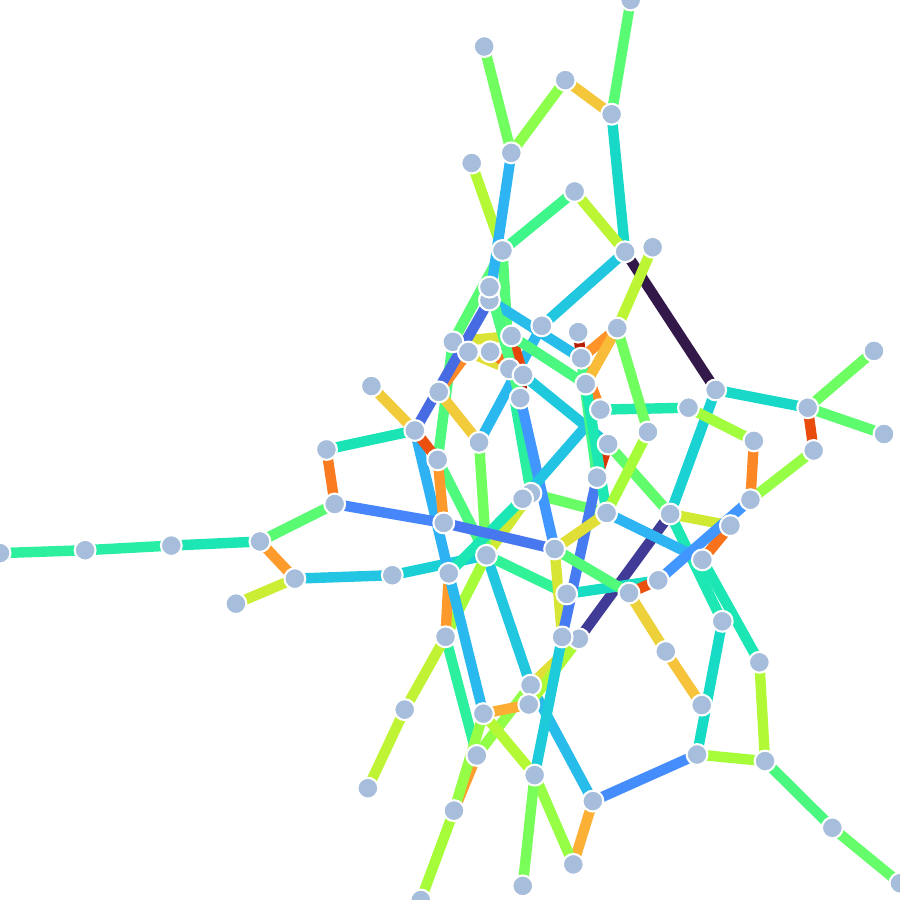} \\ \vspace{-0.0cm} \fontsize{7pt}{0pt}\selectfont -2.071,\textbf{-2.079}} & \parbox{1.7cm}{\centering \includegraphics[height=1.5cm]{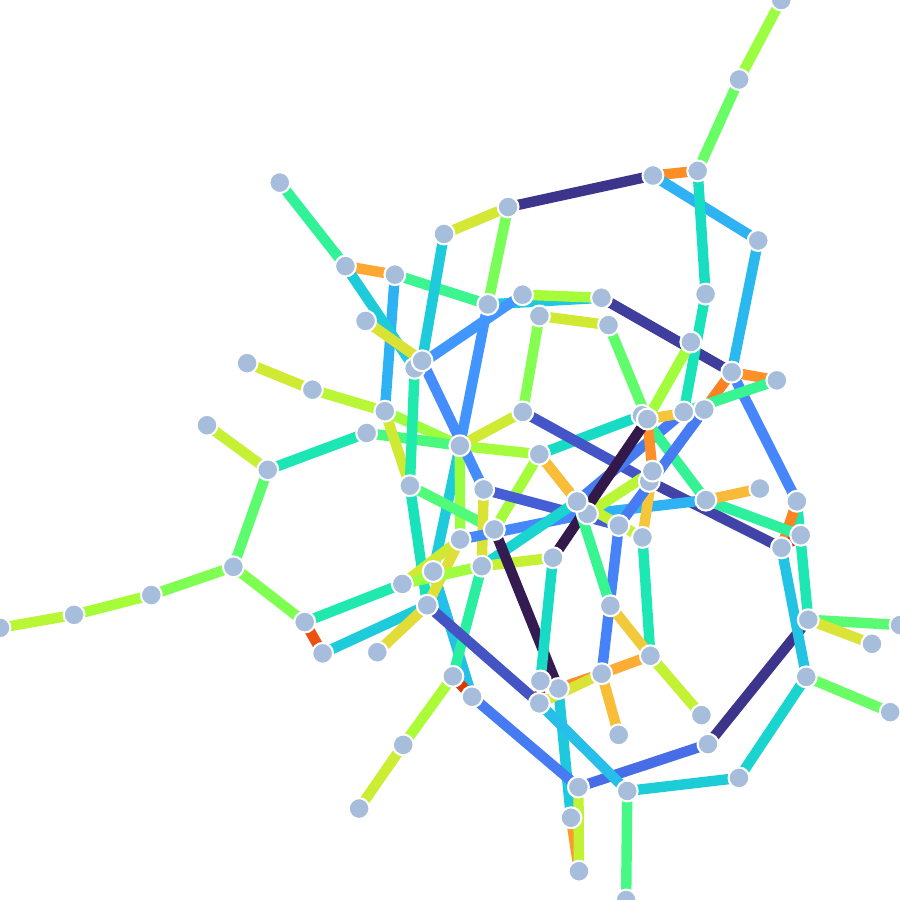} \\ \vspace{-0.0cm} \fontsize{7pt}{0pt}\selectfont \textbf{0.092},0.102} & \parbox{1.7cm}{\centering \includegraphics[height=1.5cm]{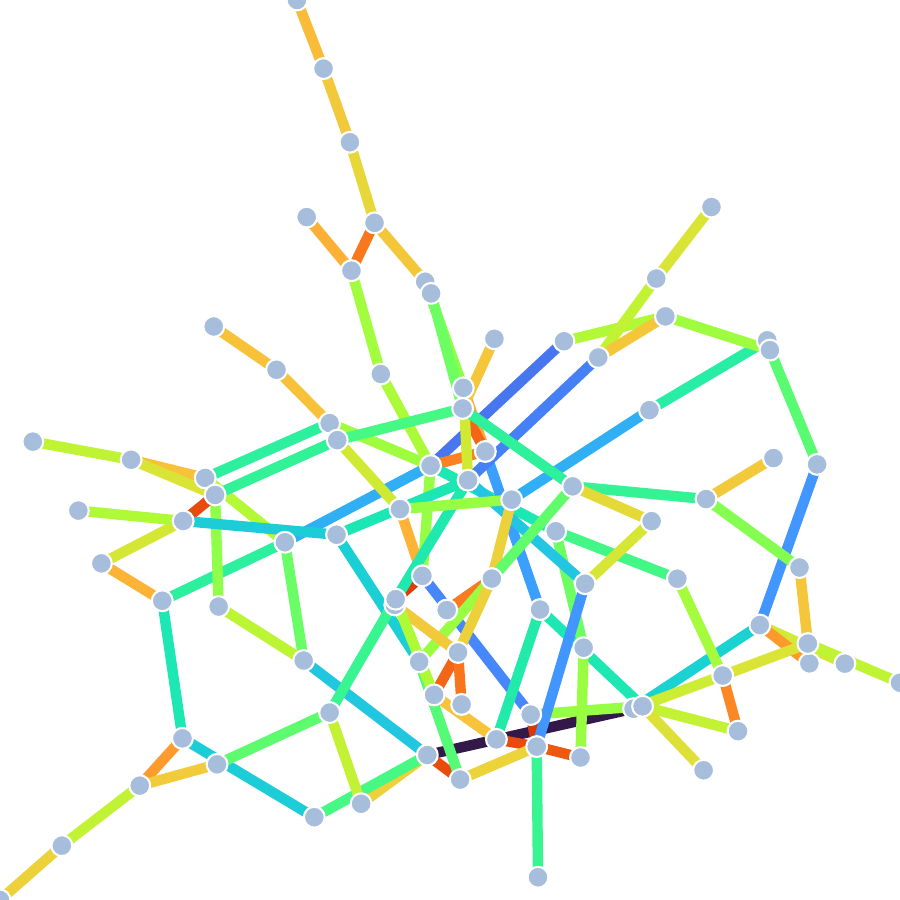} \\ \vspace{-0.0cm} \fontsize{7pt}{0pt}\selectfont 1.361,\textbf{1.257}} & \parbox{1.7cm}{\centering \includegraphics[height=1.5cm]{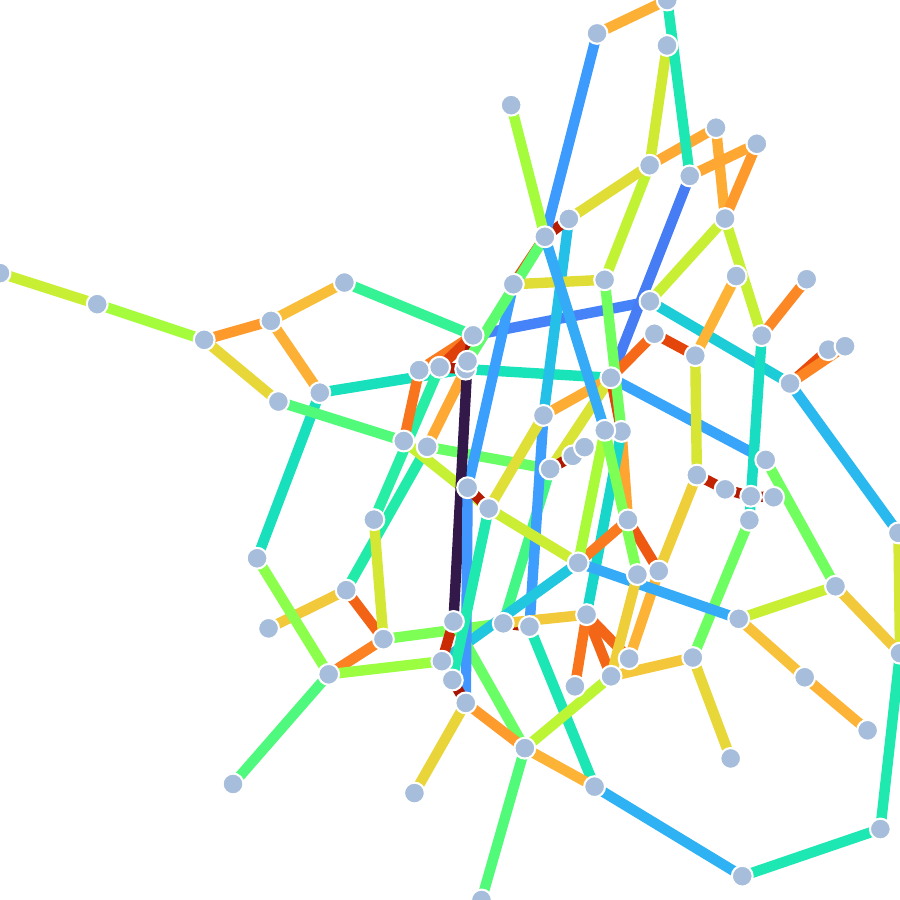} \\ \vspace{-0.0cm} \fontsize{7pt}{0pt}\selectfont 81,\textbf{66}} \\
grafo847.22 & \parbox{1.7cm}{\centering \includegraphics[height=1.5cm]{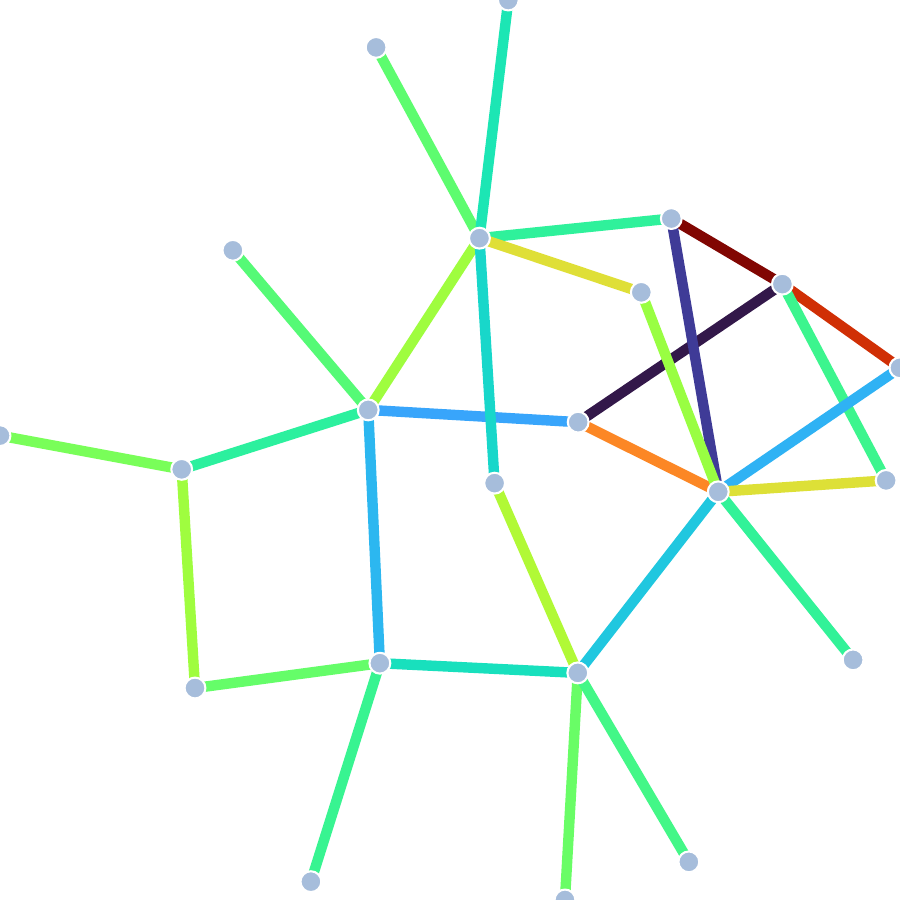} \\ \vspace{-0.0cm} \fontsize{7pt}{0pt}\selectfont } & \parbox{1.7cm}{\centering \includegraphics[height=1.5cm]{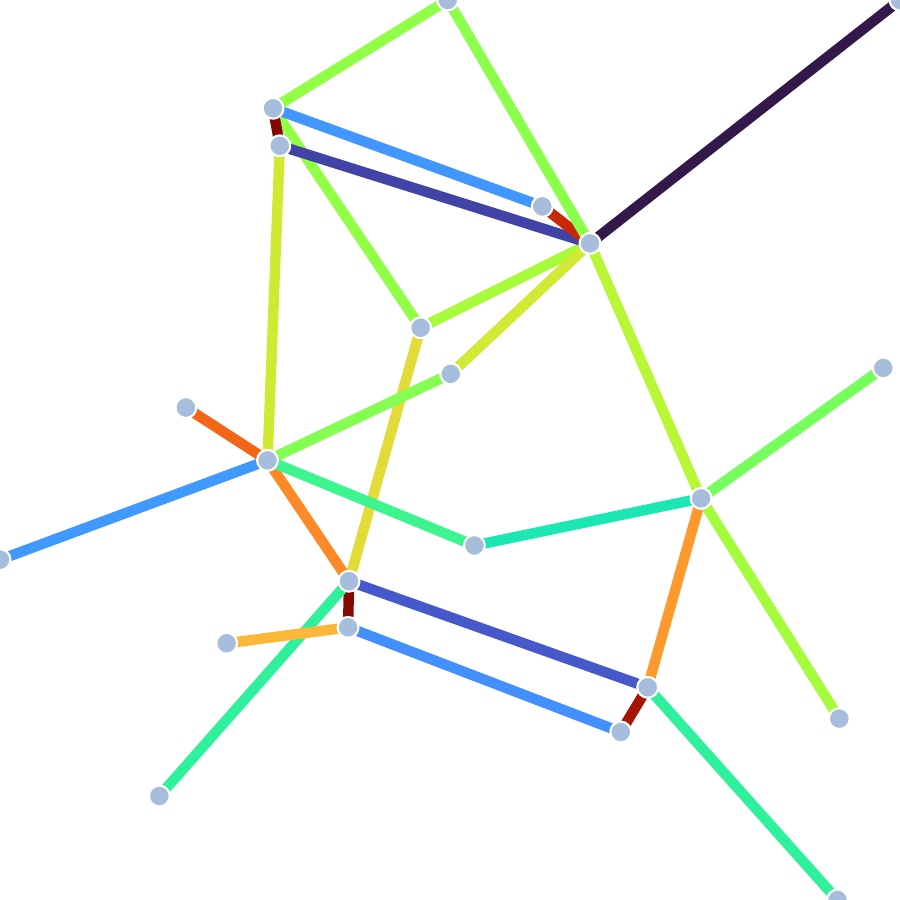} \\ \vspace{-0.0cm} \fontsize{7pt}{0pt}\selectfont 1.136,\textbf{0.622}} & \parbox{1.7cm}{\centering \includegraphics[height=1.5cm]{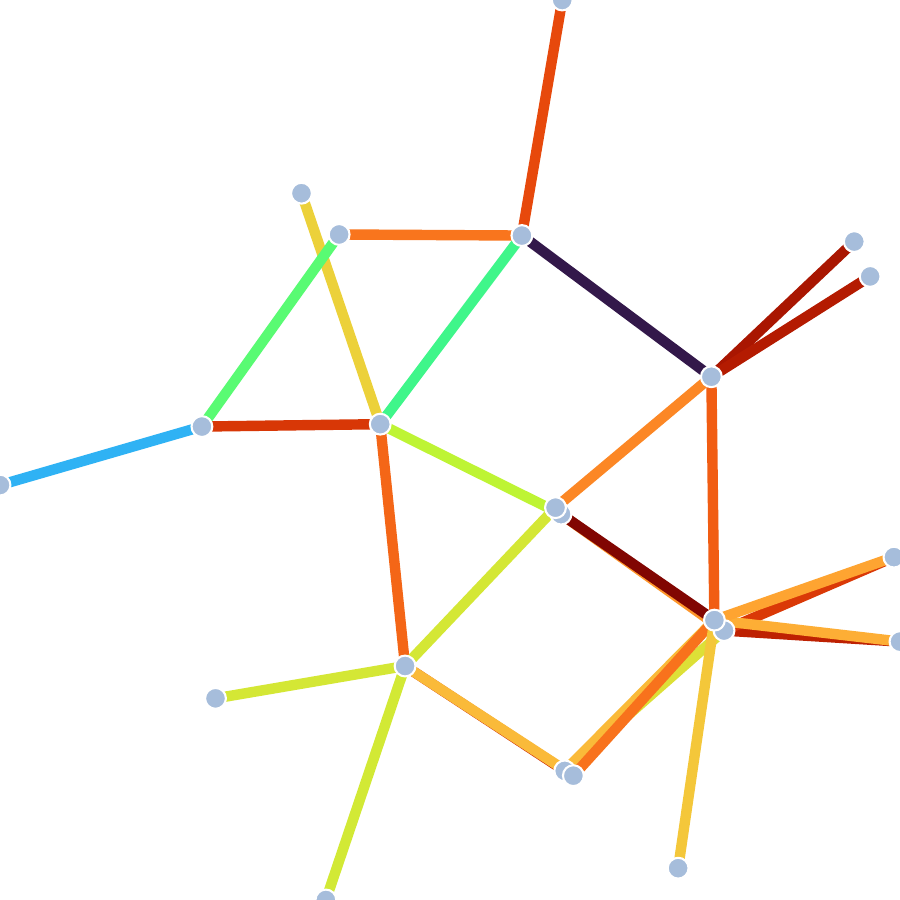} \\ \vspace{-0.0cm} \fontsize{7pt}{0pt}\selectfont 0.121,\textbf{0.051}} & \parbox{1.7cm}{\centering \includegraphics[height=1.5cm]{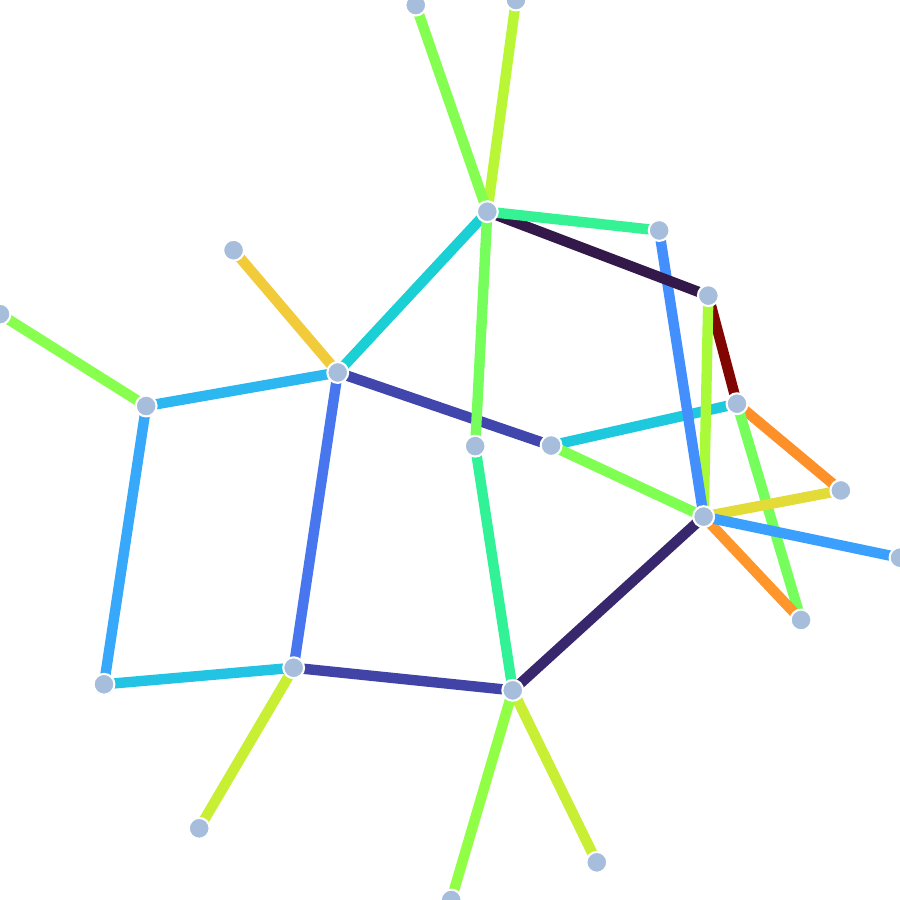} \\ \vspace{-0.0cm} \fontsize{7pt}{0pt}\selectfont \textbf{-1.257},-1.257} & \parbox{1.7cm}{\centering \includegraphics[height=1.5cm]{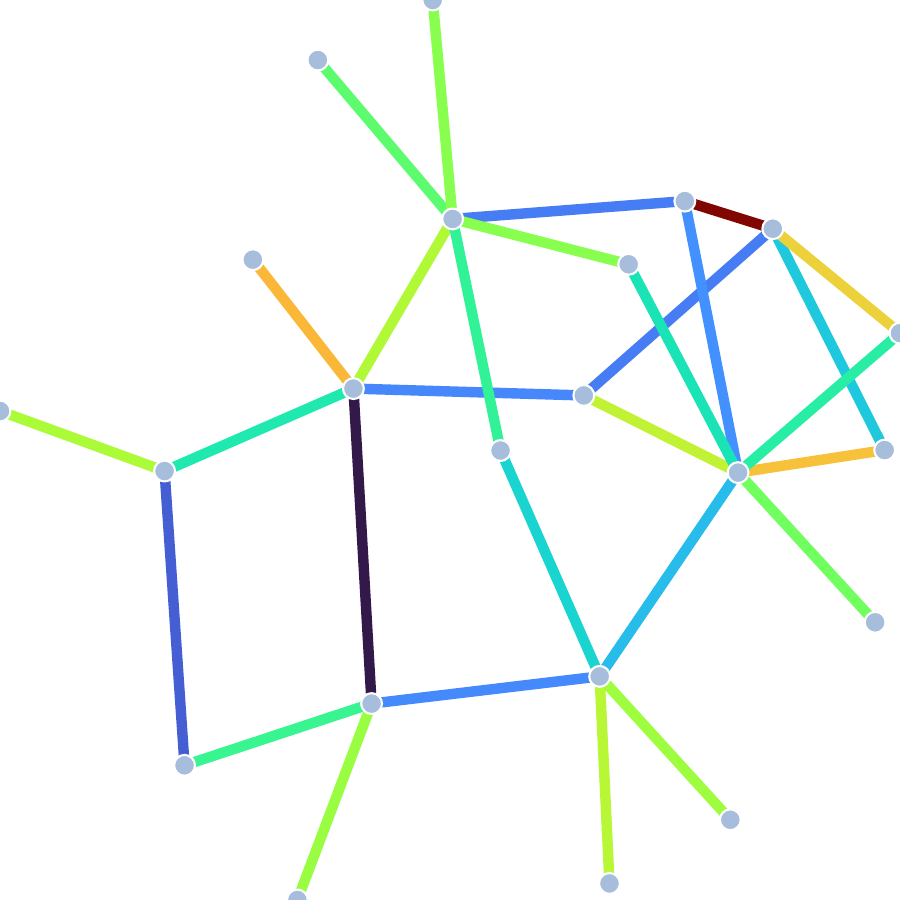} \\ \vspace{-0.0cm} \fontsize{7pt}{0pt}\selectfont \textbf{0.051},0.056} & \parbox{1.7cm}{\centering \includegraphics[height=1.5cm]{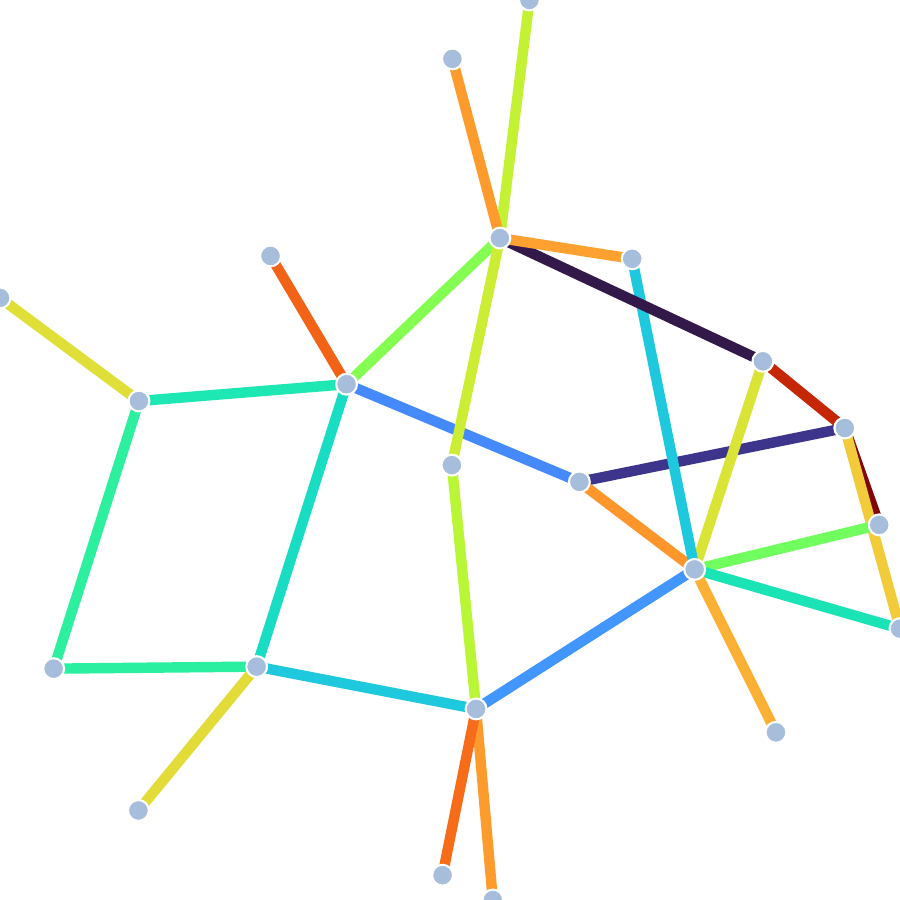} \\ \vspace{-0.0cm} \fontsize{7pt}{0pt}\selectfont 0.421,\textbf{0.373}} & \parbox{1.7cm}{\centering \includegraphics[height=1.5cm]{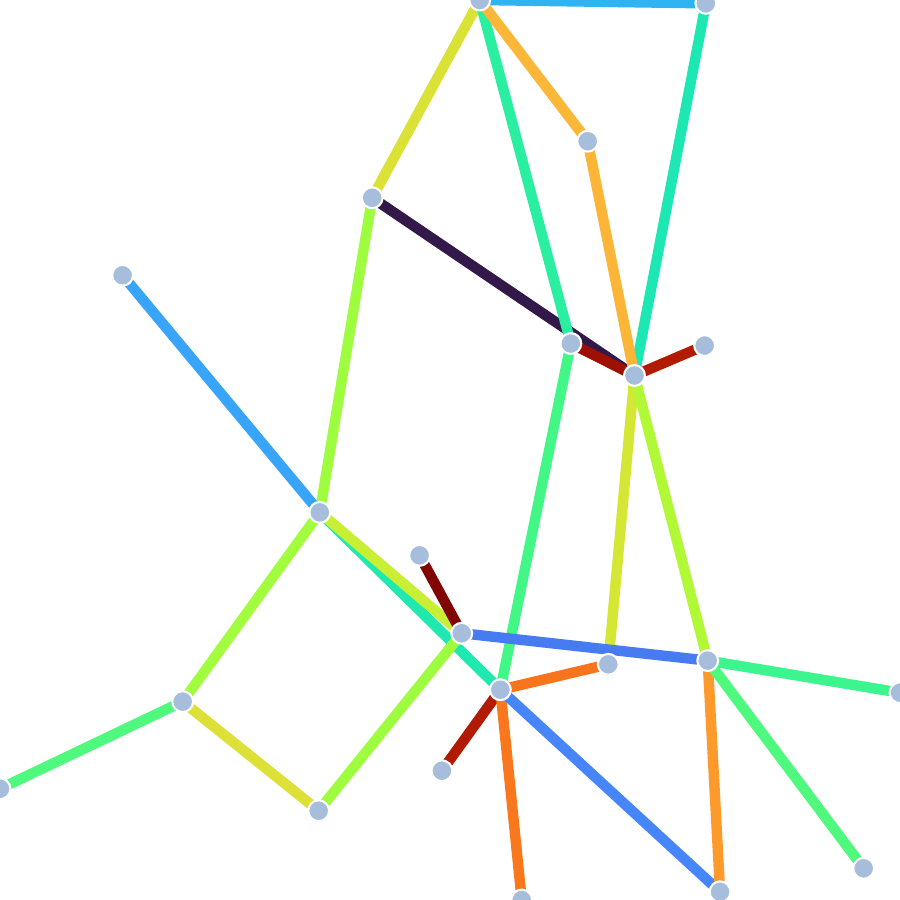} \\ \vspace{-0.0cm} \fontsize{7pt}{0pt}\selectfont 4,4} \\
grafo9033.80 & \parbox{1.7cm}{\centering \includegraphics[height=1.5cm]{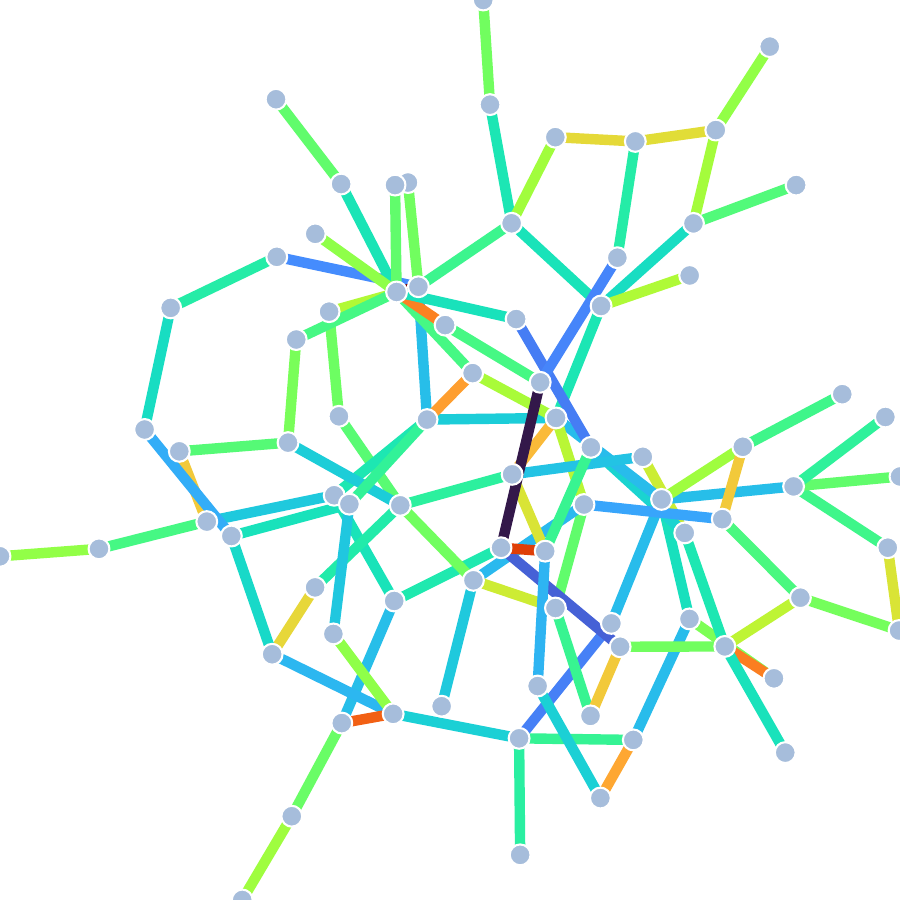} \\ \vspace{-0.0cm} \fontsize{7pt}{0pt}\selectfont } & \parbox{1.7cm}{\centering \includegraphics[height=1.5cm]{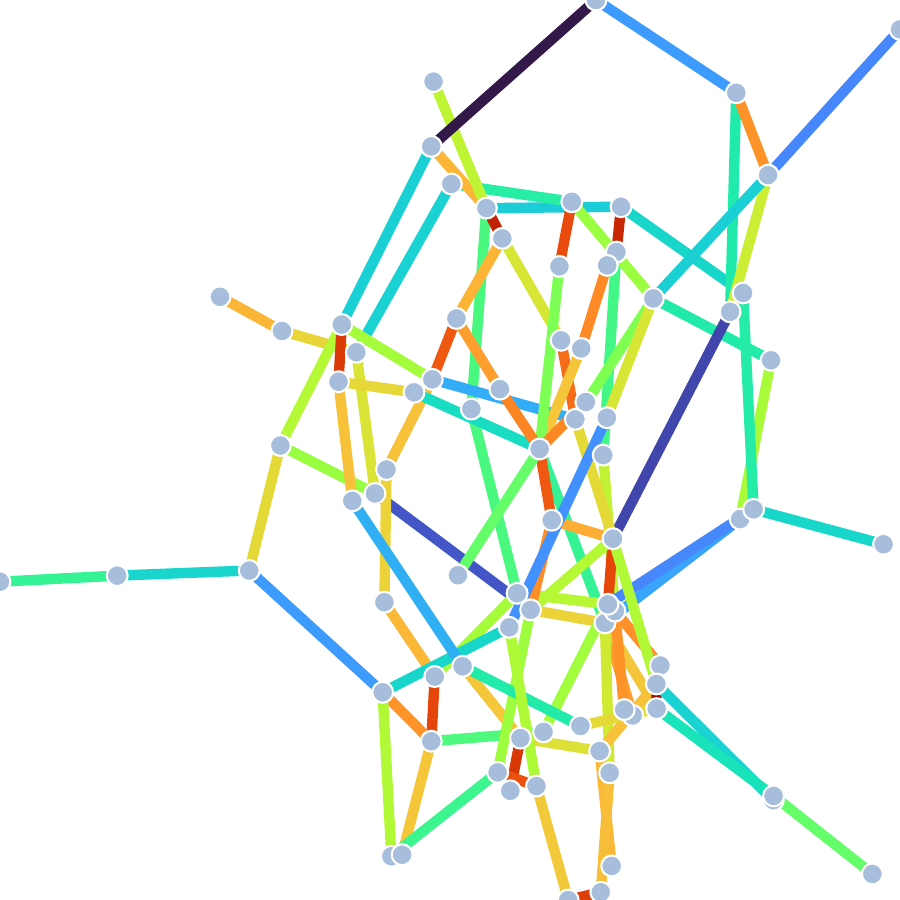} \\ \vspace{-0.0cm} \fontsize{7pt}{0pt}\selectfont 1.025,\textbf{0.778}} & \parbox{1.7cm}{\centering \includegraphics[height=1.5cm]{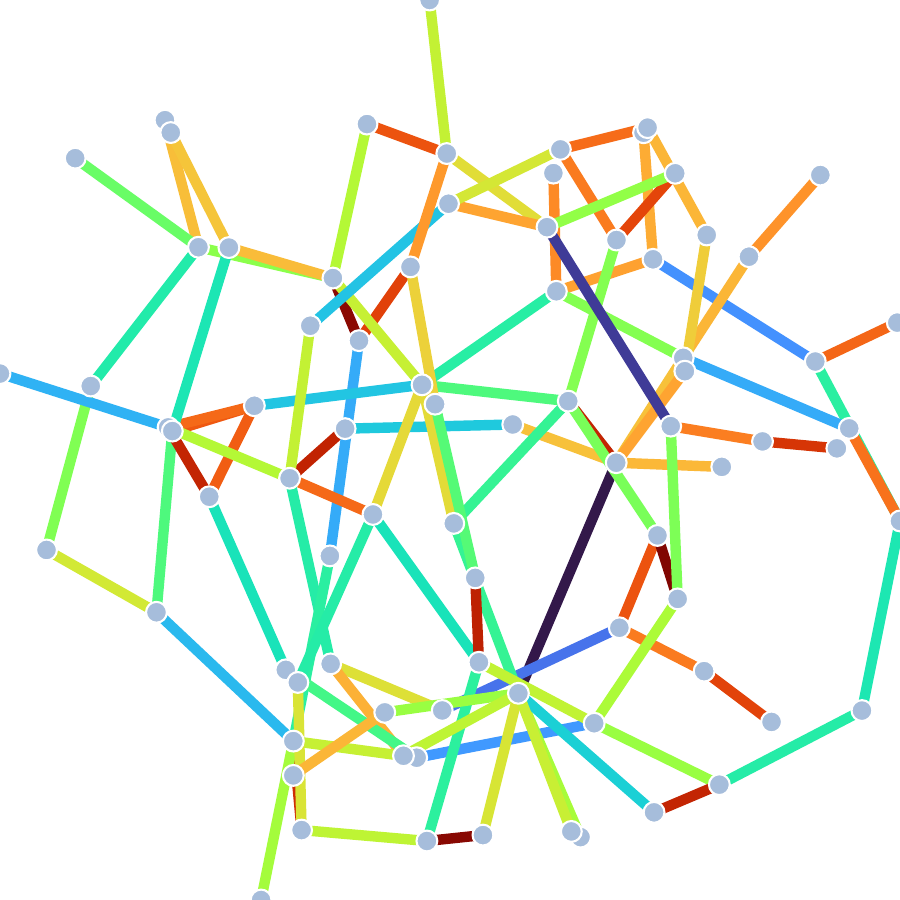} \\ \vspace{-0.0cm} \fontsize{7pt}{0pt}\selectfont \textbf{0.219},0.279} & \parbox{1.7cm}{\centering \includegraphics[height=1.5cm]{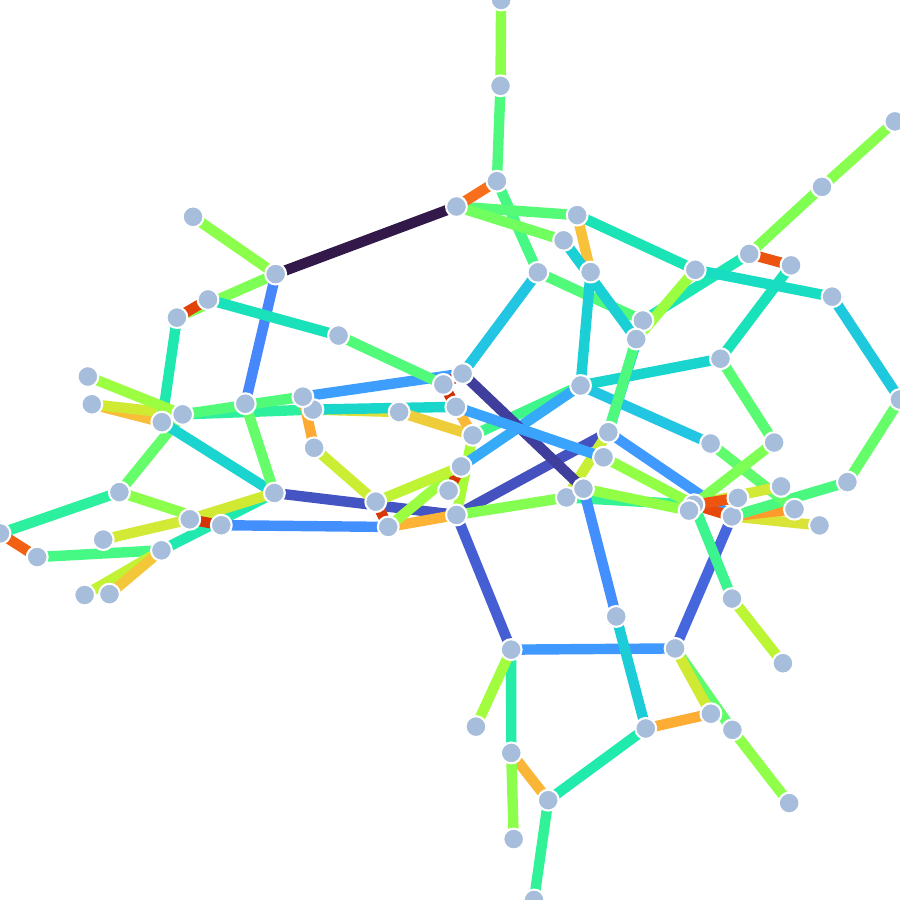} \\ \vspace{-0.0cm} \fontsize{7pt}{0pt}\selectfont -2.066,\textbf{-2.079}} & \parbox{1.7cm}{\centering \includegraphics[height=1.5cm]{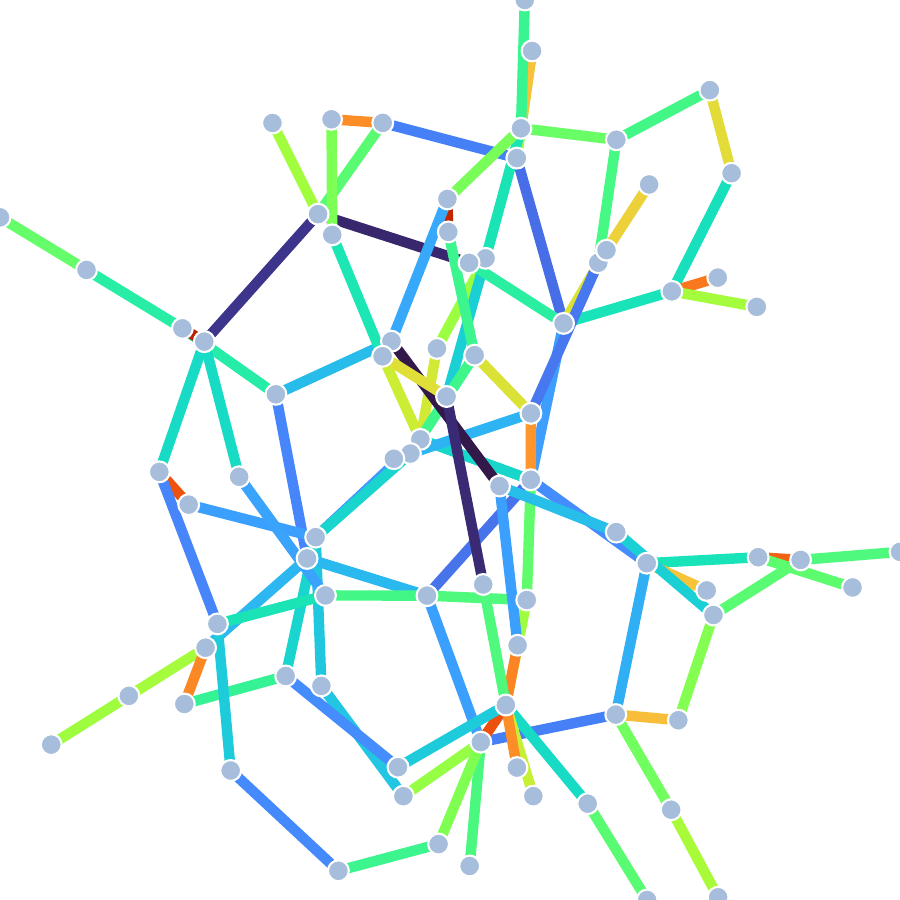} \\ \vspace{-0.0cm} \fontsize{7pt}{0pt}\selectfont \textbf{0.083},0.09} & \parbox{1.7cm}{\centering \includegraphics[height=1.5cm]{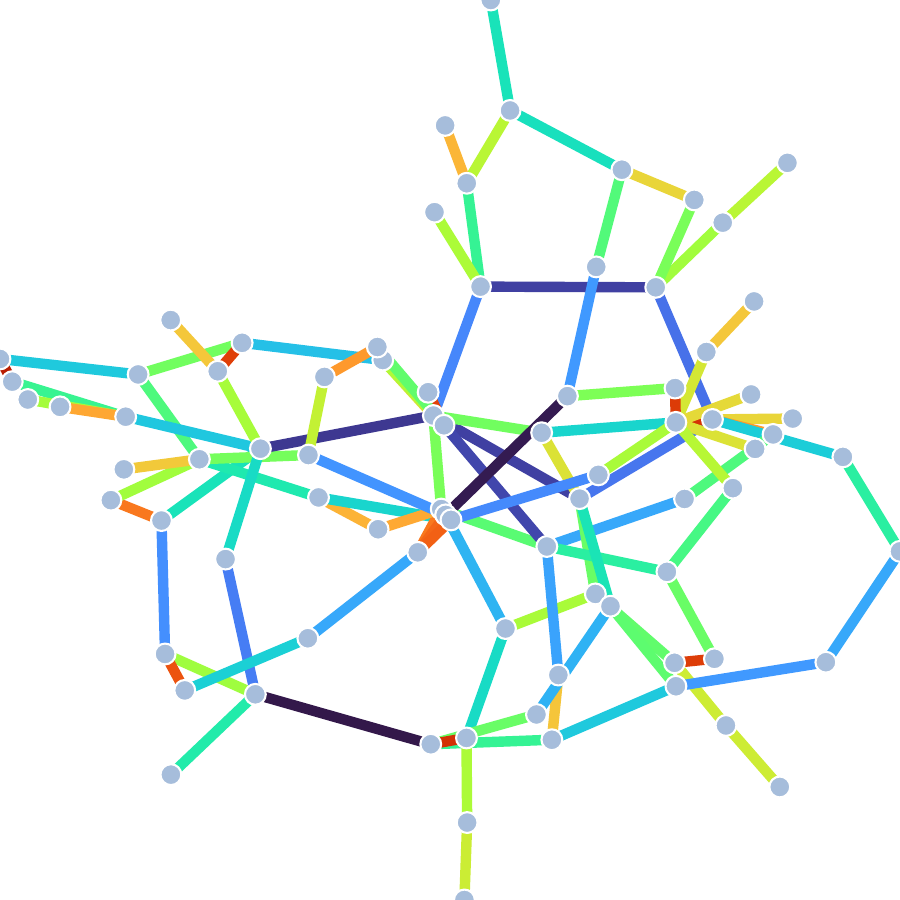} \\ \vspace{-0.0cm} \fontsize{7pt}{0pt}\selectfont 1.235,\textbf{1.138}} & \parbox{1.7cm}{\centering \includegraphics[height=1.5cm]{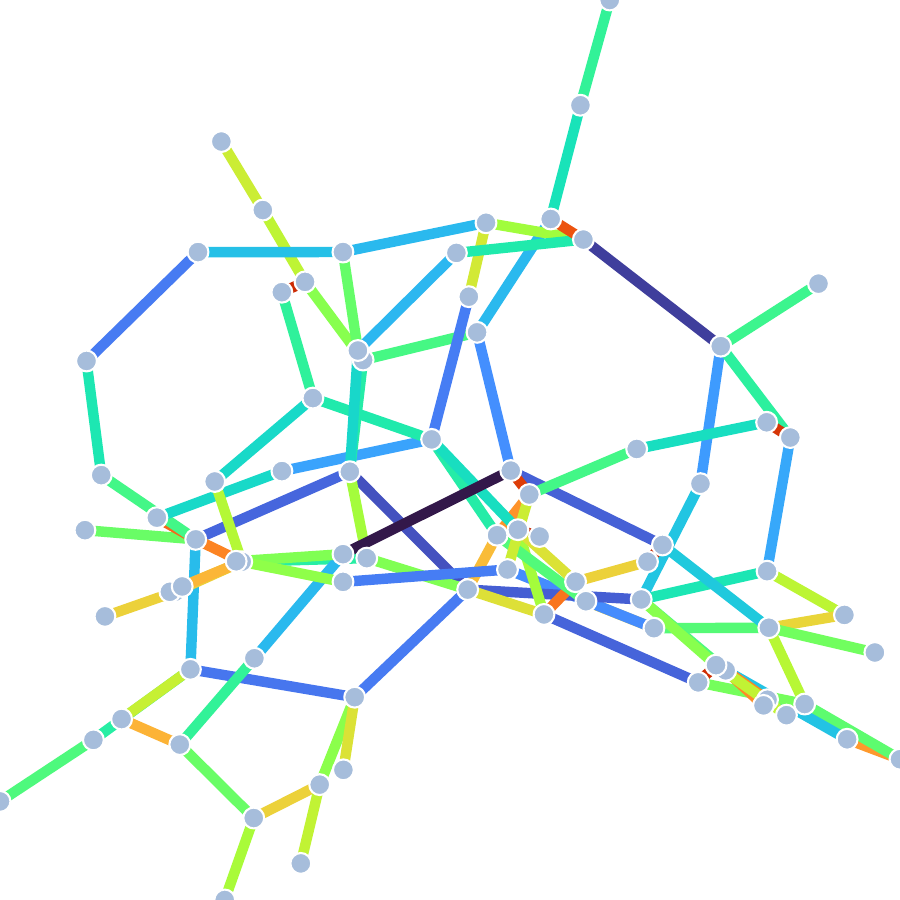} \\ \vspace{-0.0cm} \fontsize{7pt}{0pt}\selectfont 61,\textbf{54}} \\
grafo9592.72 & \parbox{1.7cm}{\centering \includegraphics[height=1.5cm]{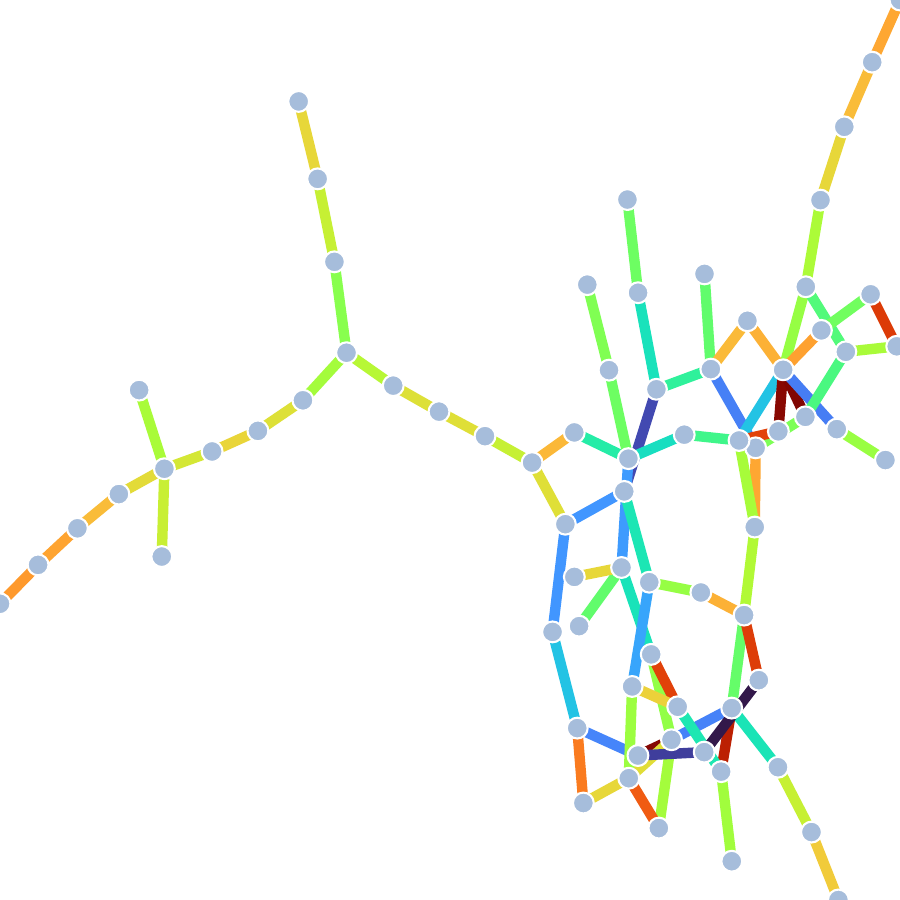} \\ \vspace{-0.0cm} \fontsize{7pt}{0pt}\selectfont } & \parbox{1.7cm}{\centering \includegraphics[height=1.5cm]{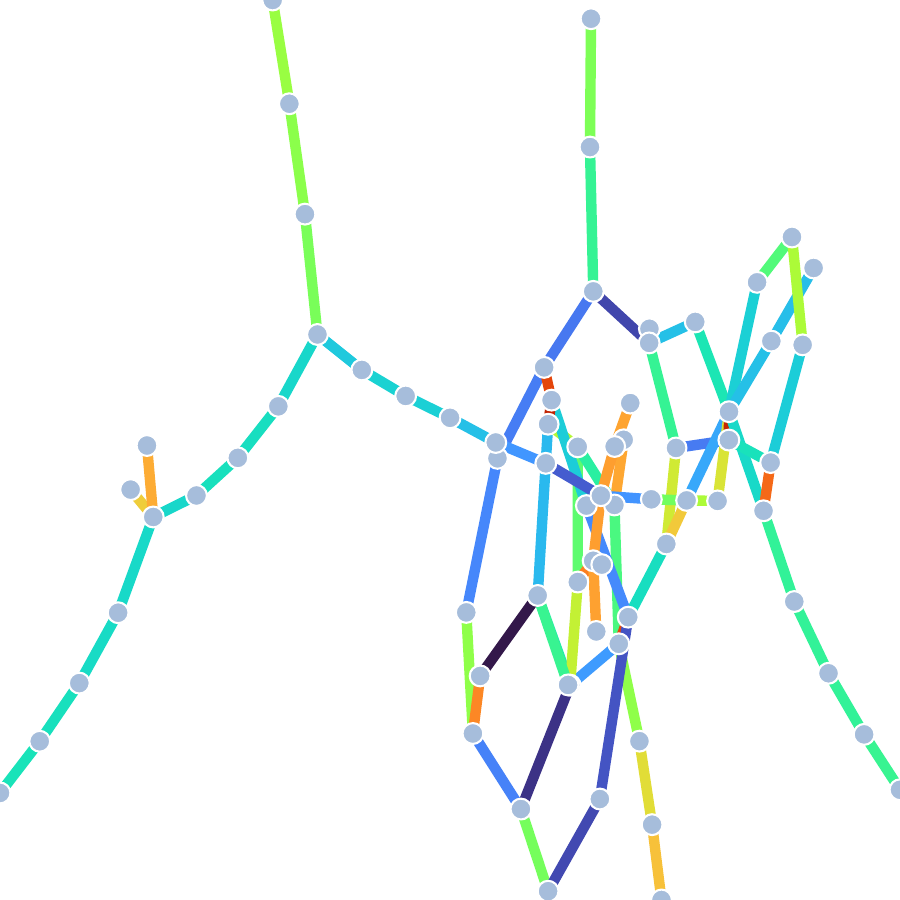} \\ \vspace{-0.0cm} \fontsize{7pt}{0pt}\selectfont 0.859,\textbf{0.692}} & \parbox{1.7cm}{\centering \includegraphics[height=1.5cm]{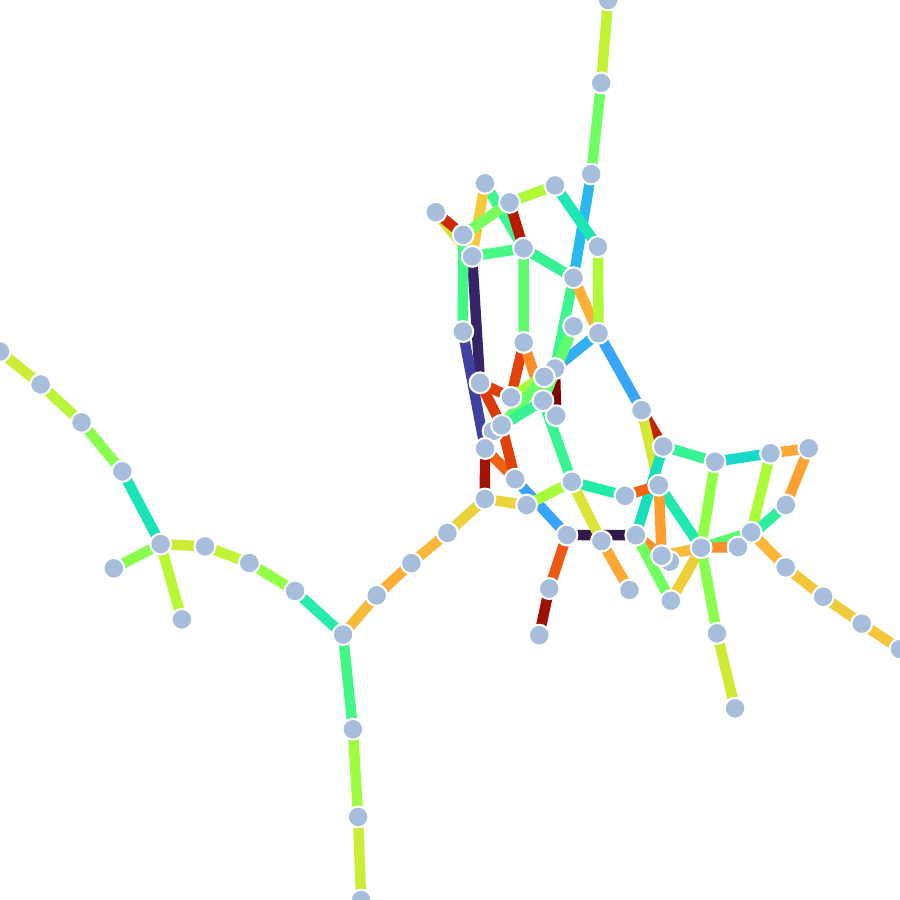} \\ \vspace{-0.0cm} \fontsize{7pt}{0pt}\selectfont \textbf{0.142},0.204} & \parbox{1.7cm}{\centering \includegraphics[height=1.5cm]{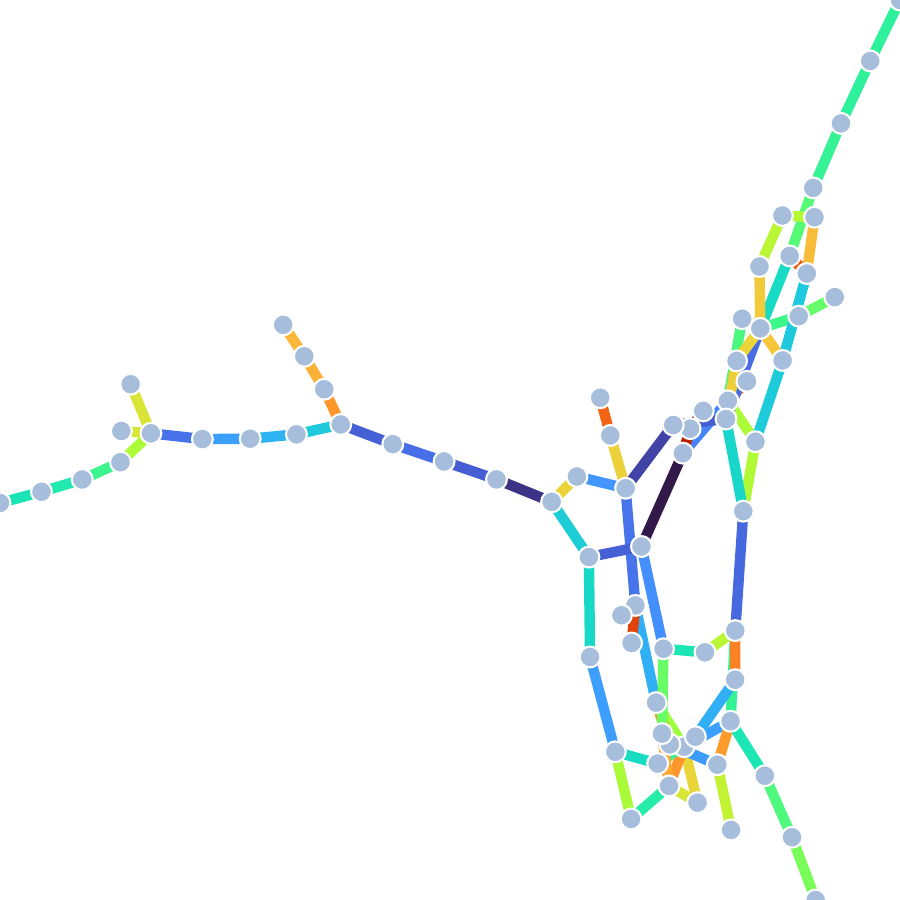} \\ \vspace{-0.0cm} \fontsize{7pt}{0pt}\selectfont -2.584,\textbf{-2.66}} & \parbox{1.7cm}{\centering \includegraphics[height=1.5cm]{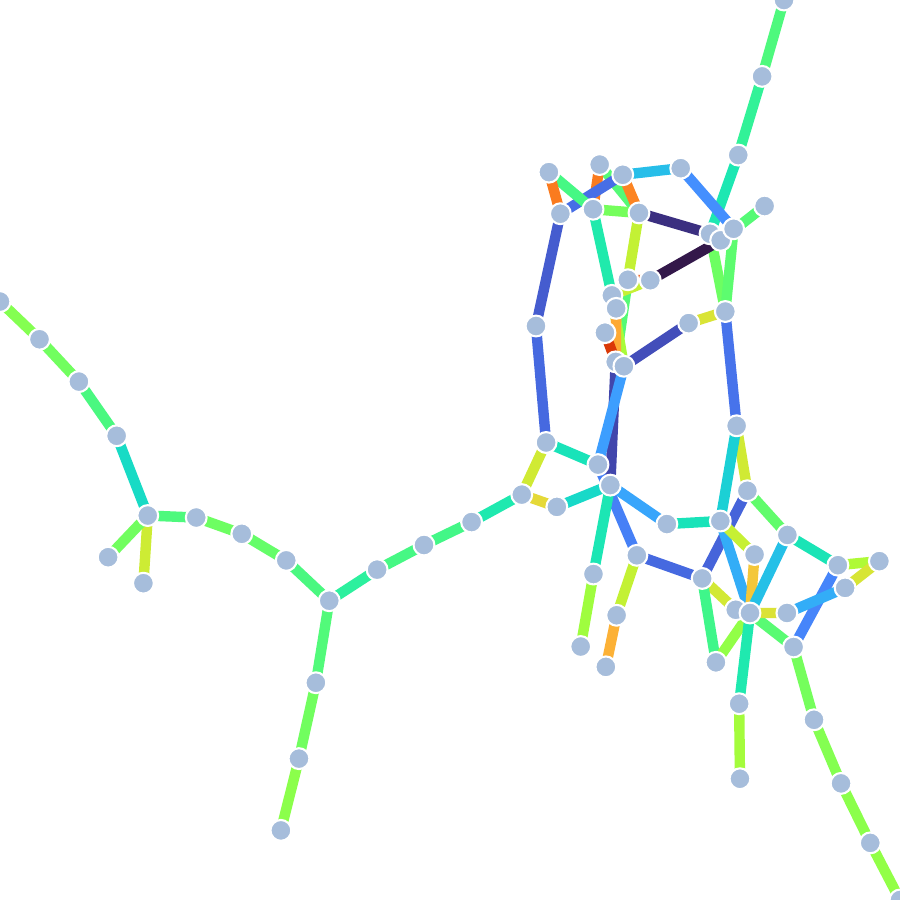} \\ \vspace{-0.0cm} \fontsize{7pt}{0pt}\selectfont \textbf{0.037},0.044} & \parbox{1.7cm}{\centering \includegraphics[height=1.5cm]{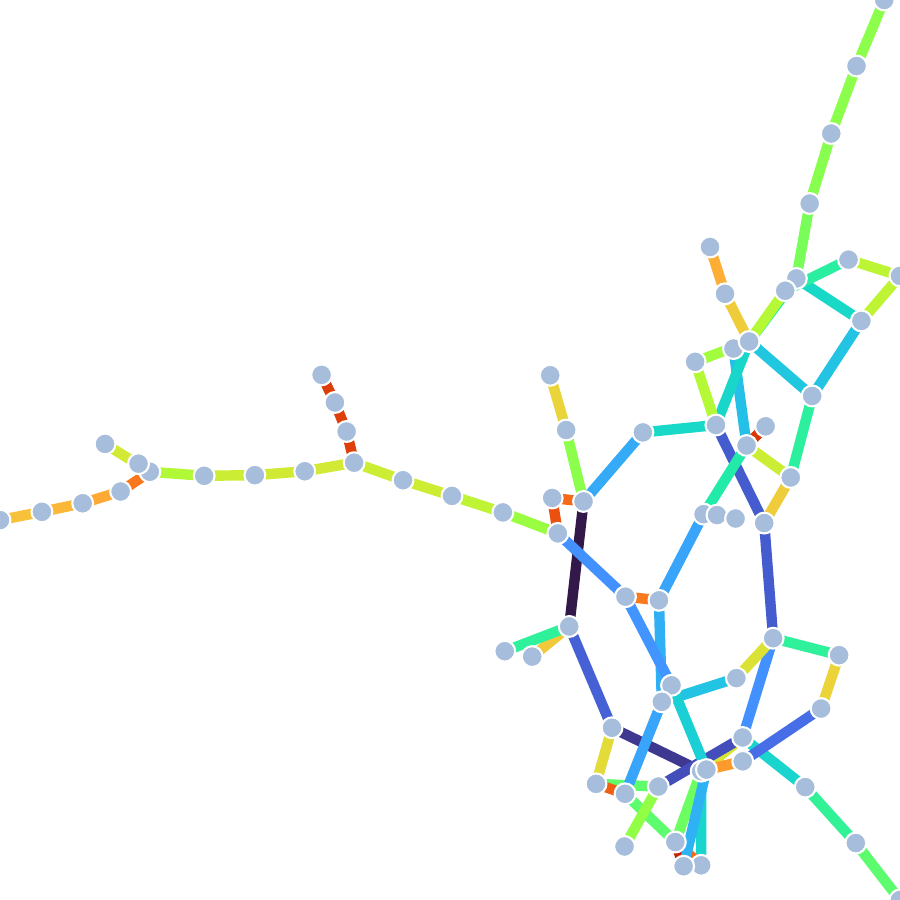} \\ \vspace{-0.0cm} \fontsize{7pt}{0pt}\selectfont 0.913,\textbf{0.845}} & \parbox{1.7cm}{\centering \includegraphics[height=1.5cm]{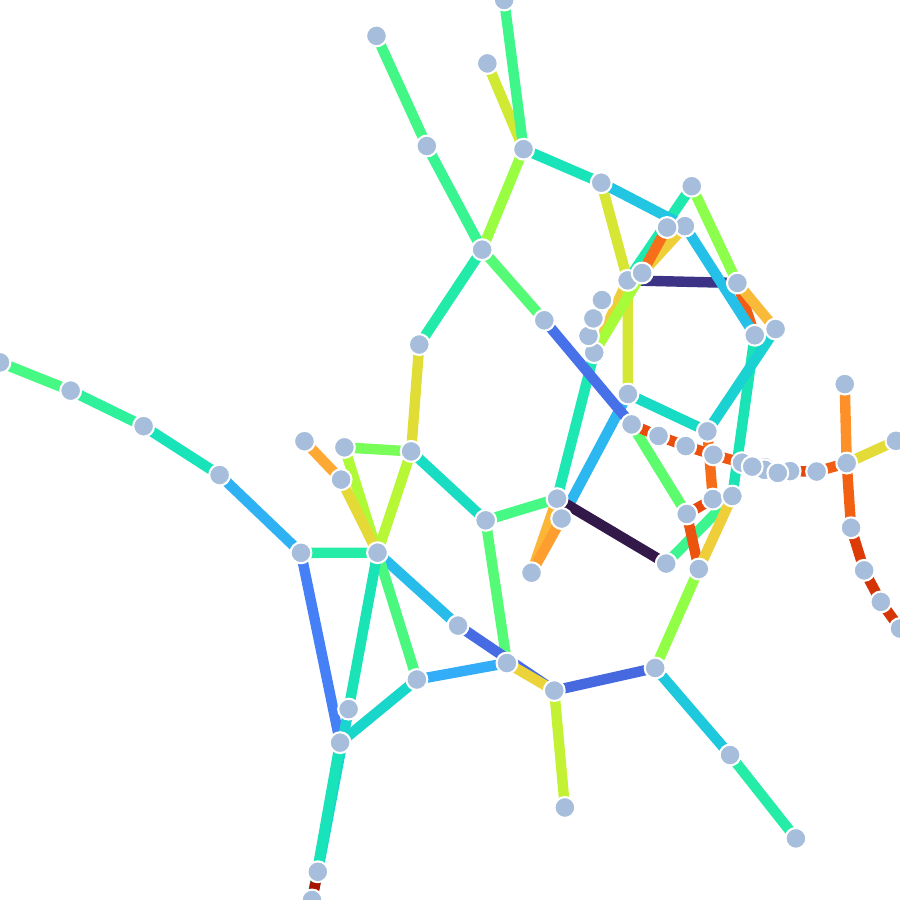} \\ \vspace{-0.0cm} \fontsize{7pt}{0pt}\selectfont 16,\textbf{13}} \\
\bottomrule
\end{tabular}
\end{table*}

\begin{table*}[htbp]
\setlength{\tabcolsep}{5pt}
\centering
\renewcommand{\arraystretch}{0.8}
\caption{Stress for Neato layouts at different dimensions, vs that for the Optimal projection to 2D, and for the best PCA projection to 2D.}
\label{tab:stress_vs_dims}

\begin{tabular}{l|lllll|llll|llll}
\toprule
 & \multicolumn{5}{c|}{neato at \highD} & \multicolumn{4}{c|}{OptProj (from \highD-D to 2D)} & \multicolumn{4}{c}{PCA (from \highD-D to 2D)} \\
 & 2 & 3 & 5 & 7 & 10 & 3 & 5 & 7 & 10 & 3 & 5 & 7 & 10 \\
\midrule
1138\_bus & 0.069 & 0.030 & 0.013 & 0.008 & \textbf{0.006} & 0.074 & 0.083 & 0.080 & 0.076 & 0.075 & 0.087 & 0.101 & 0.113 \\
bcspwr07 & 0.046 & 0.022 & 0.010 & 0.007 & \textbf{0.006} & 0.057 & 0.062 & 0.063 & 0.062 & 0.058 & 0.068 & 0.074 & 0.077 \\
email & 0.133 & 0.085 & 0.047 & 0.033 & \textbf{0.025} & 0.145 & 0.157 & 0.156 & 0.163 & 0.145 & 0.159 & 0.165 & 0.171 \\
Erdos991 & 0.109 & 0.063 & 0.033 & 0.023 & \textbf{0.017} & 0.124 & 0.131 & 0.131 & 0.129 & 0.125 & 0.143 & 0.150 & 0.161 \\
EX1 & 0.167 & 0.103 & 0.050 & 0.035 & \textbf{0.033} & 0.169 & 0.173 & 0.176 & 0.176 & 0.169 & 0.177 & 0.177 & 0.177 \\
football & 0.128 & 0.077 & 0.043 & 0.036 & \textbf{0.032} & 0.132 & 0.133 & 0.133 & 0.132 & 0.132 & 0.134 & 0.136 & 0.139 \\
hypercube10 & 0.198 & 0.135 & 0.081 & 0.058 & \textbf{0.041} & 0.199 & 0.202 & 0.204 & 0.206 & 0.199 & 0.202 & 0.205 & 0.208 \\
Journals & 0.167 & 0.114 & 0.071 & 0.055 & \textbf{0.043} & 0.176 & 0.180 & 0.185 & 0.182 & 0.178 & 0.182 & 0.194 & 0.198 \\
mobius & 0.029 & 0.017 & \textbf{0.016} & \textbf{0.016} & \textbf{0.016} & 0.035 & 0.036 & 0.036 & 0.034 & 0.042 & 0.043 & 0.043 & 0.043 \\
qh882 & 0.056 & 0.028 & 0.014 & 0.010 & \textbf{0.009} & 0.069 & 0.071 & 0.074 & 0.072 & 0.072 & 0.080 & 0.087 & 0.089 \\
Si2 & 0.159 & 0.088 & 0.050 & 0.036 & \textbf{0.027} & 0.139 & 0.142 & 0.143 & 0.143 & 0.139 & 0.142 & 0.143 & 0.144 \\
spiral & 0.044 & 0.025 & 0.017 & 0.014 & \textbf{0.012} & 0.053 & 0.064 & 0.064 & 0.062 & 0.054 & 0.064 & 0.075 & 0.078 \\
Trefethen\_700 & 0.177 & 0.118 & 0.072 & 0.055 & \textbf{0.044} & 0.178 & 0.180 & 0.181 & 0.181 & 0.180 & 0.197 & 0.207 & 0.218 \\
USpowerGrid & 0.058 & 0.028 & 0.011 & 0.007 & \textbf{0.005} & 0.066 & 0.074 & 0.077 & 0.073 & 0.066 & 0.077 & 0.082 & 0.086 \\
\midrule
SPC & 0 & -0.459 & -0.707 & -0.794 & \textbf{-0.844} & 0.088 & 0.122 & 0.127 & 0.122 & 0.088 & 0.136 & 0.163 & 0.176 \\
\bottomrule
\end{tabular}
\end{table*}

\section{Additional results for \sfdp}
\label{sec_appendix_additional_res_sfdp}

Table~\ref{tab:sfdp_vis} presents visualizations produced by \sfdp on the SuiteSparse14 dataset. Optimizing different metrics results in markedly different layouts. For example, optimizing edge crossings (xing) produces a layout that differs substantially from the default \sfdp output. Optimizing the t-SNE metric yields a visually less complex layout that appears to contain fewer edge crossings; however, closer inspection reveals many crossings hidden at finer scales.

Table~\ref{tab:sfdp_vs_dims} presents the stress achieved by \sfdp\  when optimizing directly in various dimensions from 2D to 10D. It is surprising that the spring-electrical energy obtained by \sfdp(10) is mostly similar to that for \sfdp(2), except for the Journals graph. Furthermore, the spring-electrical loss by \OptProjSfdp\ and \OptPCASfdp\ are also quite similar to that of \neato. This seems to indicate that optimal layout for this loss function can be realized well in lower dimension.

One way to decide whether a high-dimensional embedding can ``live'' happily in a lower-dimensional hyperplane is to look at the variance explained by PCA projections at a lower dimension. This is confirmed in Table~\ref{tab:variance_explained}, where we looked at variance explained for SuiteSparse14 graphs by PCA projection from 10D of \sfdp\ vs \neato.
We can see that the variance explained for \sfdp\ is indeed higher than for \neato, at the same layout dimension. E.g., PCA projection of 10D embeddings to 3D can explain 87.8\% of the original variance, vs 82.5\% for \neato.

\begin{table*}[htbp]
\setlength{\tabcolsep}{2pt}
\centering
\renewcommand{\arraystretch}{0.8}
\caption{Optimal projections of the 10D \sfdp\ layout. Numbers show  \sfdp\  vs. \OptProjSfdp metrics. \textbf{Bold} = better. Edges colored by length: {\color{red}red} = short, {\color{blue}blue} = long.}
\label{tab:sfdp_vis}

\begin{tabular}{llllllll}
\toprule
 & sfdp & ang\_res & edge\_len & spr\_elec & stress & tsne & xing \\
\midrule
1138\_bus & \parbox{1.7cm}{\centering \includegraphics[height=1.1cm]{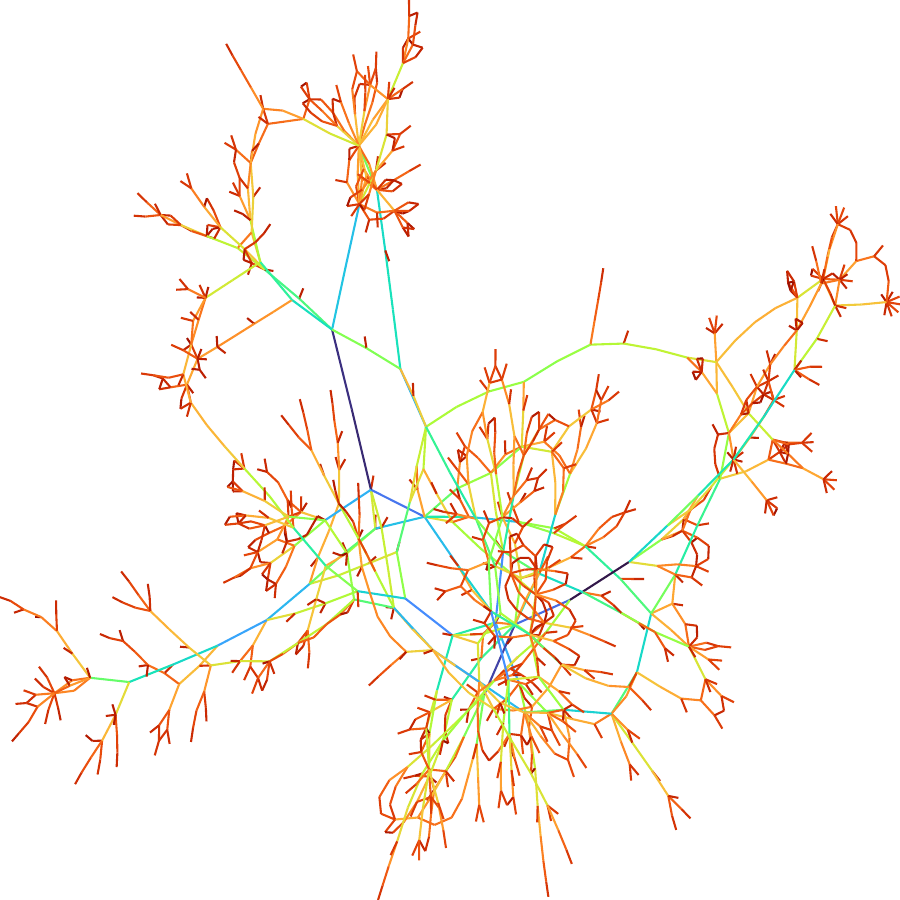} \\ \fontsize{7pt}{0pt}\selectfont } & \parbox{1.7cm}{\centering \includegraphics[height=1.1cm]{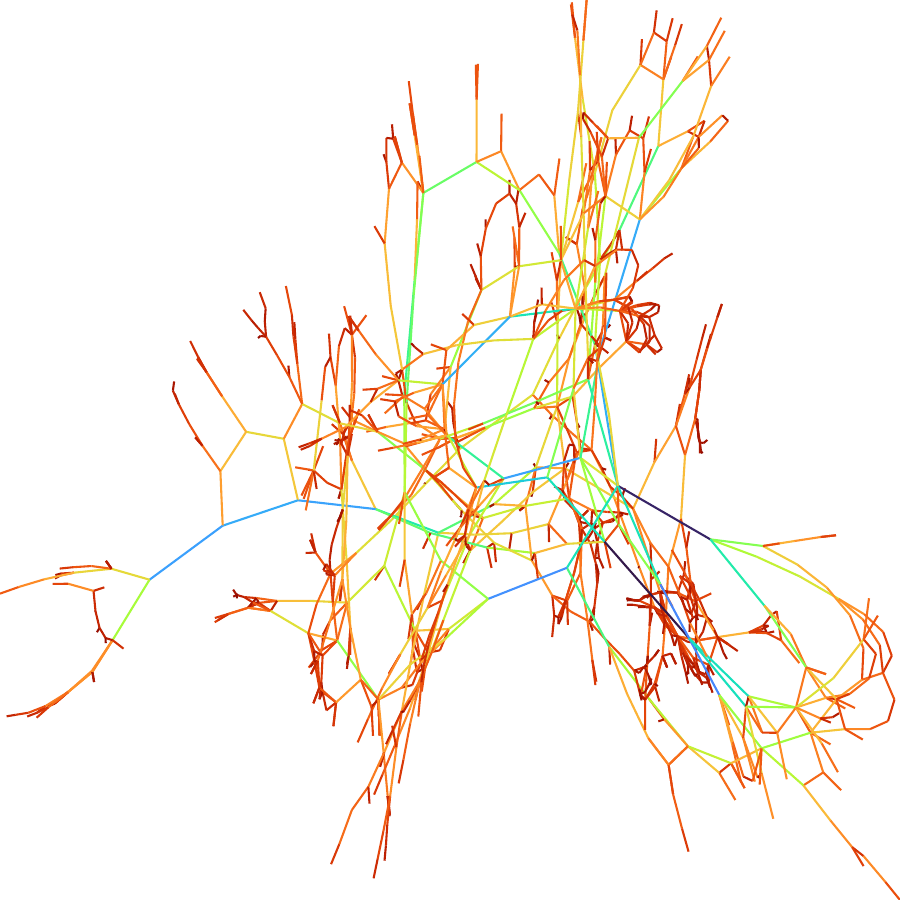} \\ \fontsize{7pt}{0pt}\selectfont \textbf{1.046},1.105} & \parbox{1.7cm}{\centering \includegraphics[height=1.1cm]{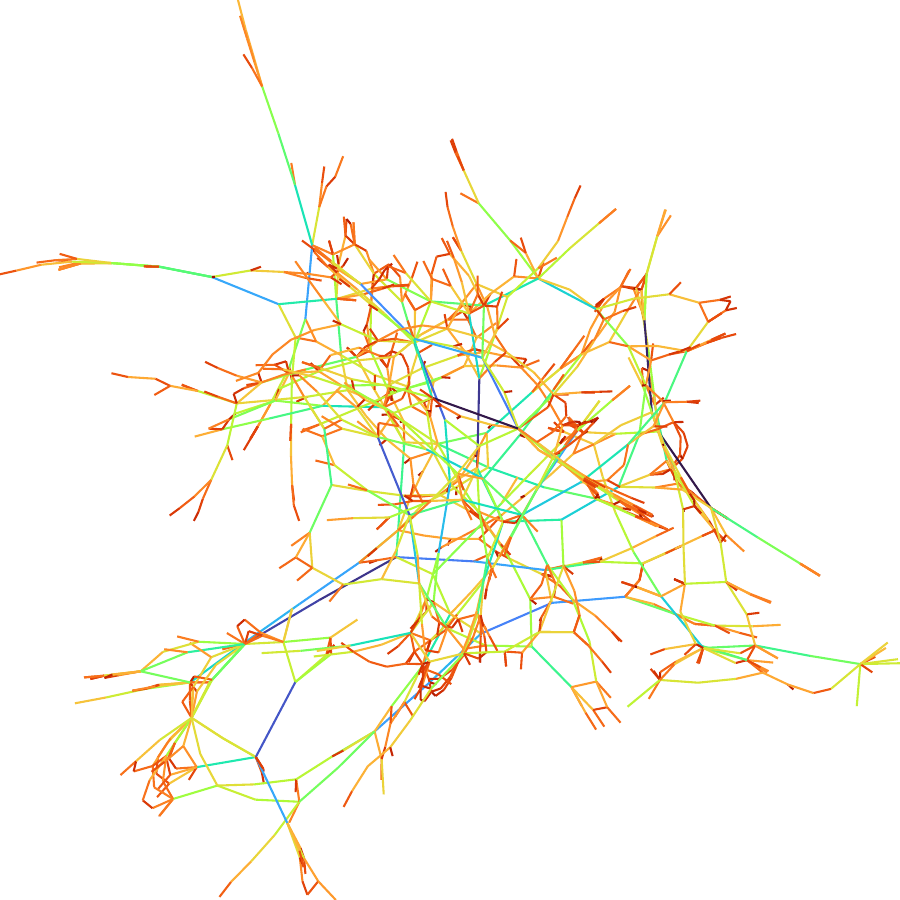} \\ \fontsize{7pt}{0pt}\selectfont \textbf{0.55},0.569} & \parbox{1.7cm}{\centering \includegraphics[height=1.1cm]{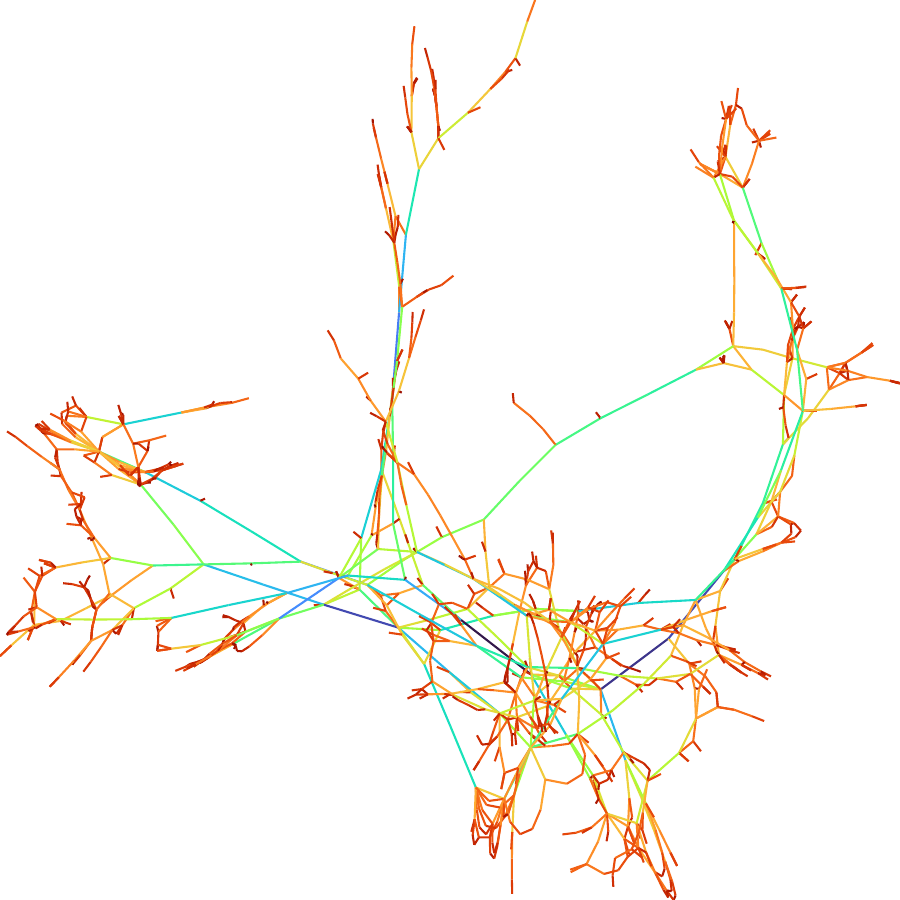} \\ \fontsize{7pt}{0pt}\selectfont \textbf{-4.369},-4.322} & \parbox{1.7cm}{\centering \includegraphics[height=1.1cm]{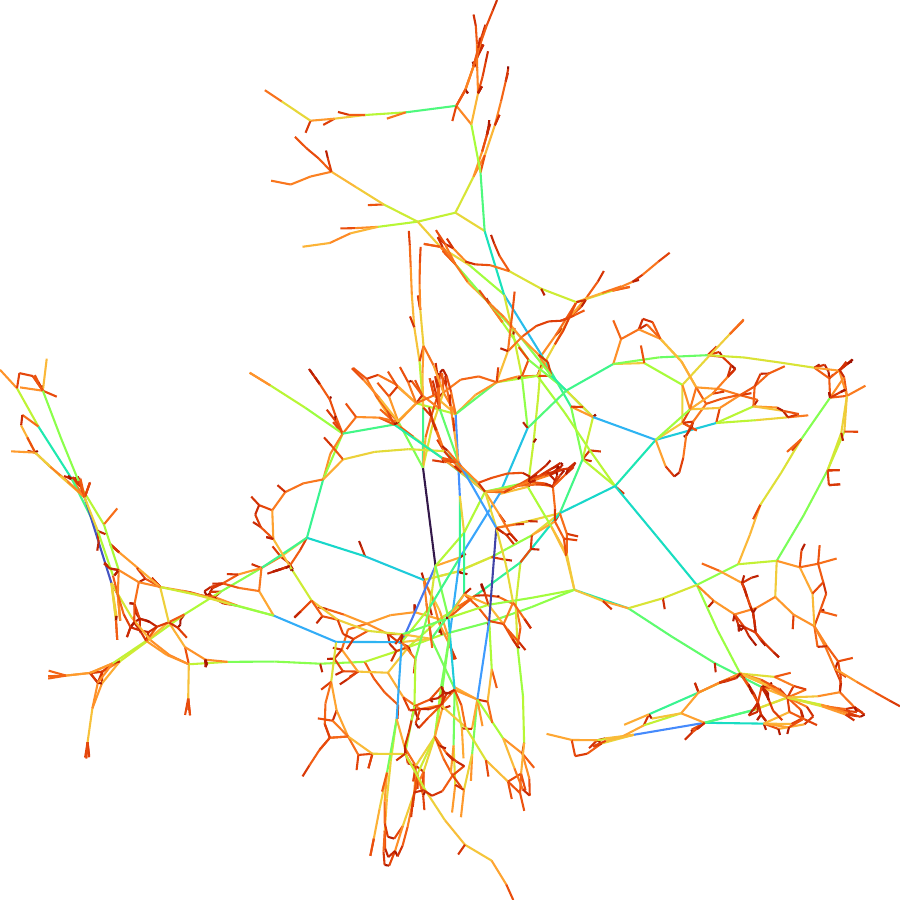} \\ \fontsize{7pt}{0pt}\selectfont \textbf{0.092},0.098} & \parbox{1.7cm}{\centering \includegraphics[height=1.1cm]{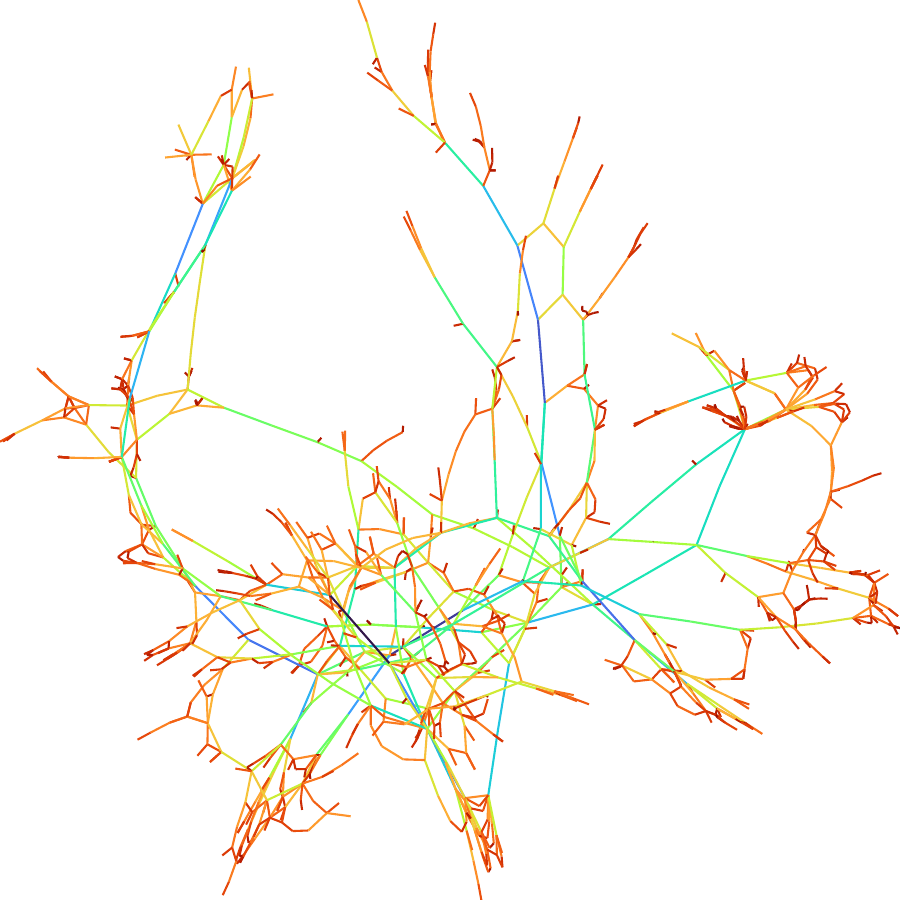} \\ \fontsize{7pt}{0pt}\selectfont \textbf{1.415},1.584} & \parbox{1.7cm}{\centering \includegraphics[height=1.1cm]{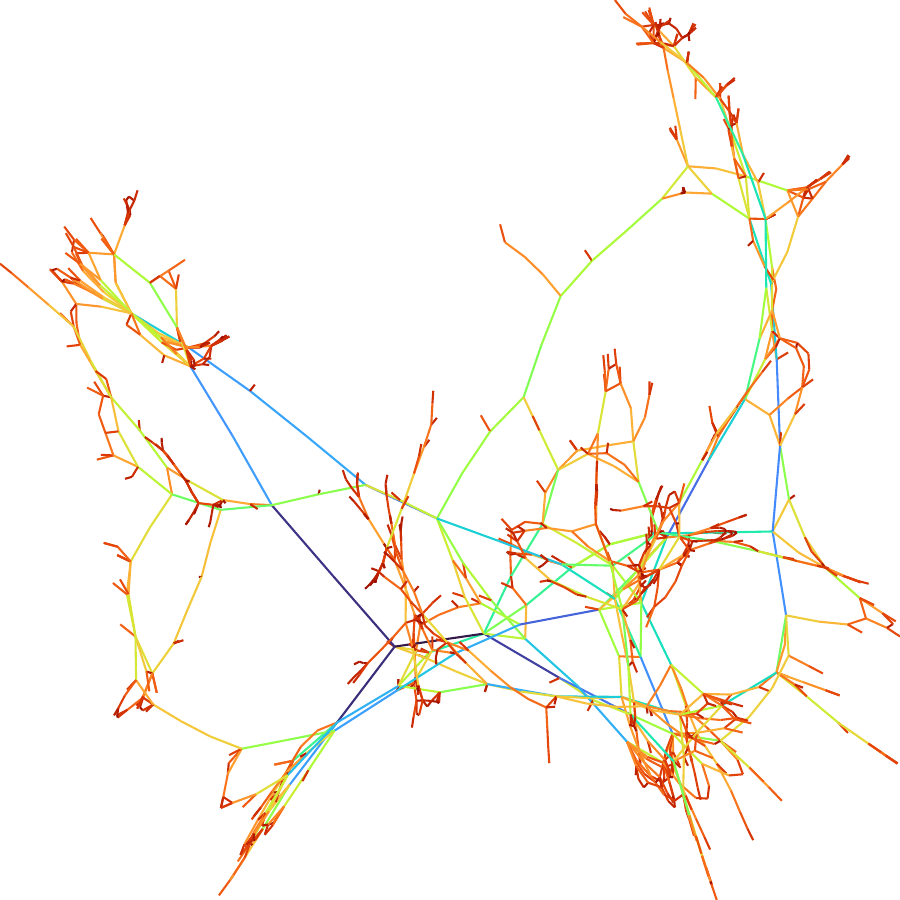} \\ \fontsize{7pt}{0pt}\selectfont \textbf{511},756} \\
bcspwr07 & \parbox{1.7cm}{\centering \includegraphics[height=1.1cm]{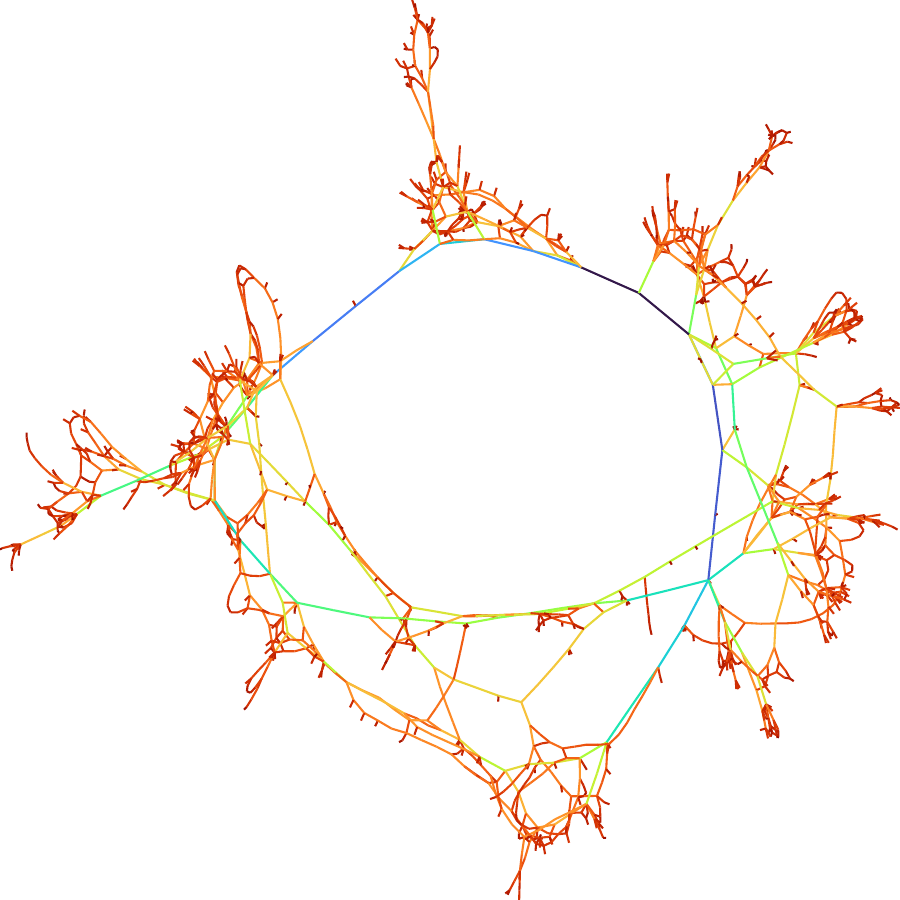} \\ \fontsize{7pt}{0pt}\selectfont } & \parbox{1.7cm}{\centering \includegraphics[height=1.1cm]{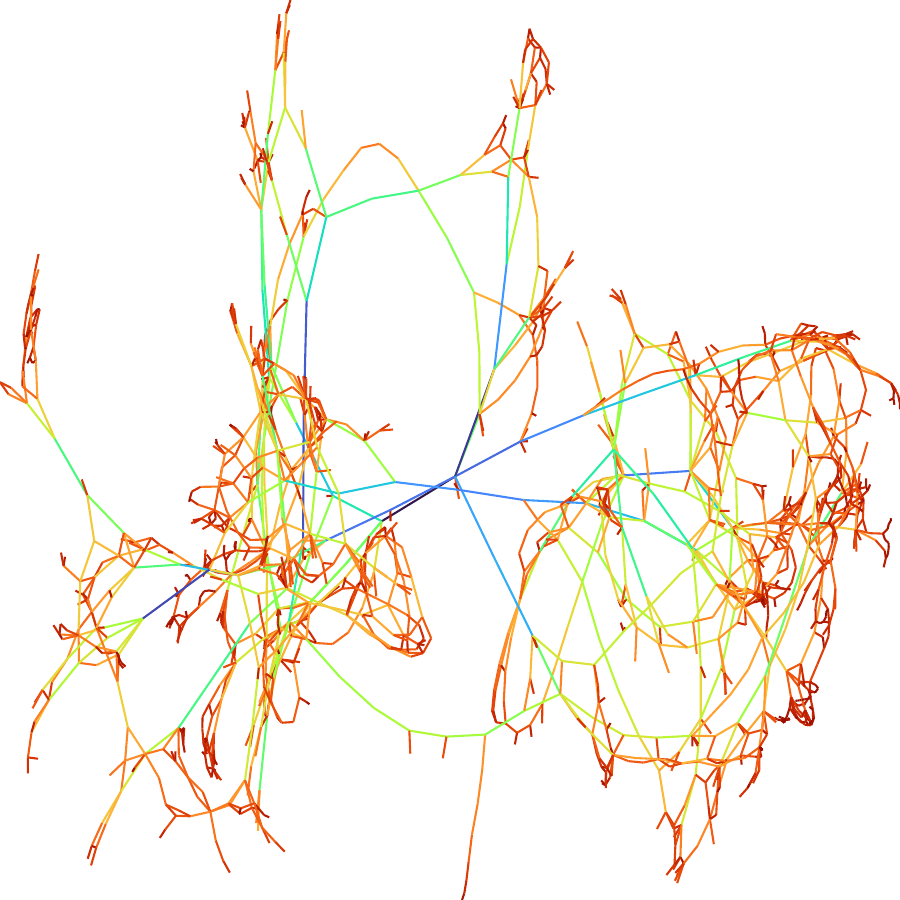} \\ \fontsize{7pt}{0pt}\selectfont 1.033,\textbf{1.03}} & \parbox{1.7cm}{\centering \includegraphics[height=1.1cm]{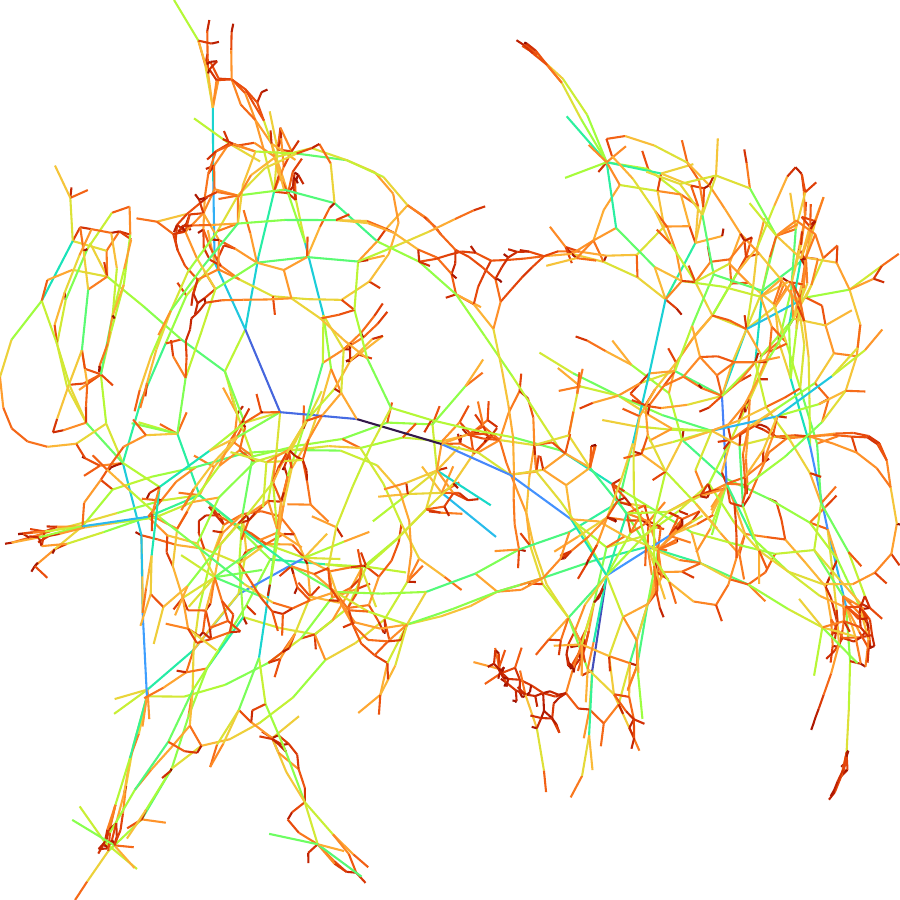} \\ \fontsize{7pt}{0pt}\selectfont 0.672,\textbf{0.568}} & \parbox{1.7cm}{\centering \includegraphics[height=1.1cm]{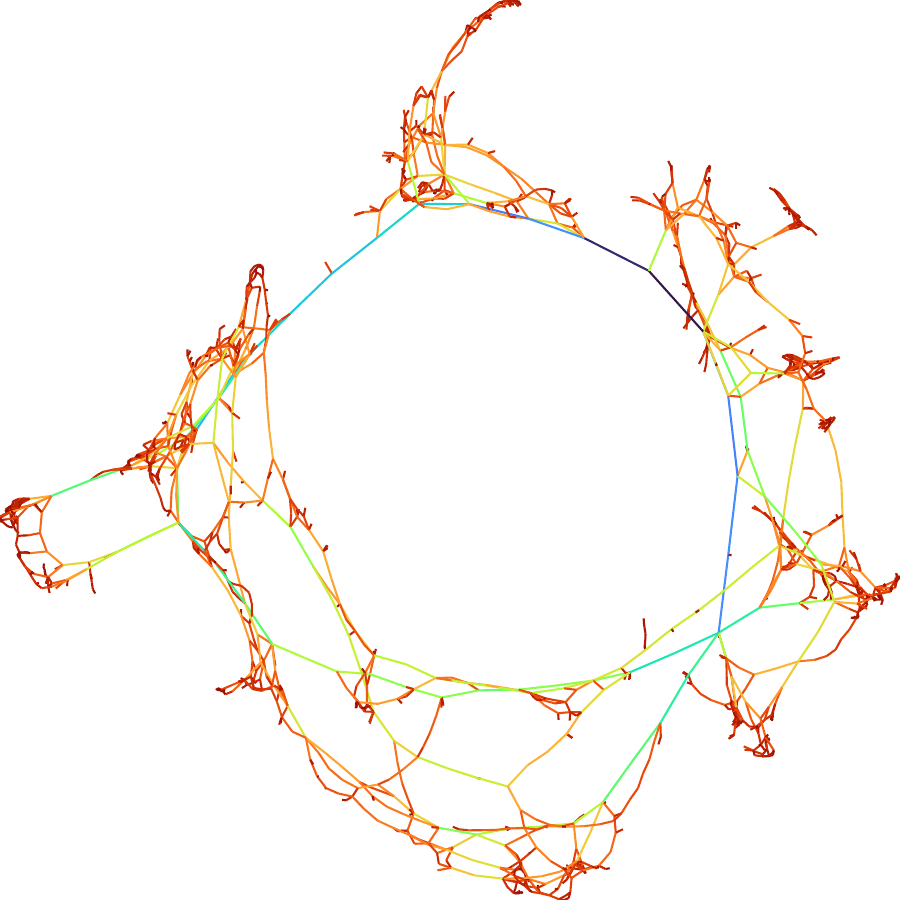} \\ \fontsize{7pt}{0pt}\selectfont \textbf{-4.998},-4.937} & \parbox{1.7cm}{\centering \includegraphics[height=1.1cm]{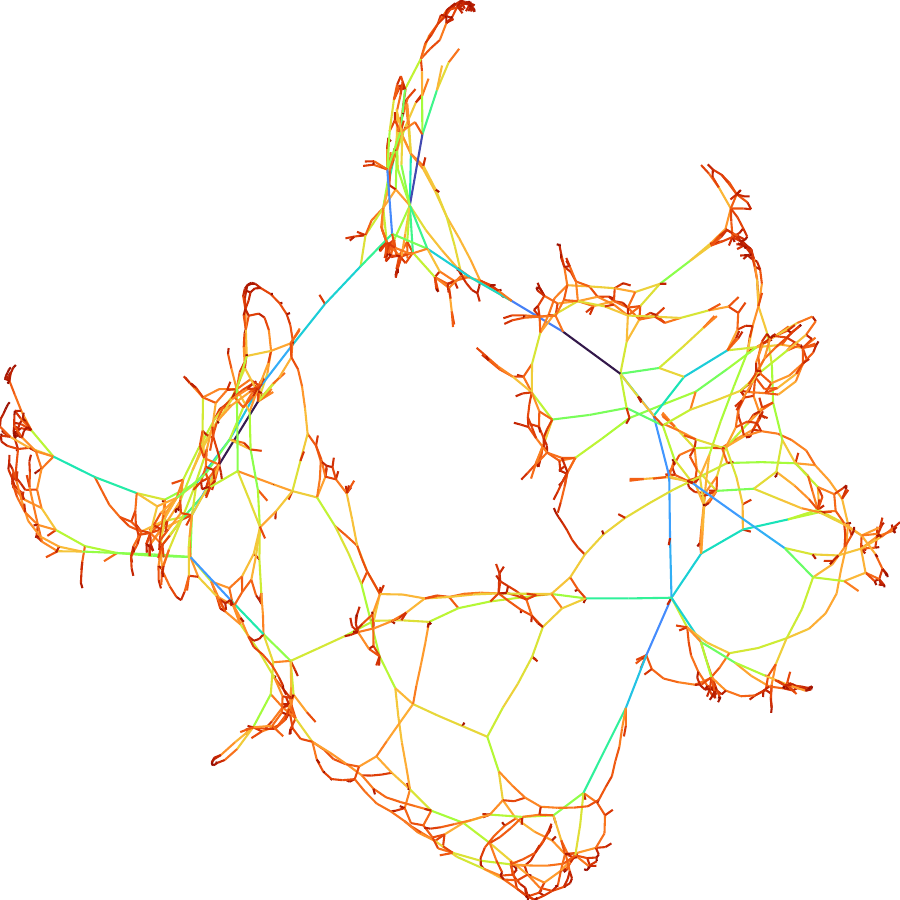} \\ \fontsize{7pt}{0pt}\selectfont 0.088,\textbf{0.084}} & \parbox{1.7cm}{\centering \includegraphics[height=1.1cm]{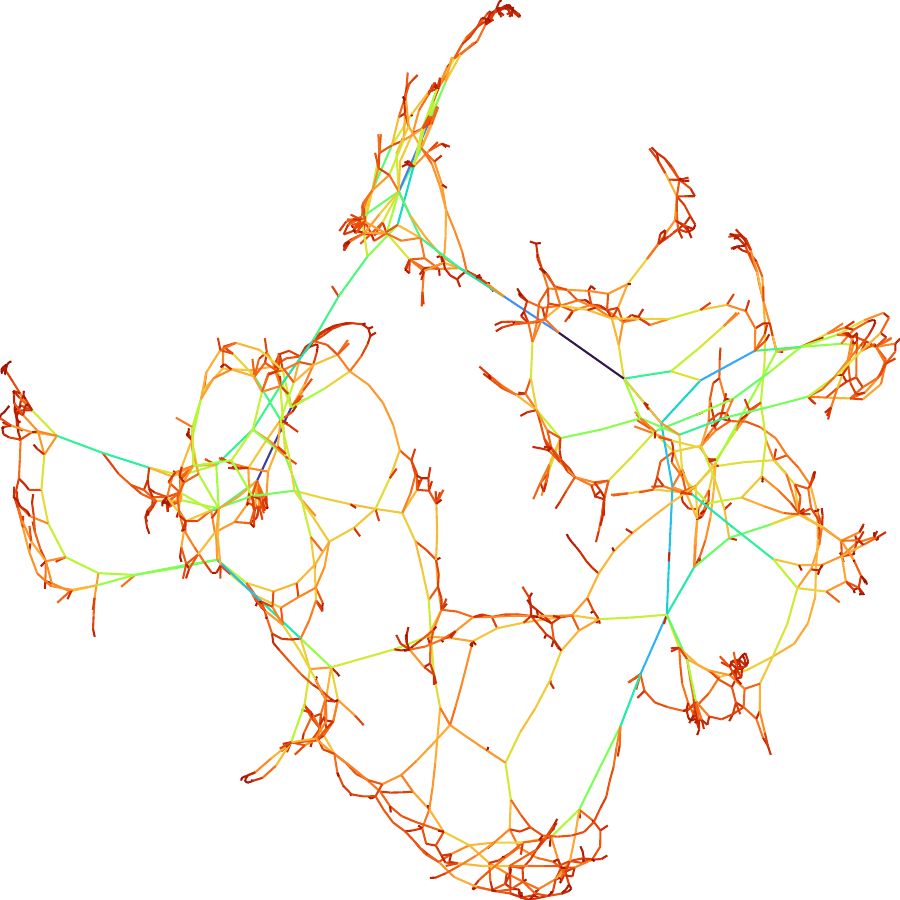} \\ \fontsize{7pt}{0pt}\selectfont \textbf{1.365},1.526} & \parbox{1.7cm}{\centering \includegraphics[height=1.1cm]{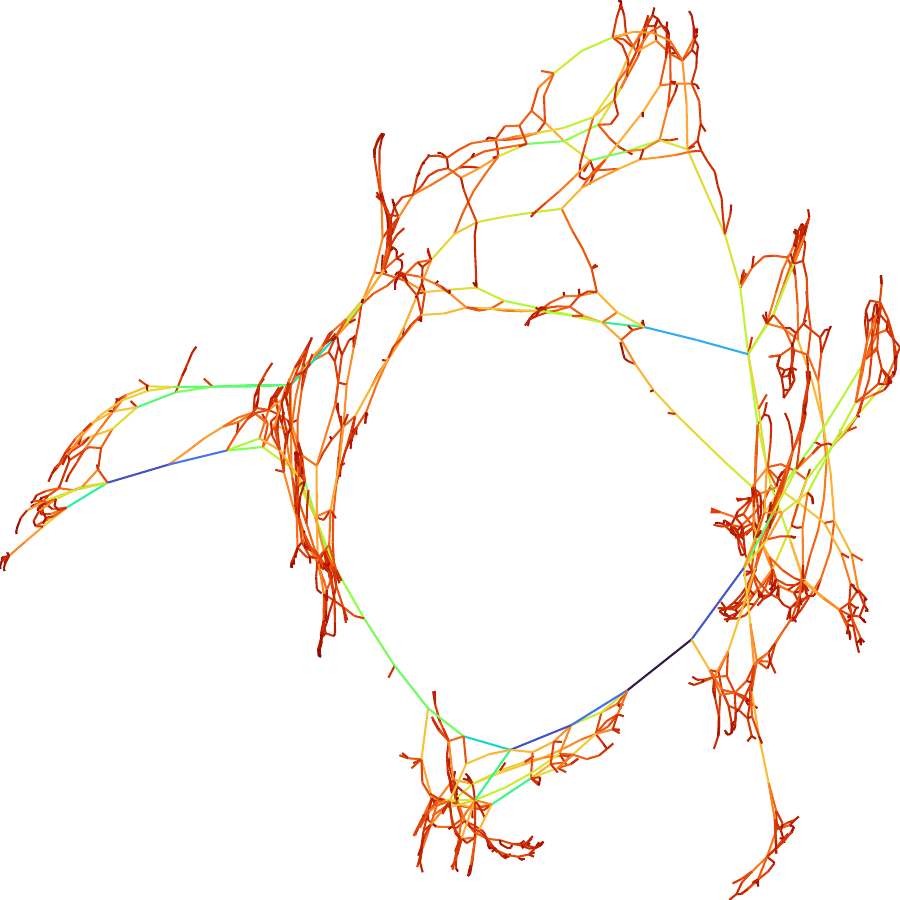} \\ \fontsize{7pt}{0pt}\selectfont \textbf{678},1047} \\
email & \parbox{1.7cm}{\centering \includegraphics[height=1.1cm]{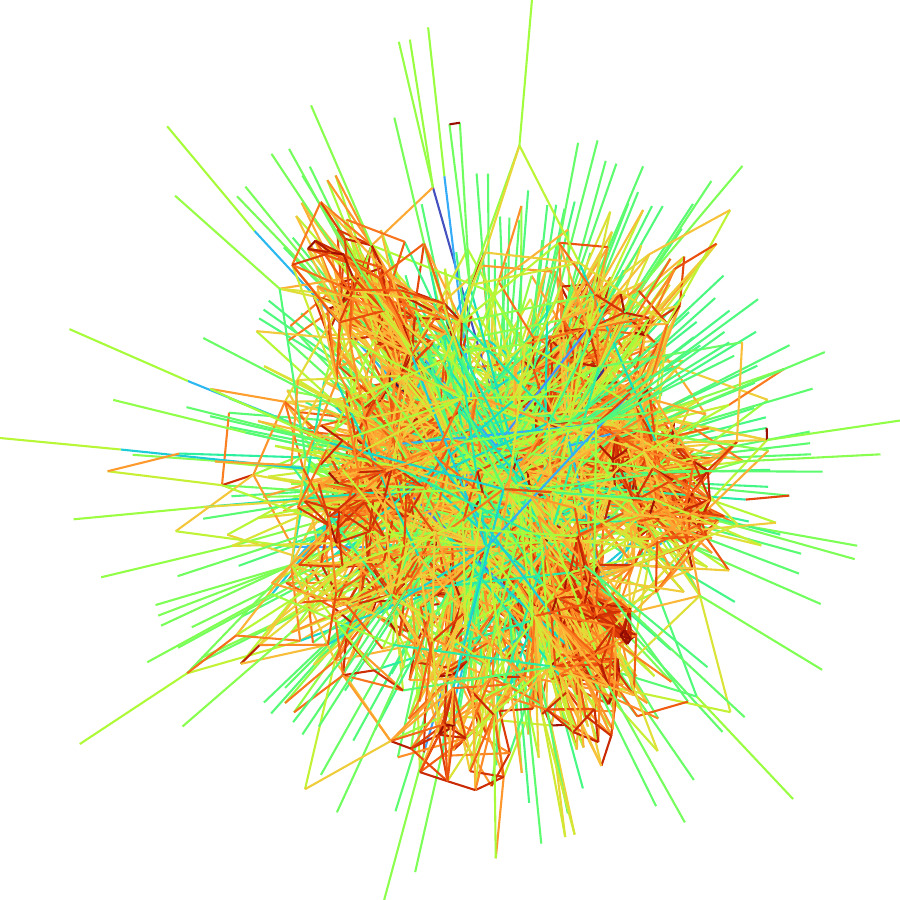} \\ \fontsize{7pt}{0pt}\selectfont } & \parbox{1.7cm}{\centering \includegraphics[height=1.1cm]{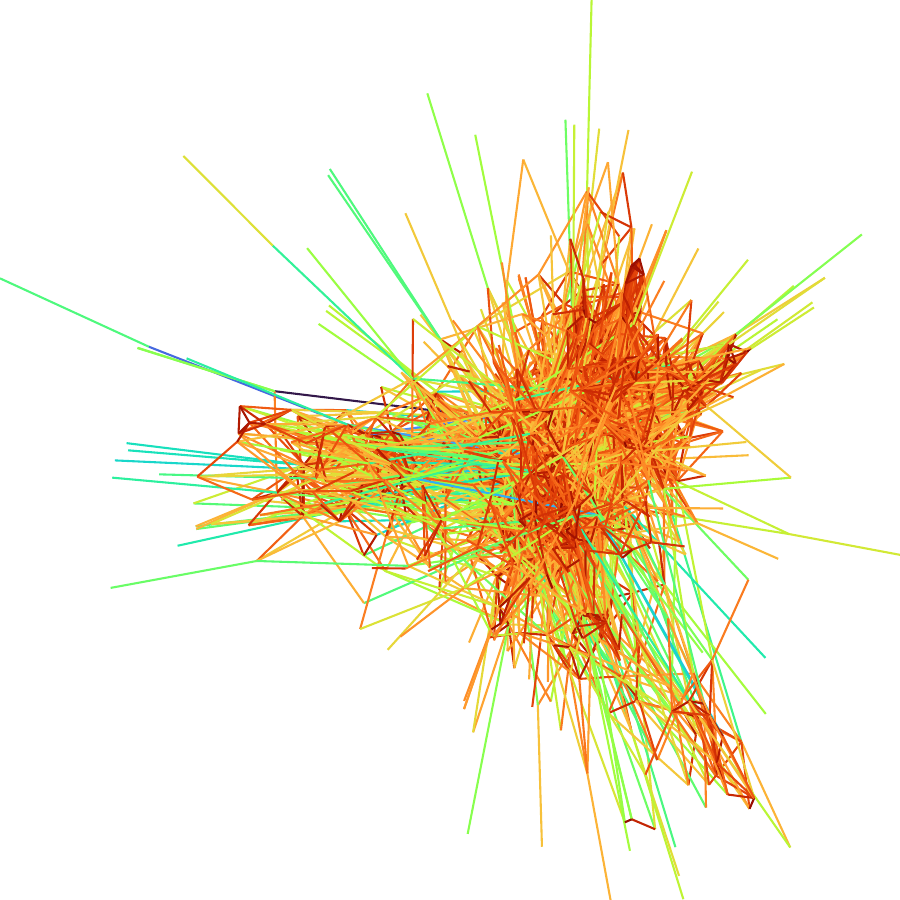} \\ \fontsize{7pt}{0pt}\selectfont 1.207,\textbf{1.097}} & \parbox{1.7cm}{\centering \includegraphics[height=1.1cm]{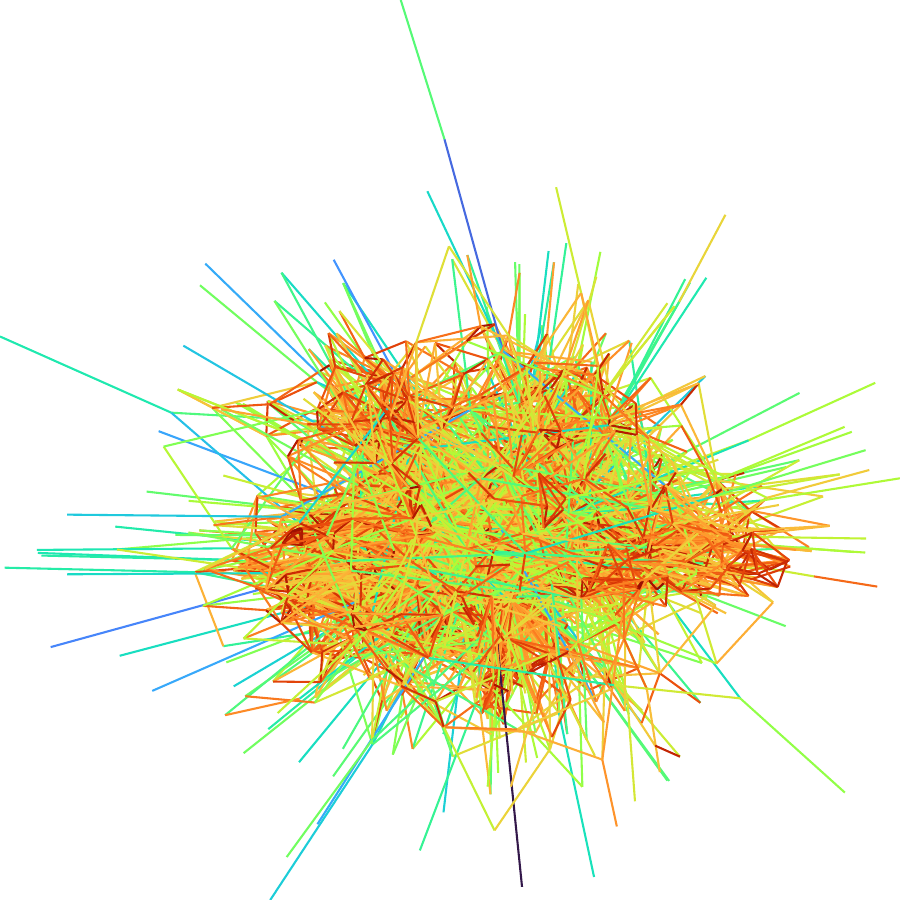} \\ \fontsize{7pt}{0pt}\selectfont 0.546,\textbf{0.529}} & \parbox{1.7cm}{\centering \includegraphics[height=1.1cm]{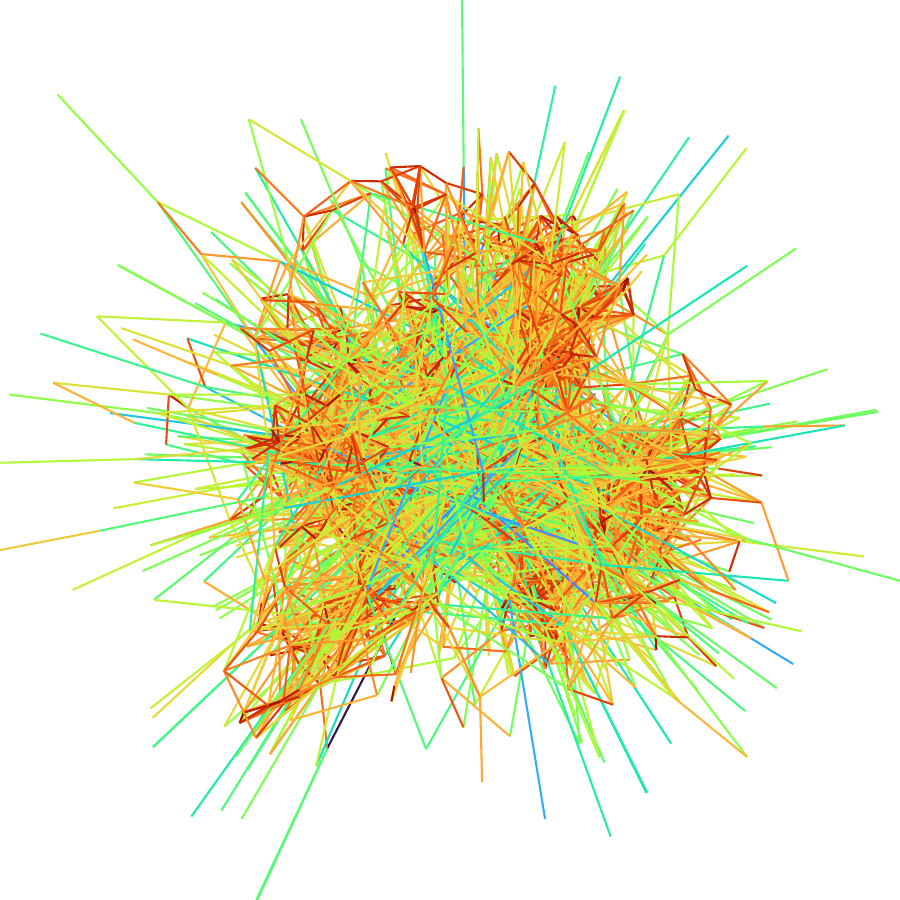} \\ \fontsize{7pt}{0pt}\selectfont \textbf{-2.306},-2.269} & \parbox{1.7cm}{\centering \includegraphics[height=1.1cm]{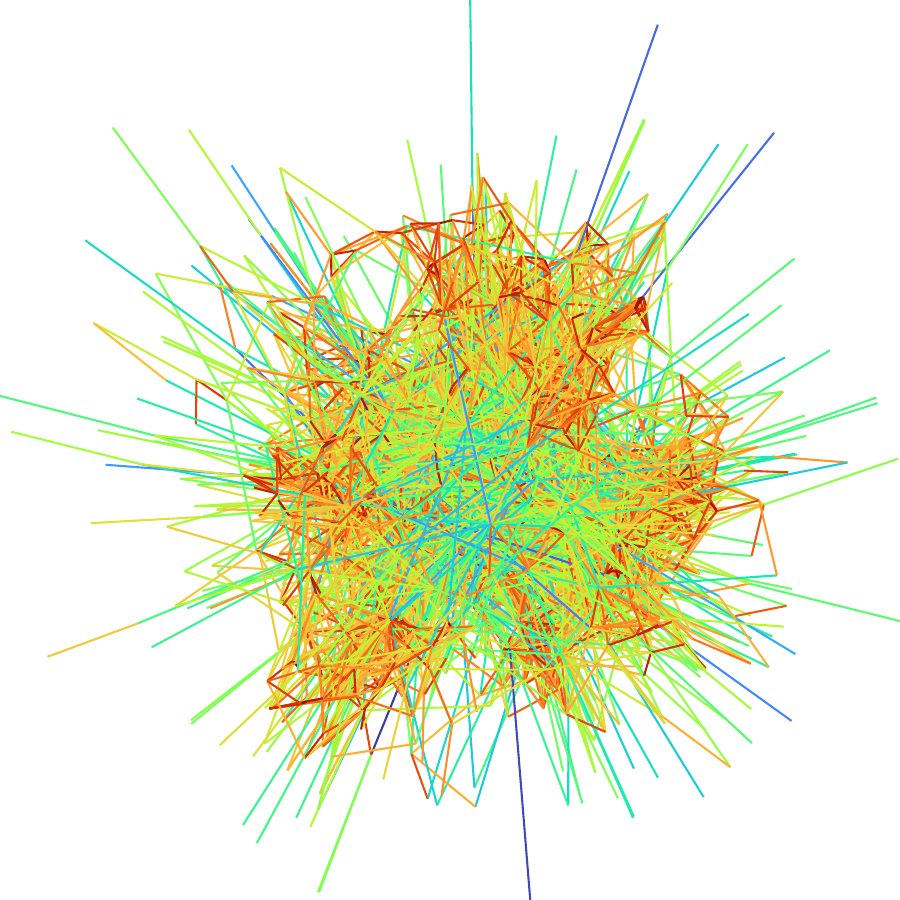} \\ \fontsize{7pt}{0pt}\selectfont \textbf{0.142},0.15} & \parbox{1.7cm}{\centering \includegraphics[height=1.1cm]{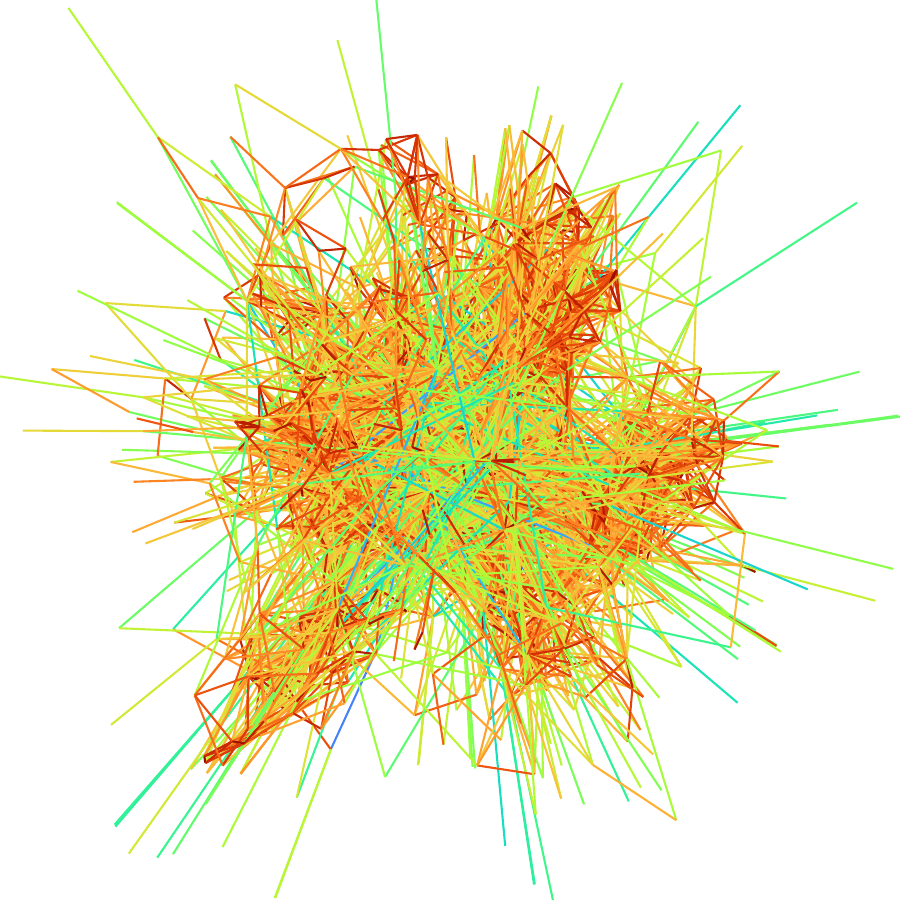} \\ \fontsize{7pt}{0pt}\selectfont \textbf{1.308},1.309} & \parbox{1.7cm}{\centering \includegraphics[height=1.1cm]{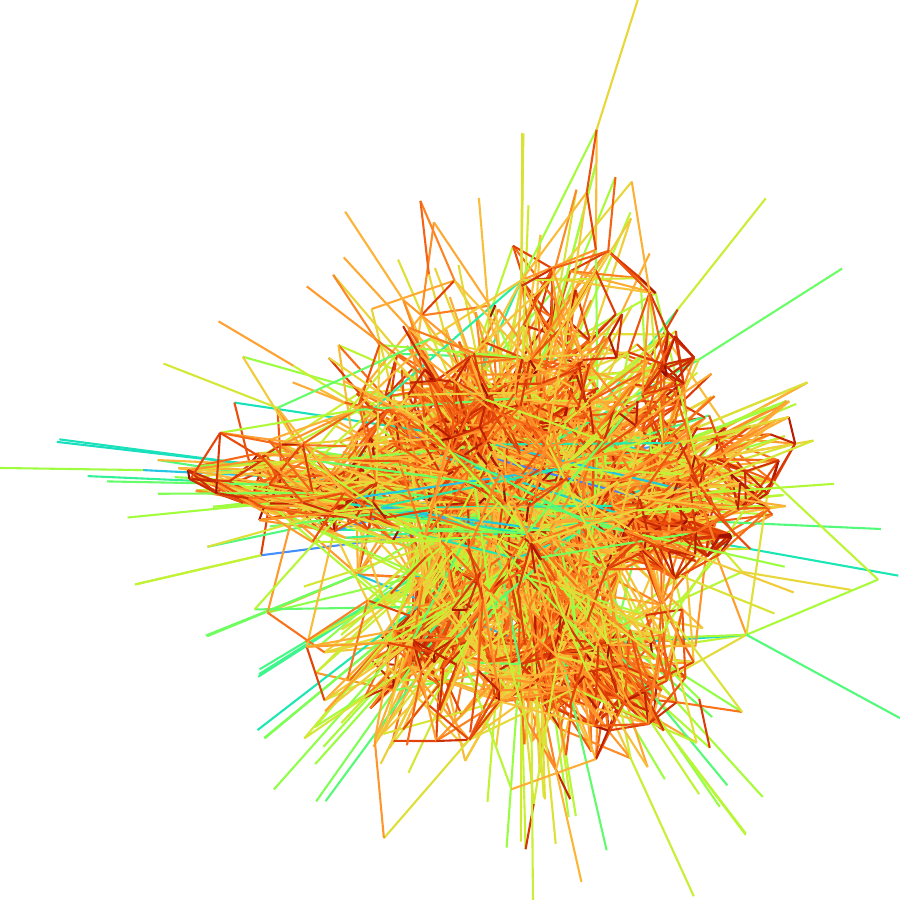} \\ \fontsize{7pt}{0pt}\selectfont 425417,\textbf{396344}} \\
Erdos991 & \parbox{1.7cm}{\centering \includegraphics[height=1.1cm]{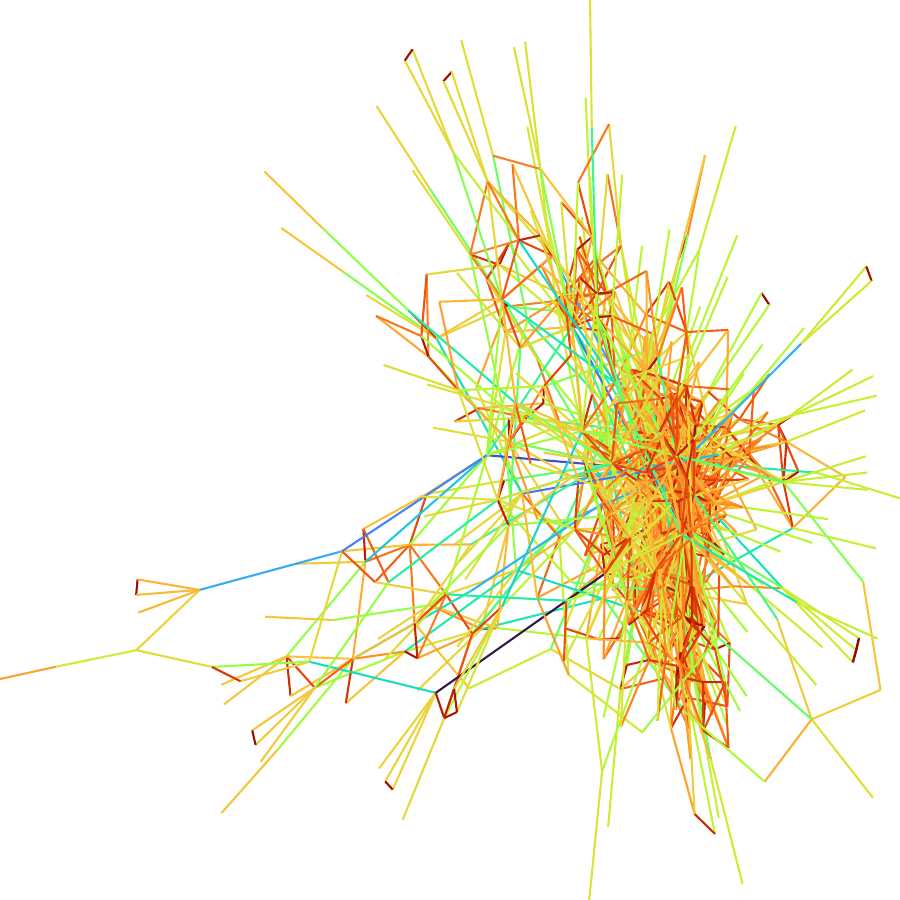} \\ \fontsize{7pt}{0pt}\selectfont } & \parbox{1.7cm}{\centering \includegraphics[height=1.1cm]{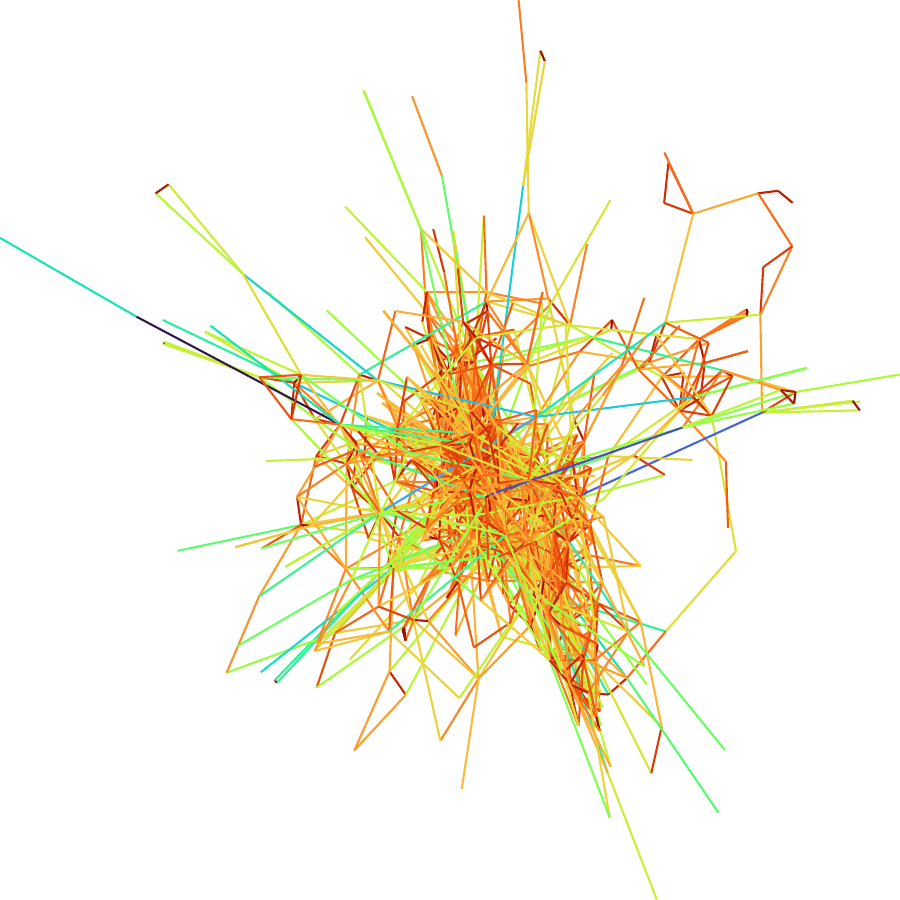} \\ \fontsize{7pt}{0pt}\selectfont 1.26,\textbf{1.183}} & \parbox{1.7cm}{\centering \includegraphics[height=1.1cm]{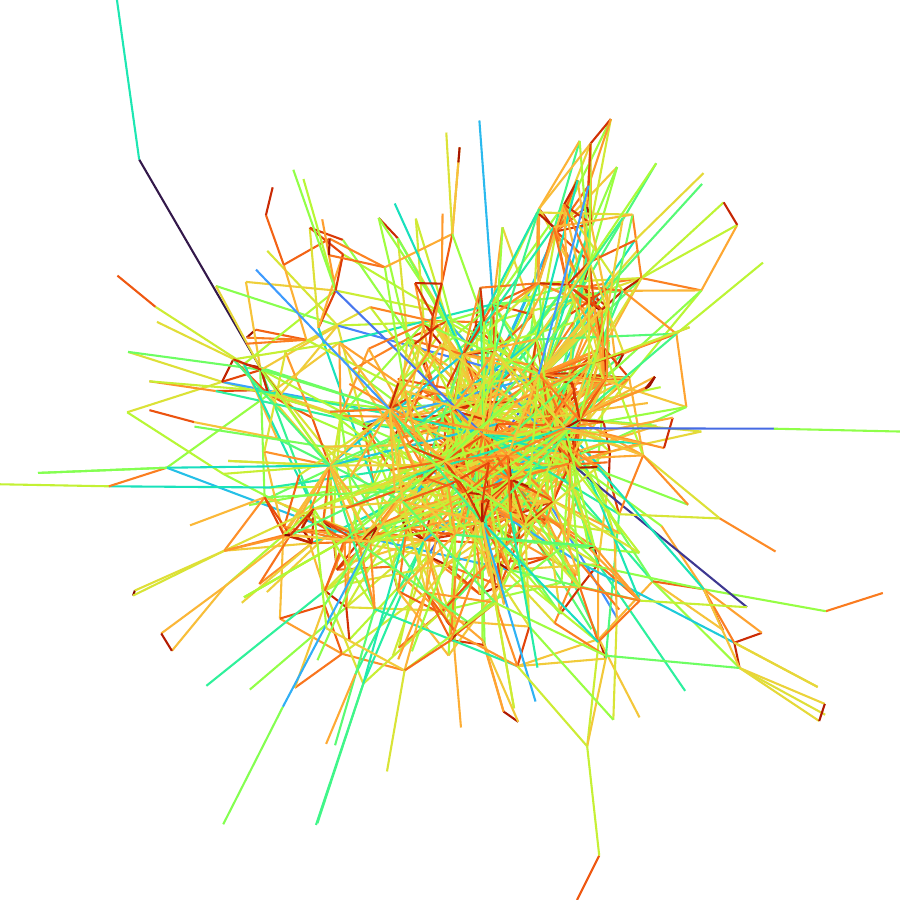} \\ \fontsize{7pt}{0pt}\selectfont 0.502,\textbf{0.465}} & \parbox{1.7cm}{\centering \includegraphics[height=1.1cm]{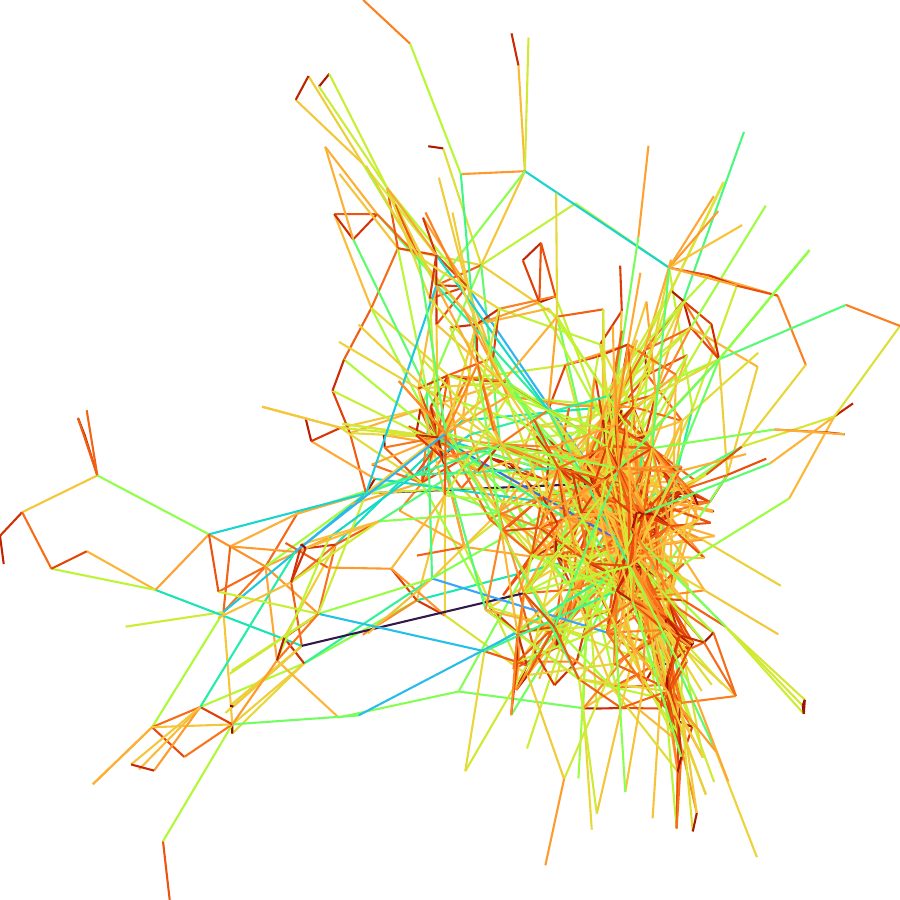} \\ \fontsize{7pt}{0pt}\selectfont \textbf{-2.313},-2.28} & \parbox{1.7cm}{\centering \includegraphics[height=1.1cm]{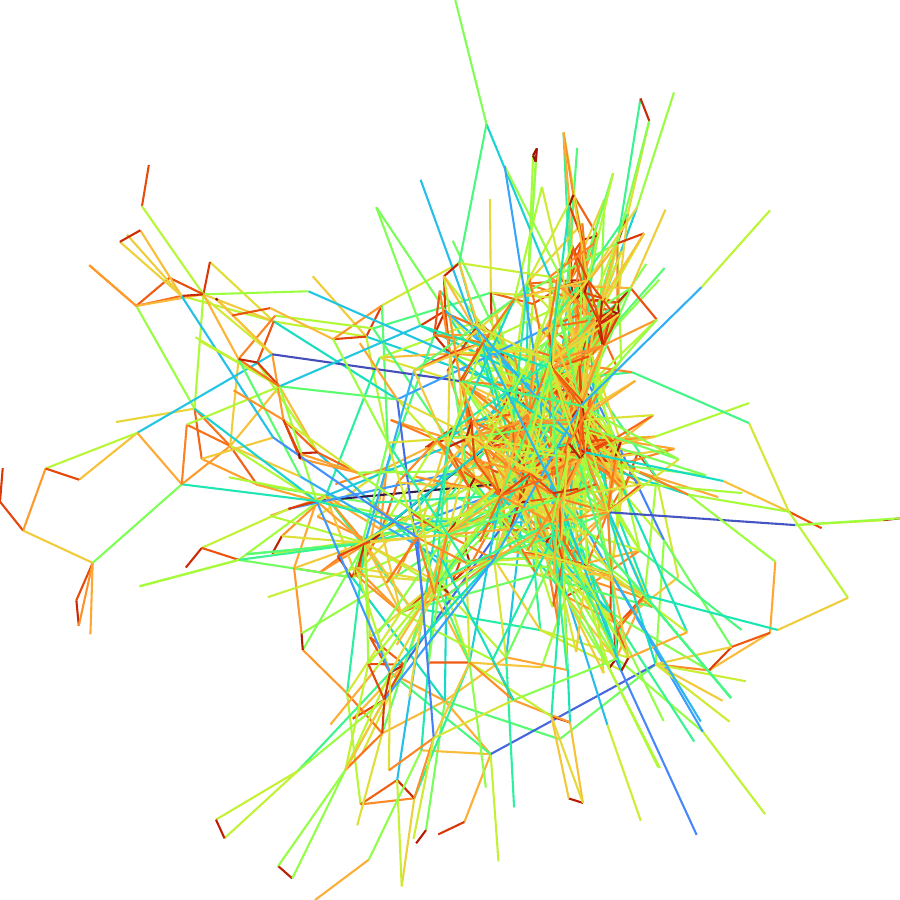} \\ \fontsize{7pt}{0pt}\selectfont 0.142,\textbf{0.131}} & \parbox{1.7cm}{\centering \includegraphics[height=1.1cm]{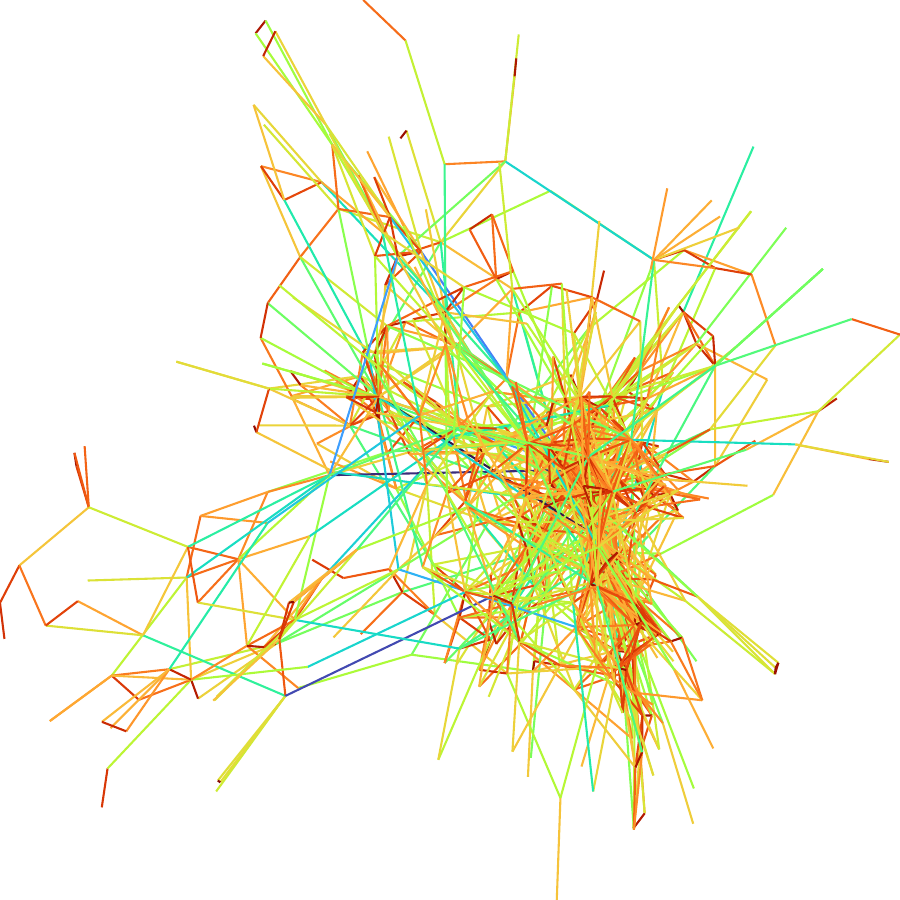} \\ \fontsize{7pt}{0pt}\selectfont 1.378,\textbf{1.334}} & \parbox{1.7cm}{\centering \includegraphics[height=1.1cm]{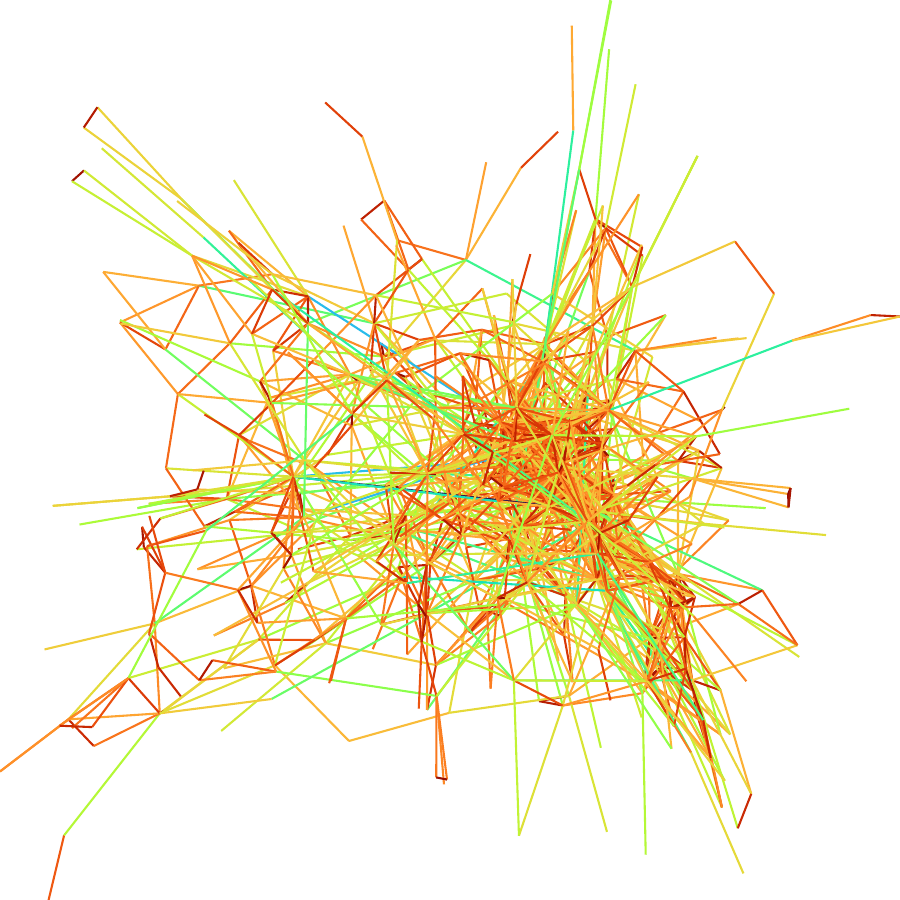} \\ \fontsize{7pt}{0pt}\selectfont 32632,\textbf{27123}} \\
EX1 & \parbox{1.7cm}{\centering \includegraphics[height=1.1cm]{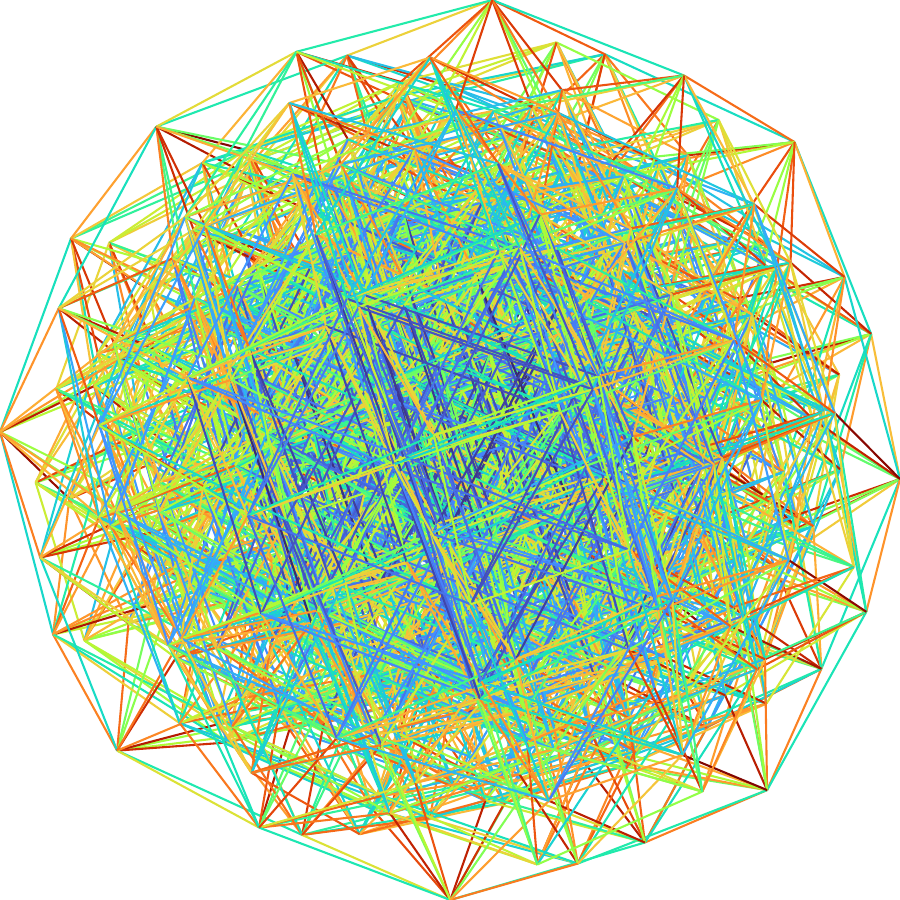} \\ \fontsize{7pt}{0pt}\selectfont } & \parbox{1.7cm}{\centering \includegraphics[height=1.1cm]{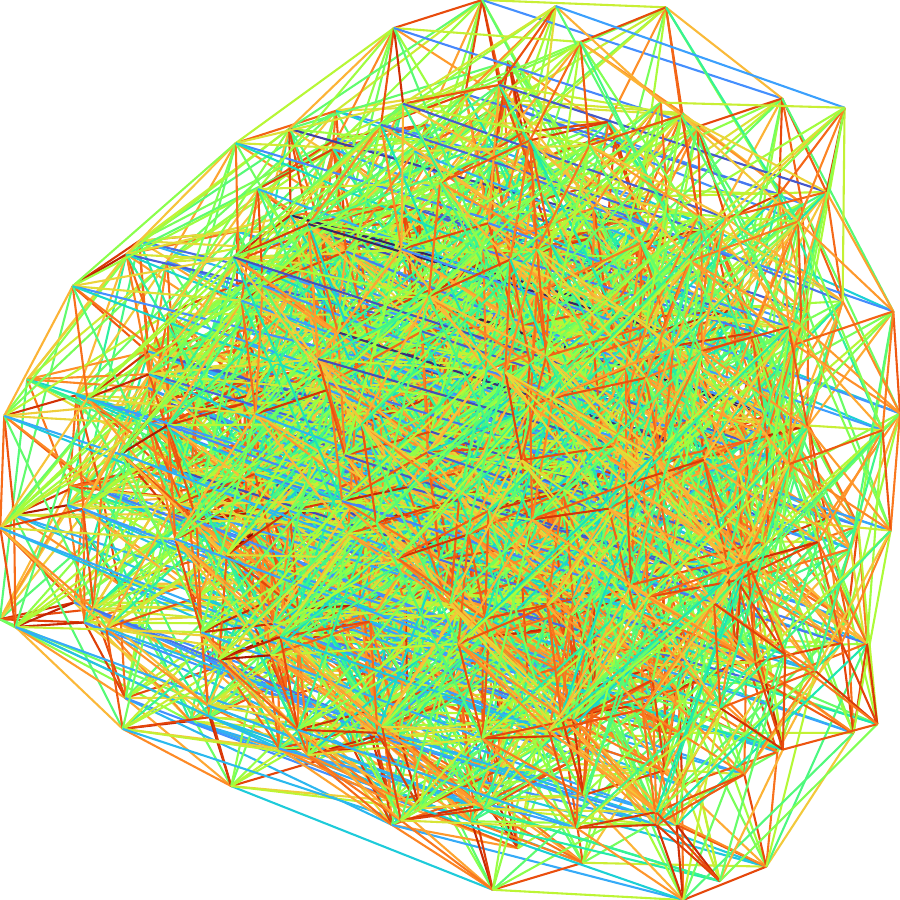} \\ \fontsize{7pt}{0pt}\selectfont 0.39,\textbf{0.367}} & \parbox{1.7cm}{\centering \includegraphics[height=1.1cm]{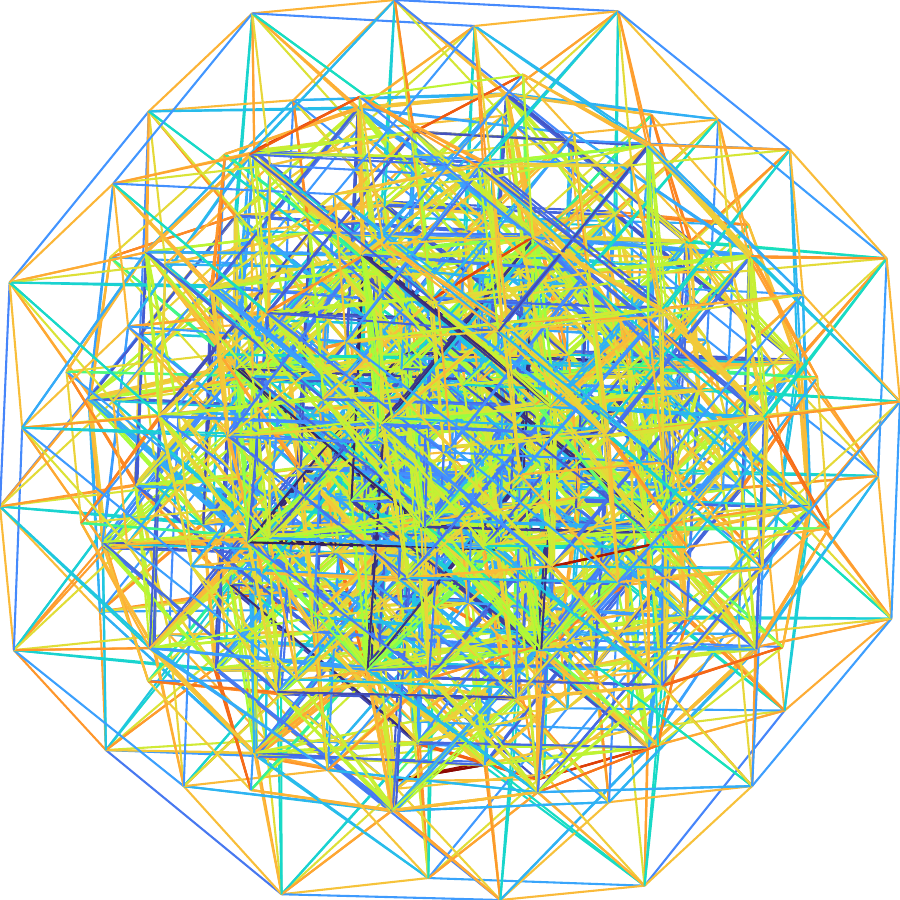} \\ \fontsize{7pt}{0pt}\selectfont 0.205,\textbf{0.183}} & \parbox{1.7cm}{\centering \includegraphics[height=1.1cm]{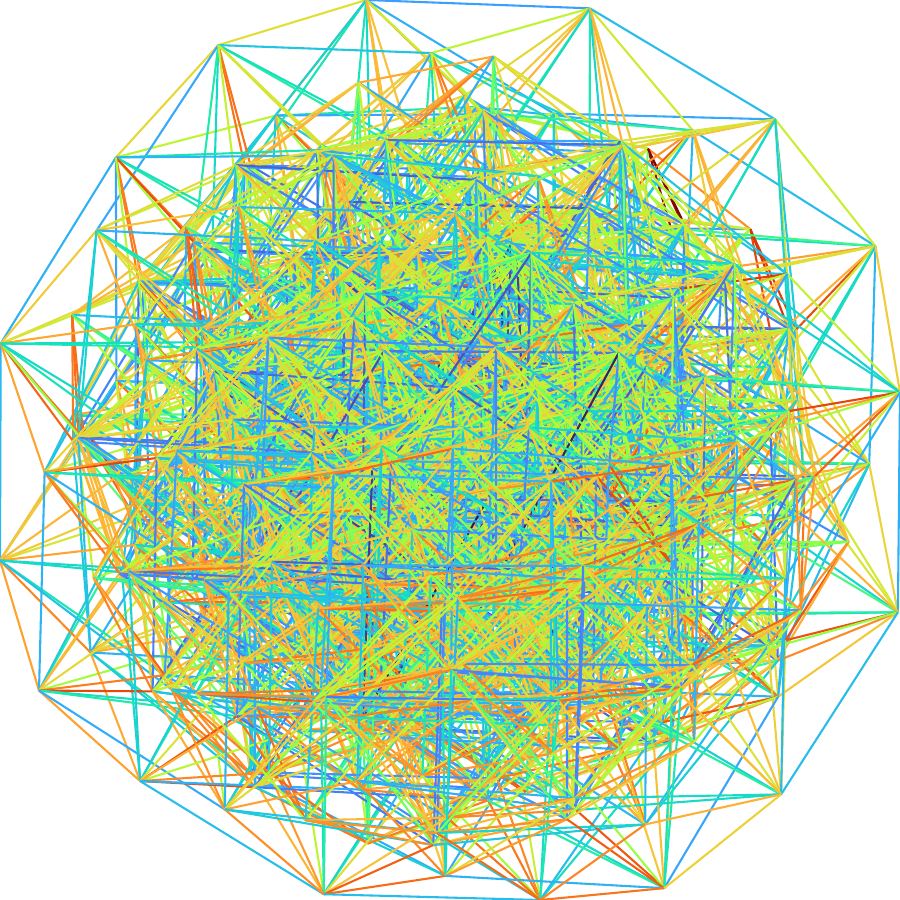} \\ \fontsize{7pt}{0pt}\selectfont \textbf{-1.493},-1.486} & \parbox{1.7cm}{\centering \includegraphics[height=1.1cm]{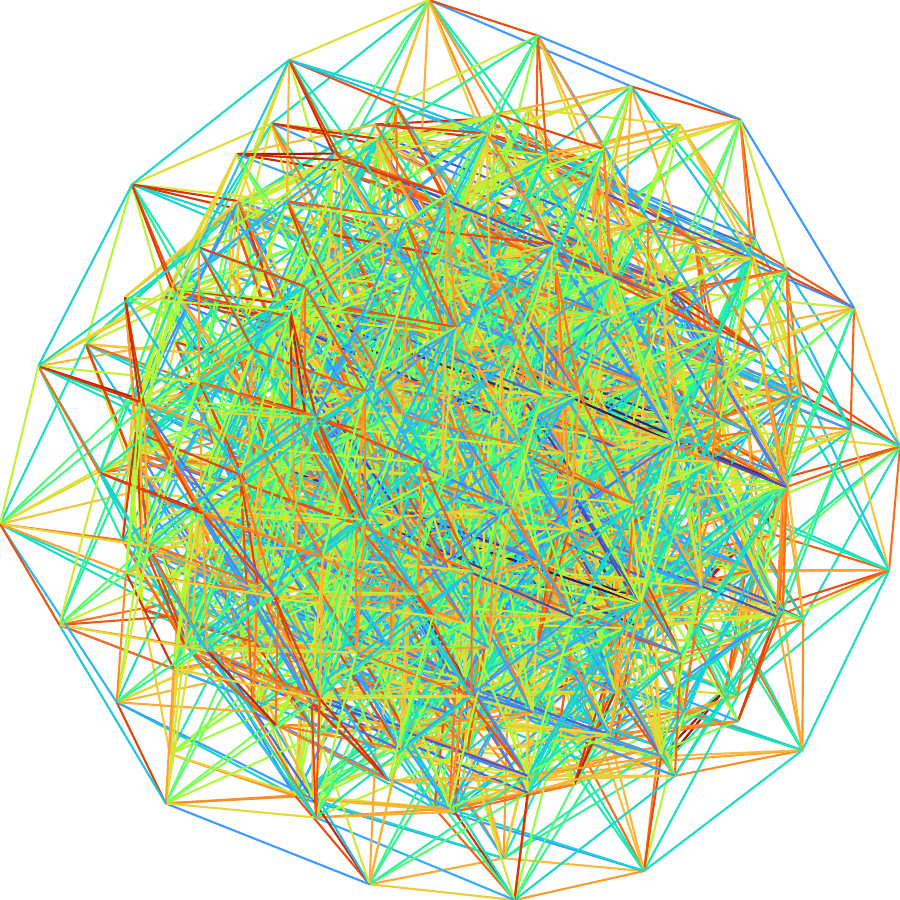} \\ \fontsize{7pt}{0pt}\selectfont \textbf{0.17},0.175} & \parbox{1.7cm}{\centering \includegraphics[height=1.1cm]{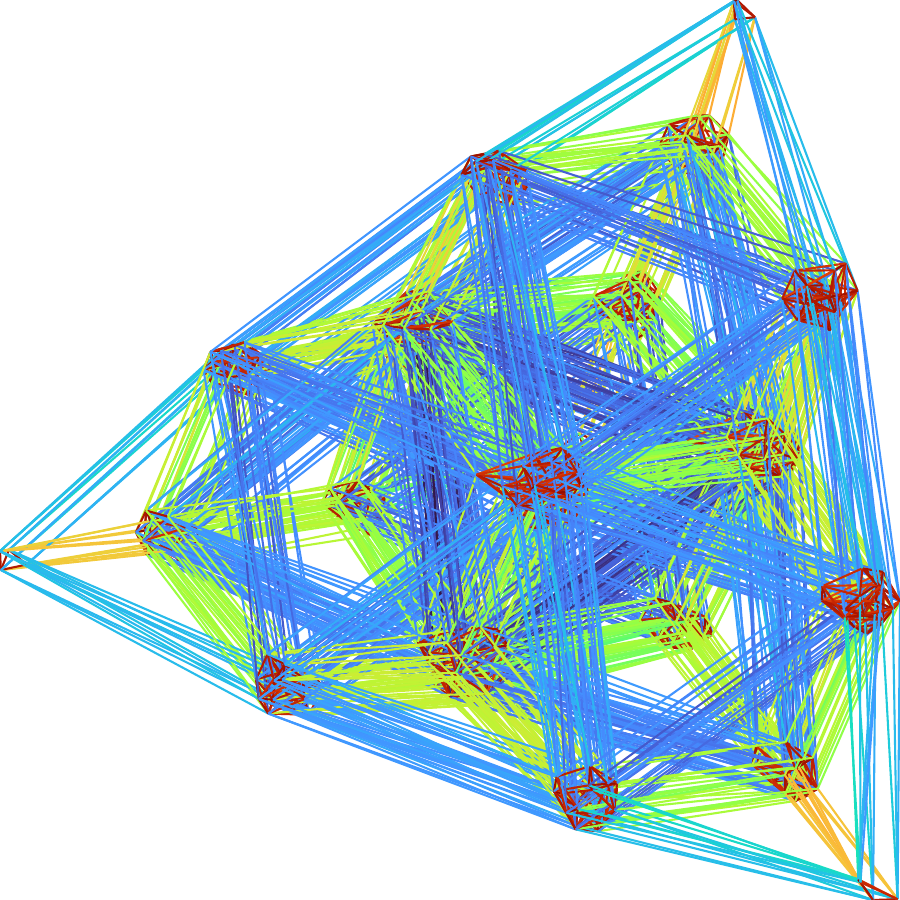} \\ \fontsize{7pt}{0pt}\selectfont 1.167,\textbf{1.089}} & \parbox{1.7cm}{\centering \includegraphics[height=1.1cm]{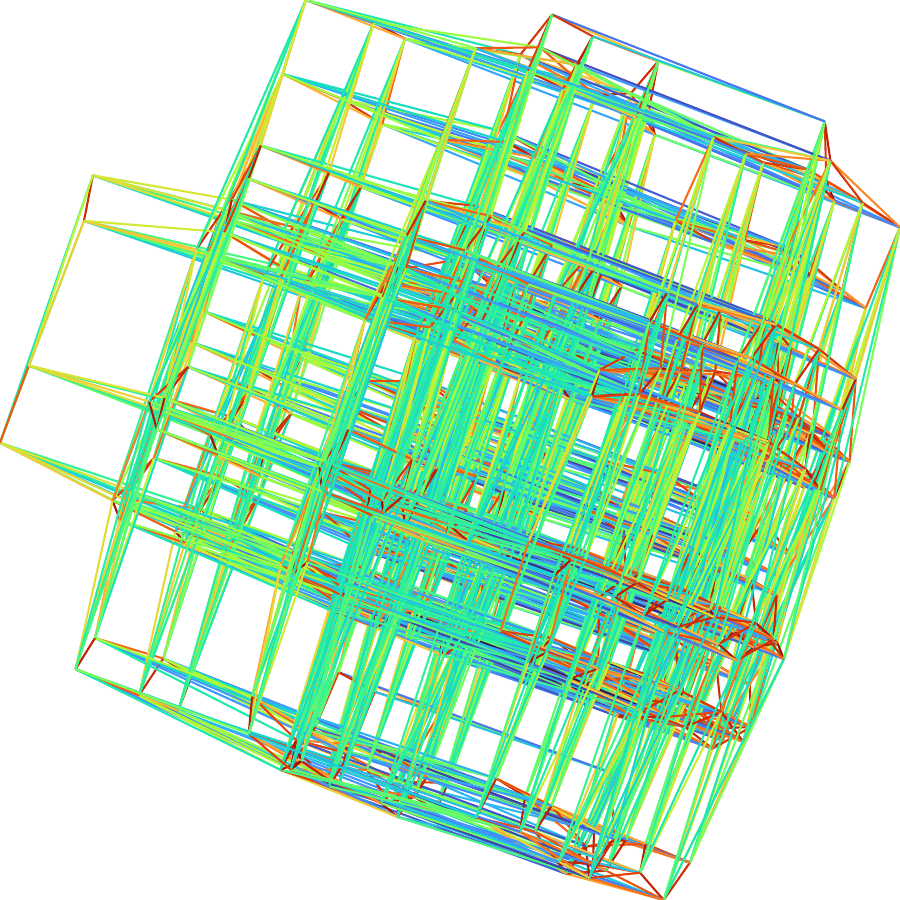} \\ \fontsize{7pt}{0pt}\selectfont 492441,\textbf{349871}} \\
football & \parbox{1.7cm}{\centering \includegraphics[height=1.1cm]{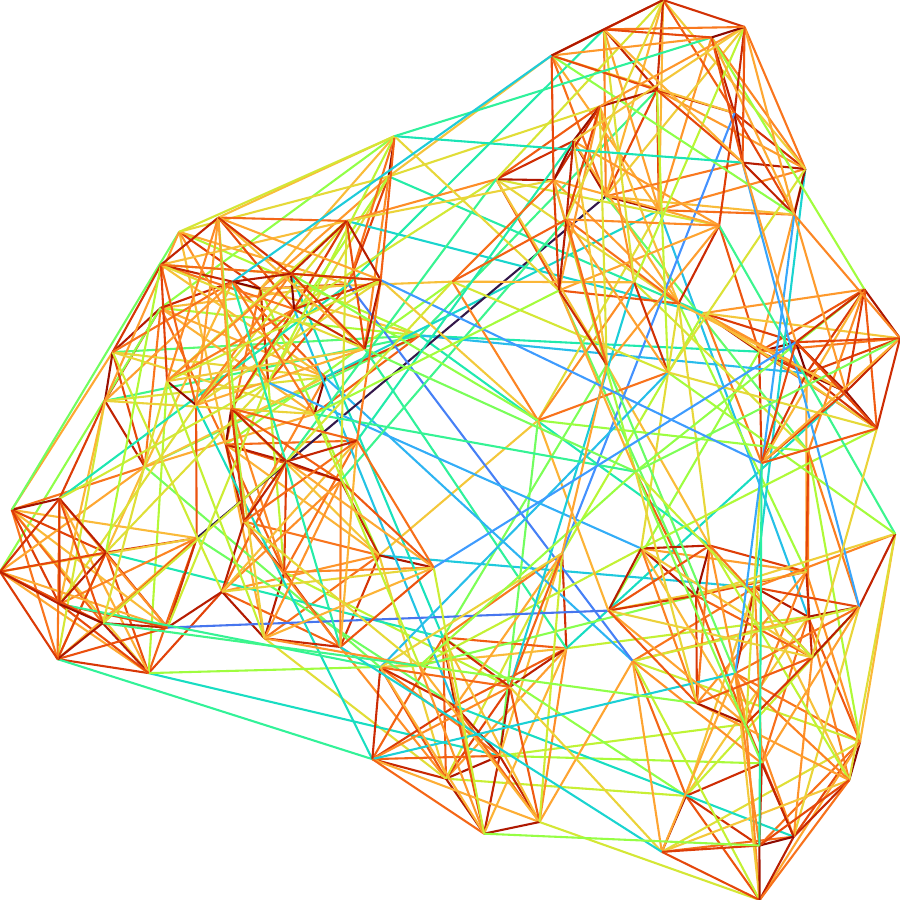} \\ \fontsize{7pt}{0pt}\selectfont } & \parbox{1.7cm}{\centering \includegraphics[height=1.1cm]{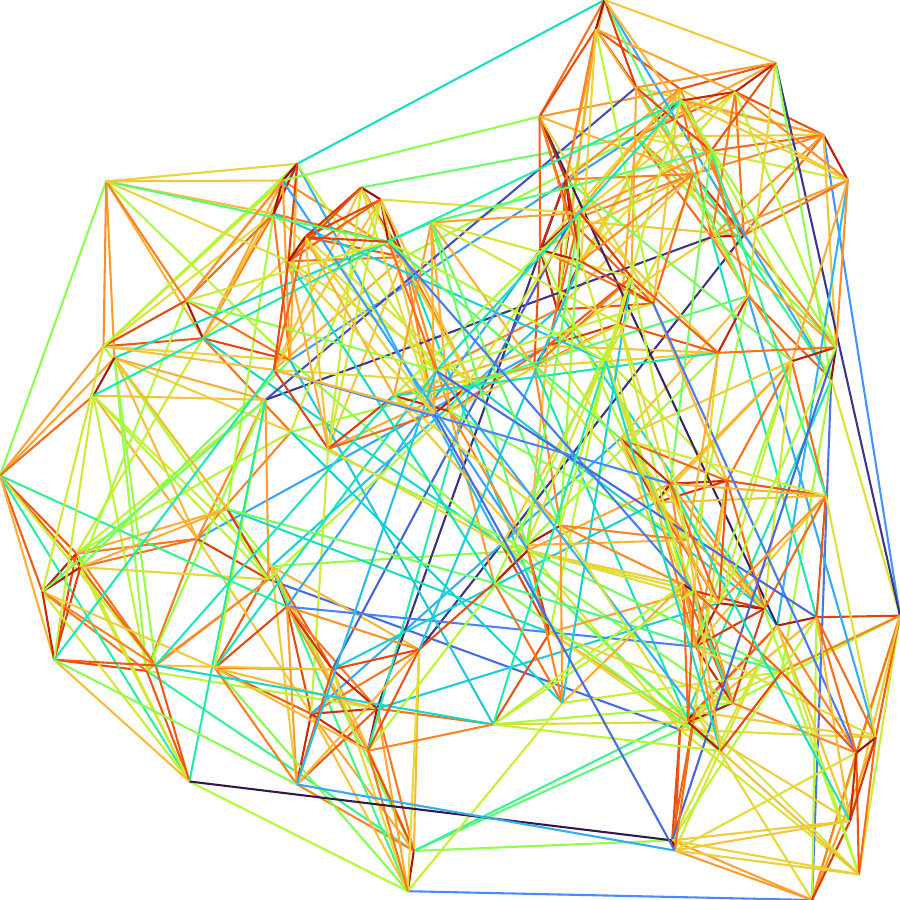} \\ \fontsize{7pt}{0pt}\selectfont 0.539,\textbf{0.526}} & \parbox{1.7cm}{\centering \includegraphics[height=1.1cm]{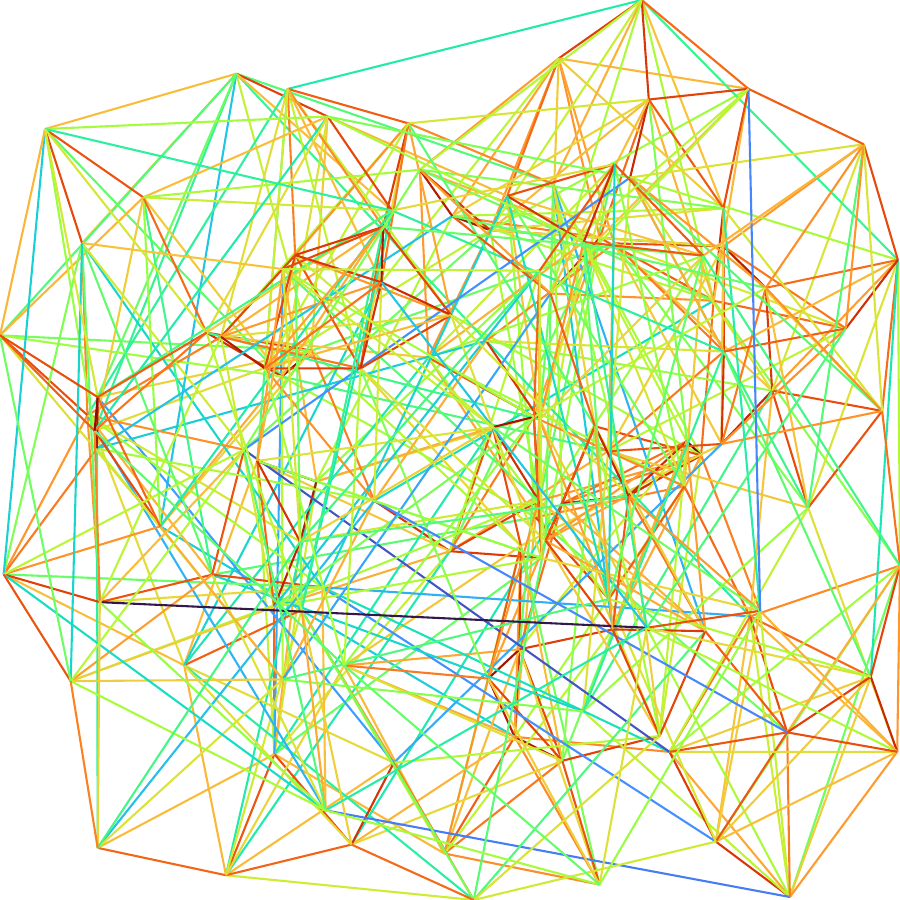} \\ \fontsize{7pt}{0pt}\selectfont 0.519,\textbf{0.438}} & \parbox{1.7cm}{\centering \includegraphics[height=1.1cm]{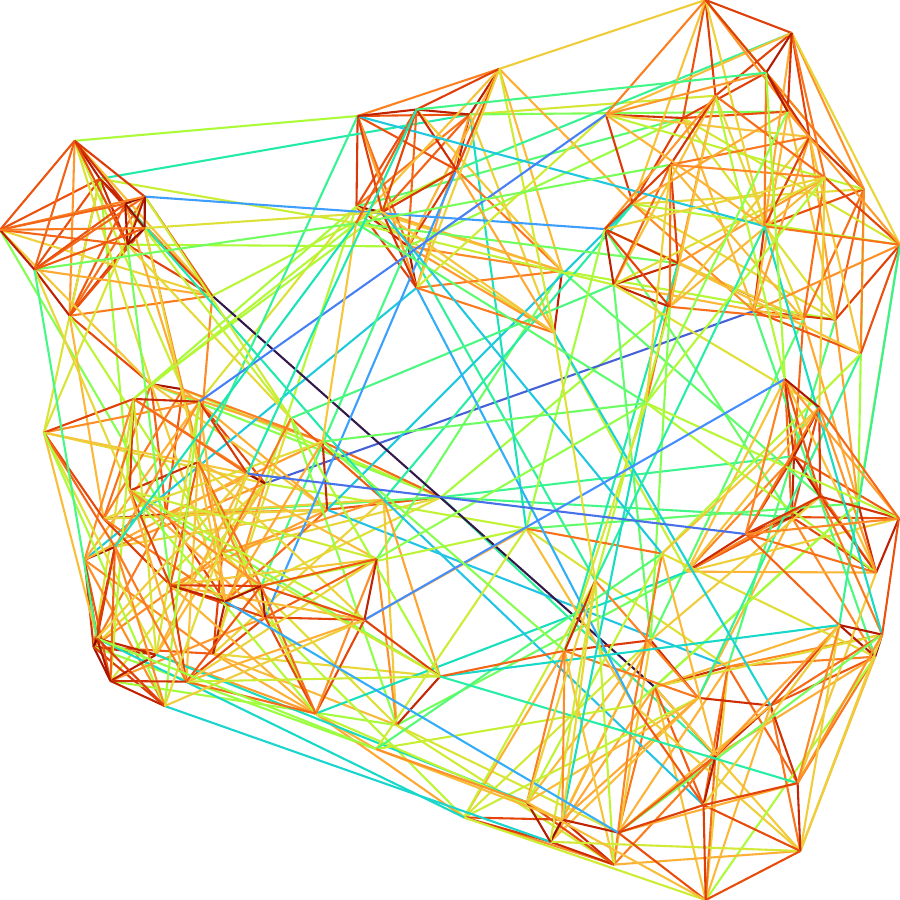} \\ \fontsize{7pt}{0pt}\selectfont \textbf{-1.196},-1.181} & \parbox{1.7cm}{\centering \includegraphics[height=1.1cm]{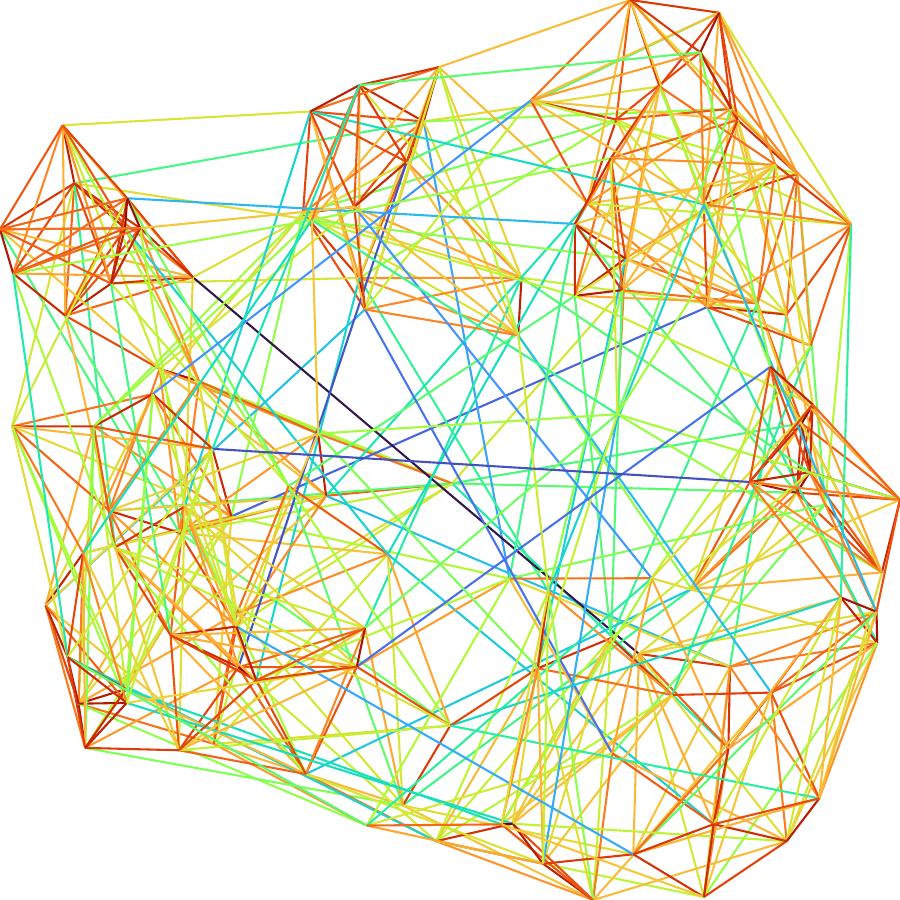} \\ \fontsize{7pt}{0pt}\selectfont 0.138,\textbf{0.135}} & \parbox{1.7cm}{\centering \includegraphics[height=1.1cm]{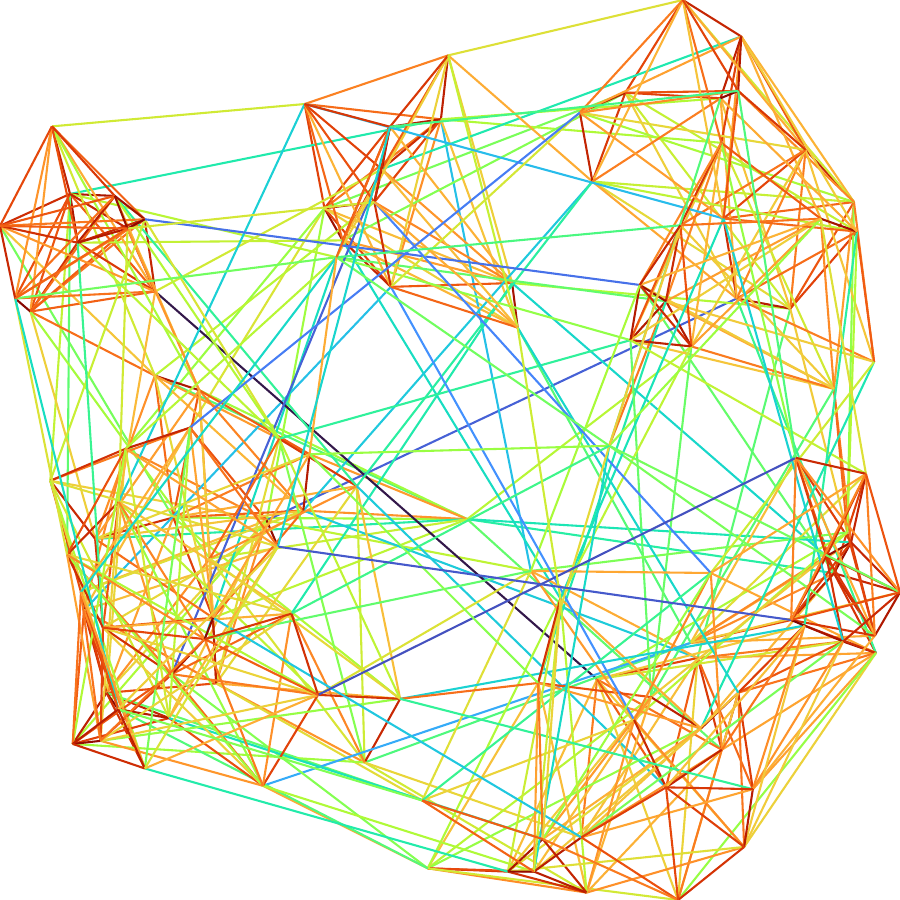} \\ \fontsize{7pt}{0pt}\selectfont 0.508,\textbf{0.507}} & \parbox{1.7cm}{\centering \includegraphics[height=1.1cm]{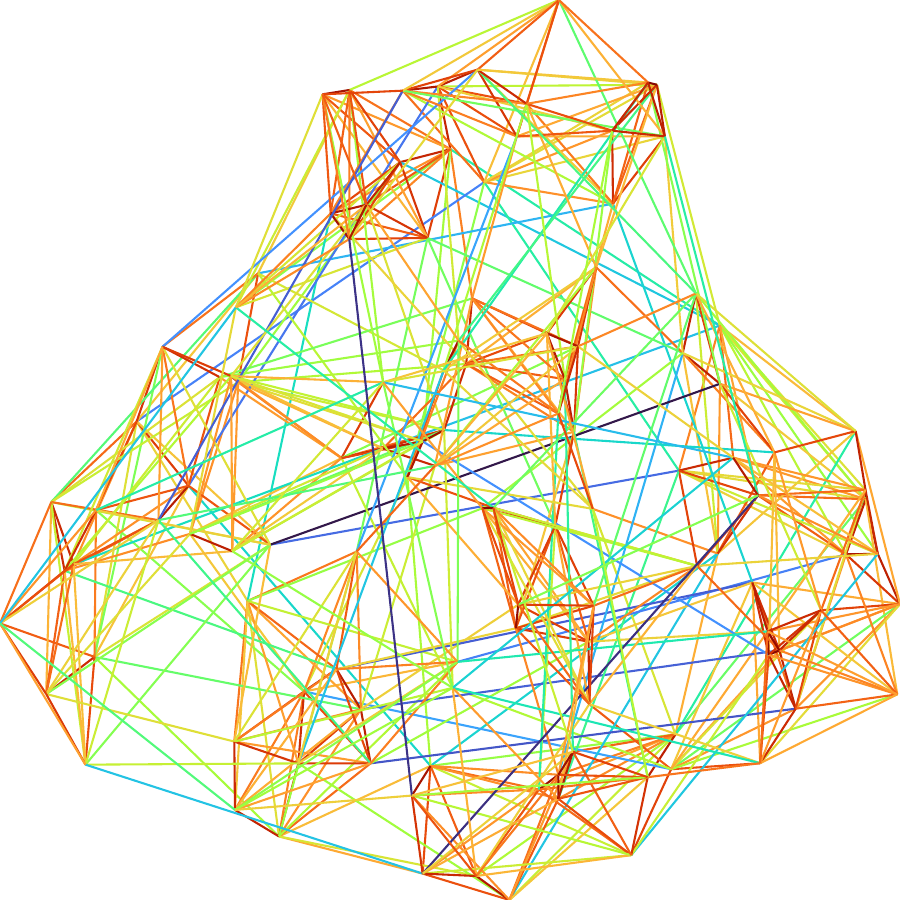} \\ \fontsize{7pt}{0pt}\selectfont 5622,\textbf{4759}} \\
hypercube10 & \parbox{1.7cm}{\centering \includegraphics[height=1.1cm]{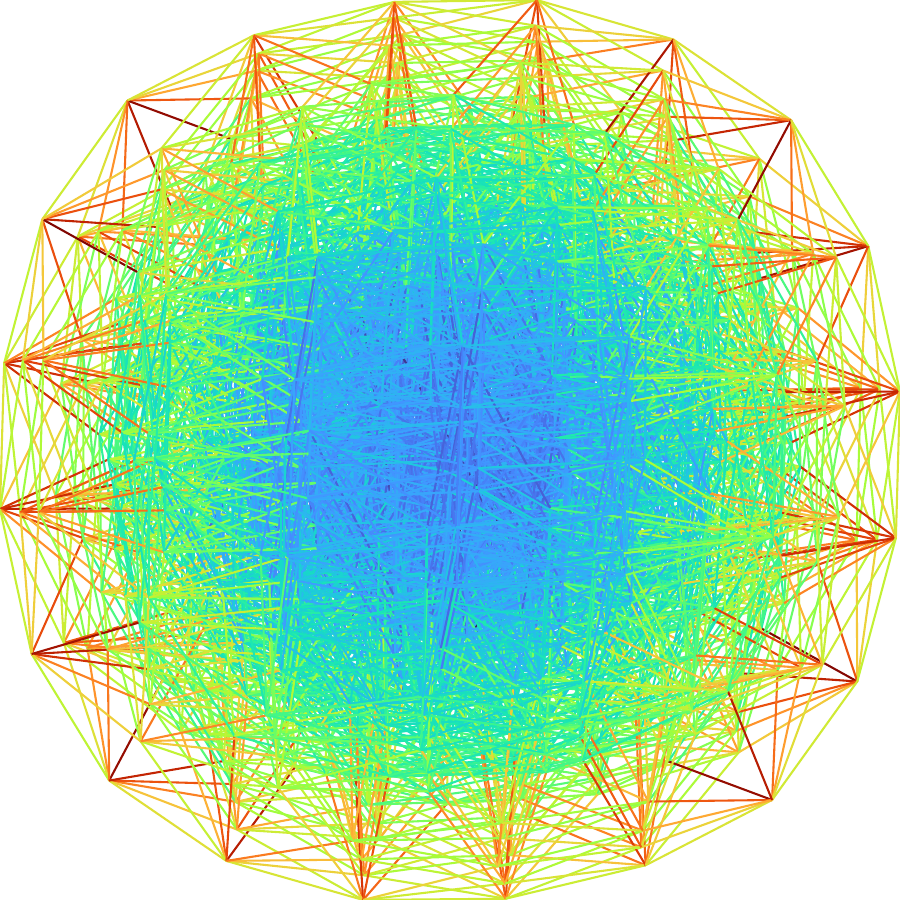} \\ \fontsize{7pt}{0pt}\selectfont } & \parbox{1.7cm}{\centering \includegraphics[height=1.1cm]{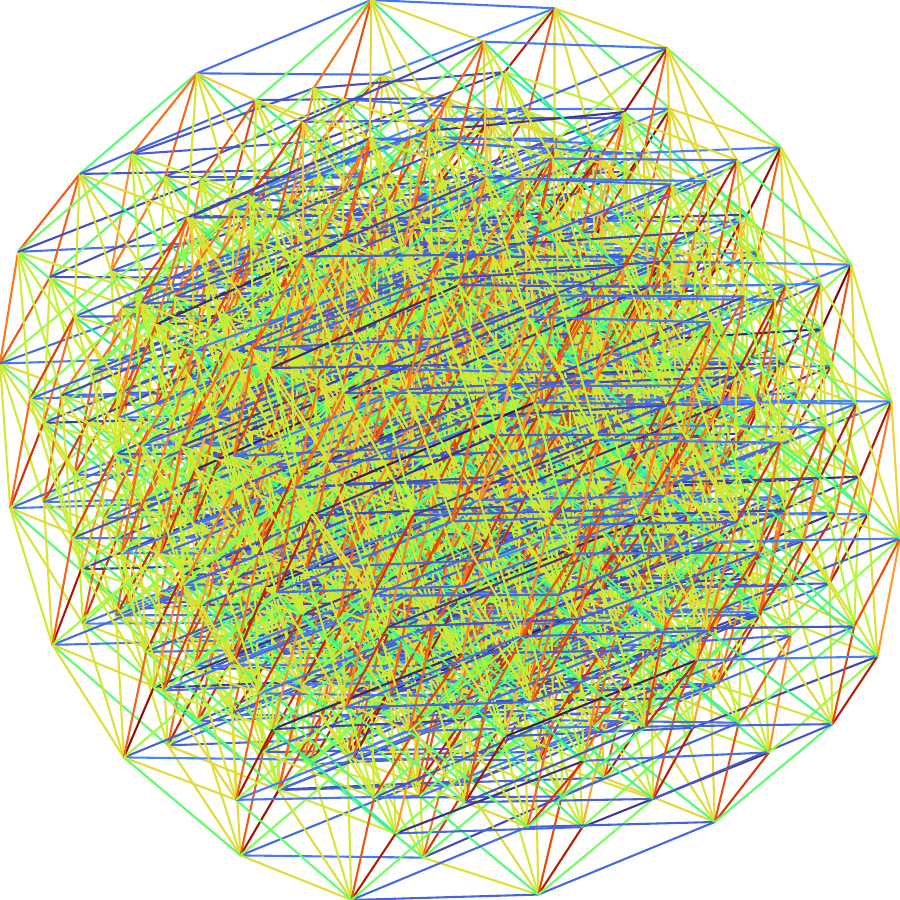} \\ \fontsize{7pt}{0pt}\selectfont 0.486,\textbf{0.356}} & \parbox{1.7cm}{\centering \includegraphics[height=1.1cm]{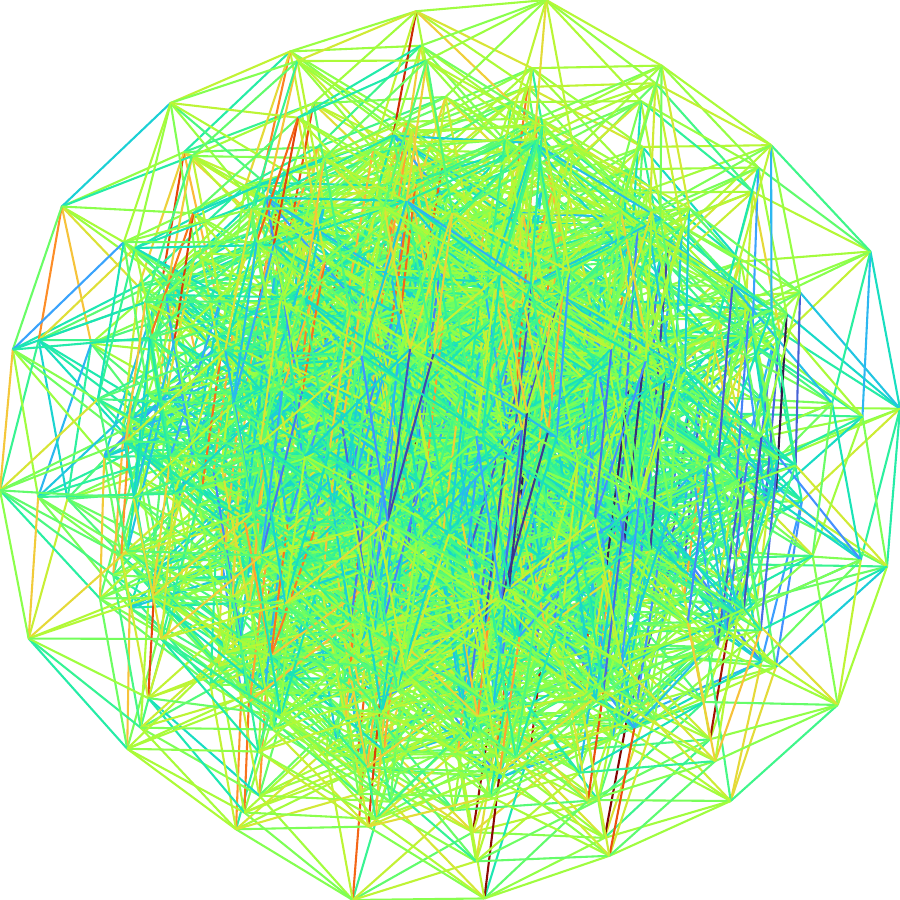} \\ \fontsize{7pt}{0pt}\selectfont 0.081,\textbf{0.04}} & \parbox{1.7cm}{\centering \includegraphics[height=1.1cm]{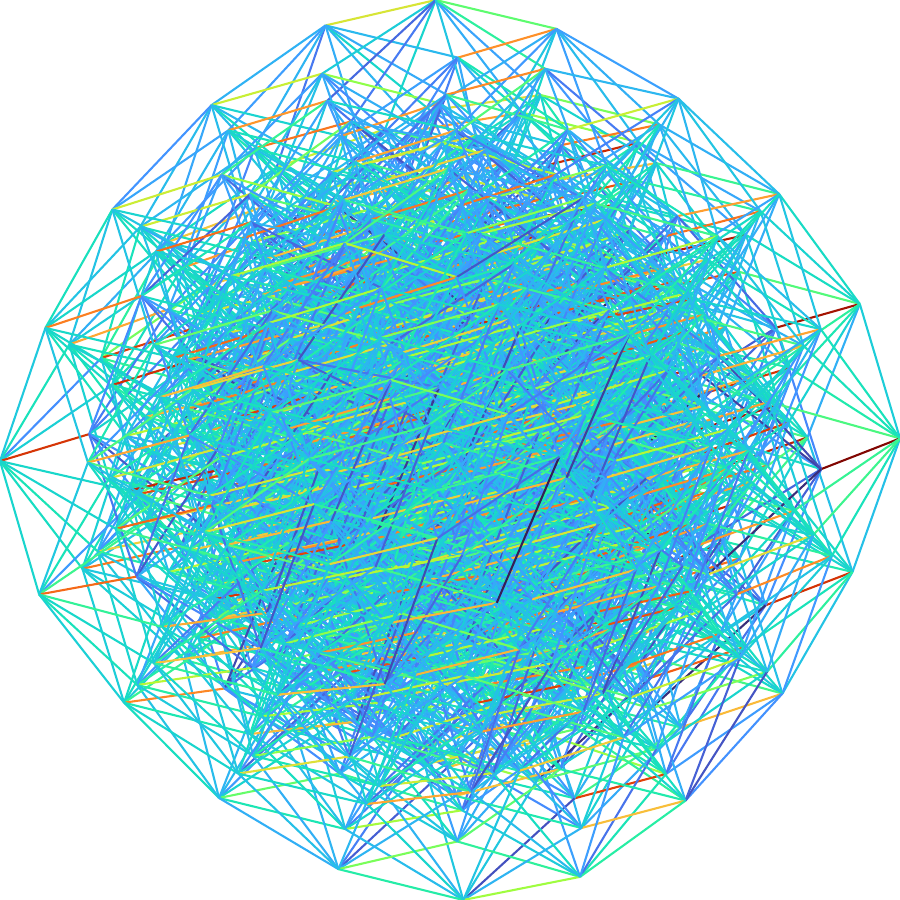} \\ \fontsize{7pt}{0pt}\selectfont \textbf{-1.98},-1.97} & \parbox{1.7cm}{\centering \includegraphics[height=1.1cm]{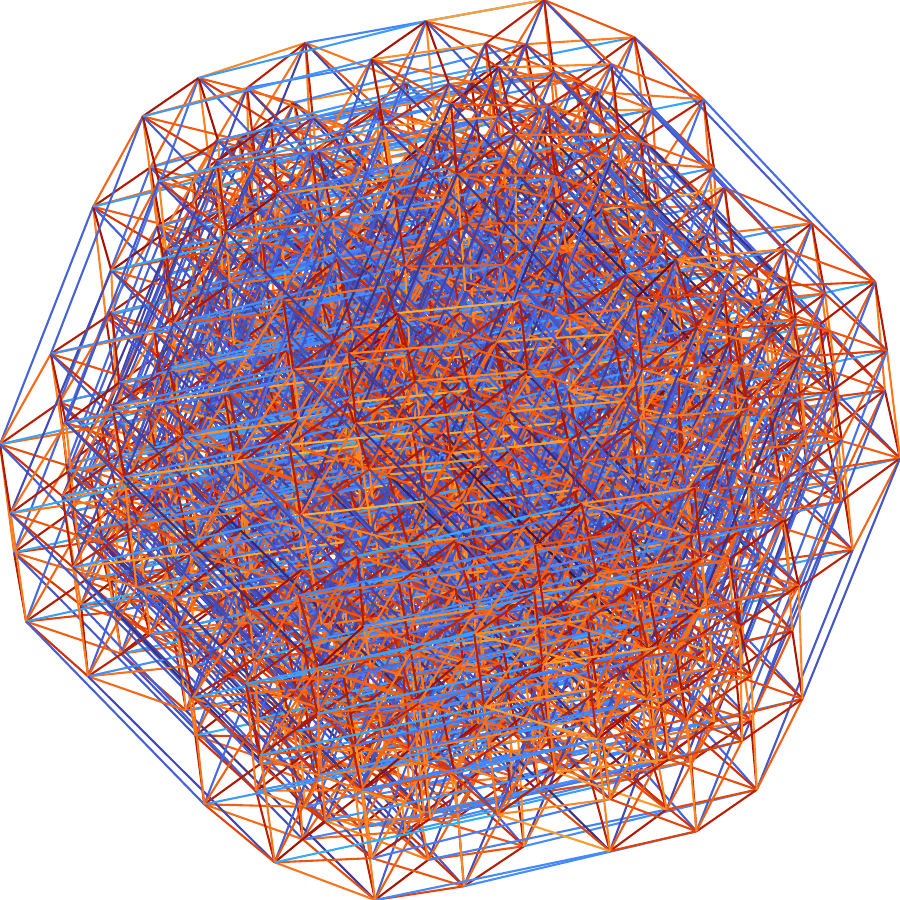} \\ \fontsize{7pt}{0pt}\selectfont \textbf{0.202},0.206} & \parbox{1.7cm}{\centering \includegraphics[height=1.1cm]{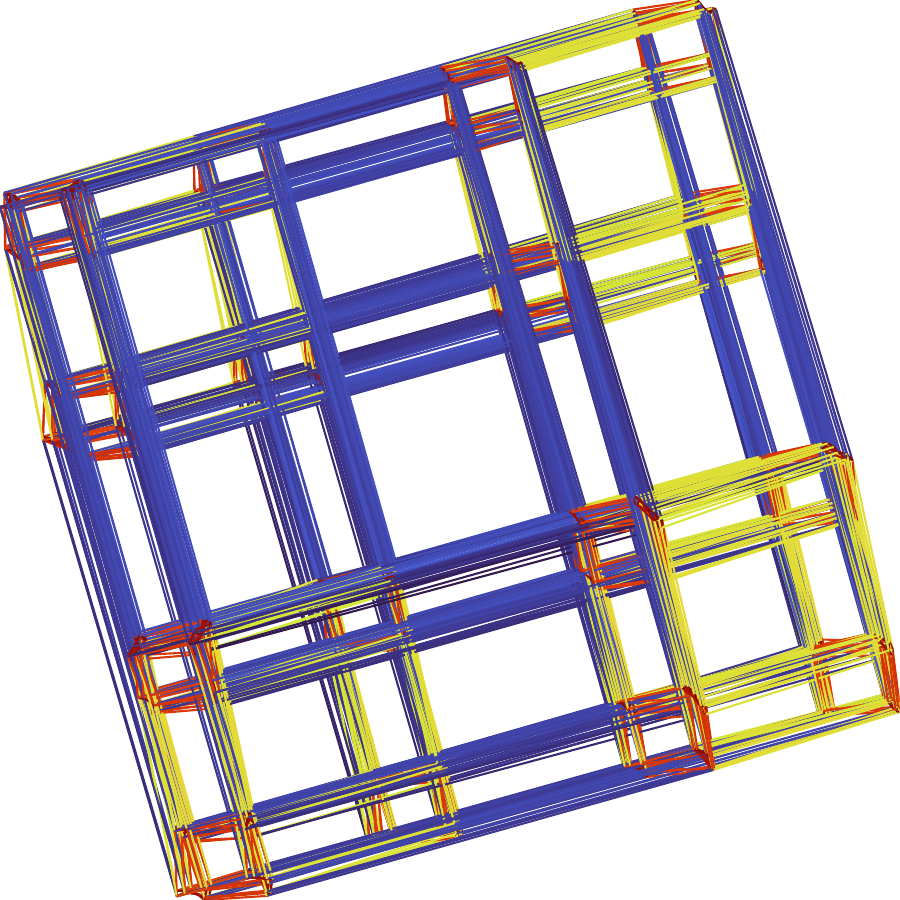} \\ \fontsize{7pt}{0pt}\selectfont 2.343,\textbf{1.991}} & \parbox{1.7cm}{\centering \includegraphics[height=1.1cm]{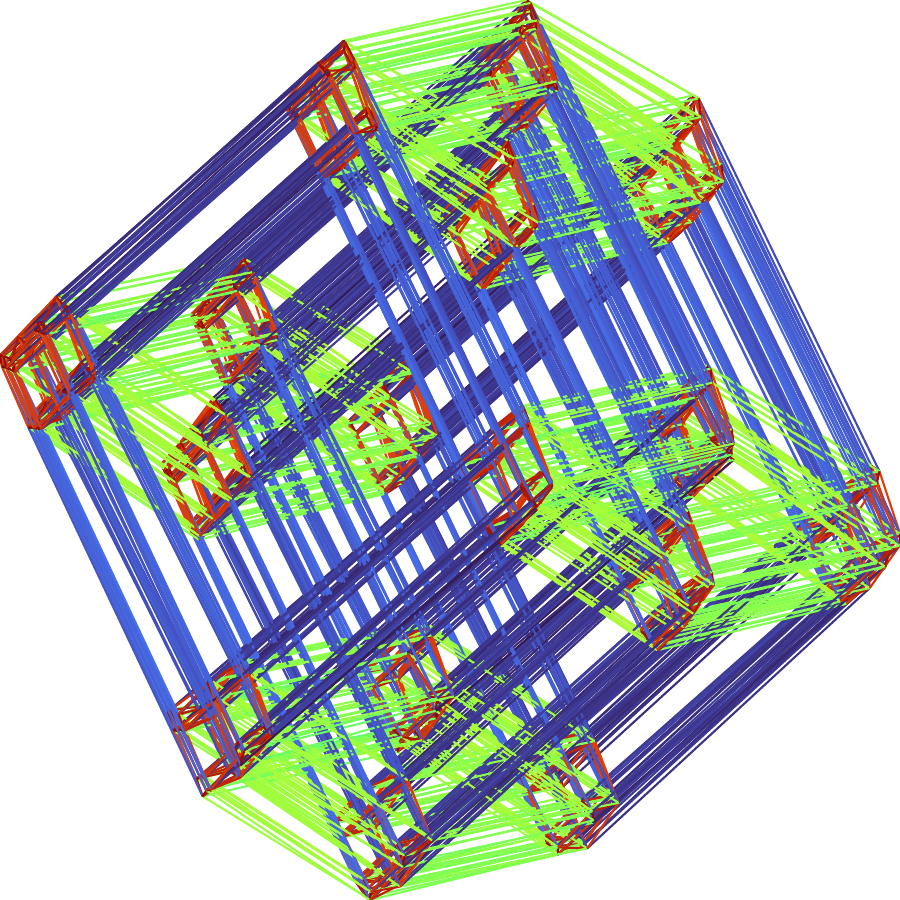} \\ \fontsize{7pt}{0pt}\selectfont 518024,\textbf{254686}} \\
Journals & \parbox{1.7cm}{\centering \includegraphics[height=1.1cm]{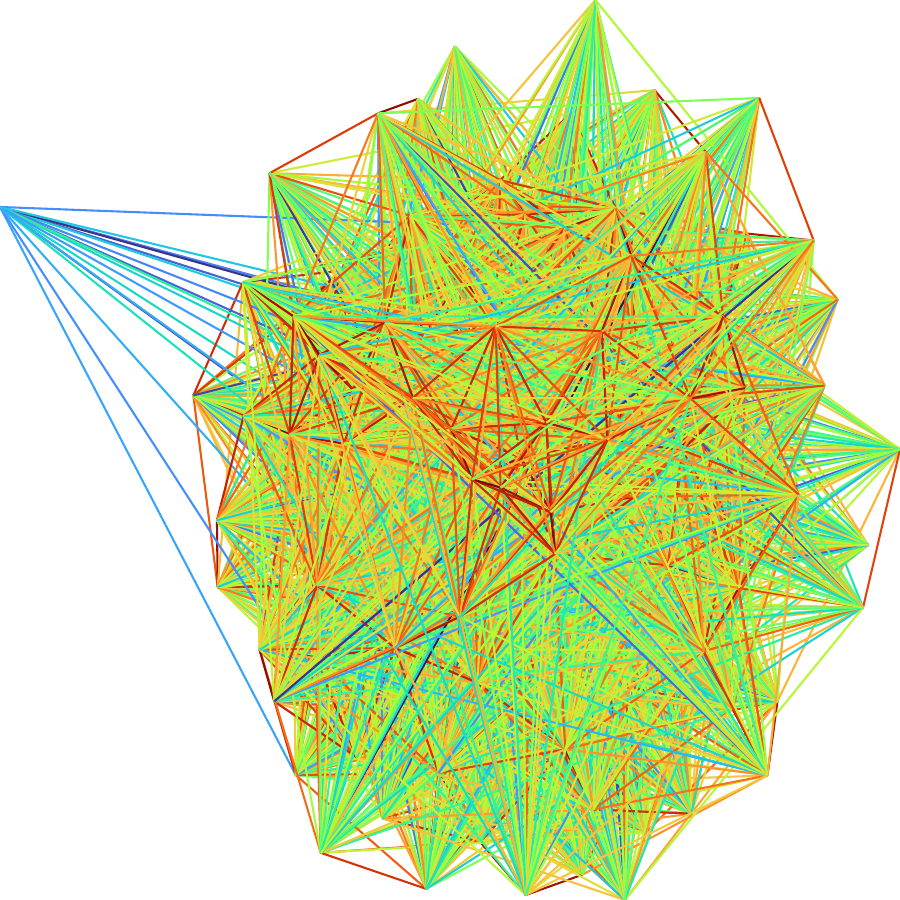} \\ \fontsize{7pt}{0pt}\selectfont } & \parbox{1.7cm}{\centering \includegraphics[height=1.1cm]{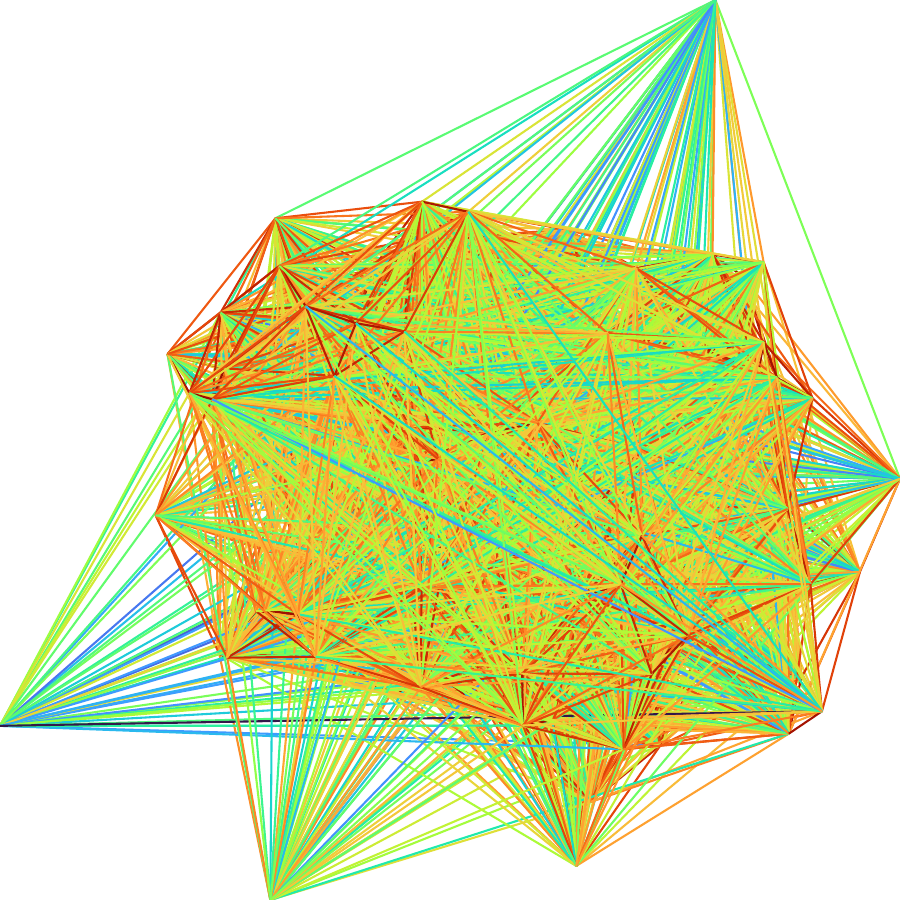} \\ \fontsize{7pt}{0pt}\selectfont 0.077,\textbf{0.074}} & \parbox{1.7cm}{\centering \includegraphics[height=1.1cm]{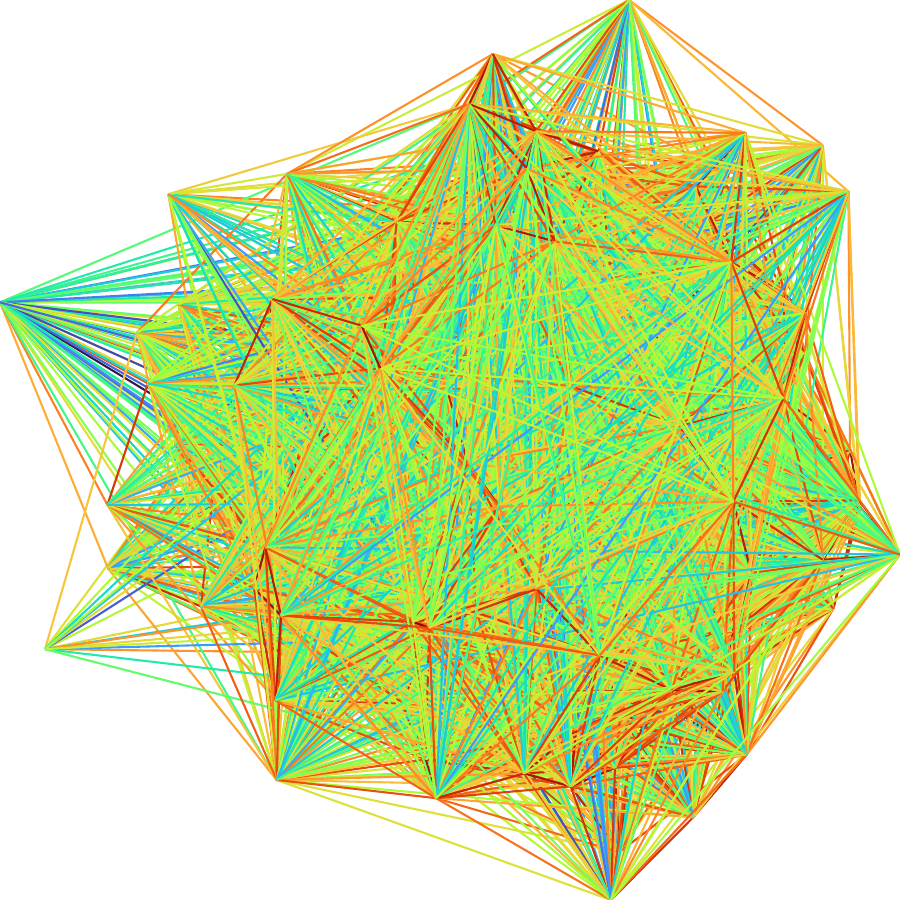} \\ \fontsize{7pt}{0pt}\selectfont \textbf{0.446},0.462} & \parbox{1.7cm}{\centering \includegraphics[height=1.1cm]{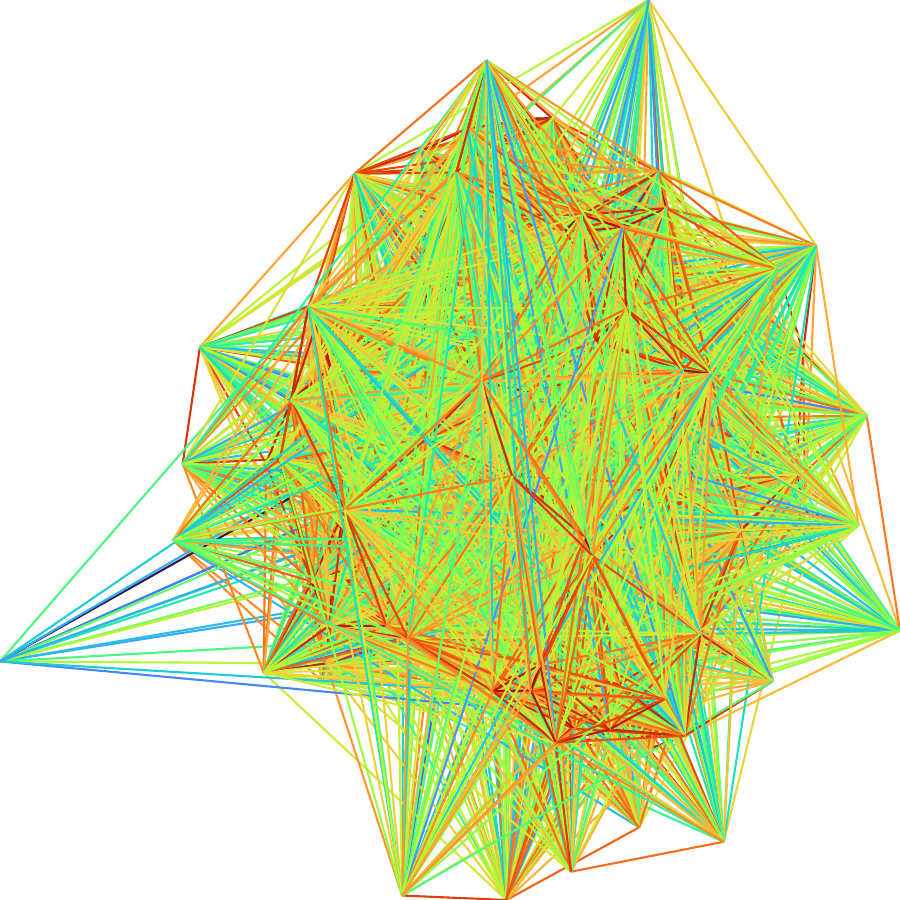} \\ \fontsize{7pt}{0pt}\selectfont \textbf{0.21},0.237} & \parbox{1.7cm}{\centering \includegraphics[height=1.1cm]{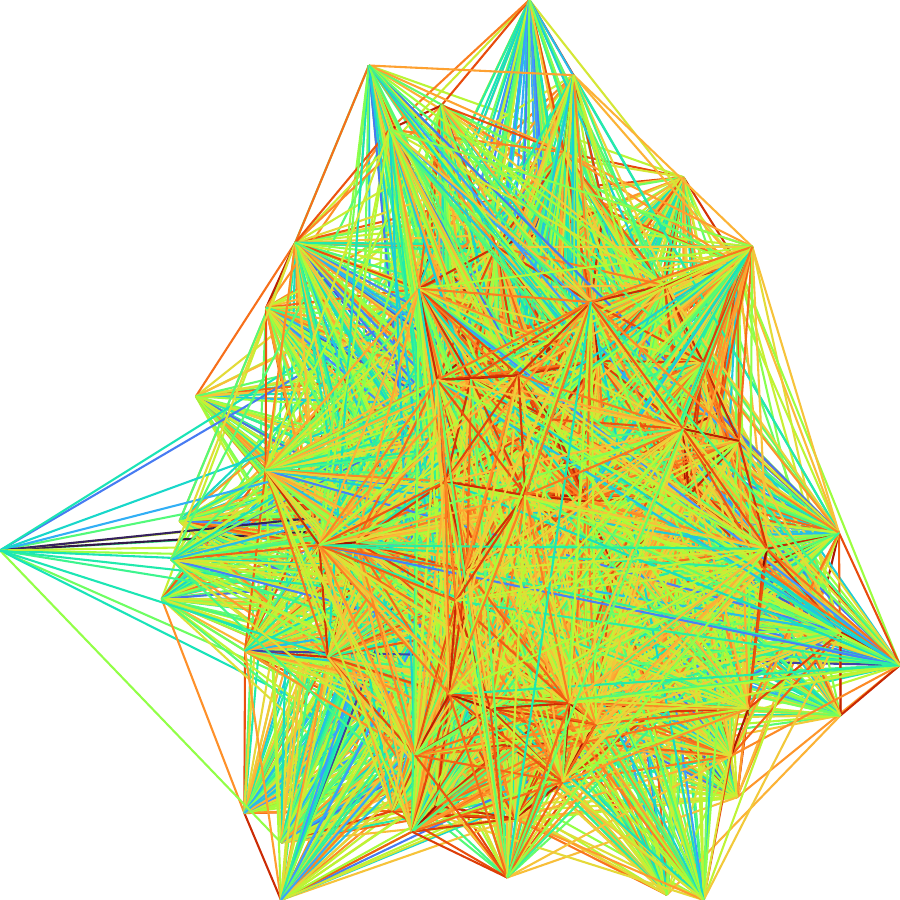} \\ \fontsize{7pt}{0pt}\selectfont \textbf{0.17},0.182} & \parbox{1.7cm}{\centering \includegraphics[height=1.1cm]{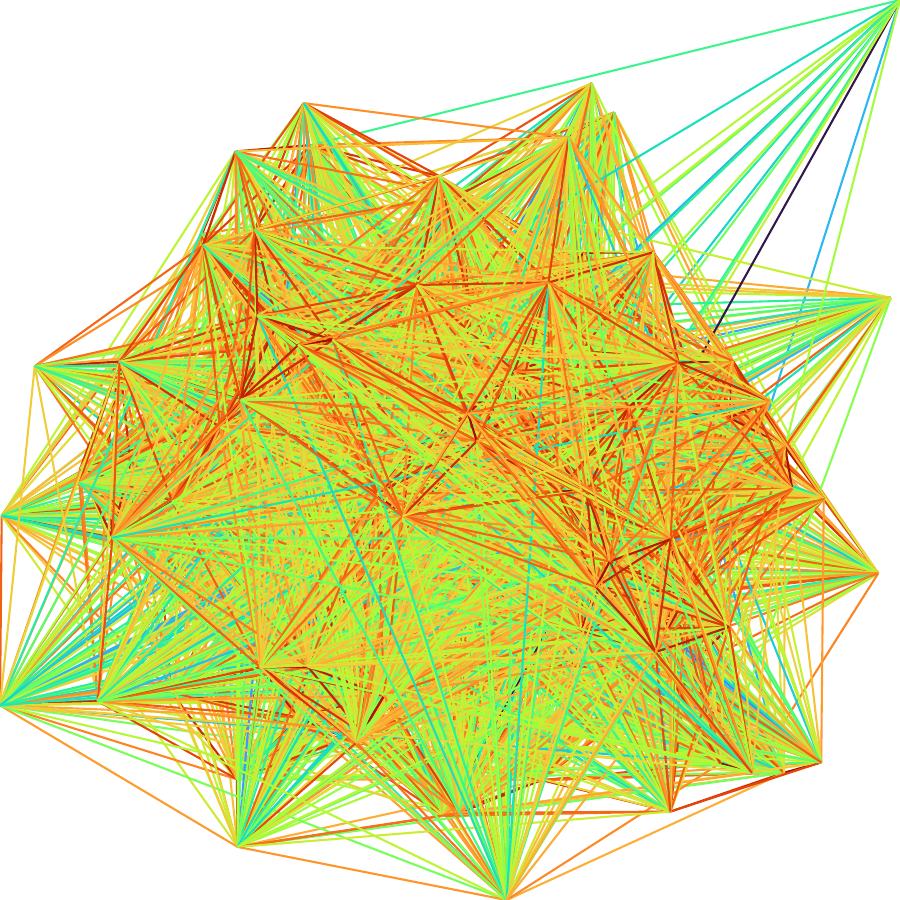} \\ \fontsize{7pt}{0pt}\selectfont 0.078,\textbf{0.077}} & \parbox{1.7cm}{\centering \includegraphics[height=1.1cm]{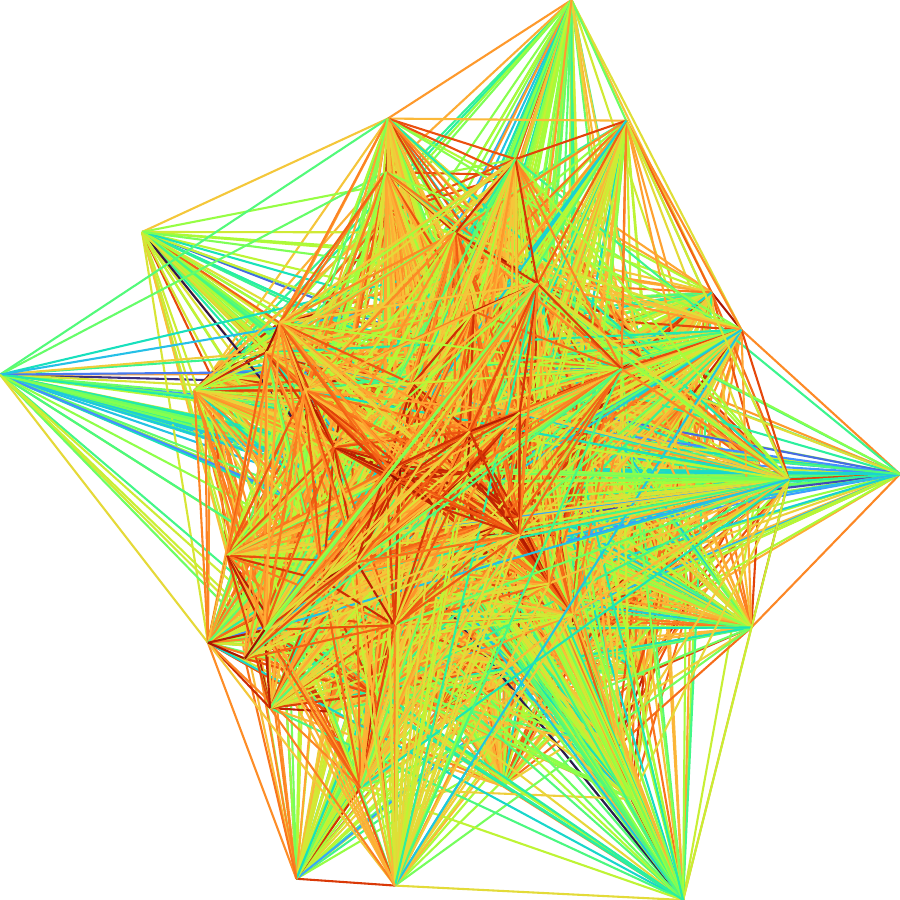} \\ \fontsize{7pt}{0pt}\selectfont 3720645,\textbf{3241000}} \\
mobius & \parbox{1.7cm}{\centering \includegraphics[height=1.1cm]{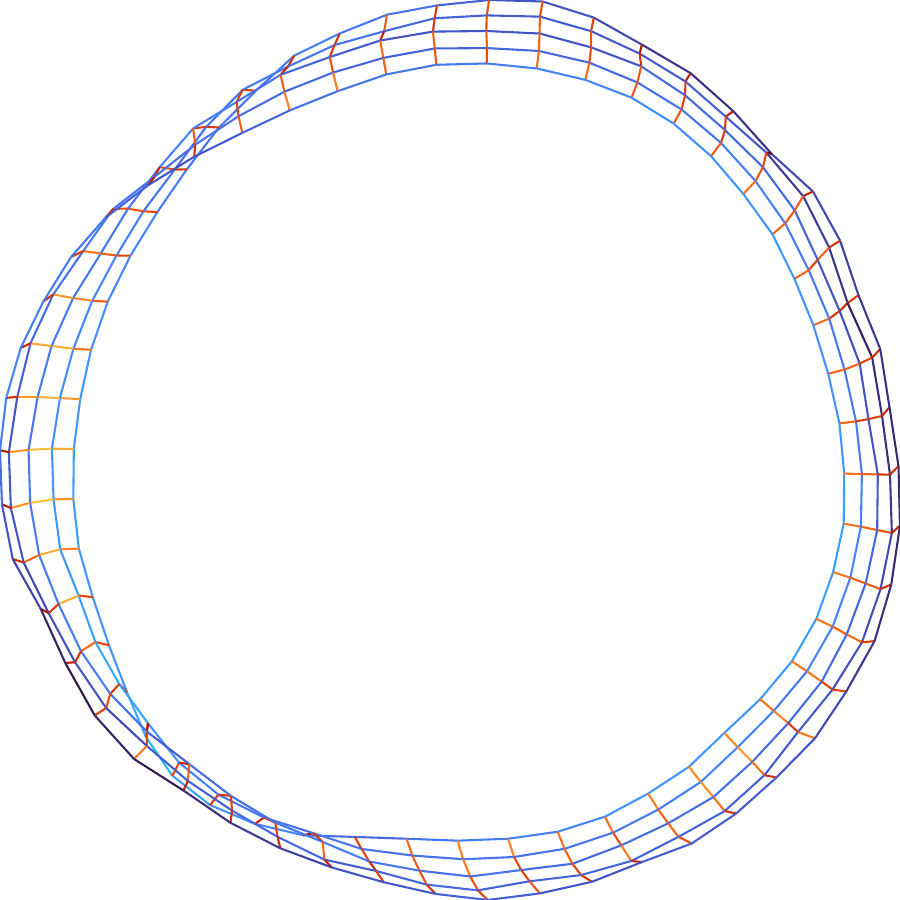} \\ \fontsize{7pt}{0pt}\selectfont } & \parbox{1.7cm}{\centering \includegraphics[height=1.1cm]{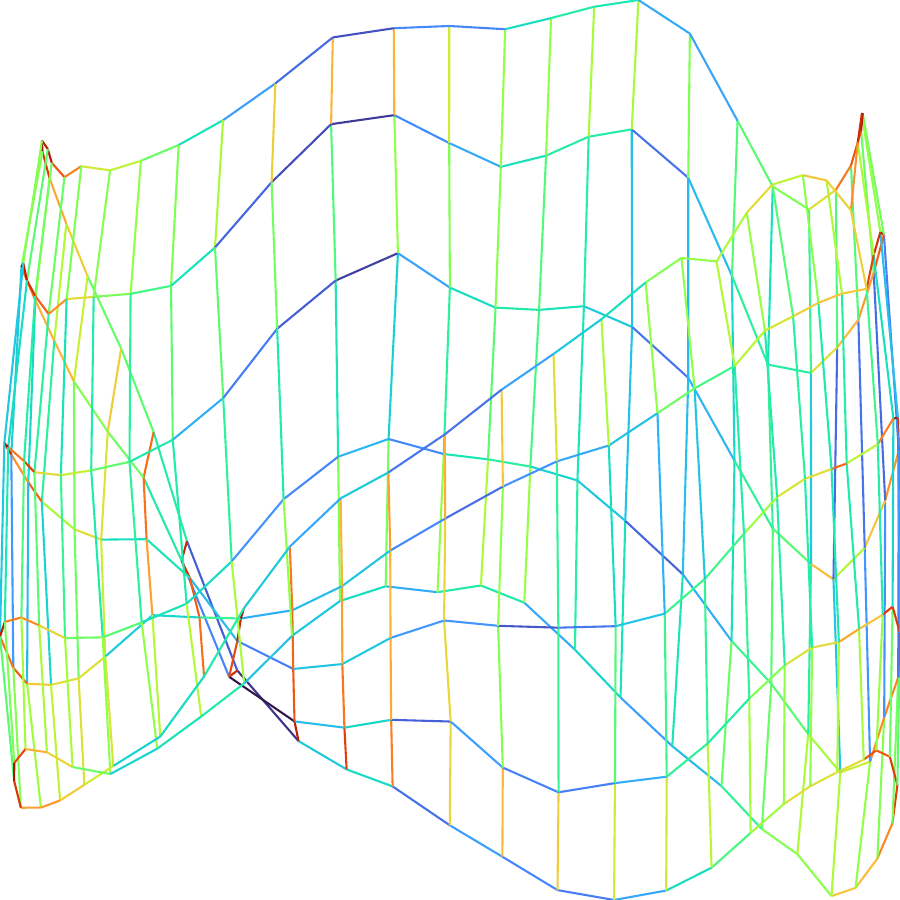} \\ \fontsize{7pt}{0pt}\selectfont 0.819,\textbf{0.65}} & \parbox{1.7cm}{\centering \includegraphics[height=1.1cm]{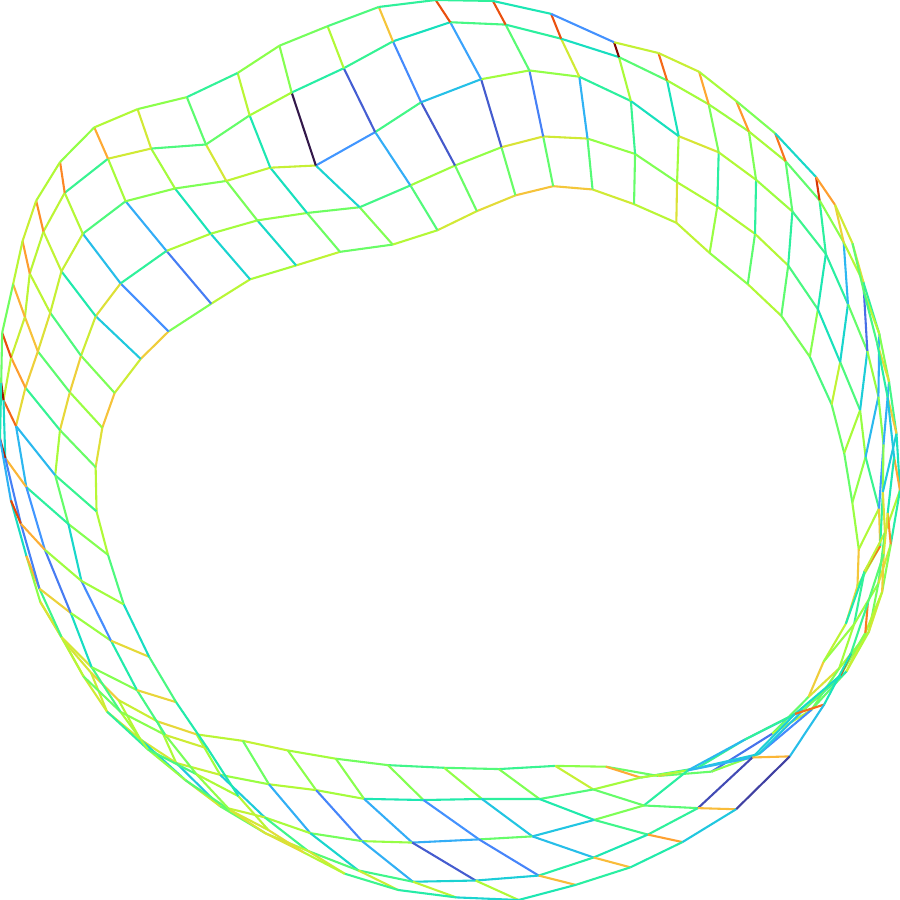} \\ \fontsize{7pt}{0pt}\selectfont 0.524,\textbf{0.193}} & \parbox{1.7cm}{\centering \includegraphics[height=1.1cm]{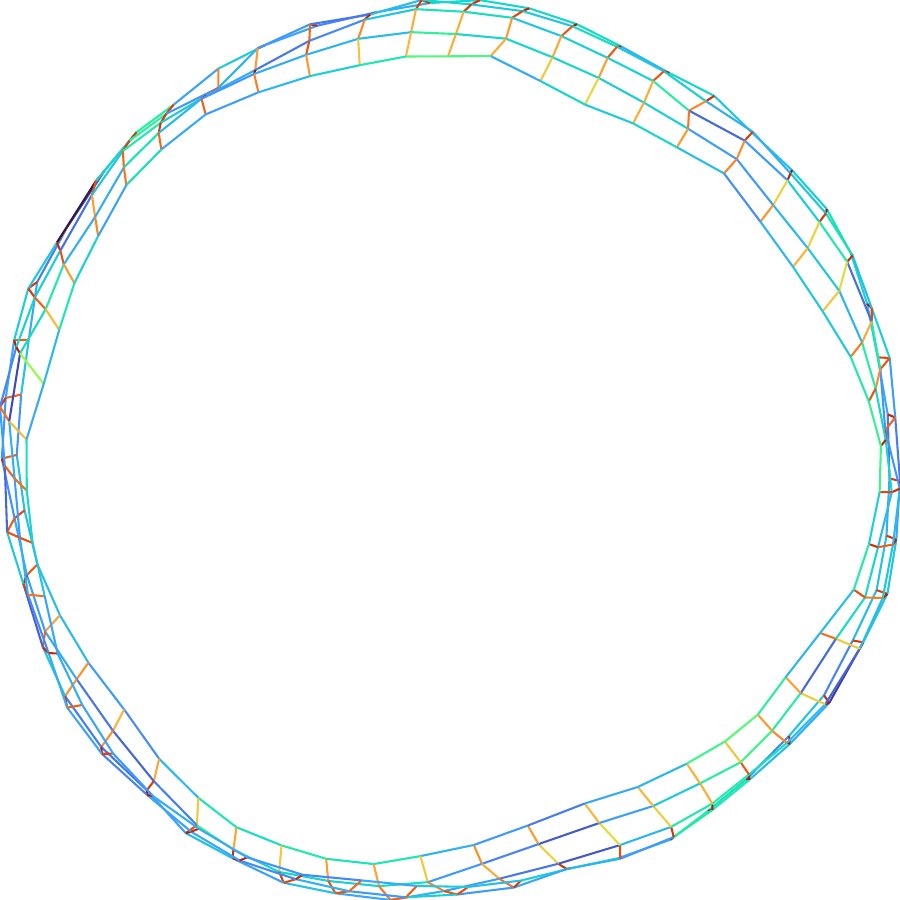} \\ \fontsize{7pt}{0pt}\selectfont \textbf{-3.607},-3.592} & \parbox{1.7cm}{\centering \includegraphics[height=1.1cm]{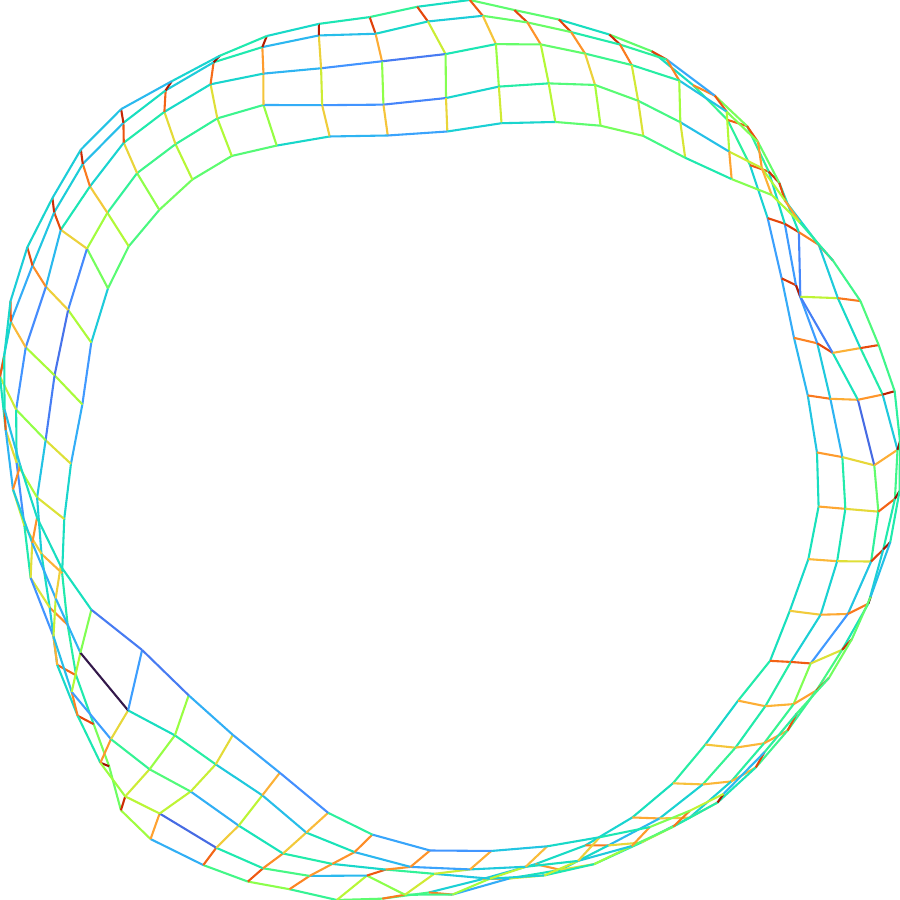} \\ \fontsize{7pt}{0pt}\selectfont \textbf{0.038},0.038} & \parbox{1.7cm}{\centering \includegraphics[height=1.1cm]{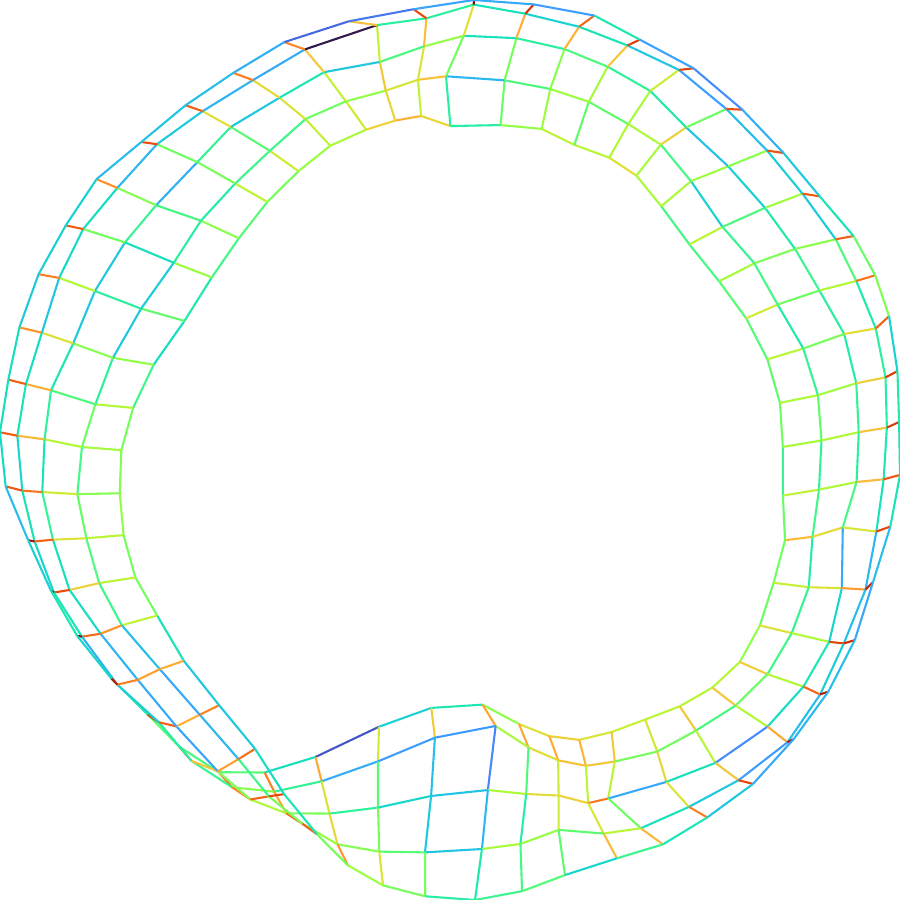} \\ \fontsize{7pt}{0pt}\selectfont 0.812,\textbf{0.73}} & \parbox{1.7cm}{\centering \includegraphics[height=1.1cm]{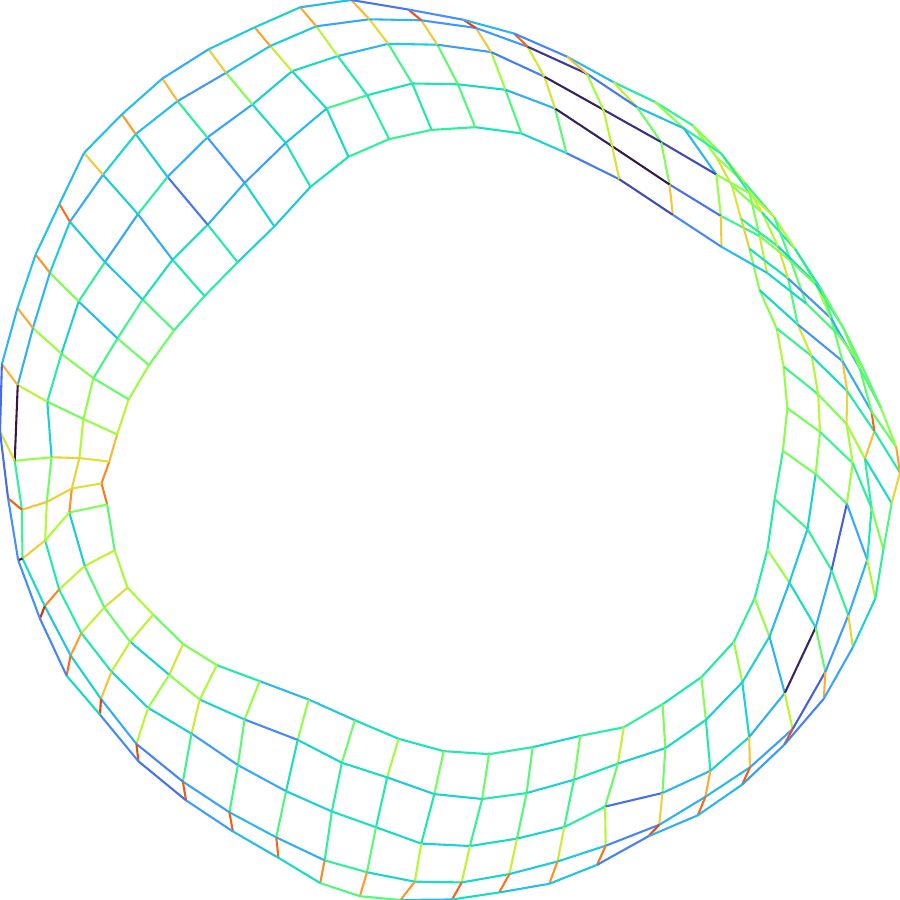} \\ \fontsize{7pt}{0pt}\selectfont 37,\textbf{21}} \\
qh882 & \parbox{1.7cm}{\centering \includegraphics[height=1.1cm]{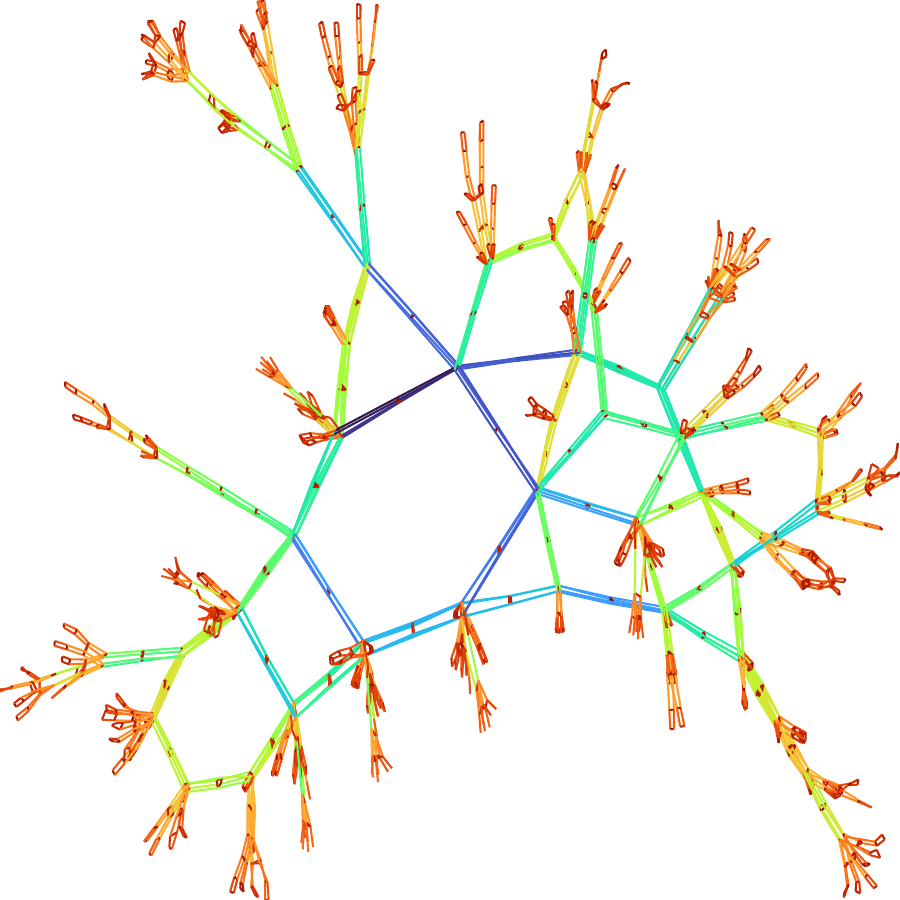} \\ \fontsize{7pt}{0pt}\selectfont } & \parbox{1.7cm}{\centering \includegraphics[height=1.1cm]{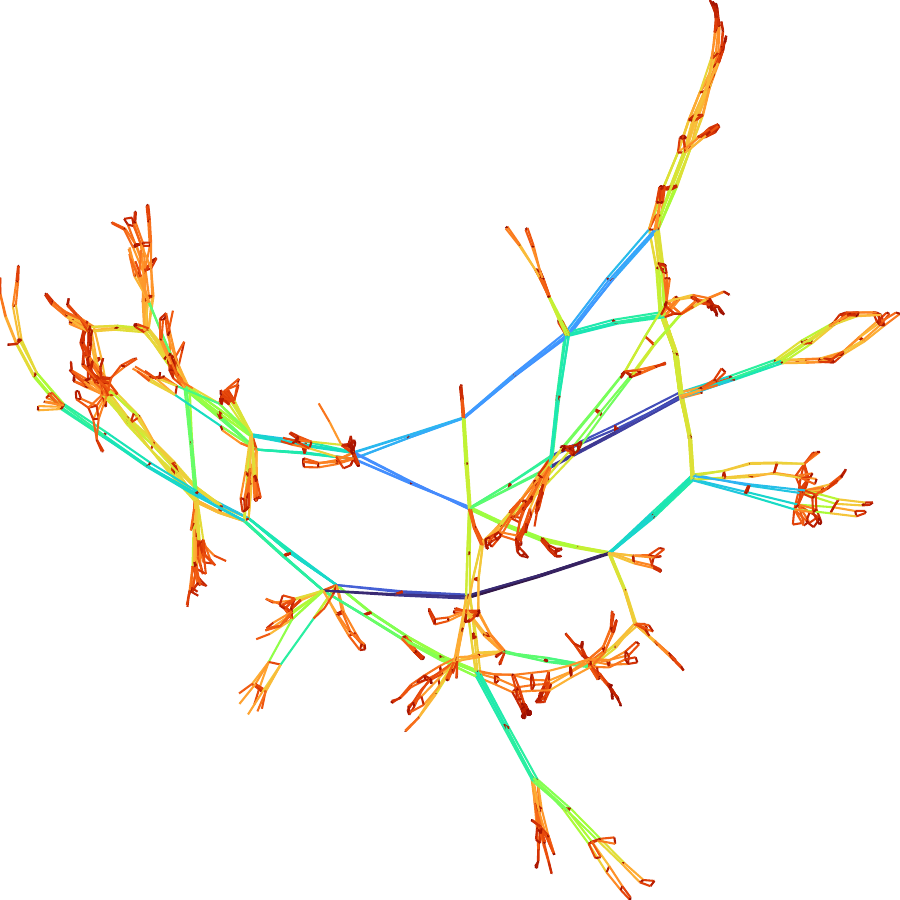} \\ \fontsize{7pt}{0pt}\selectfont \textbf{1.311},1.356} & \parbox{1.7cm}{\centering \includegraphics[height=1.1cm]{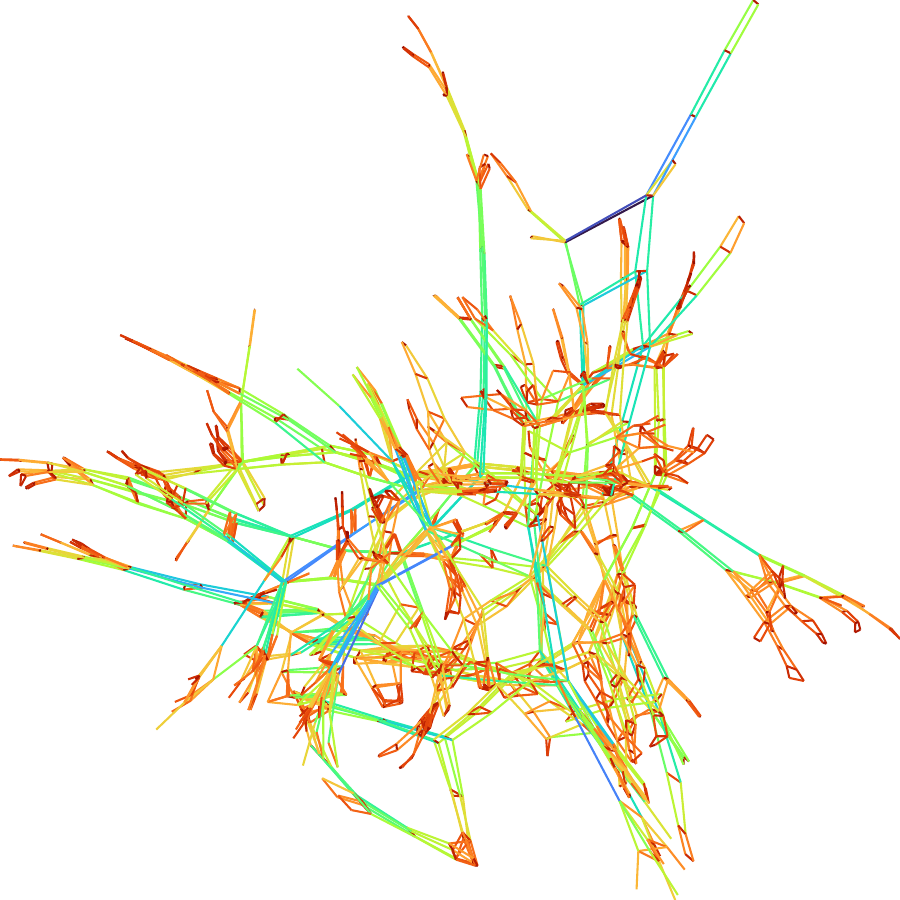} \\ \fontsize{7pt}{0pt}\selectfont 0.874,\textbf{0.752}} & \parbox{1.7cm}{\centering \includegraphics[height=1.1cm]{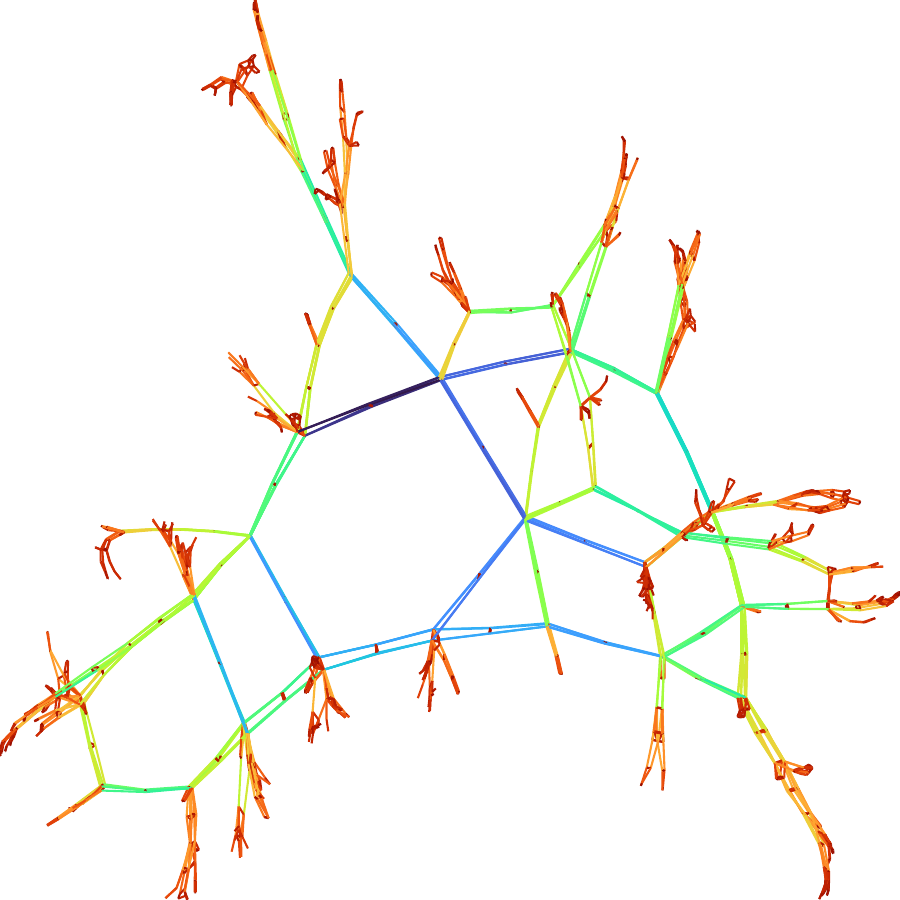} \\ \fontsize{7pt}{0pt}\selectfont \textbf{-4.484},-4.442} & \parbox{1.7cm}{\centering \includegraphics[height=1.1cm]{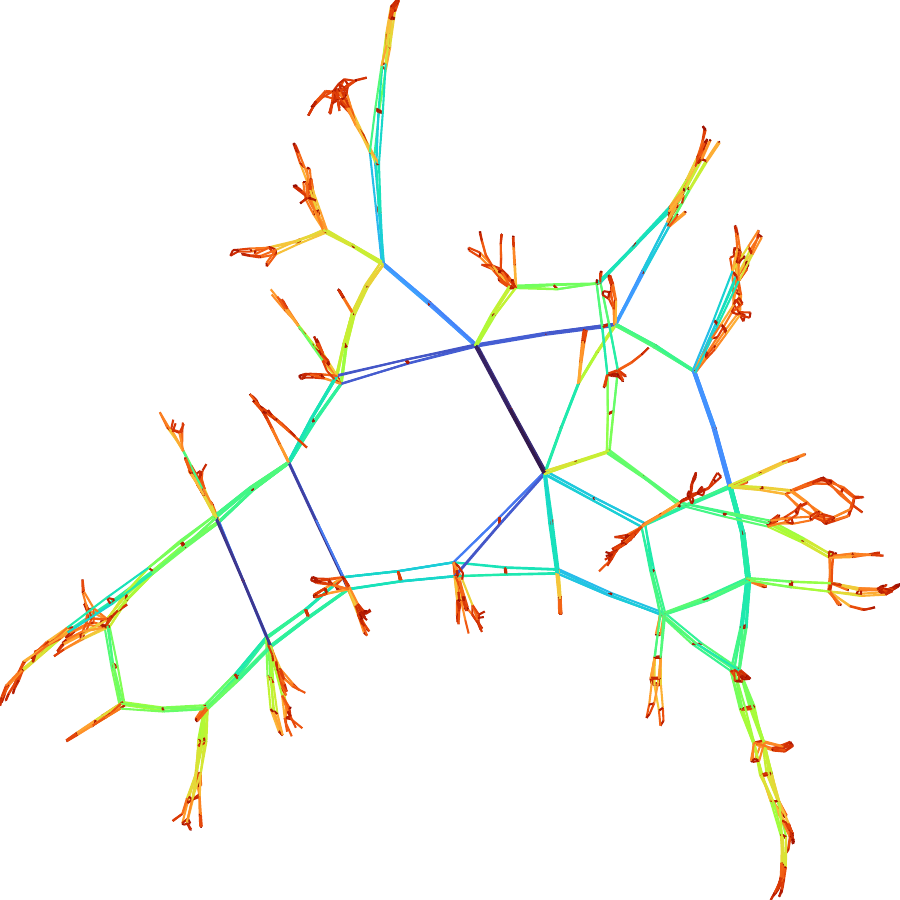} \\ \fontsize{7pt}{0pt}\selectfont \textbf{0.083},0.087} & \parbox{1.7cm}{\centering \includegraphics[height=1.1cm]{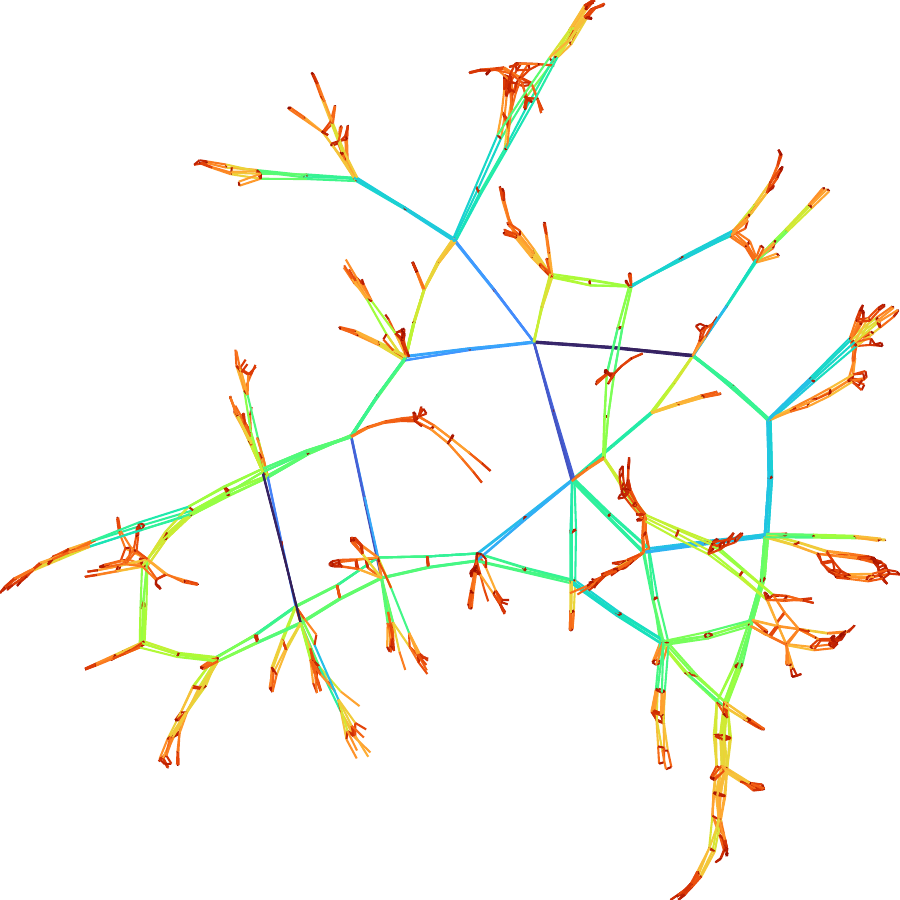} \\ \fontsize{7pt}{0pt}\selectfont \textbf{1.399},1.515} & \parbox{1.7cm}{\centering \includegraphics[height=1.1cm]{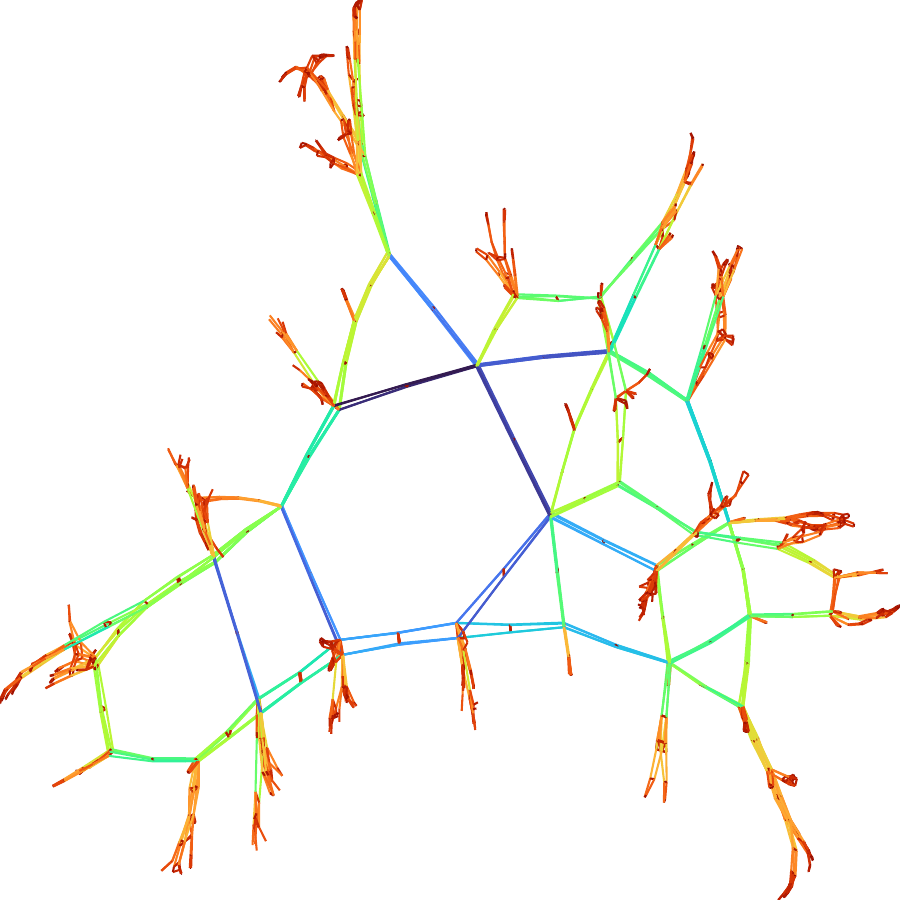} \\ \fontsize{7pt}{0pt}\selectfont \textbf{4337},5566} \\
Si2 & \parbox{1.7cm}{\centering \includegraphics[height=1.1cm]{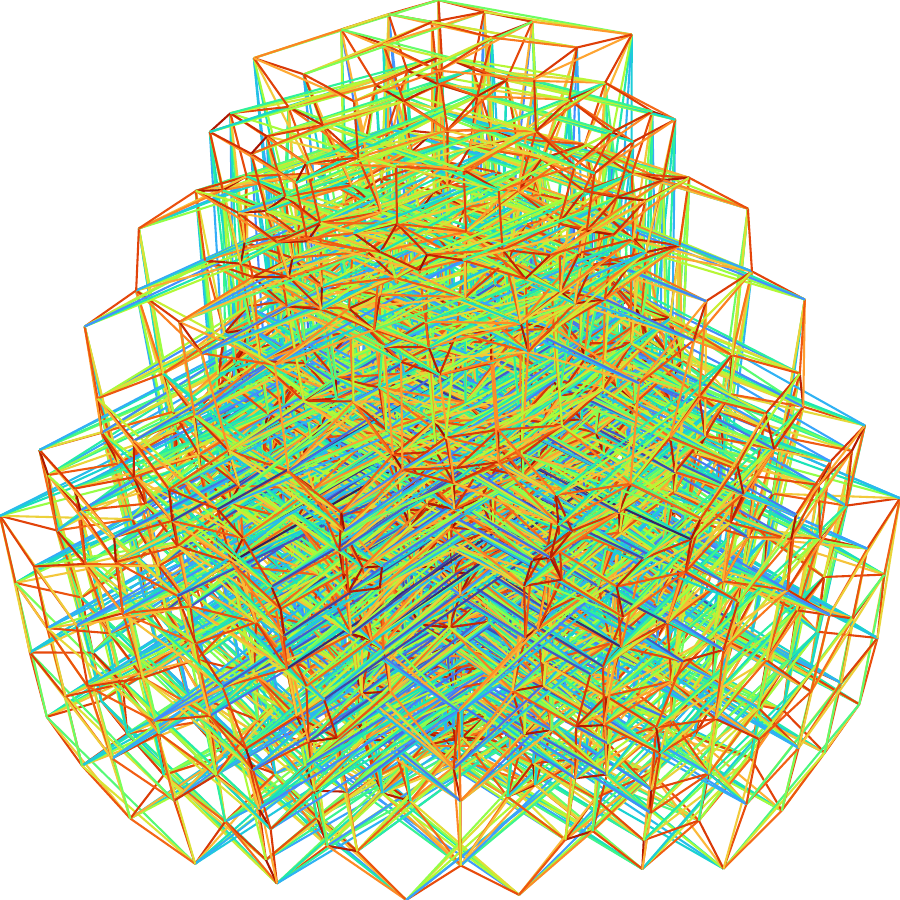} \\ \fontsize{7pt}{0pt}\selectfont } & \parbox{1.7cm}{\centering \includegraphics[height=1.1cm]{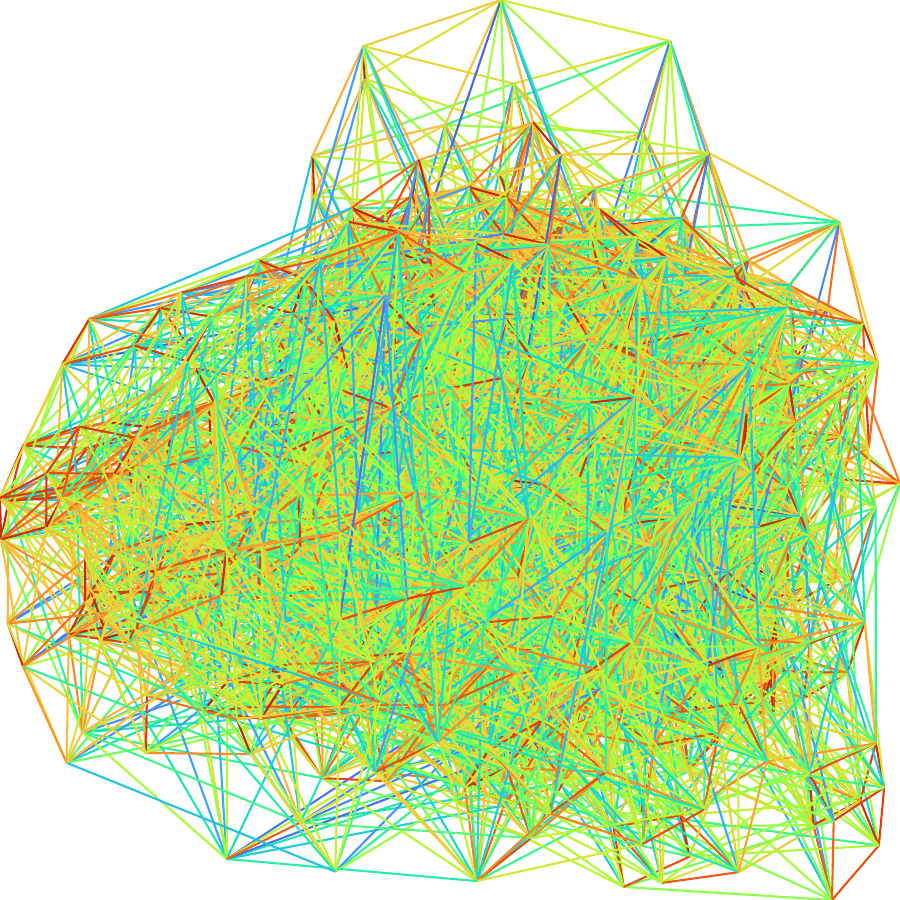} \\ \fontsize{7pt}{0pt}\selectfont 0.326,\textbf{0.307}} & \parbox{1.7cm}{\centering \includegraphics[height=1.1cm]{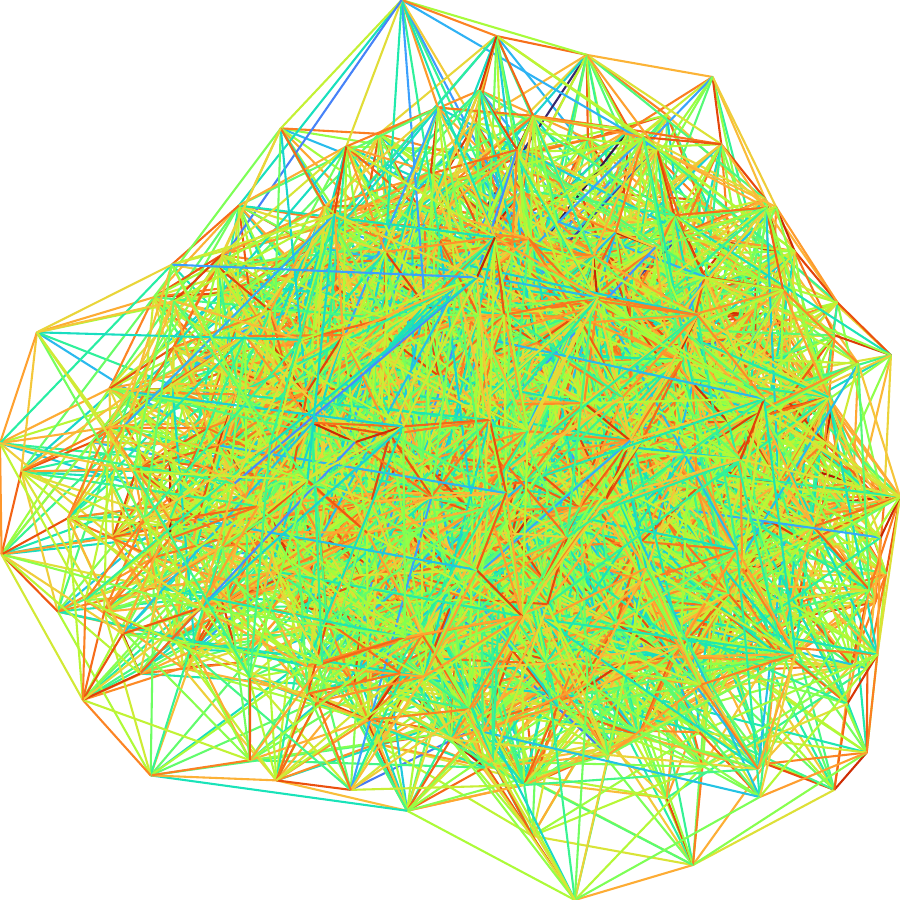} \\ \fontsize{7pt}{0pt}\selectfont 0.569,\textbf{0.393}} & \parbox{1.7cm}{\centering \includegraphics[height=1.1cm]{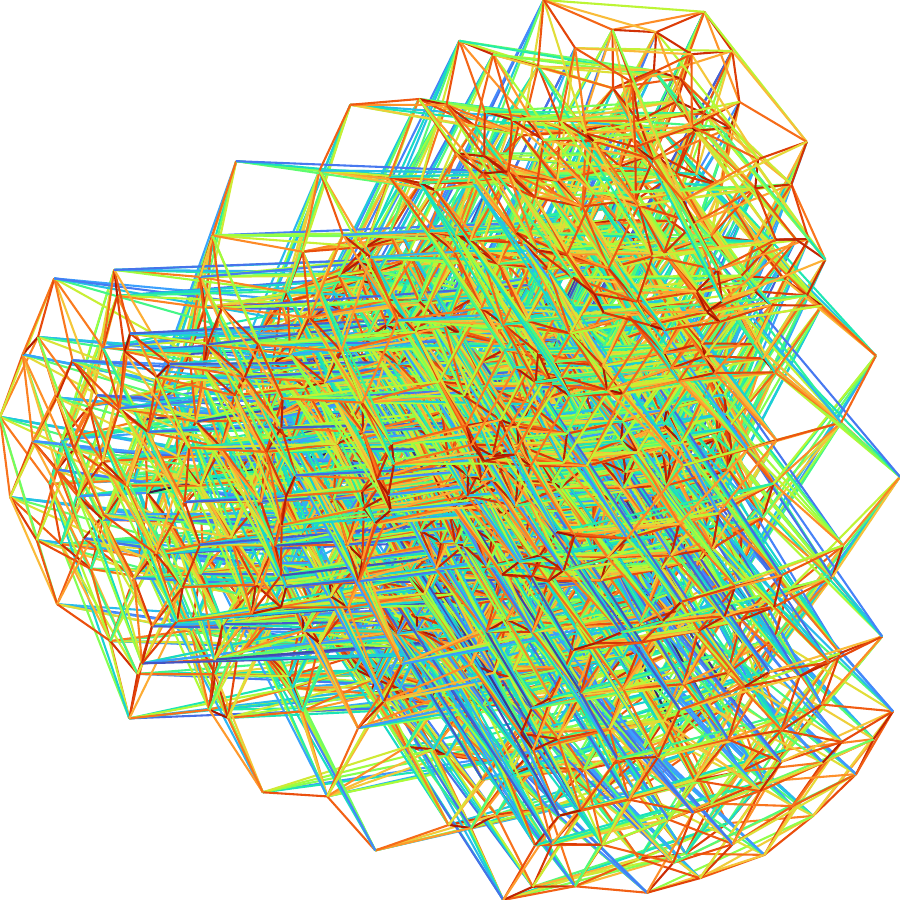} \\ \fontsize{7pt}{0pt}\selectfont \textbf{-1.568},-1.561} & \parbox{1.7cm}{\centering \includegraphics[height=1.1cm]{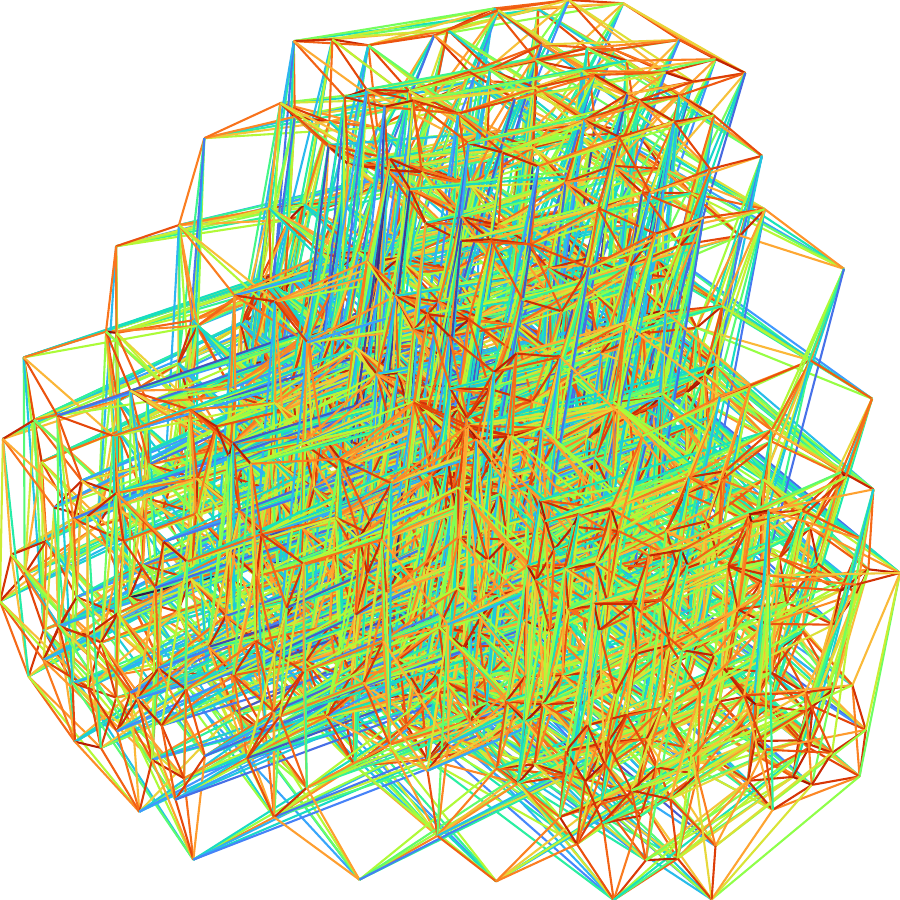} \\ \fontsize{7pt}{0pt}\selectfont \textbf{0.141},0.146} & \parbox{1.7cm}{\centering \includegraphics[height=1.1cm]{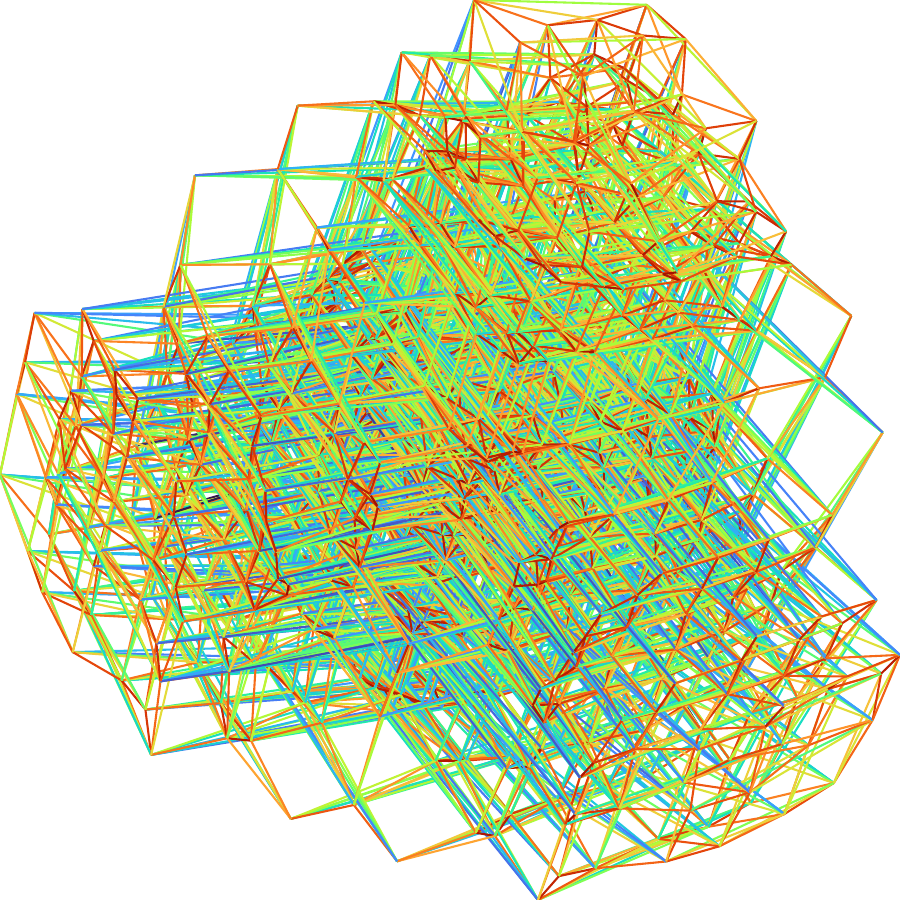} \\ \fontsize{7pt}{0pt}\selectfont \textbf{0.825},0.83} & \parbox{1.7cm}{\centering \includegraphics[height=1.1cm]{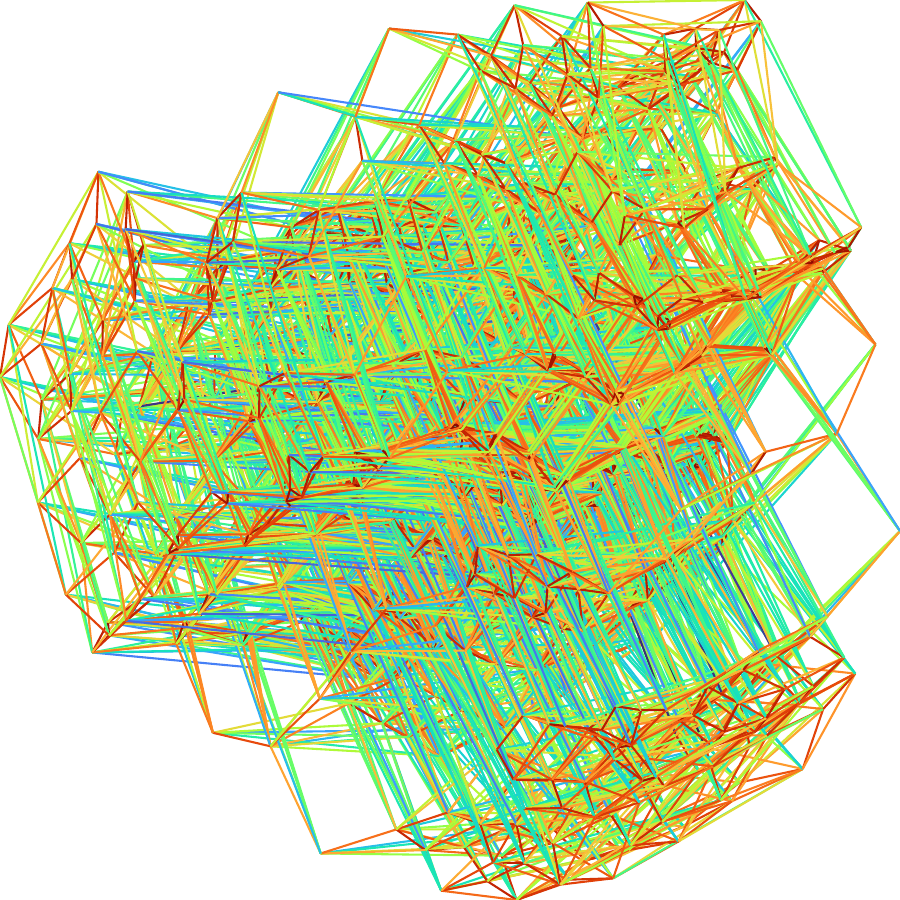} \\ \fontsize{7pt}{0pt}\selectfont 1242149,\textbf{1180389}} \\
spiral & \parbox{1.7cm}{\centering \includegraphics[height=1.1cm]{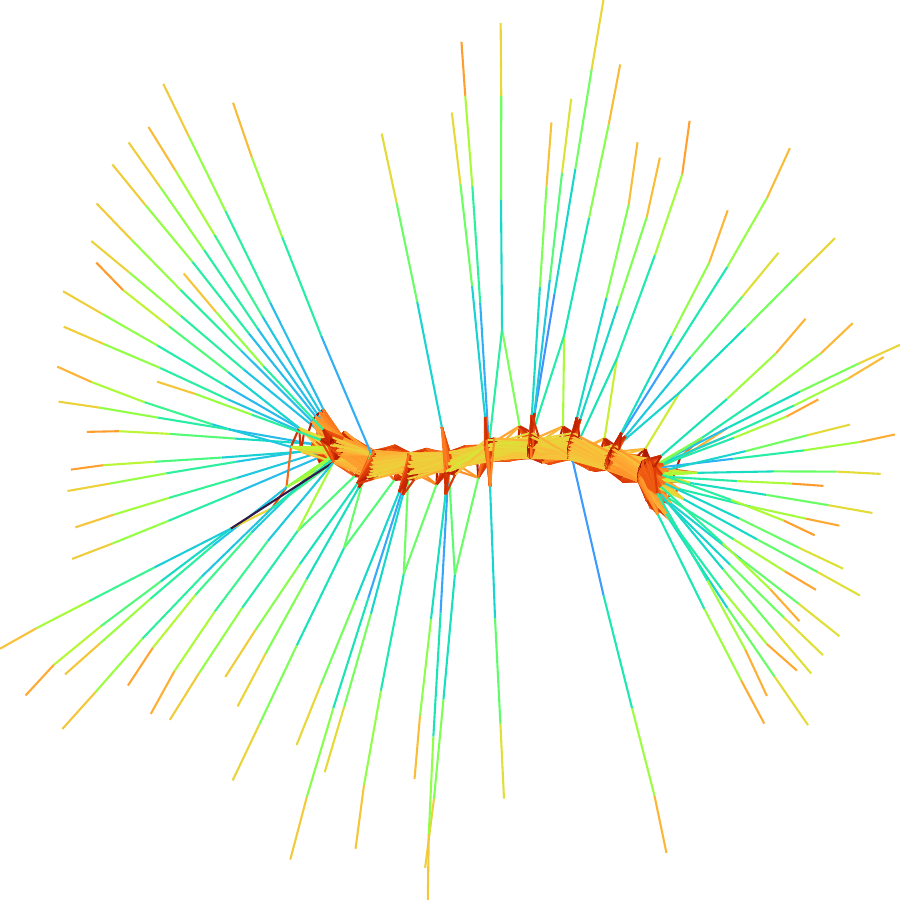} \\ \fontsize{7pt}{0pt}\selectfont } & \parbox{1.7cm}{\centering \includegraphics[height=1.1cm]{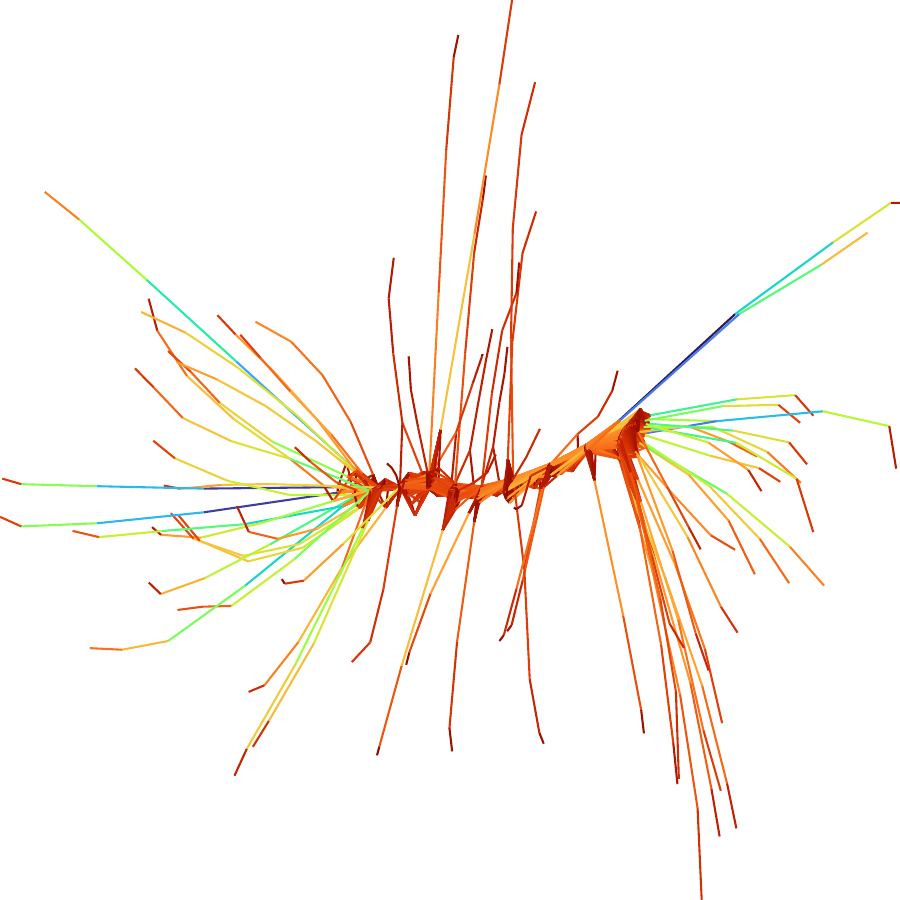} \\ \fontsize{7pt}{0pt}\selectfont 0.349,\textbf{0.34}} & \parbox{1.7cm}{\centering \includegraphics[height=1.1cm]{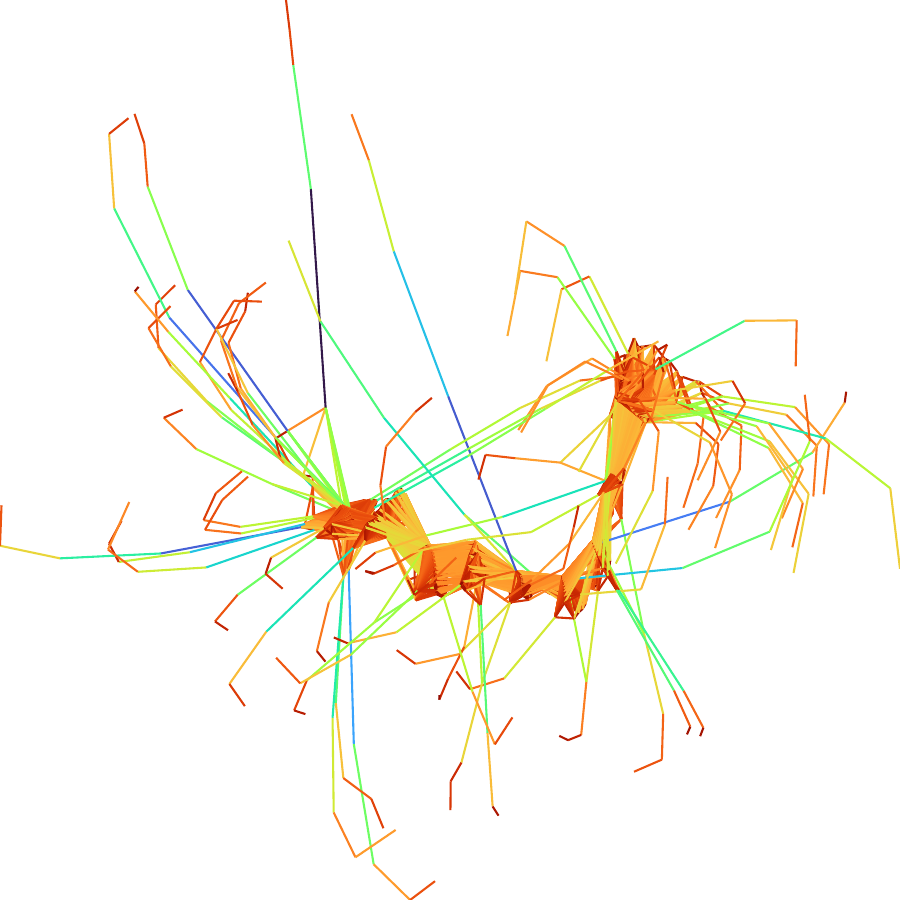} \\ \fontsize{7pt}{0pt}\selectfont 0.685,\textbf{0.618}} & \parbox{1.7cm}{\centering \includegraphics[height=1.1cm]{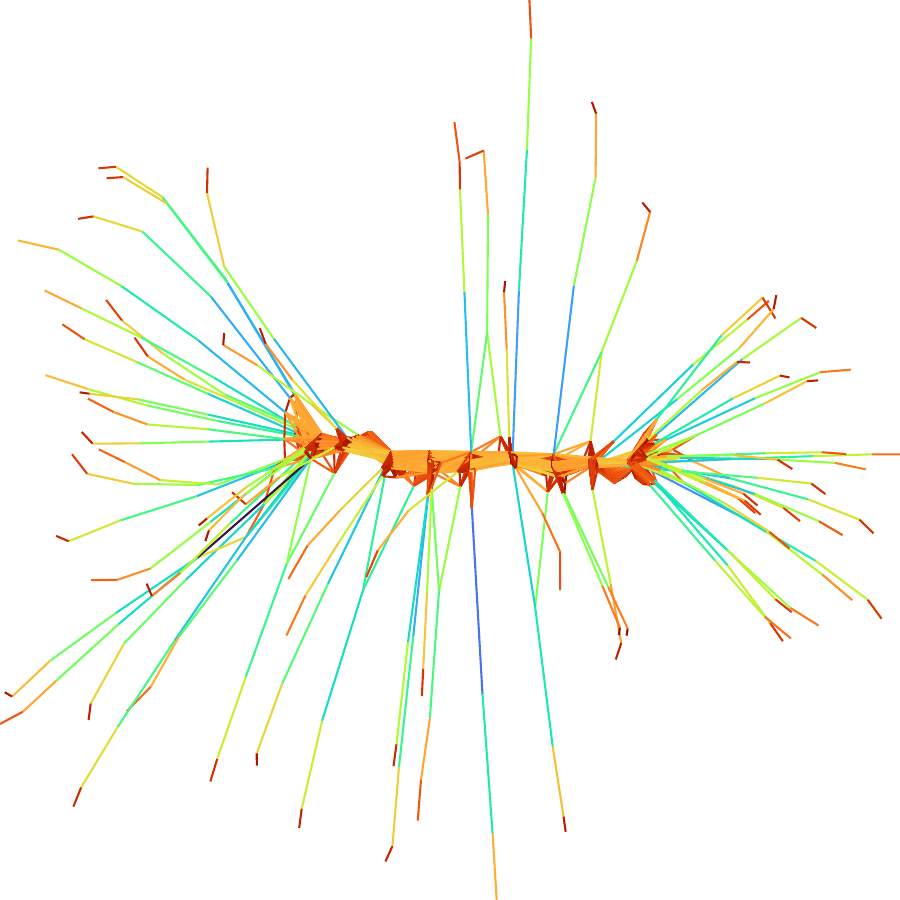} \\ \fontsize{7pt}{0pt}\selectfont \textbf{-2.857},-2.815} & \parbox{1.7cm}{\centering \includegraphics[height=1.1cm]{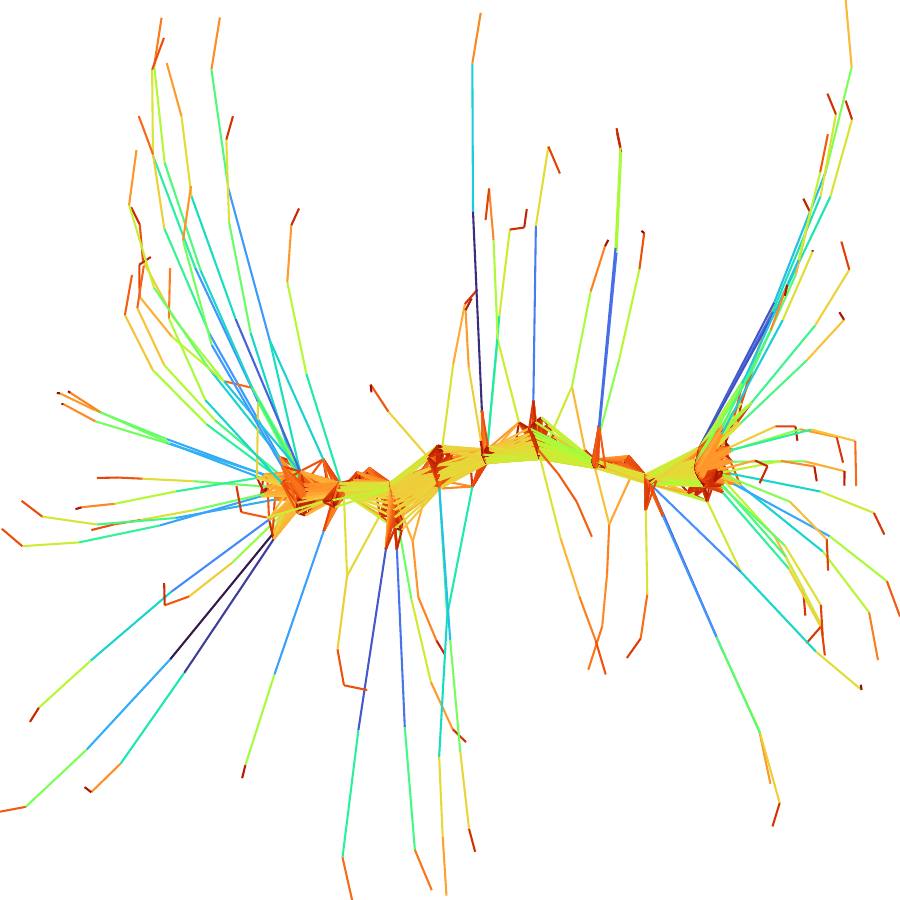} \\ \fontsize{7pt}{0pt}\selectfont \textbf{0.076},0.079} & \parbox{1.7cm}{\centering \includegraphics[height=1.1cm]{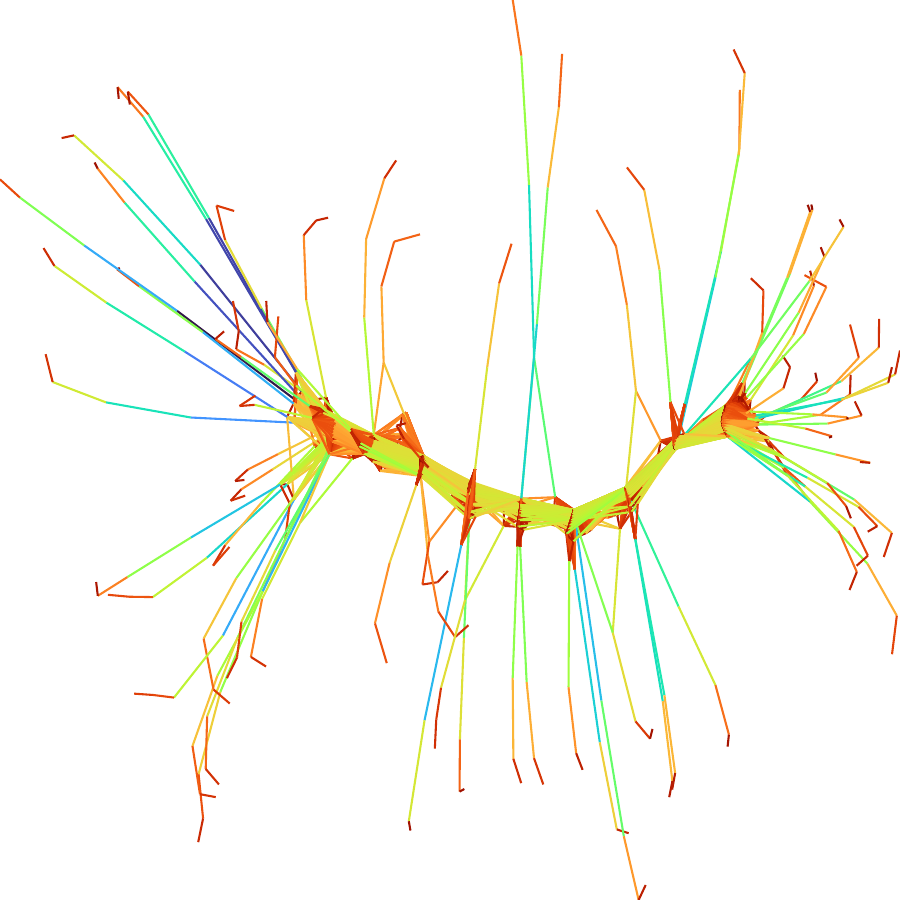} \\ \fontsize{7pt}{0pt}\selectfont 1.424,\textbf{1.193}} & \parbox{1.7cm}{\centering \includegraphics[height=1.1cm]{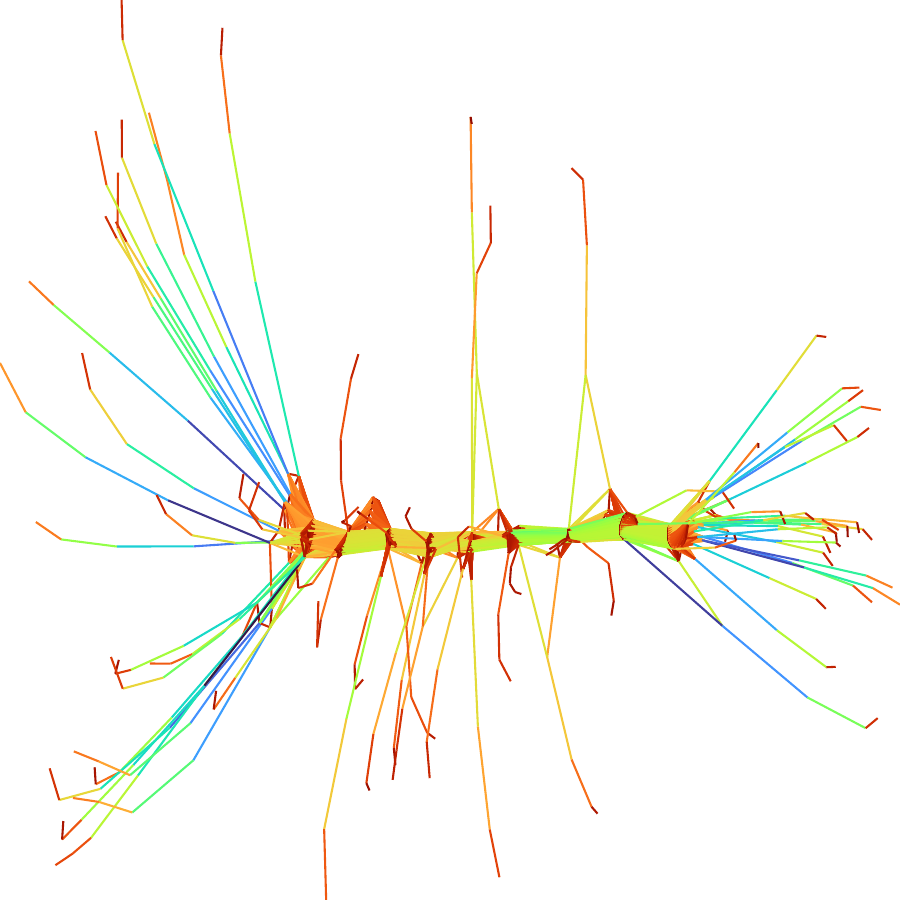} \\ \fontsize{7pt}{0pt}\selectfont 725528,\textbf{615614}} \\
Trefethen\_700 & \parbox{1.7cm}{\centering \includegraphics[height=1.1cm]{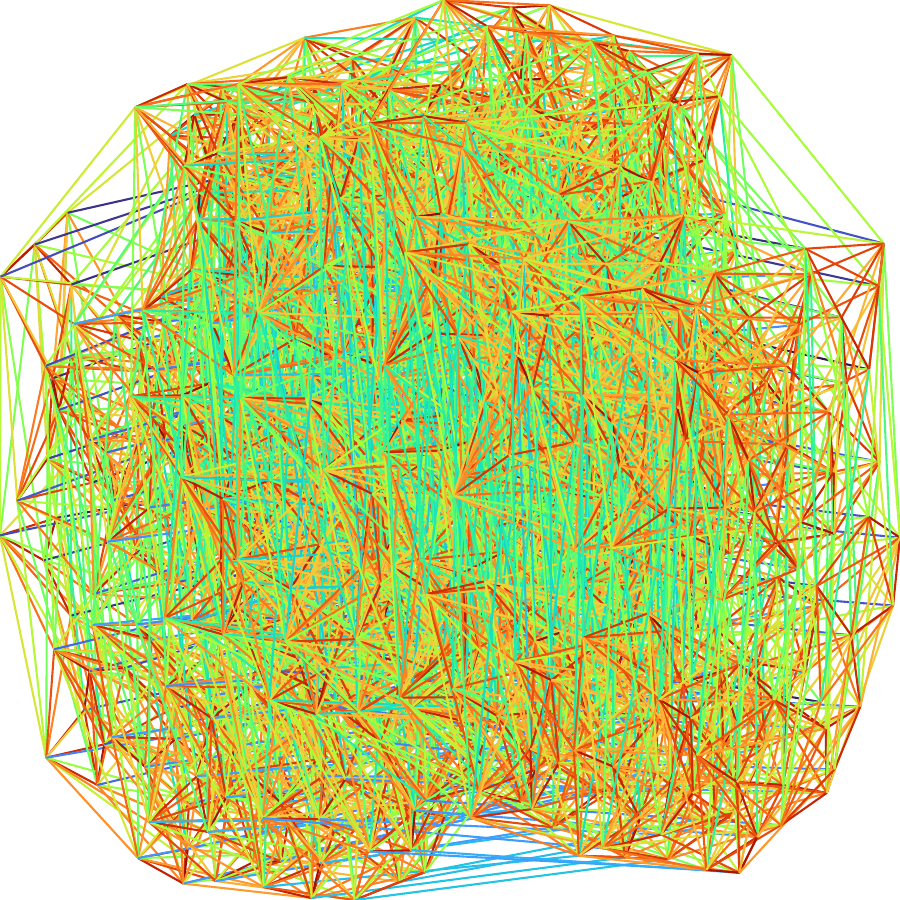} \\ \fontsize{7pt}{0pt}\selectfont } & \parbox{1.7cm}{\centering \includegraphics[height=1.1cm]{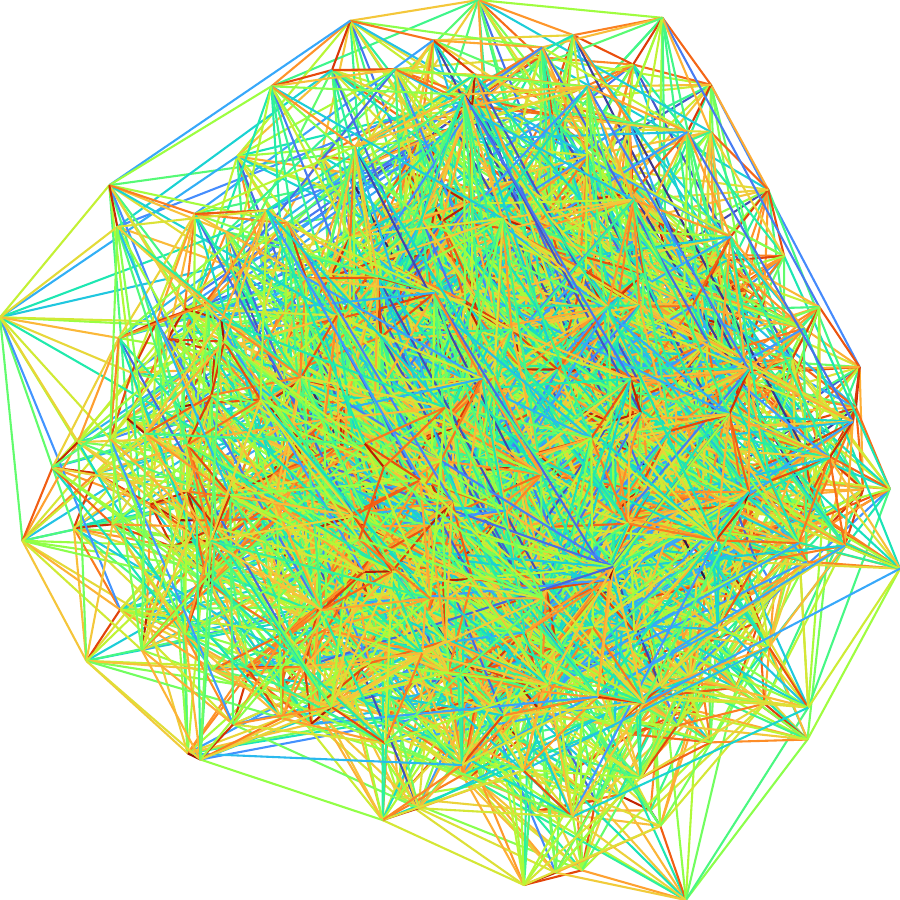} \\ \fontsize{7pt}{0pt}\selectfont 0.352,\textbf{0.348}} & \parbox{1.7cm}{\centering \includegraphics[height=1.1cm]{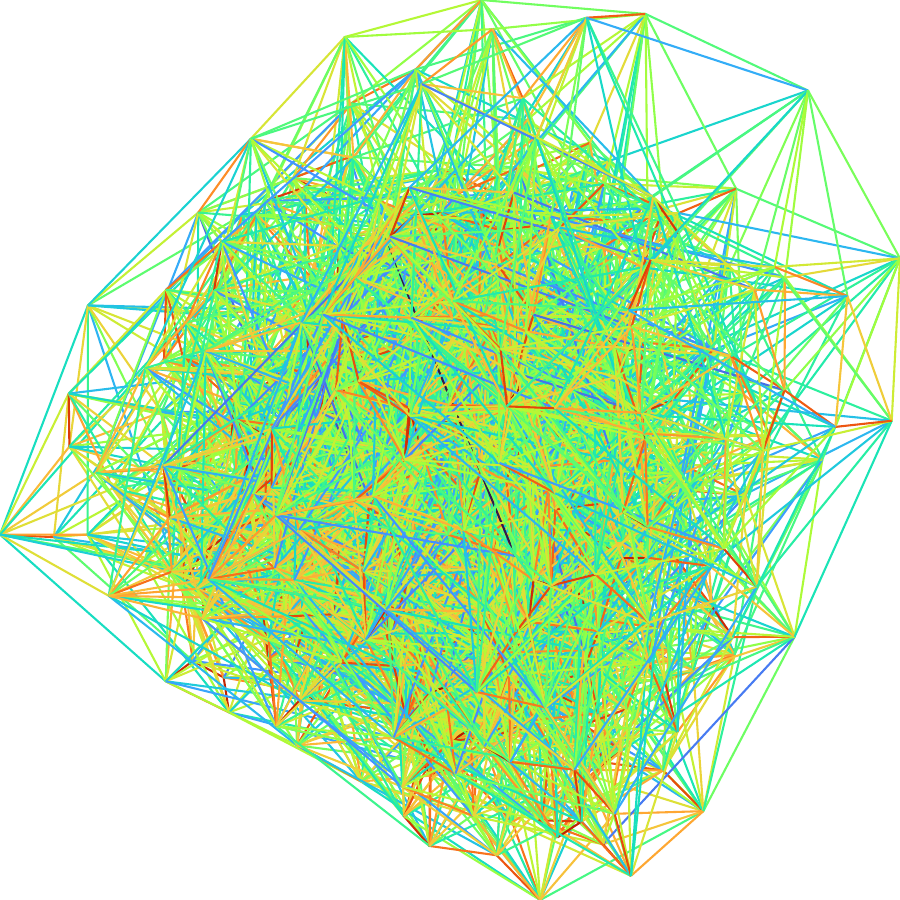} \\ \fontsize{7pt}{0pt}\selectfont 0.519,\textbf{0.348}} & \parbox{1.7cm}{\centering \includegraphics[height=1.1cm]{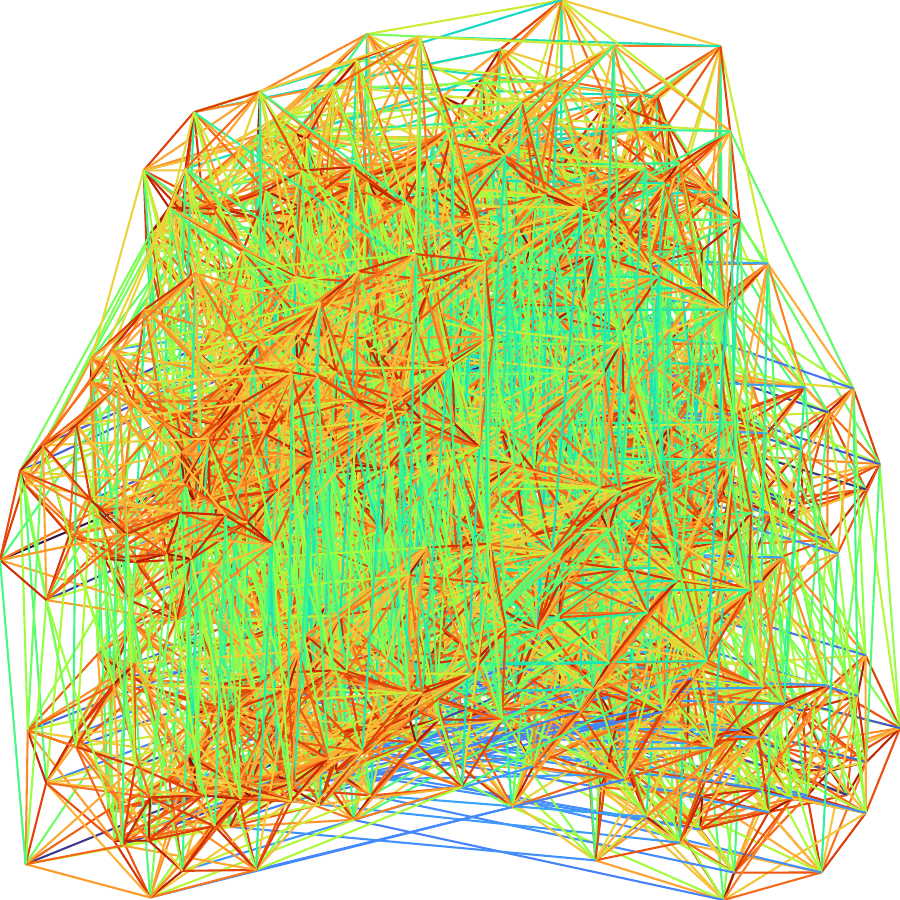} \\ \fontsize{7pt}{0pt}\selectfont \textbf{-1.512},-1.503} & \parbox{1.7cm}{\centering \includegraphics[height=1.1cm]{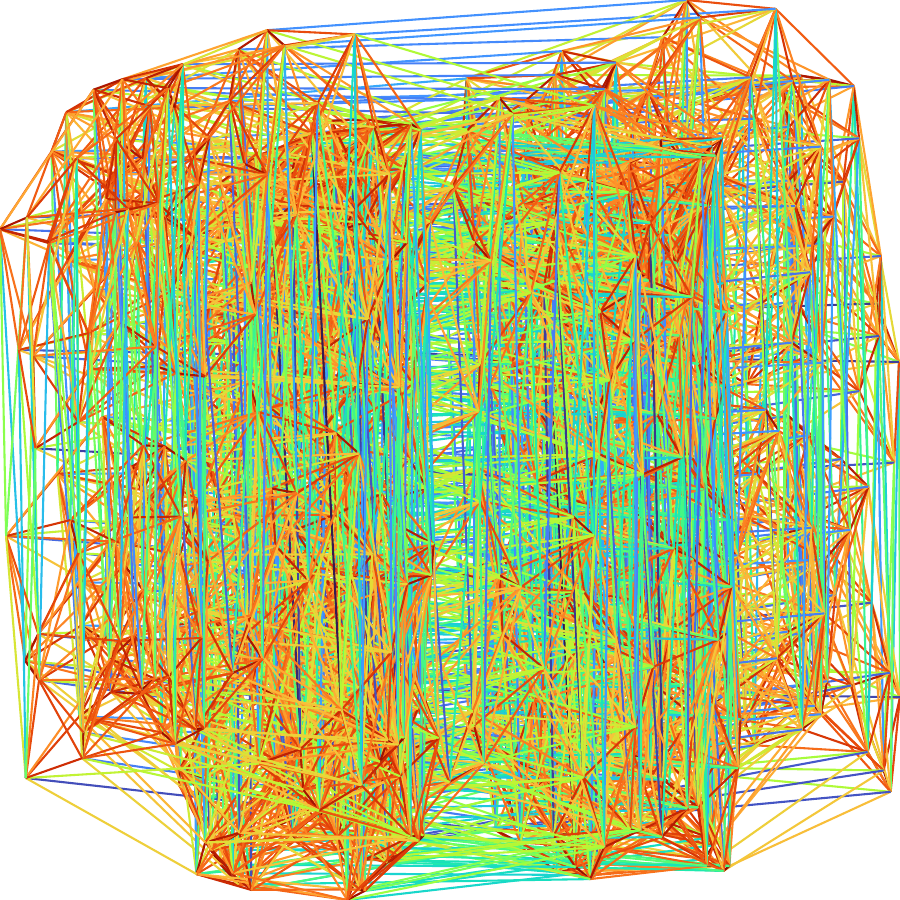} \\ \fontsize{7pt}{0pt}\selectfont 0.18,\textbf{0.179}} & \parbox{1.7cm}{\centering \includegraphics[height=1.1cm]{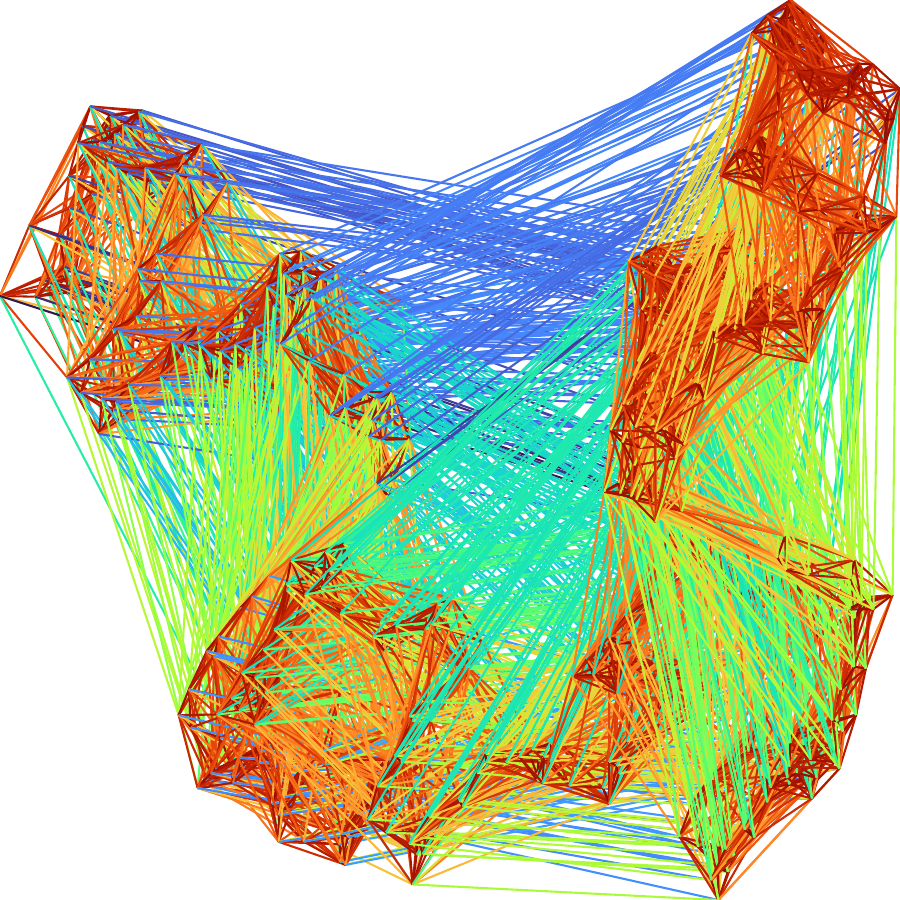} \\ \fontsize{7pt}{0pt}\selectfont 1.239,\textbf{1.132}} & \parbox{1.7cm}{\centering \includegraphics[height=1.1cm]{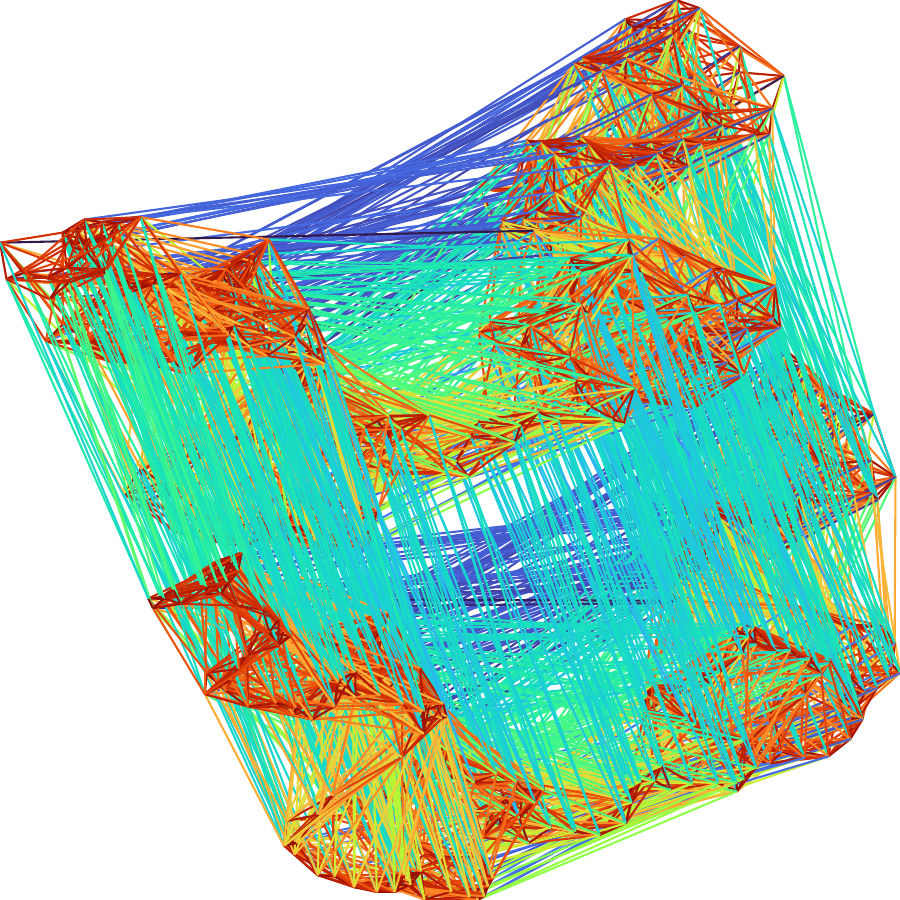} \\ \fontsize{7pt}{0pt}\selectfont 696237,\textbf{504843}} \\
USpowerGrid & \parbox{1.7cm}{\centering \includegraphics[height=1.1cm]{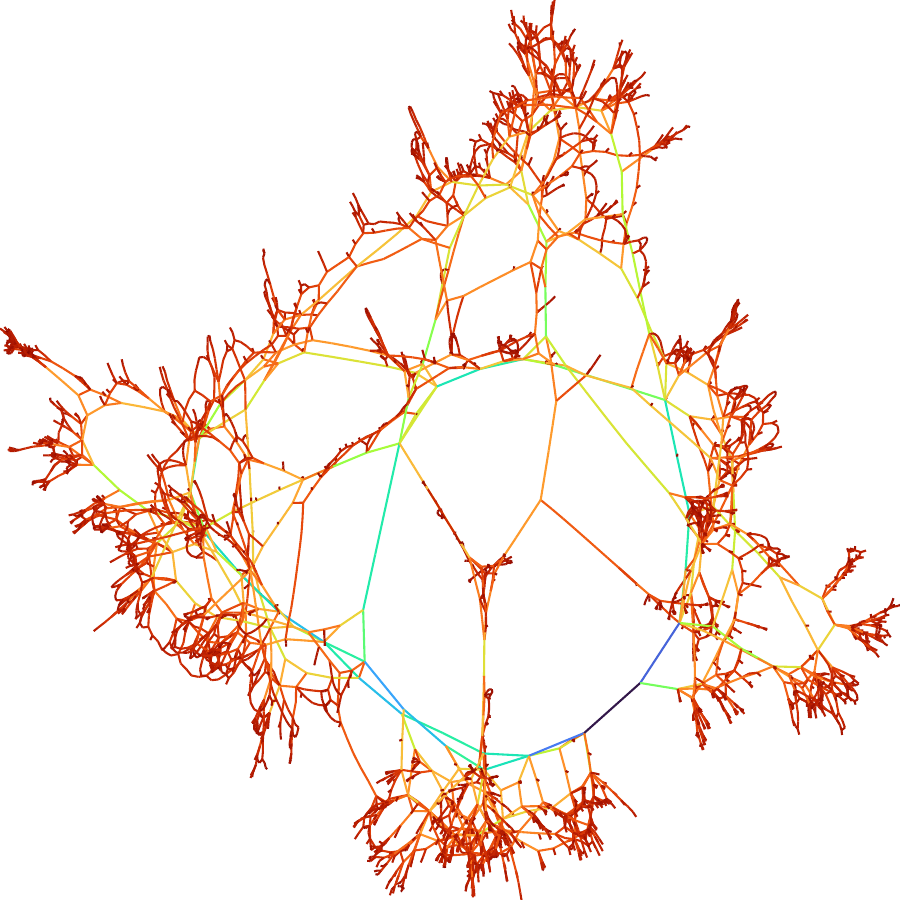} \\ \fontsize{7pt}{0pt}\selectfont } & \parbox{1.7cm}{\centering \includegraphics[height=1.1cm]{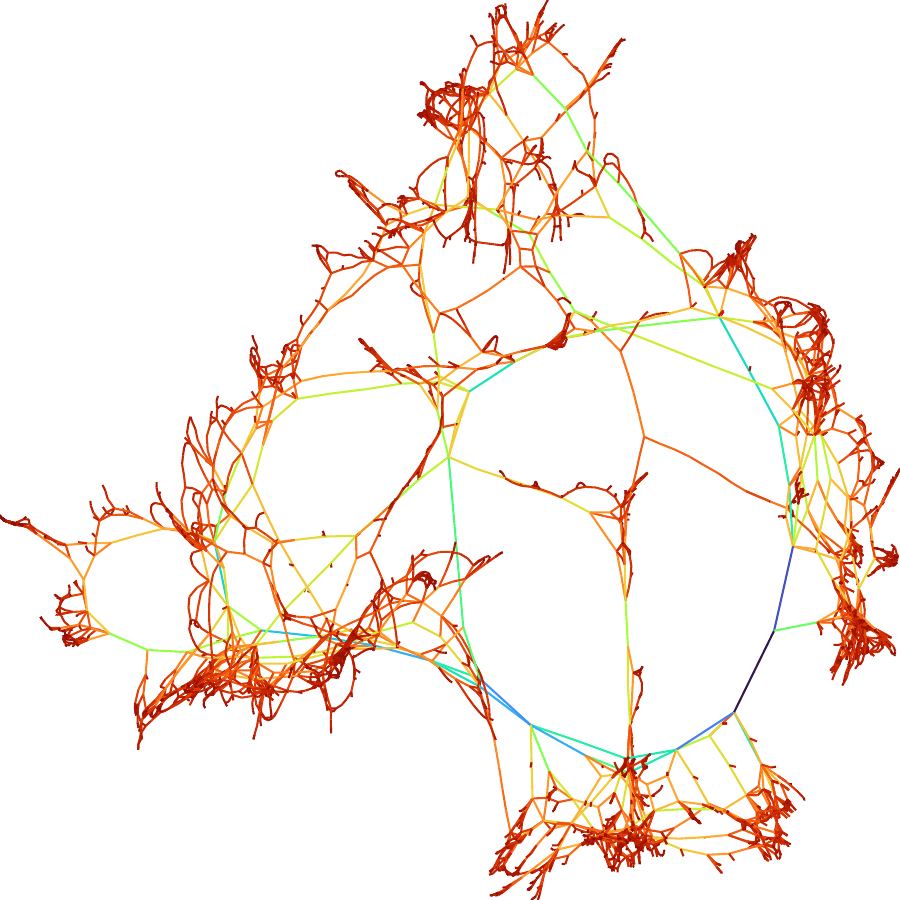} \\ \fontsize{7pt}{0pt}\selectfont \textbf{1.066},1.129} & \parbox{1.7cm}{\centering \includegraphics[height=1.1cm]{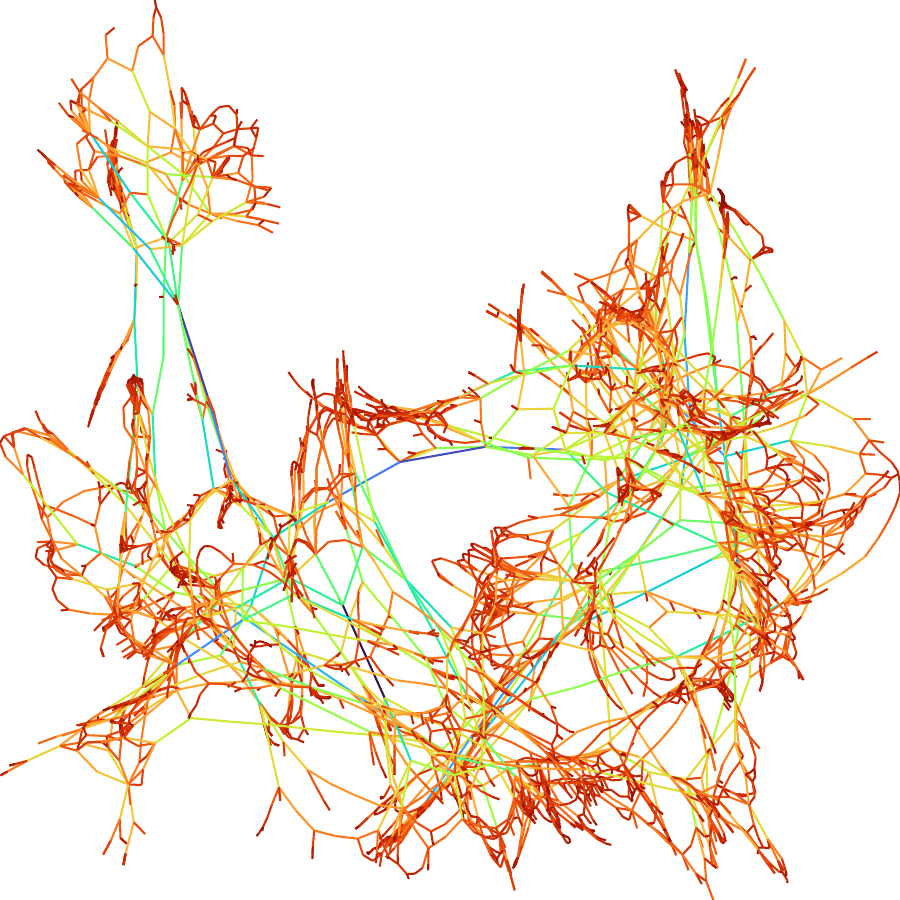} \\ \fontsize{7pt}{0pt}\selectfont 0.795,\textbf{0.773}} & \parbox{1.7cm}{\centering \includegraphics[height=1.1cm]{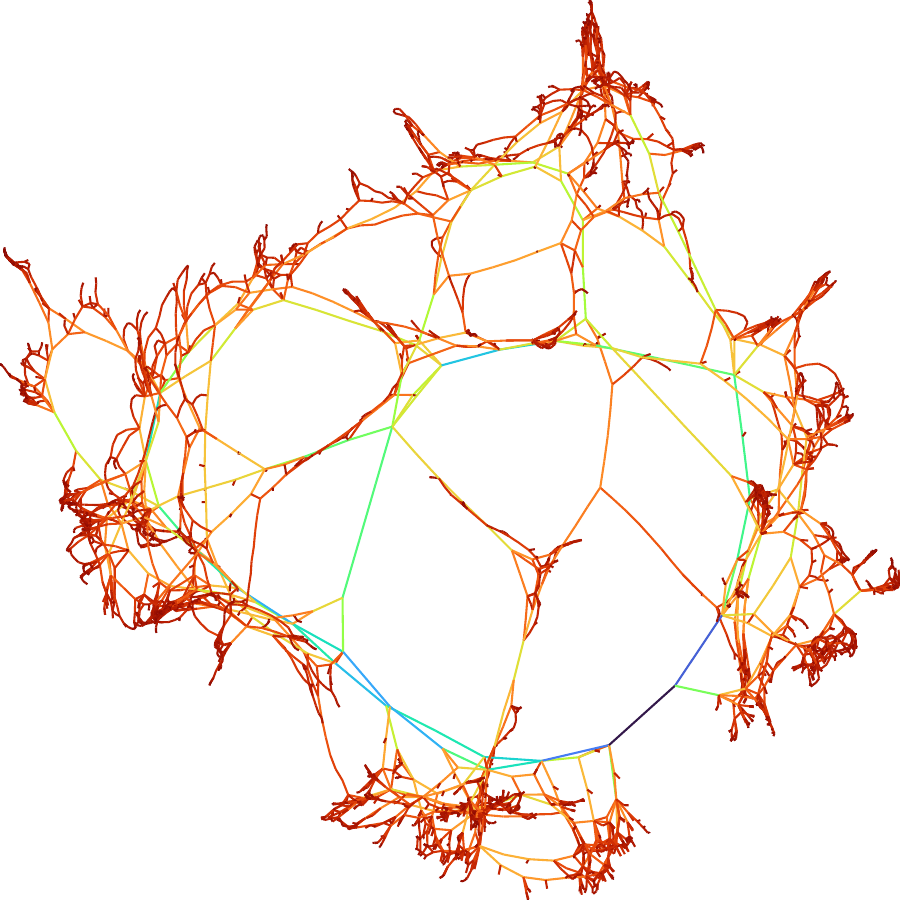} \\ \fontsize{7pt}{0pt}\selectfont \textbf{-5.633},-5.579} & \parbox{1.7cm}{\centering \includegraphics[height=1.1cm]{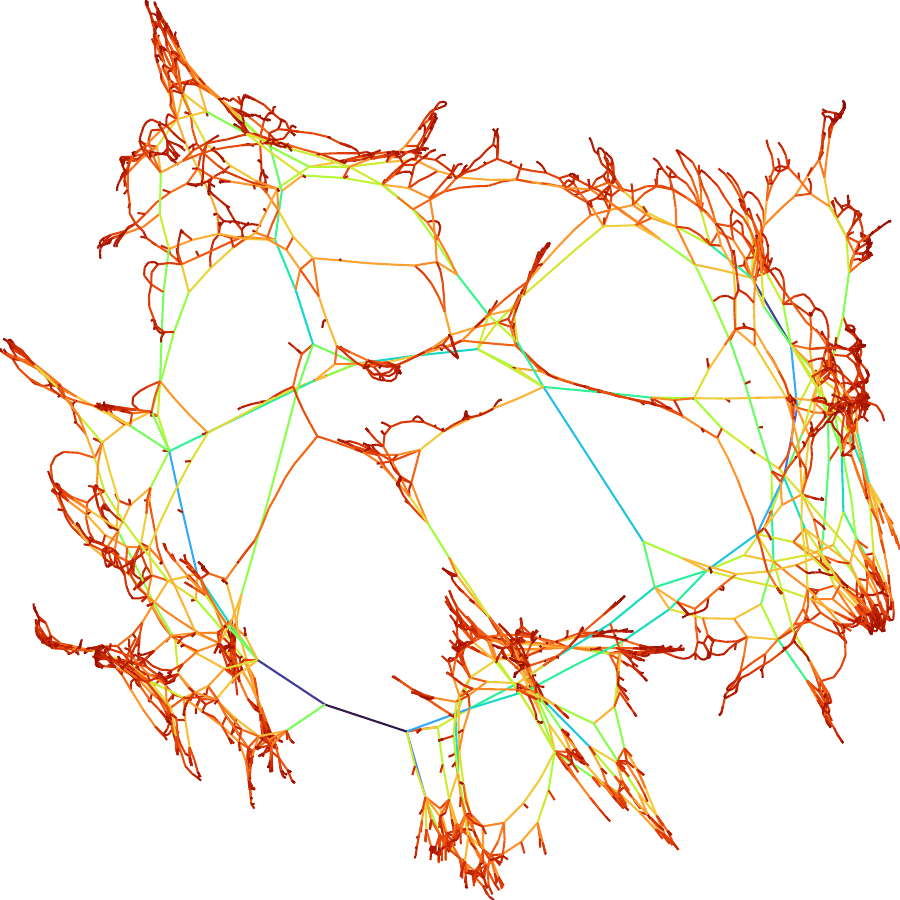} \\ \fontsize{7pt}{0pt}\selectfont \textbf{0.098},0.101} & \parbox{1.7cm}{\centering \includegraphics[height=1.1cm]{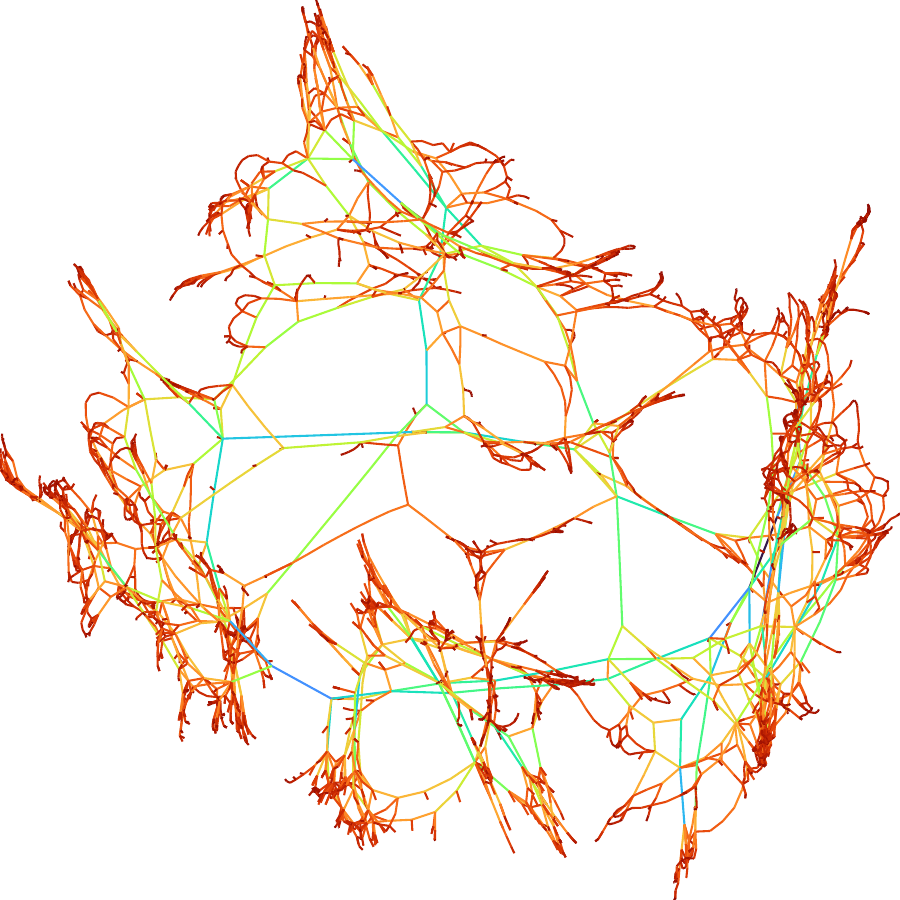} \\ \fontsize{7pt}{0pt}\selectfont \textbf{1.712},2.044} & \parbox{1.7cm}{\centering \includegraphics[height=1.1cm]{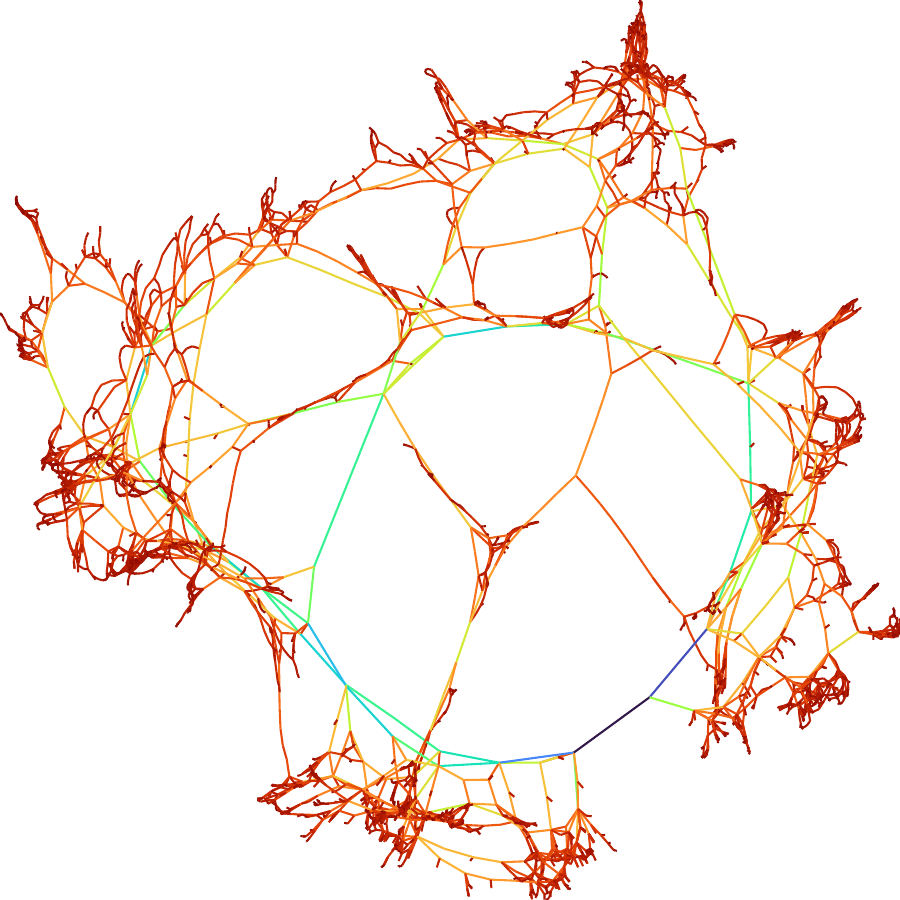} \\ \fontsize{7pt}{0pt}\selectfont \textbf{3614},5207} \\
\bottomrule
\end{tabular}

\end{table*}

\begin{table*}[htbp]
\setlength{\tabcolsep}{4pt}
\centering
\renewcommand{\arraystretch}{0.8}
\caption{Spring-electrical loss for SFDP layouts at different dimensions, vs that for the Optimal projection to 2D, and for the best PCA projection to 2D.}
\label{tab:sfdp_vs_dims}

\begin{tabular}{l|lllll|llll|llll}
\toprule
 & \multicolumn{5}{c|}{sfdp at \highD} & \multicolumn{4}{c|}{OptProj (from \highD-D to 2D)} & \multicolumn{4}{c}{PCA  (from \highD-D to 2D)} \\
 & 2 & 3 & 5 & 7 & 10 & 3 & 5 & 7 & 10 & 3 & 5 & 7 & 10 \\
\midrule
1138\_bus & -4.369 & \textbf{-4.425} & -4.396 & -4.362 & -4.337 & -4.417 & -4.352 & -4.374 & -4.322 & -4.415 & -4.336 & -4.339 & -4.264 \\
bcspwr07 & \textbf{-4.998} & -4.977 & -4.970 & -4.950 & -4.852 & -4.989 & -4.992 & -4.984 & -4.937 & -4.984 & -4.984 & -4.976 & -4.918 \\
email & -2.306 & -2.368 & -2.392 & -2.402 & \textbf{-2.406} & -2.274 & -2.258 & -2.262 & -2.269 & -2.273 & -2.237 & -2.227 & -2.234 \\
Erdos991 & -2.313 & -2.377 & -2.369 & \textbf{-2.392} & -2.389 & -2.291 & -2.260 & -2.283 & -2.280 & -2.287 & -2.227 & -2.258 & -2.235 \\
EX1 & -1.493 & -1.606 & -1.678 & -1.700 & \textbf{-1.701} & -1.487 & -1.486 & -1.486 & -1.486 & -1.476 & -1.459 & -1.469 & -1.458 \\
football & -1.196 & -1.260 & -1.292 & \textbf{-1.294} & -1.285 & -1.189 & -1.188 & -1.190 & -1.181 & -1.187 & -1.182 & -1.177 & -1.172 \\
hypercube10 & -1.980 & -2.084 & -2.160 & -2.192 & \textbf{-2.203} & -1.966 & -1.975 & -1.973 & -1.970 & -1.961 & -1.936 & -1.909 & -1.951 \\
Journals & 0.210 & 0.105 & 0.041 & 0.013 & \textbf{-0.007} & 0.225 & 0.246 & 0.238 & 0.237 & 0.228 & 0.260 & 0.272 & 0.282 \\
mobius & \textbf{-3.607} & -3.606 & -3.591 & -3.590 & -3.523 & -3.605 & -3.603 & -3.605 & -3.592 & -3.596 & -3.594 & -3.594 & -3.588 \\
qh882 & -4.484 & \textbf{-4.493} & -4.473 & -4.457 & -4.446 & -4.474 & -4.459 & -4.447 & -4.442 & -4.472 & -4.454 & -4.436 & -4.429 \\
Si2 & -1.568 & -1.628 & -1.666 & -1.679 & \textbf{-1.686} & -1.566 & -1.562 & -1.562 & -1.561 & -1.566 & -1.562 & -1.561 & -1.560 \\
spiral & -2.857 & \textbf{-2.865} & -2.828 & -2.805 & -2.819 & -2.839 & -2.799 & -2.791 & -2.815 & -2.833 & -2.786 & -2.753 & -2.761 \\
Trefethen\_700 & -1.512 & -1.608 & -1.667 & -1.692 & \textbf{-1.703} & -1.502 & -1.503 & -1.504 & -1.503 & -1.492 & -1.471 & -1.411 & -1.420 \\
USpowerGrid & \textbf{-5.633} & -5.619 & -5.593 & -5.564 & -5.532 & -5.629 & -5.614 & -5.595 & -5.579 & -5.626 & -5.609 & -5.576 & -5.573 \\
\midrule
SPC(\sfdpNoSpace)$\downarrow$ & 0.000 & -0.059 & -0.090 & -0.102 & \textbf{-0.106} & 0.009 & 0.018 & 0.015 & 0.017 & 0.011 & 0.028 & 0.036 & 0.039 \\
\bottomrule
\end{tabular}

\end{table*}

\begin{table}[h!]
\centering
\caption{Average variance explained for SuiteSparse14 graphs by PCA projection from 10D of two algorithms, with \sfdp\ showing higher explained variance than \neato.
This suggests that minimum spring-electrical energy is better realized in lower dimensions than minimum stress energy.\label{tab:variance_explained}}
\begin{tabular}{c|cc}
\toprule
Dimension & \neato & \sfdp \\
\midrule
1  & 0.471 & 0.497 \\
2  & 0.702 & 0.754 \\
3  & 0.825 & 0.878 \\
4  & 0.901 & 0.940 \\
5  & 0.948 & 0.972 \\
\bottomrule
\end{tabular}
\end{table}

\section{Statistical Significance of Outperformance of \OptProj\ and \OptPCASfdp}

\begin{figure}
    \centering
    \includegraphics[width=0.9\linewidth]{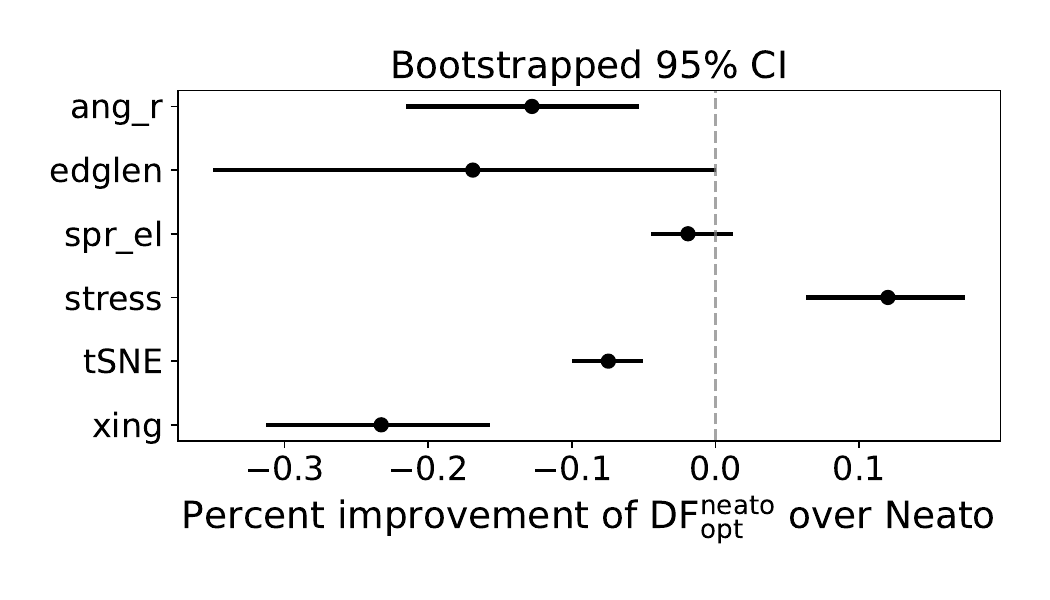}
    \includegraphics[width=0.9\linewidth]{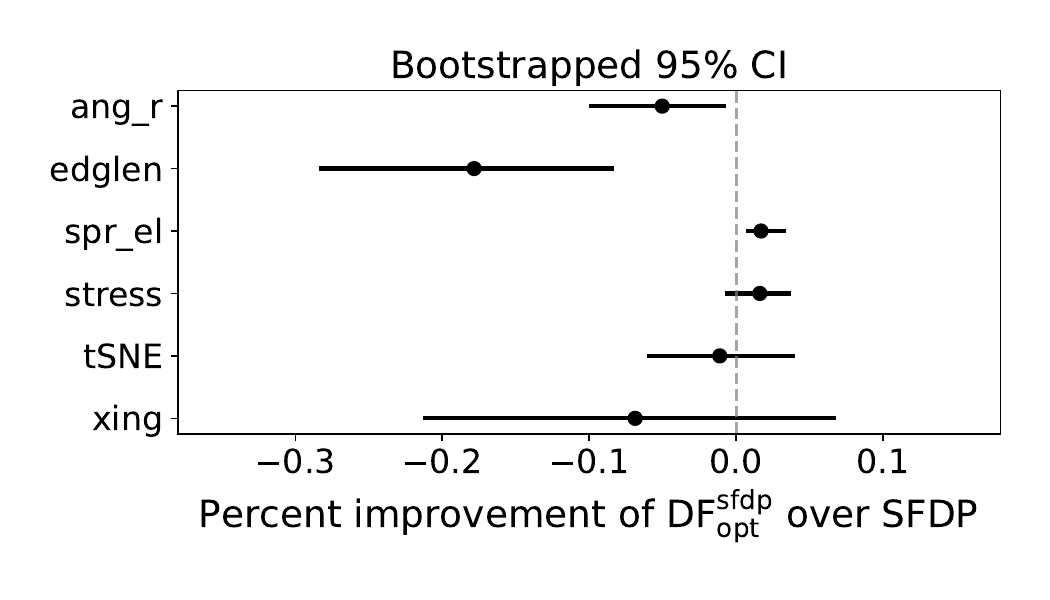}
    \caption{Bootstrapped 95\% confidence intervals for \OptProj\ (top) and \OptProjSfdp\ (bottom) improvement over Neato and SFDP, respectively, on SuiteSparse14 graphs. Negative values favor \OptProj\ and \OptProjSfdp.}
    \label{fig:CI_plots}
\end{figure}

Figure~\ref{fig:CI_plots} 
(top) shows paired-bootstrap ($n=5000$) 95\% confidence intervals (CIs) for the SPC metric across various evaluation measures. When a CI lies entirely below 0, it indicates that \OptProj\ outperforms \neato\ with statistical significance (confirmed via a two-sided bootstrap test). Under this view, \OptProj\ clearly outperforms in angular resolution, t-SNE score, and edge crossings. Paired-bootstrap 95\% confidence intervals (CIs) for the SPC metric across various evaluation measures are given in Figure~A2 in the Appendix. We see that the outperformance of \OptProjSfdp\ over \sfdp\ in angular resolution and edge length variance is statistically significant. 

Additional CIs for the other comparisons are shown in Figures~A3 and~A4. Most of the differences are clear, except for \OptProj\ and \smartgd\ in crossings compared to \sgd. Here, the \smartgd\ interval intersects 0, suggesting that there may be no difference in overall performance in crossing minimization between these graphs. The \OptProj\ interval is close, but does not contain 0. If we compute a $p$-value for whether the mean SPC between the algorithms is 0, we get $0.0394$.

\begin{figure}
    \centering
    \includegraphics[width=0.9\linewidth]{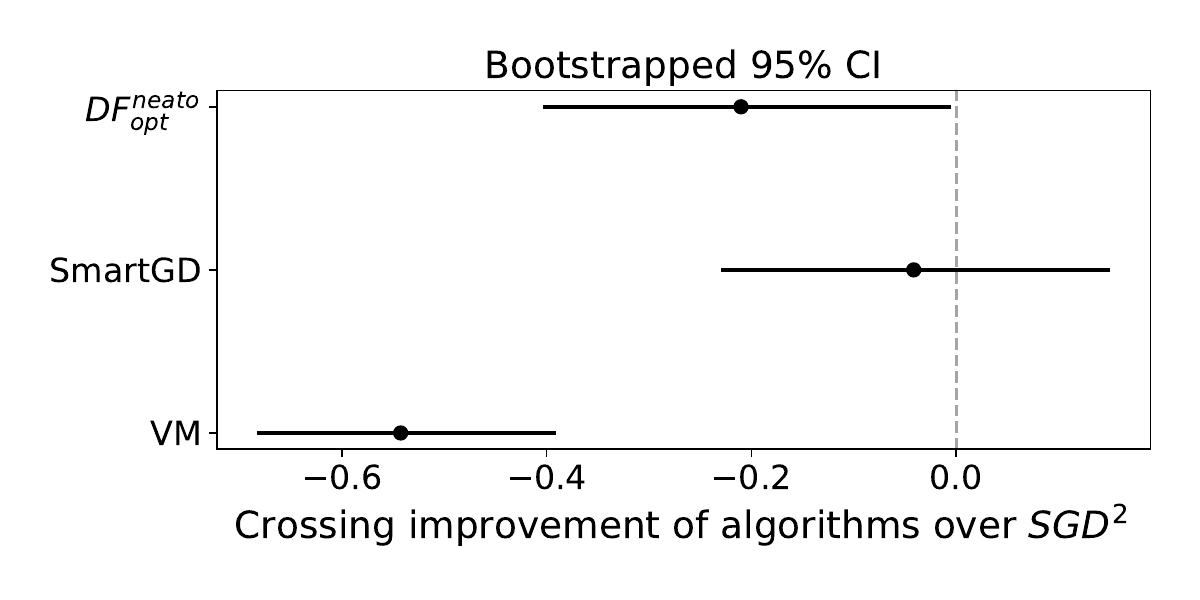}
    \includegraphics[width=0.9\linewidth]{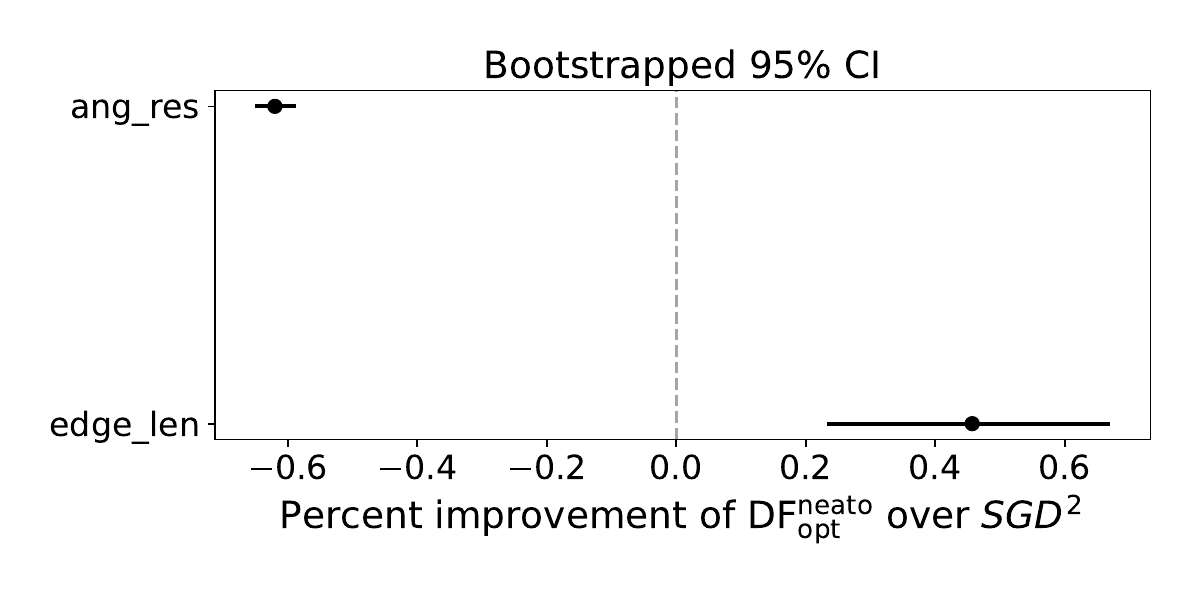}
    \caption{(Top) Bootstrapped CIs for crossing improvement of algorithms compared to $SGD^2$. (Bottom) Bootstrapped CIs for \OptProj improvement over \sgd in angular resolution and edge length variance. Both on the Rome20 graphs.
    }
    \label{fig:additional-ci1}
\end{figure}

\begin{figure}
    \centering
    \includegraphics[width=0.9\linewidth]{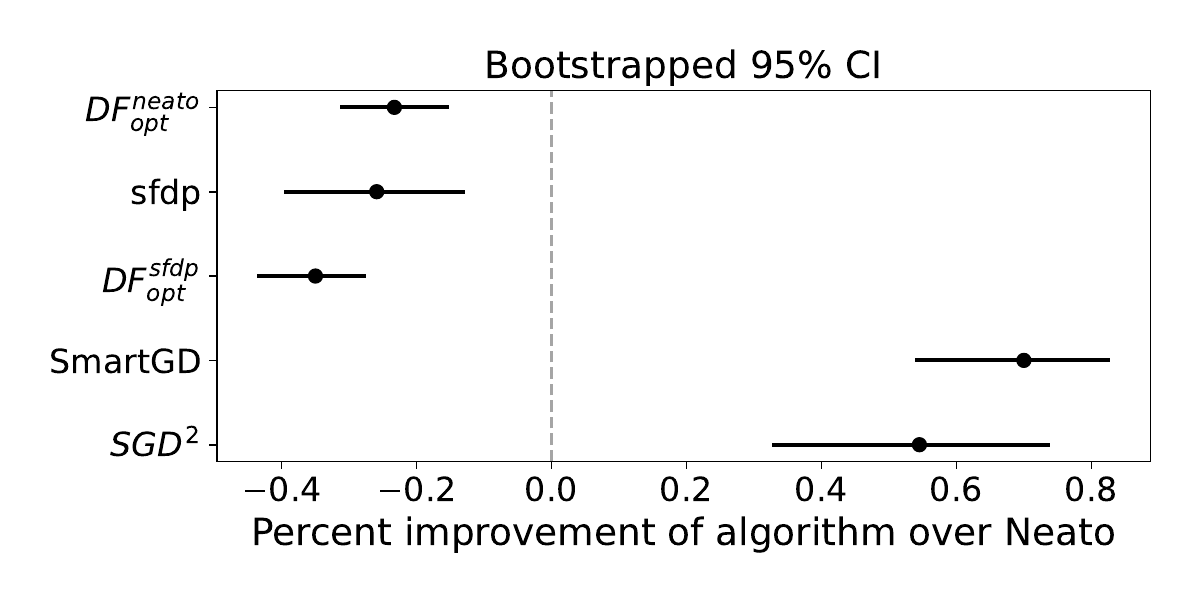}
    \includegraphics[width=0.9\linewidth]{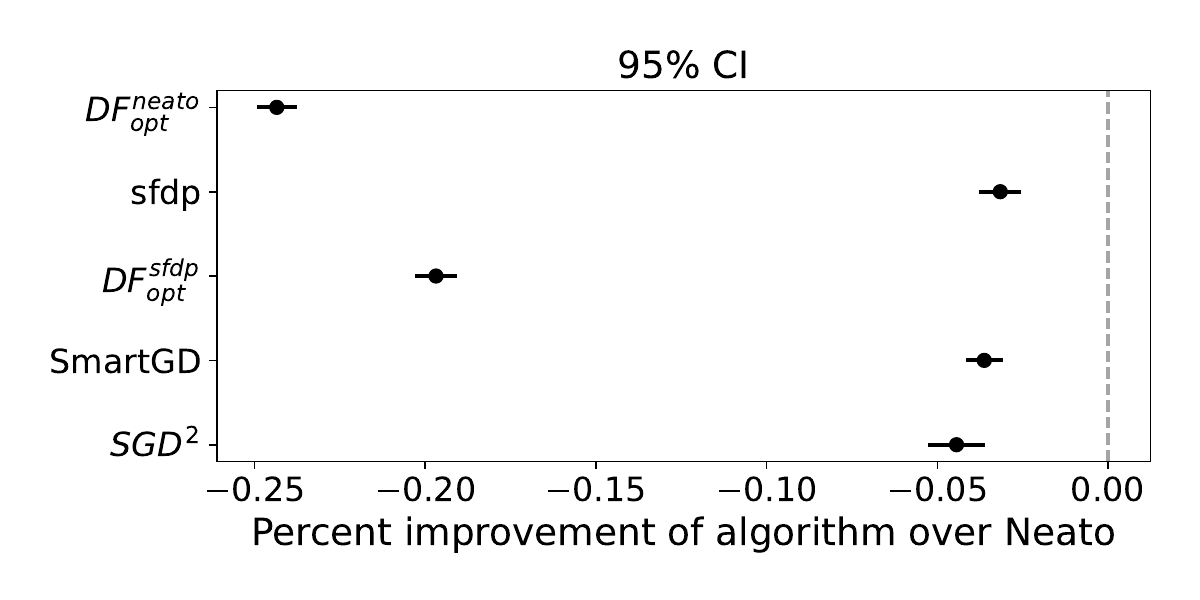}    
    \caption{(Top) Bootstrapped CIs for crossing improvement of algorithms compared to \neato, on the 14 SuiteSparse graphs in our benchmark. 
    (Bottom) Directly computed CIs across all Rome graphs for each algorithm's improvement over Neato in edge crossings.
    }
    \label{fig:additional-ci2}
\end{figure}

\begin{figure}
    \centering
    \includegraphics[width=0.9\linewidth]{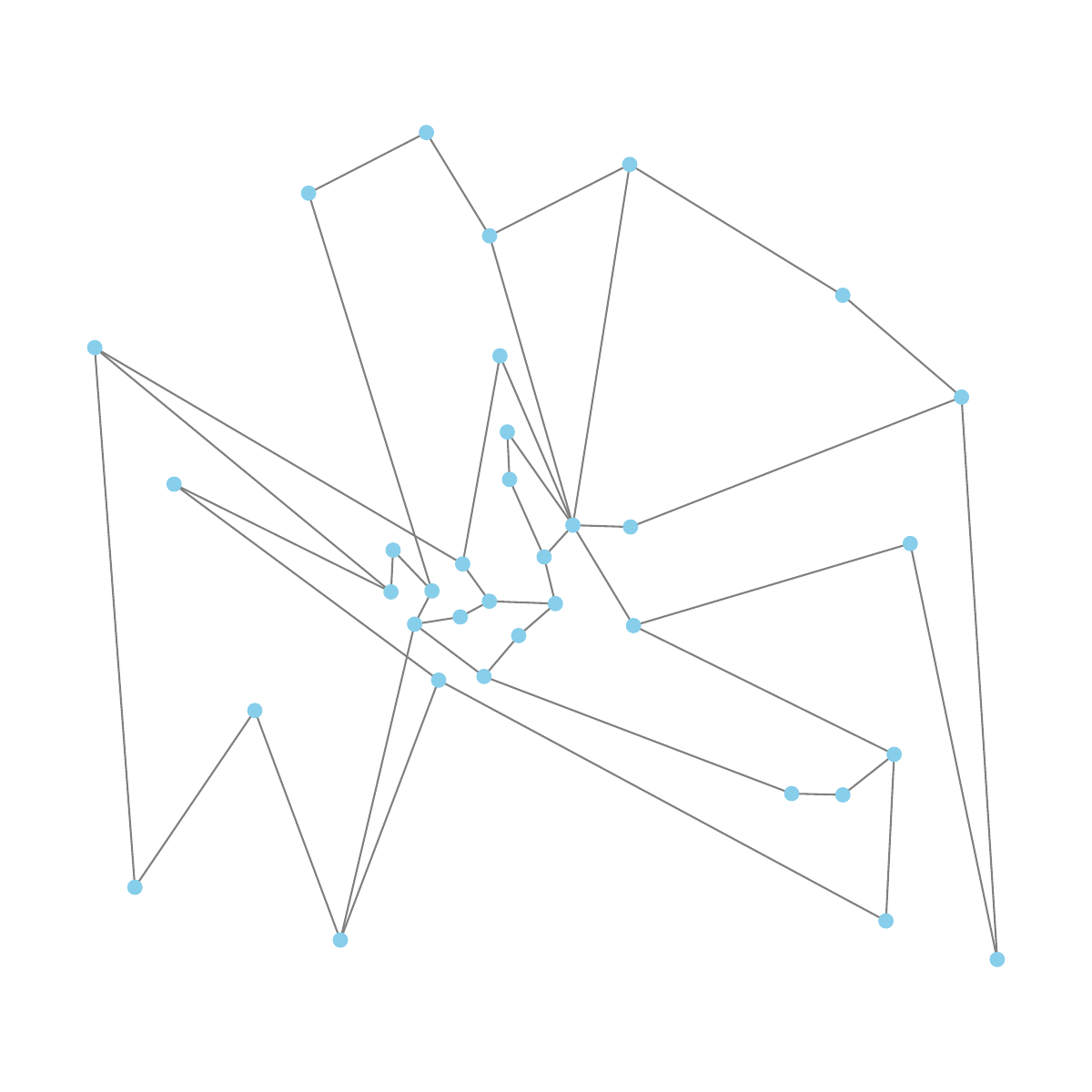}
    \caption{Drawing of the rome graph grafo7023 obtained by the Vertex Movement algorithm. 
            The drawing has a crossing number of 2, but ignores other aesthetic criteria. 
            See Table~\ref{tab:fly_neato_dim10_vis_rome20_b} row 5, last column, for a drawing of the same graph by \OptProj, which is aesthetically much more pleasing but with a crossing number of 8.}
    \label{fig:VM_example}
\end{figure}

\section{System Details}

The system uses a Python/Flask backend with SocketIO for real-time communication, and a JavaScript/HTML/CSS frontend featuring 3D graph visualization via Three.js and interactive metric charts with Chart.js.

We improve efficiency and flexibility through on-demand metric computation and lazy loading.
Expensive metrics like t-SNE are computed and loaded only when requested via asynchronous API calls, allowing users to explore the visualization and basic metrics immediately while more expensive metrics (e.g., t-SNE, edge crossings) are used as needed, keeping the system responsive and preventing blocking. 
Smart caching and cancellation further enhance usability: users can cancel in-progress jobs, toggle metric curves via the legend, and reuse cached results to avoid redundant work.

\section{Scalability of the system}
\label{sec:scalability}

\begin{figure}[htbp]
    \centering
    \setlength{\tabcolsep}{1pt}

    \begin{tabular}{cc}
        \includegraphics[width=0.2\textwidth]{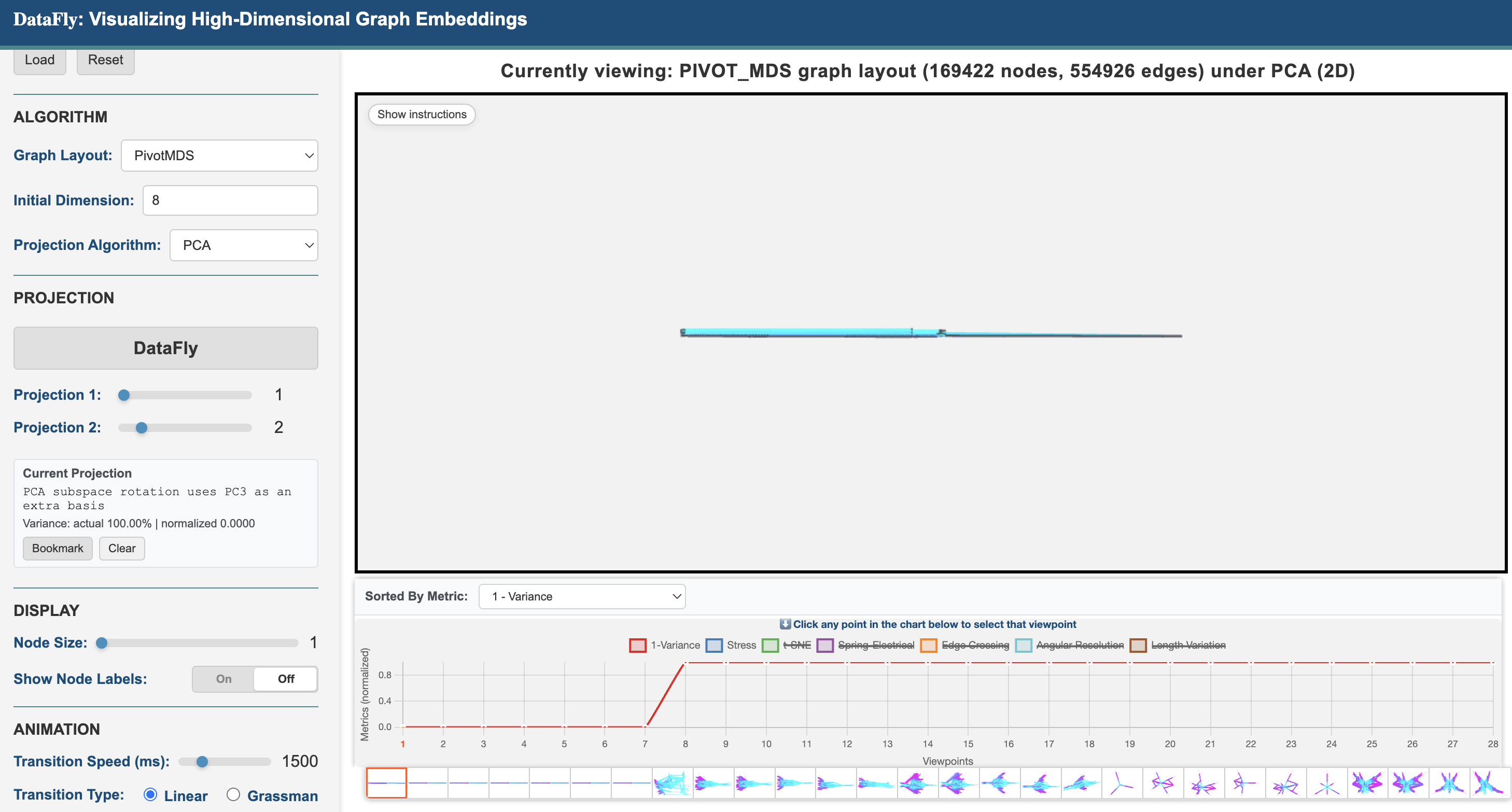} &
        \includegraphics[width=0.2\textwidth]{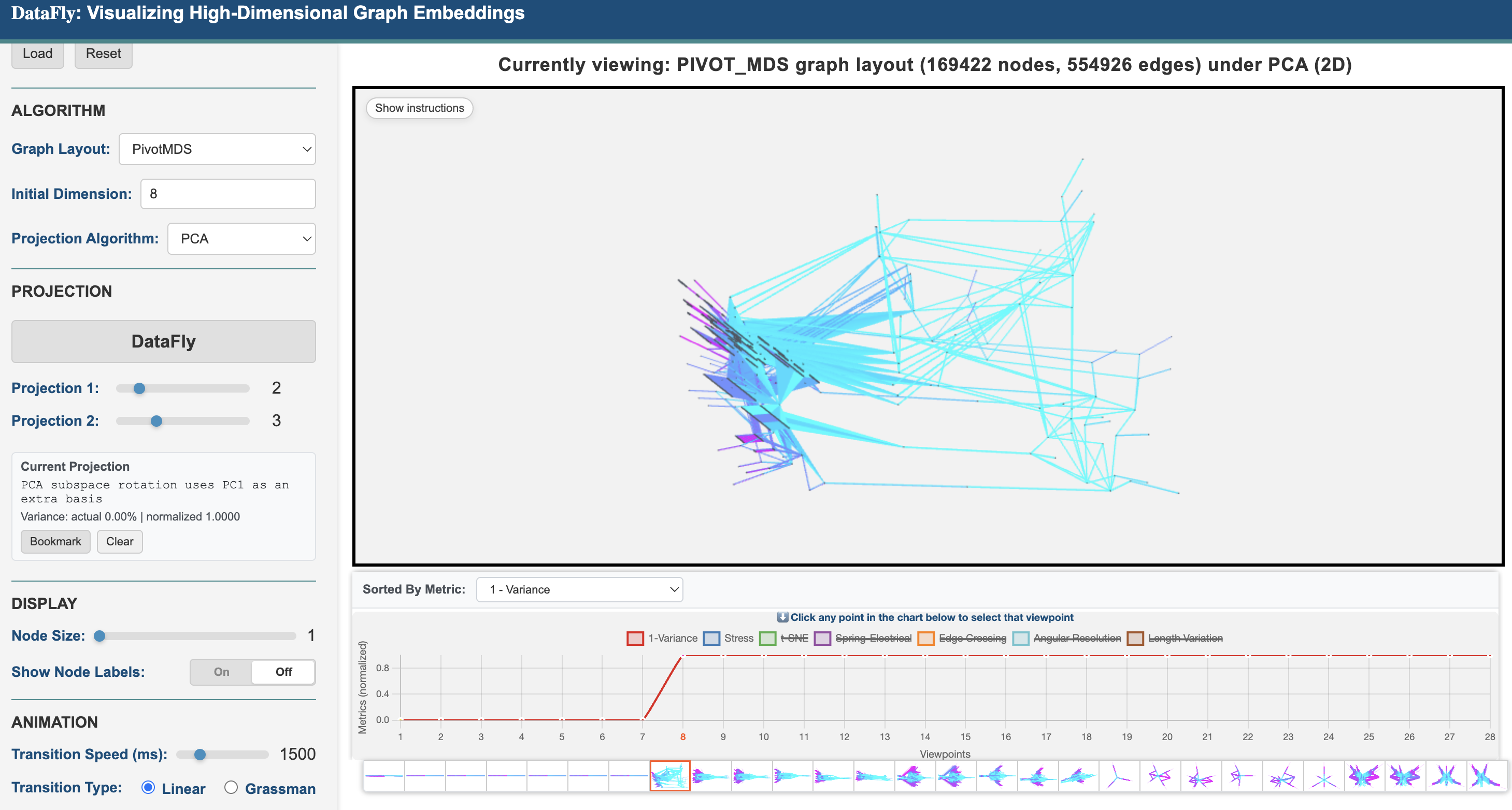} \\
        \includegraphics[width=0.2\textwidth]{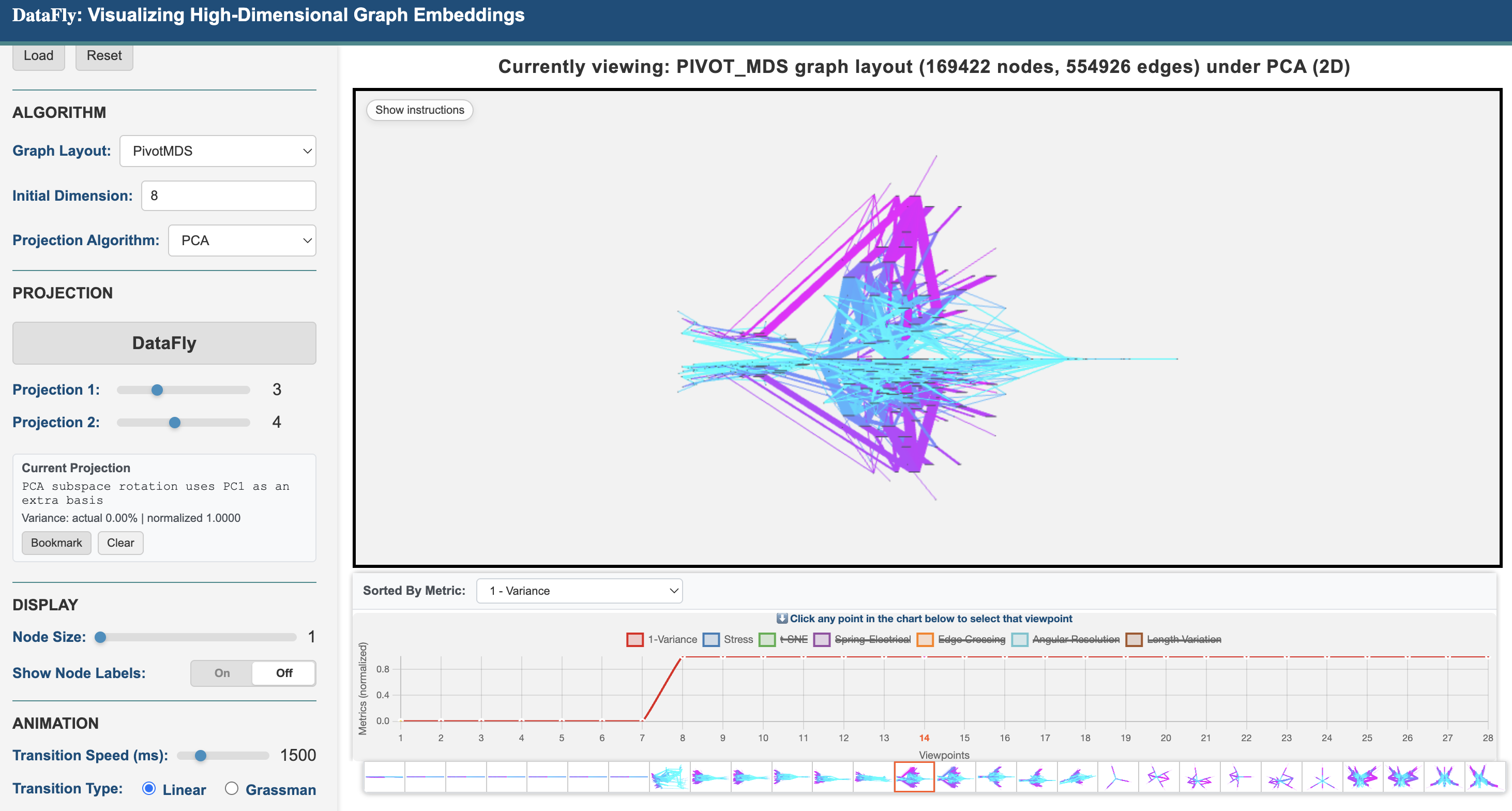} &
        \includegraphics[width=0.2\textwidth]{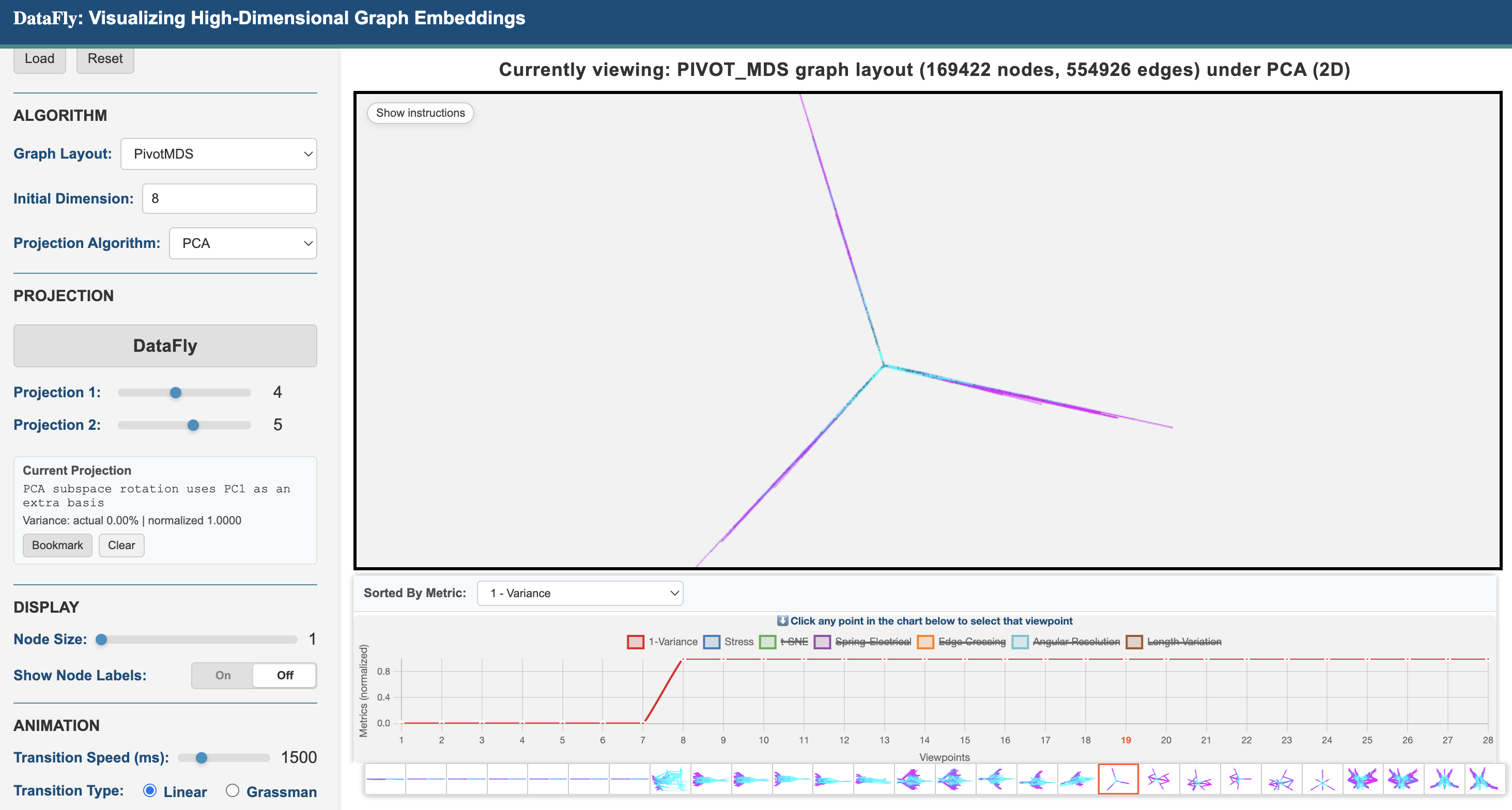}
    \end{tabular}
    \caption{The c-73 graph (169422 nodes, 554926 edges) from four viewpoints in \DataFly.}
    \label{fig:cca_grid}
\end{figure}

Built on \texttt{Three.js}, an OpenGL-based library, \DataFly\ leverages GPU acceleration to enable smooth, real-time interaction with large graphs. On the backend, the server-side pipeline is also optimized for scalability. These, combined with scalable layout algorithms, e.g., \texttt{sfdp} or PivotMDS, allow \DataFly\ to quickly visualize the layouts of large graphs. \Cref{fig:cca_grid} demonstrates the system rendering a  169K-node graph using the \texttt{sfdp} layout from four different viewpoints.

\begin{figure}[htbp]
    \centering
    \includegraphics[width=0.9\linewidth]{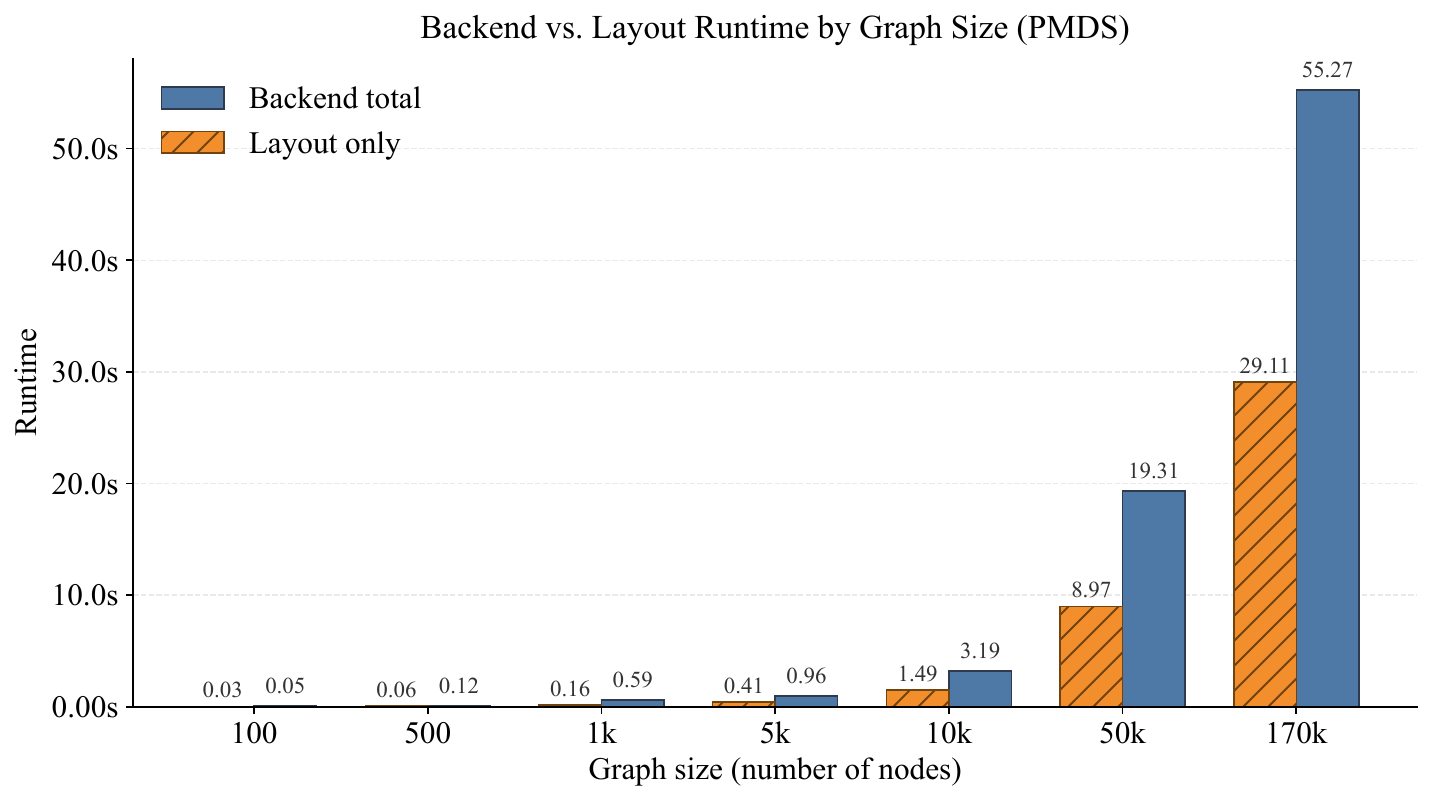}
    \caption{Computation time comparison between Pivot MDS only and Pivot MDS + system processing}
    \label{fig:pmds_time}
\end{figure}

\begin{figure}[htbp]
    \centering
    \includegraphics[width=0.9\linewidth]{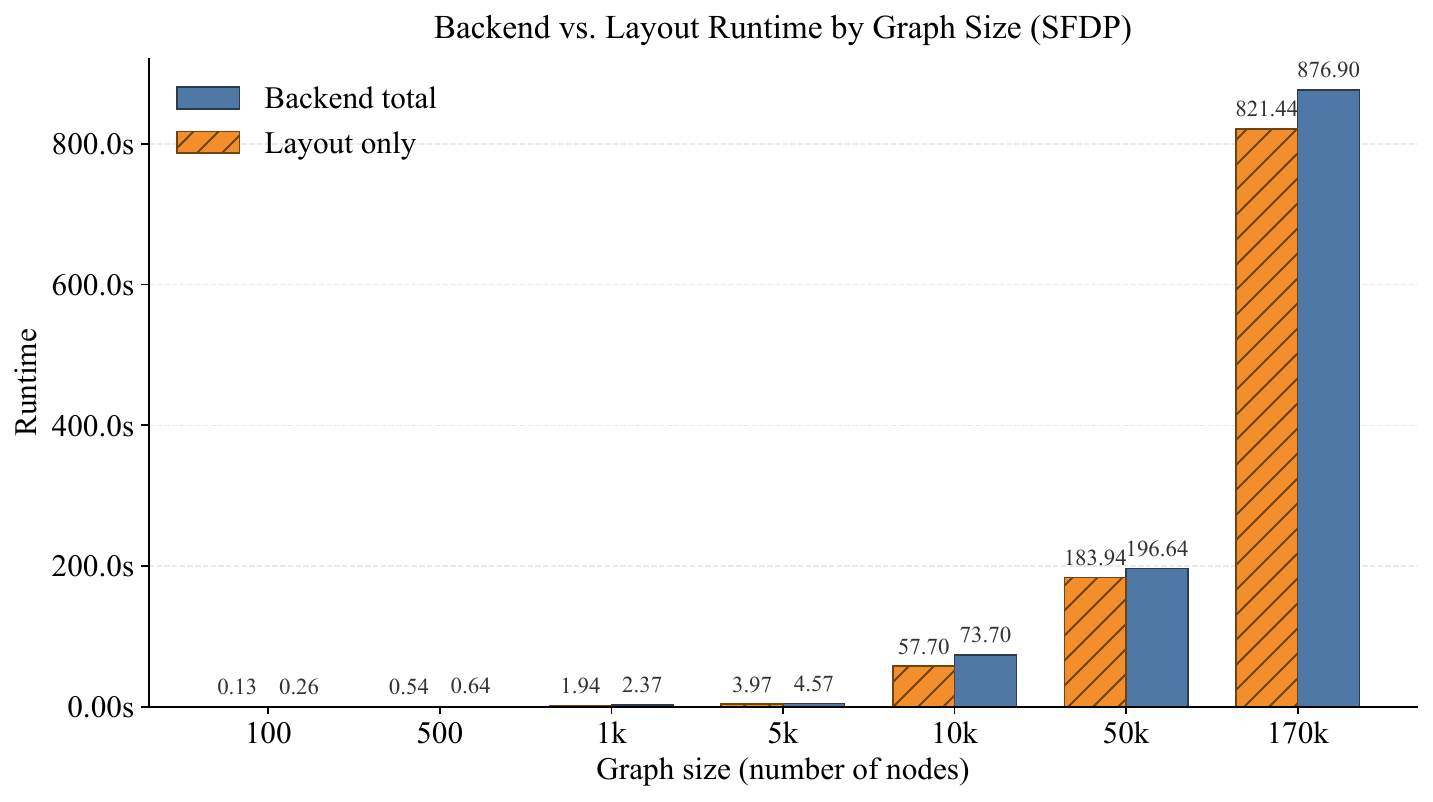}
    \caption{Computation time comparison between sfdp only and sfdp + system processing}
    \label{fig:sfdp_time}
\end{figure}

To quantitatively evaluate the practical scalability of our system, we measured both the total backend runtime and the time spent computing layouts across graphs of increasing size. We selected graphs from the SuiteSparse matrix collection, ranging from 115 to 170k nodes: football (115 nodes), Erdos991 (446 nodes), email (1133 nodes), USpower (4941 nodes), c-40 (9941 nodes), c-66 (49989 nodes) and c-73 (169422 nodes).  Figures \ref{fig:pmds_time} and \ref{fig:sfdp_time} show that layout computation dominates the overall runtime, especially for larger graphs. The gap between the total backend time and the layout-only time remains relatively modest compared with the layout cost itself, indicating that the main computational bottleneck lies in the layout generation procedure rather than in backend overhead.
This suggests that the system is scalable to large graphs.

\section{Usability study, improvements, and future work} \label{sec:appendix_improvements}
\begin{figure*}
    \centering
    \includegraphics[width=0.32\linewidth]{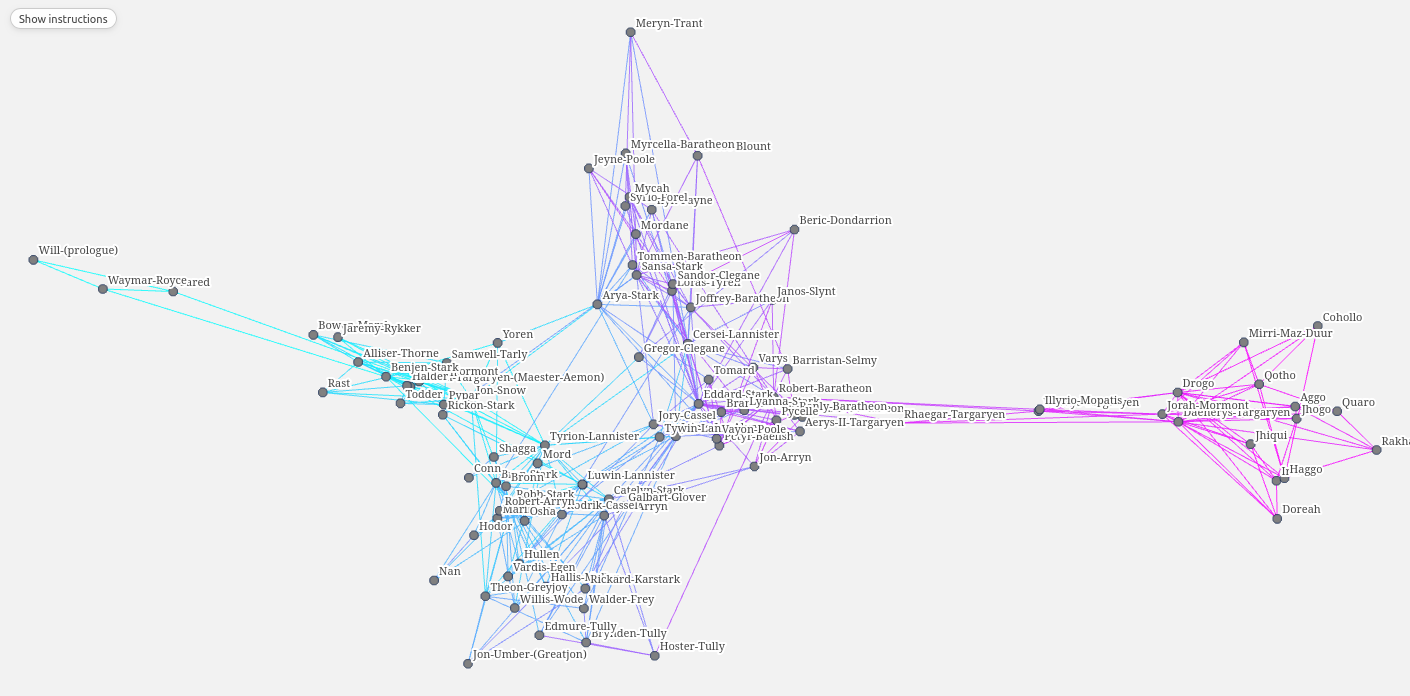}
    \includegraphics[width=0.32\linewidth]{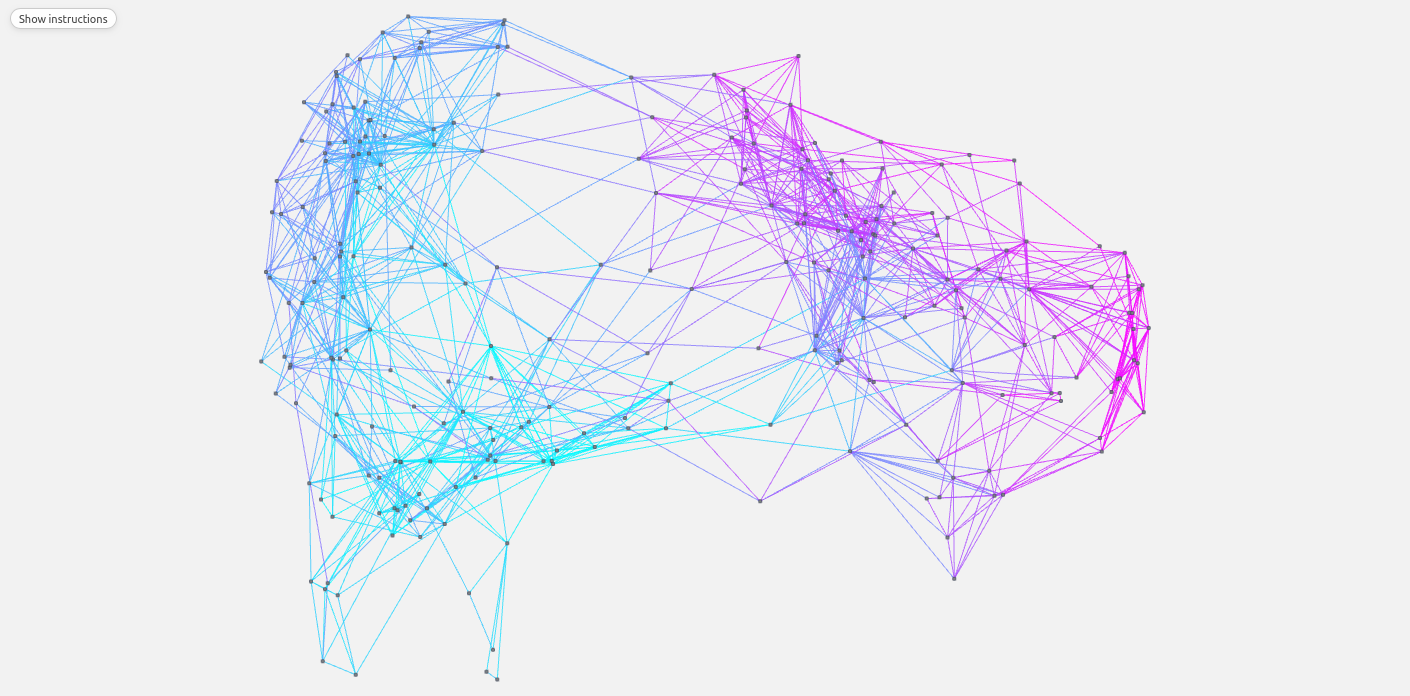}
    \includegraphics[width=0.32\linewidth]{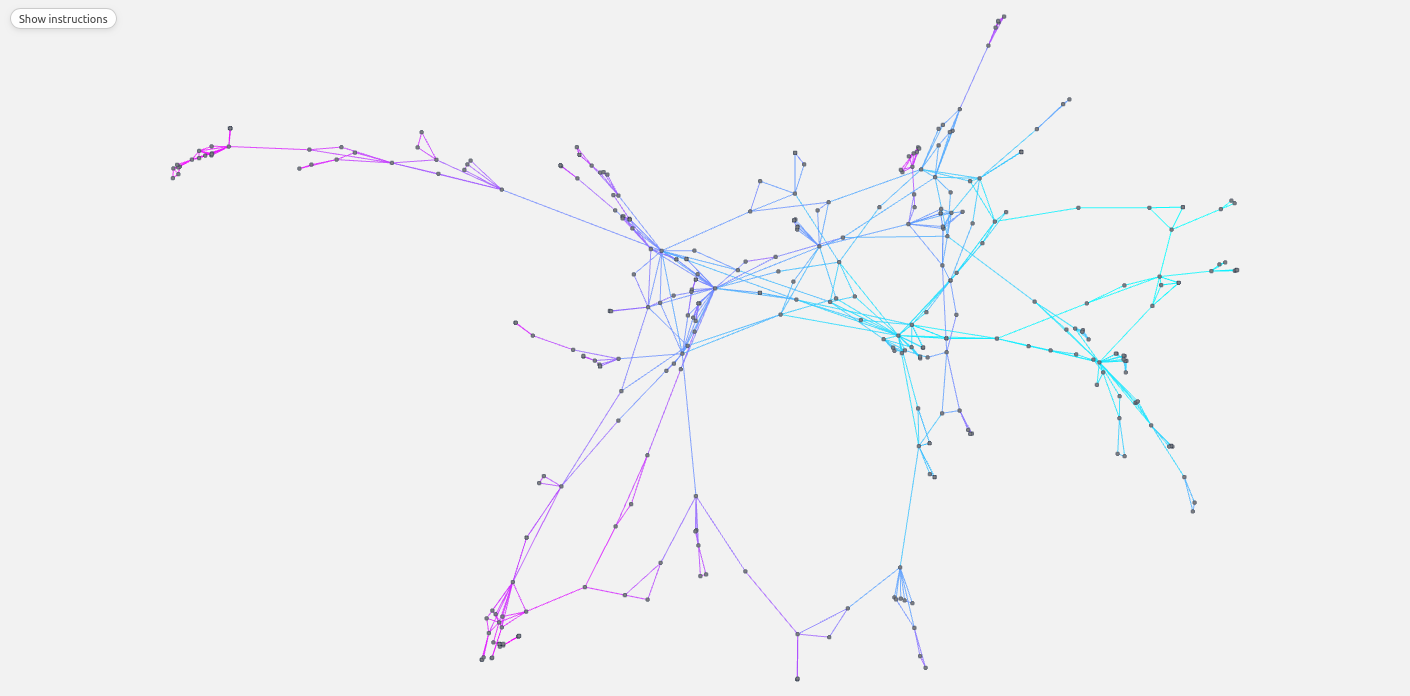}    

    \parbox[c]{0.32\linewidth}{\centering (a)}
    \parbox[c]{0.32\linewidth}{\centering (b)}
    \parbox[c]{0.32\linewidth}{\centering (c)}    
    \caption{(a) the Game of Thrones network, (b) the Last.fm network, (c) the Vis coauthorship network}
    \label{fig:supplemental-figs}
\end{figure*}

\Cref{fig:supplemental-figs} shows the remaining graphs used by participants in the usability study; the Game of Thrones network, Last.fm music network, and Vis coauthorship network. The three graphs show different types of structure and cover different domains.

The tool received consistently high marks for visual appeal (median 4.5) and user interaction quality (median 4.5). Intuitiveness ratings were high among experts and N1 (all rated 4) but lower for N2 (rated 2), who struggled with interpreting the axes and viewpoints. Suitability scores were consistently high (median 4). Likelihood of use for work or personal interest varied (range 2--4). Exact responses are found in Table~\ref{tab:likert}.

Expert E1 and non-expert N2 from the usability study suggested additional interaction features. E1 proposed automatic neighborhood highlighting (selecting a node and its immediate neighbors) and edge-based tracking through transitions, while N2 suggested a ``lock-rotation'' button and highlight-on-hover functionality.  

In response, we have implemented several improvements in the live tool. A pause-and-resume function for animated transitions allows users to freeze the rotation at any viewpoint using the space key, supporting detailed inspection. Dynamic node and label highlighting, with enlargement on hover, improves visual clarity, particularly in dense projections.  

Building on participant feedback, we have enhanced the system's interaction support, including neighboring nodes and edges highlighting, to reduce manual effort during neighbor-tracing tasks. To bridge the gap between expert and non-expert users, we have included a guided walkthrough video, and an instruction panel. We also plan to introduce contextual tool-tips, and progressive disclosure of advanced features to manage interaction complexity. 

\begin{figure}
    \centering
    \includegraphics[width=0.49\linewidth]{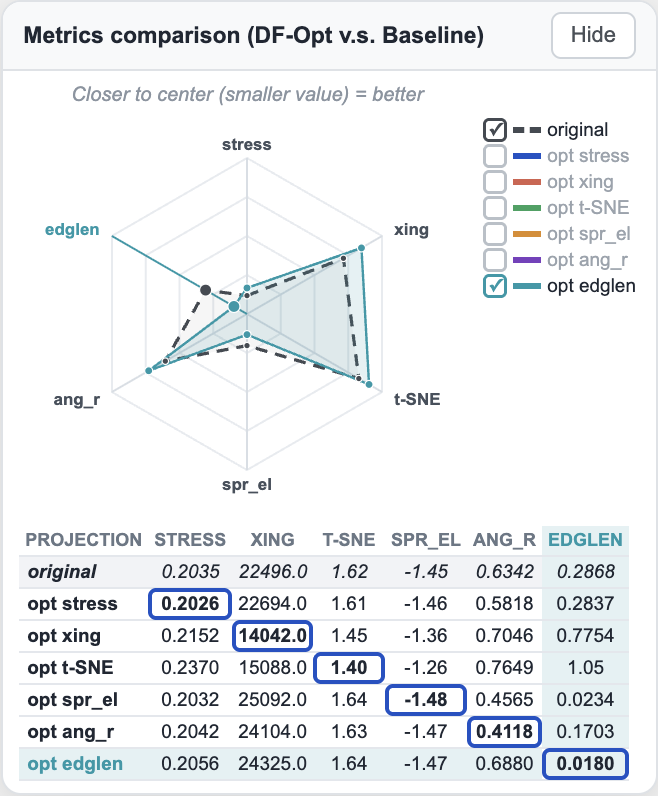}
    \includegraphics[width=0.49\linewidth]{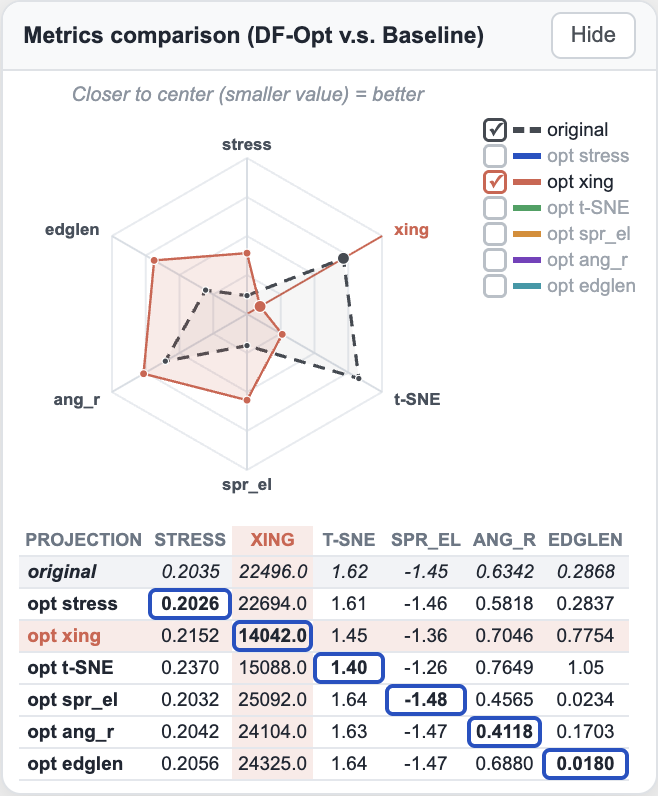}
    \parbox[c]{0.49\linewidth}{\centering (a)}
    \parbox[c]{0.49\linewidth}{\centering (b)}
    \caption{(a) Metrics comparison panel with the metrics of edge length variance-optimized layout (opt edglen, teal) vs. the metrics of baseline (2D sfdp, dashed), plotted on a radar chart; the table below reports raw values, with the per-metric best value boxed. (b) The same panel with the edge crossing-optimized layout (opt xing, orange) selected, highlighting its axis in the radar chart and its column in the table; values closer to the center of the radar chart are considered better (smaller).}
    \label{fig:metrics_comparison_component}
\end{figure}

We also added a metrics-comparison panel (as shown in \cref{fig:metrics_comparison_component} and \cref{fig:system_showcase} \circmark{L}) to the ``optimal projection" viewport, where users can directly compare the optimized layout against standard baseline layouts (2D \neato, \sfdp, \pivotmds, or \spectral) across all six quality metrics (stress, crossings, t-SNE, spring-electrical energy, angular resolution, and edge-length variance). The comparison is presented through a table, analogous to \cref{tab:combined_suitesparse} but more visually expressive, together with an interactive radar chart. This lets users visually assess and quantitatively compare where the optimized result improves over the baselines, making the framework's benefits more concrete and improving user understanding and effectiveness when reasoning about projection quality.

For future work, we plan to explore optimal ``tours'' between viewpoints—extending current linear or Grassmann geodesic transitions—to minimize both travel distance and a chosen quality metric (e.g., integrated stress), inspired by projection pursuit~\cite{cook1995grand}. 

\begin{table}[htbp]
\setlength{\tabcolsep}{4pt}
\centering
\caption{Likert-scale ratings (1 = poor, 5 = excellent) from all four participants.}
\label{tab:likert}
\small
\begin{tabular}{lccccc}
\toprule
\textbf{Participant} & \textbf{Intuitive} & \textbf{Would Use} & \textbf{Appeal} & \textbf{Suitable} & \textbf{Interaction} \\
\midrule
Expert 1 (E1) & 4 & 4 & 4 & 4 & 5 \\
Expert 2 (E2) & 4 & 3 & 5 & 4 & 4 \\
Non-expert 1 (N1) & 4 & 4 & 4 & 5 & 4 \\
Non-expert 2 (N2) & 2 & 2 & 5 & 4 & 5 \\
\midrule
Median & 4 & 3.5 & 4.5 & 4 & 4.5 \\
\bottomrule
\end{tabular}
\end{table}

\section{Additional Experiment on Embedding Dimensionality}
\label{sec_appendix_30D}
Graphviz's layout algorithm implementation restricts the maximum layout dimension to \(K=10\). To examine whether this implementation limit affects our conclusions, we compiled Graphviz 15.0.0 from source with the internal dimensionality increased from 10 to 30. We then repeated the same projection-and-evaluation procedure used in the main experiments on the Rome-20 graphs for
\[
K \in \{2,10,15,20,25,30\}.
\]

Overall, the results in Tables~\ref{tab:add-dim-stress}, \ref{tab:add-dim-spr_elec}, and \ref{tab:add-dim-xing} do not show a consistent monotonic improvement from using embedding dimensions beyond \(K=10\).
\begin{table}[htbp]
\scriptsize\setlength{\tabcolsep}{5pt}\renewcommand{\arraystretch}{1}
\centering
\caption{Stress for optimized projections of \texttt{neato} layouts beyond 10 dimensions.}
\label{tab:add-dim-stress}
\begin{tabular}{lrrrrrr}
\toprule
 & \multicolumn{6}{c}{\texttt{neato} at $K$} \\
\cmidrule(lr){2-7}
Graph & 2 & 10 & 15 & 20 & 25 & 30 \\
\midrule
grafo10379.95 & \textbf{0.0819} & 0.0891 & 0.0861 & 0.0861 & 0.0848 & 0.0848 \\
grafo10451.95 & \textbf{0.0923} & 0.0991 & 0.0966 & 0.0957 & 0.0956 & 0.0953 \\
grafo10480.95 & \textbf{0.0804} & 0.0878 & 0.0991 & 0.0870 & 0.0868 & 0.0977 \\
grafo11311.40 & \textbf{0.0645} & 0.0699 & 0.0751 & 0.0733 & 0.0761 & 0.0755 \\
grafo11412.32 & 0.0343 & 0.0328 & 0.0373 & 0.0349 & 0.0325 & \textbf{0.0324} \\
grafo11543.34 & \textbf{0.0360} & 0.0416 & 0.0417 & 0.0415 & 0.0416 & 0.0415 \\
grafo11674.33 & 0.0757 & 0.0574 & \textbf{0.0574} & 0.0628 & 0.0614 & 0.0624 \\
grafo1583.66 & \textbf{0.0595} & 0.0664 & 0.0652 & 0.0649 & 0.0645 & 0.0647 \\
grafo2747.16 & \textbf{0.0097} & 0.0105 & 0.0107 & 0.0101 & 0.0106 & 0.0107 \\
grafo2815.35 & \textbf{0.0297} & 0.0338 & 0.0333 & 0.0332 & 0.0331 & 0.0332 \\
grafo2955.38 & \textbf{0.0676} & 0.0683 & 0.0754 & 0.0677 & 0.0695 & 0.0697 \\
grafo3138.60 & 0.1026 & 0.0942 & 0.0929 & 0.0926 & \textbf{0.0924} & 0.0930 \\
grafo3587.36 & \textbf{0.0409} & 0.0443 & 0.0484 & 0.0475 & 0.0449 & 0.0474 \\
grafo5780.48 & \textbf{0.0835} & 0.0888 & 0.0889 & 0.0900 & 0.0891 & 0.0900 \\
grafo7023.45 & \textbf{0.0652} & 0.0685 & 0.0685 & 0.0675 & 0.0672 & 0.0675 \\
grafo762.27 & \textbf{0.0521} & 0.0567 & 0.0569 & 0.0604 & 0.0579 & 0.0657 \\
grafo8296.86 & 0.1006 & 0.1067 & \textbf{0.0992} & 0.1042 & 0.1020 & 0.1010 \\
grafo847.22 & 0.0752 & \textbf{0.0540} & 0.0542 & 0.0542 & 0.0542 & 0.0543 \\
grafo9033.80 & \textbf{0.0817} & 0.0903 & 0.0898 & 0.0987 & 0.0891 & 0.0892 \\
grafo9592.72 & \textbf{0.0413} & 0.0465 & 0.0432 & 0.0434 & 0.0474 & 0.0440 \\
\midrule
SPC & \textbf{0} & +0.032 & +0.044 & +0.038 & +0.031 & +0.041 \\
\bottomrule
\end{tabular}
\end{table}

\begin{table}[htbp]
\scriptsize\setlength{\tabcolsep}{3.5pt}
\renewcommand{\arraystretch}{1}
\centering
\caption{Spring-electrical energy for optimized projections of \texttt{neato} layouts beyond 10 dimensions.}
\label{tab:add-dim-spr_elec}
\begin{tabular}{lrrrrrr}
\toprule
 & \multicolumn{6}{c}{\texttt{neato} at $K$} \\
\cmidrule(lr){2-7}
Graph & 2 & 10 & 15 & 20 & 25 & 30 \\
\midrule
grafo10379.95 & -2.1561 & -2.1620 & -2.1747 & -2.1770 & -2.1816 & \textbf{-2.1826} \\
grafo10451.95 & -2.0617 & -2.0652 & -2.0805 & -2.0836 & -2.0848 & \textbf{-2.0849} \\
grafo10480.95 & -2.2346 & -2.2471 & -2.2599 & -2.2544 & \textbf{-2.2614} & -2.2602 \\
grafo11311.40 & \textbf{-1.7590} & -1.7454 & -1.7481 & -1.7507 & -1.7580 & -1.7473 \\
grafo11412.32 & -1.9866 & -2.0332 & -2.0273 & -2.0344 & \textbf{-2.0358} & -2.0354 \\
grafo11543.34 & -1.8307 & -1.8381 & \textbf{-1.8388} & -1.8388 & -1.8381 & -1.8384 \\
grafo11674.33 & -1.6260 & \textbf{-1.7255} & -1.7227 & -1.7180 & -1.7167 & -1.7247 \\
grafo1583.66 & -2.2531 & -2.2486 & -2.2537 & -2.2558 & \textbf{-2.2578} & -2.2463 \\
grafo2747.16 & -1.4619 & -1.4737 & \textbf{-1.4798} & -1.4737 & -1.4783 & -1.4794 \\
grafo2815.35 & -1.8821 & -1.8938 & -1.8949 & -1.8983 & -1.8983 & \textbf{-1.8986} \\
grafo2955.38 & -1.6858 & \textbf{-1.7049} & -1.6926 & -1.7025 & -1.7008 & -1.7030 \\
grafo3138.60 & -1.7618 & -1.8119 & \textbf{-1.8167} & -1.8154 & -1.8139 & -1.8162 \\
grafo3587.36 & -1.7521 & -1.7181 & -1.7574 & \textbf{-1.7586} & -1.7236 & -1.7214 \\
grafo5780.48 & \textbf{-1.6879} & -1.6631 & -1.6678 & -1.6650 & -1.6654 & -1.6654 \\
grafo7023.45 & -1.8336 & -1.8462 & \textbf{-1.8496} & -1.8330 & -1.8427 & -1.8455 \\
grafo762.27 & -1.5619 & -1.5678 & \textbf{-1.5690} & -1.5670 & -1.5610 & -1.5671 \\
grafo8296.86 & -2.0366 & -2.0693 & -2.0834 & -2.0826 & -2.0620 & \textbf{-2.0847} \\
grafo847.22 & -1.1943 & \textbf{-1.2585} & -1.2545 & -1.2576 & -1.2558 & -1.2537 \\
grafo9033.80 & -2.0719 & -2.0765 & -2.0832 & \textbf{-2.0868} & -2.0853 & -2.0810 \\
grafo9592.72 & -2.5773 & -2.6607 & -2.6635 & -2.6659 & -2.6659 & \textbf{-2.6667} \\
\midrule
SPC & 0 & -0.011 & \textbf{-0.013} & -0.013 & -0.013 & -0.013 \\
\bottomrule
\end{tabular}
\end{table}

\begin{table}[htbp]
\scriptsize\setlength{\tabcolsep}{5pt}\renewcommand{\arraystretch}{1}
\centering
\caption{Crossings for optimized projections of \texttt{neato} layouts beyond 10 dimensions.}
\label{tab:add-dim-xing}
\begin{tabular}{lrrrrrr}
\toprule
 & \multicolumn{6}{c}{\texttt{neato} at $K$} \\
\cmidrule(lr){2-7}
Graph & 2 & 10 & 15 & 20 & 25 & 30 \\
\midrule
grafo10379.95 & 90 & 71 & \textbf{68} & 76 & 80 & 76 \\
grafo10451.95 & 113 & 84 & 95 & 77 & \textbf{75} & 102 \\
grafo10480.95 & 72 & 61 & 55 & 64 & 61 & \textbf{50} \\
grafo11311.40 & 10 & \textbf{5} & 12 & 11 & 12 & 7 \\
grafo11412.32 & 3 & 1 & 1 & 1 & \textbf{0} & 1 \\
grafo11543.34 & 4 & 1 & 3 & \textbf{0} & 4 & 1 \\
grafo11674.33 & 12 & \textbf{4} & 4 & 6 & 5 & 4 \\
grafo1583.66 & \textbf{13} & 15 & 16 & 18 & 21 & 18 \\
grafo2747.16 & \textbf{0} & 0 & 0 & 0 & 0 & 0 \\
grafo2815.35 & 3 & \textbf{1} & 3 & 1 & 3 & 1 \\
grafo2955.38 & 10 & 11 & 6 & \textbf{5} & 10 & 7 \\
grafo3138.60 & 53 & 36 & 42 & \textbf{29} & 46 & 49 \\
grafo3587.36 & \textbf{6} & 7 & 7 & 6 & 7 & 7 \\
grafo5780.48 & 31 & 29 & 31 & 24 & 26 & \textbf{22} \\
grafo7023.45 & 15 & \textbf{7} & 8 & 10 & 8 & 8 \\
grafo762.27 & 3 & 2 & \textbf{0} & 2 & 2 & 2 \\
grafo8296.86 & 79 & 78 & 75 & 82 & 71 & \textbf{62} \\
grafo847.22 & 7 & 4 & 4 & 3 & \textbf{1} & 4 \\
grafo9033.80 & 53 & 56 & 54 & \textbf{39} & 40 & 44 \\
grafo9592.72 & 17 & \textbf{8} & 8 & 15 & 14 & 9 \\
\midrule
SPC & 0 & -0.284 & -0.240 & -0.291 & -0.198 & \textbf{-0.297} \\
\bottomrule
\end{tabular}
\end{table}

\section{Additional Related Work}
Our work connects to broader themes in graph and network visualization quality assessment. Bertini and Santucci~\cite{bertini2006visual} proposed a systematization of visual quality metrics for information visualization, classifying them into size, visual effectiveness, and feature preservation categories, justifying the principled use of quality metrics that we adopt for evaluating viewpoints. In a similar spirit, diagnostic approaches have been developed for specific visual representations: Gove~\cite{gove2019gragnostics} introduced \emph{Gragnostics}, a set of fast, interpretable graph-level features for comparing graph topologies, while Behrisch et al.~\cite{behrisch2017magnostics} proposed \emph{Magnostics}, an image-based approach for ranking matrix views of networks according to the presence of visual patterns such as blocks and lines, and identified extending their diagnostics to hybrid representations such as \emph{NodeTrix}~\cite{henry2007nodetrix} as future work. \emph{NodeTrix}, introduced by Henry et al., combines node-link diagrams with adjacency matrices to simultaneously convey global network structure and local community detail, highlighting the long-standing challenge of representing dense substructures and reinforcing the case for our multi-view approach.

\end{document}